\documentclass[a4paper, 12pt]{article}
\usepackage{varwidth}
\usepackage{lipsum}
\usepackage{threeparttable}
\usepackage{makecell}
\usepackage{amsfonts}
\usepackage{amsmath}
\usepackage{mathrsfs}
\usepackage{fancyhdr}
\usepackage{epsfig,morefloats}
\usepackage[round]{natbib}
\usepackage{booktabs,multirow,multicol,longtable}
\usepackage{amsthm}
\usepackage{amssymb}
\usepackage{lscape}
\usepackage{rotating}
\usepackage{caption}
\usepackage{subfigure}
\usepackage{graphicx}
\usepackage{wrapfig}
\usepackage[compact]{titlesec}
\titlespacing*{\subsection}{0pt}{10pt}{0pt}
\titlespacing*{\subsubsection}{0pt}{10pt}{0pt}
\usepackage[breaklinks,colorlinks,linkcolor=black,citecolor=black,urlcolor=black]{hyperref}
\usepackage{lineno}
\usepackage{cancel}
\usepackage{multirow}
\usepackage{xcolor}
\usepackage{setspace}

\usepackage{algorithm}
\usepackage{algpseudocode}
\usepackage{enumitem}
\usepackage{mathtools}  

\definecolor{mygray}{gray}{0.6}

\newcommand{\Date}[1]{\def\@Date{#1}}
\def\today{\number\day~\ifcase\month\or
	January\or February\or March\or April\or May\or June\or
	July\or August\or September\or October\or November\or December\fi~\number\year}

\allowdisplaybreaks

\graphicspath{{Fig/}}

\newtheorem{theorem}{Theorem}
\newtheorem{remark}{Remark}%
\newcommand{\romaninline}[1]{\textup{(\romannumeral #1)}}
\newtheorem{definition}{Definition}
\newtheorem{assumption}{Assumption}
\newcommand{\E}{\mathbb{E}}

\newcommand{\N}{\mathcal{N}}

\newcommand{\T}{\mathcal{T}}
\newcommand{\F}{\mathcal{F}}

\newcommand{\R}{\mathbb{R}}

\definecolor{cbblue}{HTML}{0173b2}
\definecolor{cborange}{HTML}{de8f05}
\definecolor{cbgreen}{HTML}{029e73}
\definecolor{cbred}{HTML}{d55e00}
\definecolor{cbpurple}{HTML}{cc78bc}
\definecolor{cbyellow}{HTML}{ca9161}  

\definecolor{cbgreen2}{HTML}{018571}     
\definecolor{cbblack}{HTML}{252525}   
\definecolor{cbdarkblue}{HTML}{091e75}   
\definecolor{cbbrown}{HTML}{662506}      
\definecolor{cbmagenta}{HTML}{850157}
\definecolor{cbgrey}{HTML}{5c5c5c} 
\definecolor{cbblue2}{HTML}{023eff}  
\definecolor{cborange2}{HTML}{FA8072} 
\definecolor{cbblue3}{HTML}{87CEEB}

\newtheorem{lemma}{Lemma}

\usepackage{authblk}
\usepackage{tikz}
\usetikzlibrary{shapes.geometric}

\usepackage{geometry}
\usepackage[scaled=0.95]{helvet}  
\usepackage{mathpazo}             

\usepackage{microtype}            
\SetTracking{encoding=*, shape=*}{25}  
\selectfont      
\usepackage[compact]{titlesec}
\titleformat{\section}
  {\normalfont\large\bfseries}{\thesection}{1em}{}
\titlespacing{\section}{0pt}{1.2ex}{0.8ex}

\usepackage{etoolbox} 

\usepackage{xr-hyper}
\usepackage{pdfpages} 
\usepackage{pgffor} 

\makeatletter
\AtBeginDocument{\let\LS@rot\@undefined}
\makeatother

\def\supplementfilename{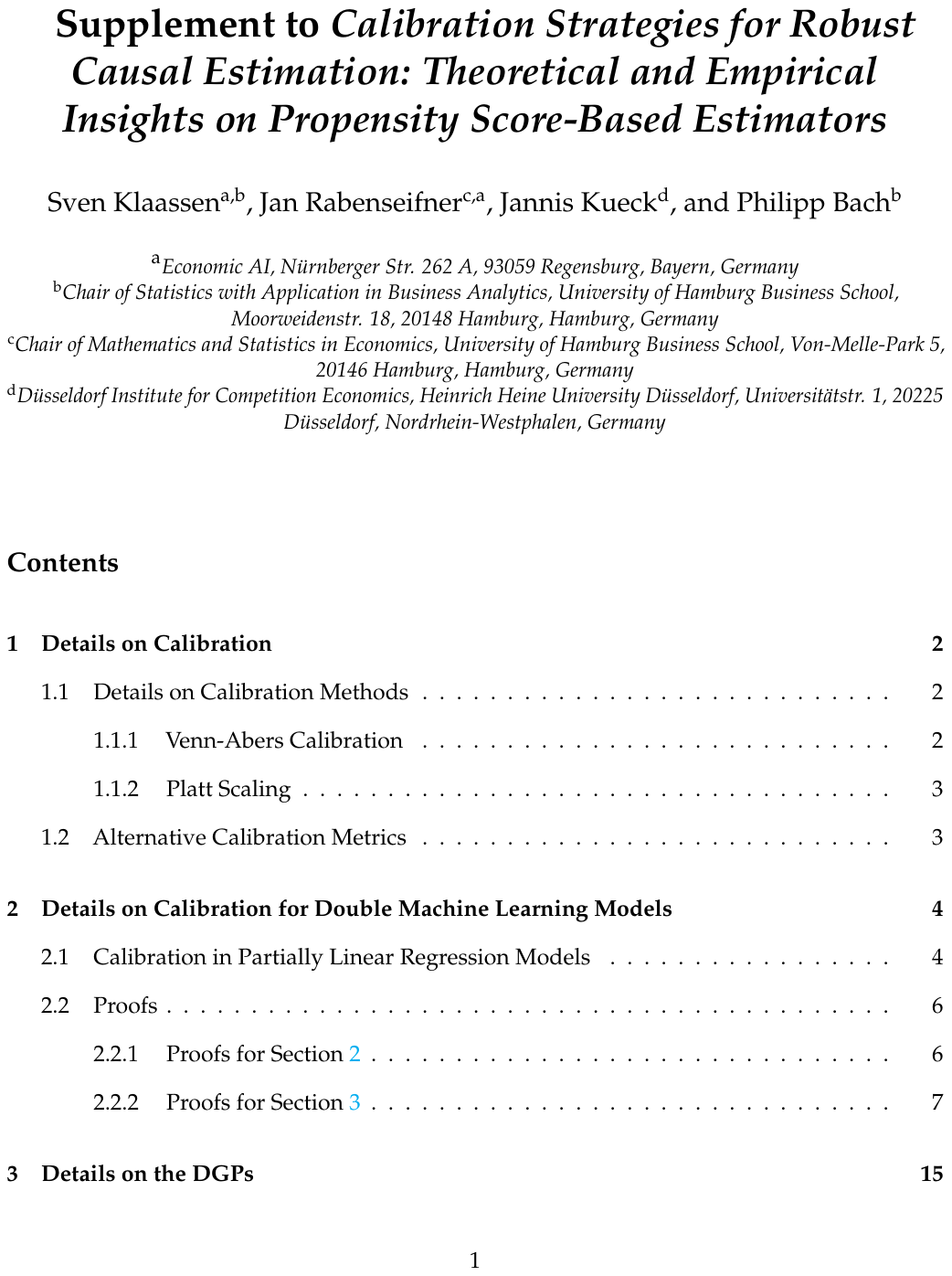}

\pdfximage{\supplementfilename}
\def\numbersupplementpages{\the\pdflastximagepages}

\newif\ifarXiv
\arXivtrue

\newcommand{\blind}{1}
\allowdisplaybreaks

\normalsize
\begin{document}







	
	\if1\blind
	{ \singlespacing
		\title{\bf \hspace{.2cm}
			 Calibration Strategies for Robust Causal Estimation: Theoretical and Empirical Insights on Propensity Score-Based Estimators 
			\\}

		\author[a,b]{Sven Klaassen}
		\author[c,a,$\dagger$]{Jan Rabenseifner}
		\author[d]{Jannis Kueck}
            \author[b]{Philipp Bach}
		\affil[a]{\it \small Economic AI, Nürnberger Str. 262 A, 93059 Regensburg, Bayern, Germany}
		\affil[b]{\it \small Chair of Statistics with Application in Business Analytics, 
                                University of Hamburg Business School, Moorweidenstr. 18, 20148 Hamburg, Hamburg, Germany}
		\affil[c]{\it \small Chair of Mathematics and Statistics in Economics, University of Hamburg Business School, Von-Melle-Park 5, 20146 Hamburg, Hamburg, Germany}
		\affil[d]{\it \small Düsseldorf Institute for Competition Economics, Heinrich Heine University Düsseldorf, Universitätstr. 1, 20225 Düsseldorf, Nordrhein-Westphalen, Germany
		}		\date{}
		
		\maketitle
	} \fi

    	\if0\blind
	{
		\bigskip
		\bigskip
		\bigskip
		\begin{center}
			{\LARGE\bf Calibration Approaches for Causal Estimation: Theoretical and Empirical Insights on Propensity Score-Based Estimators}
		\end{center}
	} \fi
	



\singlespacing{
	\begin{abstract}
            \footnotesize
            The partitioning of data for estimation and calibration critically impacts the performance of propensity score based estimators like inverse probability weighting (IPW) and double/debiased machine learning (DML) frameworks. We extend recent advances in calibration techniques for propensity score estimation, improving the robustness of propensity scores in challenging settings such as limited overlap, small sample sizes, or unbalanced data. Our contributions are twofold: First, we provide a theoretical analysis of the properties of calibrated estimators in the context of DML. To this end, we refine existing calibration frameworks for propensity score models, with a particular emphasis on the role of sample-splitting schemes in ensuring valid causal inference. Second, through extensive simulations, we show that calibration reduces variance of inverse-based propensity score estimators while also mitigating bias in IPW, even in small-sample regimes. Notably, calibration improves stability for flexible learners (e.g., gradient boosting) while preserving the doubly robust properties of DML. A key insight is that, even when methods perform well without calibration, incorporating a calibration step does not degrade performance, provided that an appropriate sample-splitting approach is chosen.

        \end{abstract}

	\bigskip
	\baselineskip 24pt
}

\begingroup
\renewcommand\thefootnote{}
\footnotetext{$\dagger$ Corresponding author email: jan.rabenseifner@uni-hamburg.de}
\addtocounter{footnote}{-0}
\endgroup


\doublespacing
\setlength{\abovedisplayskip}{3pt}
\setlength{\belowdisplayskip}{3pt}
\setlength{\abovedisplayshortskip}{3pt}
\setlength{\belowdisplayshortskip}{3pt}

\section{Introduction}\label{sec:intro}
\subsection{Motivation}
In many settings of the causal inference literature,
researchers are interested in the effect of a (binary) treatment $D\in\{0,1\}$ on an outcome $Y\in \mathbb{R}$. If the treatment is not assigned randomly, a common assumption is the so-called unconfoundedness assumption $ Y(0), Y(1) \perp D \mid X$ assuming that the potential outcomes $Y(1)$ and $Y(0)$ are independent of the actual treatment status $D$ conditional on control variables $X$. Let
\begin{align}\label{def::prop_score}
    m_0(x) := P(D=1|X=x) = \E[D|X=x].
\end{align}
be the propensity score. As famously shown in \citet{rosenbaum1983central} conditioning on the propensity score is sufficient to effectively account for the confounding through $X$
\begin{equation*}
    Y(0), Y(1) \perp D \mid  m_0(X).
\end{equation*}
Consequently, propensity scores are a cornerstone of modern causal inference for addressing confounding in observational studies. They enable balancing treatment and control groups, allowing for unbiased estimation of treatment effects under unconfoundedness. Commonly, propensity scores are used in methods such as inverse probability weighting (IPW), matching, stratification, Bayesian causal inference and more recently double machine learning (DML). Moreover, effective propensity adjustment requires sufficient overlap for propensity score-based estimators. When overlap between treatment and control groups is limited, or treatment assignment is unbalanced, propensity scores can become extreme (i.e. close to 0 or 1) leading to instability of the causal estimates. In such cases, estimators can suffer from inflated variance as extreme weights in IPW disproportionately amplify small errors in propensity score estimation. Similarly, matching algorithms may struggle to find suitable matches, resulting in biased estimates. These challenges highlight the importance of robust and well-calibrated propensity score models to maintain the reliability of causal estimates.

To mitigate instability, researchers often enforce common support by trimming or bounding propensity scores. For example, observations with propensity scores below a certain threshold or outside a predefined range are excluded from analysis. While this approach can reduce variance, it does so at the expense of bias, as it discards valuable information and reduces sample size. Such trade-offs are particularly problematic in small-sample settings, where the exclusion of even a few observations can significantly impact the precision and validity of treatment effect estimates.

To use the propensity score for balancing, the property
\begin{align}\label{eq::calibration_prop_score}
    m_0(X) = \E[D|m_0(X)]
\end{align}
is crucial, e.g. see Theorem 2 in \citet{rosenbaum1983central}. In the classification literature, Equation (\ref{eq::calibration_prop_score}) is known as the so-called calibration property.\footnote{To be precise, this notion of calibration is also referred to as conditional calibration, which is equivalent to probabilistic calibration for binary outcomes \citep{Gneiting2011probcalib}.} Intuitively, a (binary) classifier $\hat{m}(\cdot)$ is well calibrated if the percentage of positive labels ($D=1$) is approximately $m$ for all instances with $\hat{m}(X)\approx m$. As the true propensity score is the conditional expectation it is calibrated 
\begin{equation*}
m_0(X) = \E[D|X] = \E[\E[D|X]|\E[D|X]] = \E[D|m_0(X)]  .\quad P\text{-a.s.}
\end{equation*}
In most settings the true propensity score is not known such that it is typically estimated via some classification algorithm such as logistic regression, random forest, boosting methods or even deep neural networks. Consequently, the resulting estimator $\hat{m}(\cdot)$ of $m_0(\cdot)$ might not be calibrated, e.g. the percentage of treated units with $\hat{m}(X)\approx m$ might differ substantially from $m$ for certain values of $m\in (0,1)$. Accurately estimating treatment probabilities is crucial for valid causal inference. Inverse propensity score estimators aim to balance covariates between treatment and control groups. This balancing ensures more reliable and unbiased treatment effect estimates by aligning the covariate distributions, improving the comparability of the treatment groups in terms of observable characteristics.  Whereas methods like isotonic regression and Platt scaling refer to the calibration of predictions, critical questions remain about the optimal integration of these calibration techniques in causal estimation workflows.

\subsection{Related Literature}
Recent advances in calibration for propensity score estimation and causal inference emphasize three interconnected themes: the adaptation of machine learning calibration techniques to causal settings, stabilization strategies for inverse probability weighting (IPW), and theoretical insights into finite-sample performance. Initial studies by \citet{deshpande2023calibrated} showed significant improvements using single split calibration on data with deterministic treatment assignments and complex settings with hidden confounders. They specifically highlight variance reduction properties and analyze the regret of recalibration of propensity scores. Their proofs are independent of the sample size \(N\), as their calibration framework consists of splitting the data into training and calibration sets. This distinction is crucial for our work, as we focus on sample size dependent calibration algorithms.

\citet{Gutman2024platt} demonstrated that post-processing propensity scores using methods such as Platt scaling, referred to as post-calibration, improves treatment effect estimation by correcting propensity score distortions beyond covariate balancing. Calibration significantly benefits miscalibrated models like tree-based methods (e.g. gradient boosting), while logistic regression, contrary to the findings of \citet{deshpande2023calibrated}, shows minimal gains. This suggests that post-calibration allows flexible learners to retain their predictive accuracy while enhancing robustness to data challenges such as small sample sizes, model misspecification, class imbalance, or limited overlap, thus challenging the conventional trade-off between model complexity and calibration. \citet{ballinari2025improving} use nested cross-fitting, reserving distinct data folds for propensity estimation and calibration, which is a straightforward application of the double machine learning theory of \citet{Chernozhukov2018dml}. They reported instability in small samples, a limitation attributed to reduced effective sample sizes for both steps. These studies collectively identify a tension between calibration flexibility and stability, particularly in finite-sample regimes. Efforts to address these challenges have taken different paths. For instance,  \citet{vanderlaan2024stabilized} proposed a stratified calibration for treated and control groups to stabilize IPW weights. In another work, they extended calibration frameworks to estimate heterogeneous treatment effects \citep{van2023causal}. Meanwhile, \citet{vanderlaan2024automaticdoublyrobustinference} extend automatic debiased machine learning (autoDML) \citep{chernozhukov2022autoDML, chernozhukov2024automaticdebiasedmachinelearning} by calibrating both outcome regression and Riesz representer. They demonstrate that the calibration step yields doubly robust asymptotically linear estimators, reducing nuisance rate requirements.

Theoretical work by \citet{gamarnik1998efficient}, \citet{mammen2007additive} and \citet{wuthrich2023isotonic} established consistency guarantees for isotonic regression, but practical implementations often struggle with convergence rates in small samples, as shown by \citet{yang2019contraction}. Furthermore, \citet{yang2019contraction} established that isotonic projection is non-contractive under the $\ell_{\infty}$ norm, exacerbating edge instability in propensity score estimates. This theoretical insight justifies the empirical requirement of clipping extreme probabilities, which explains the unstable treatment effect estimates in \citet{ballinari2025improving} using isotonic regression under limited overlap.

This paper systematically evaluates how calibration performance depends on data partitioning for propensity estimation and calibration. While existing studies fix specific splitting strategies (e.g., single-split or nested cross-fitting), we show that the choice of partitioning interacts critically with sample size, clipping thresholds, and complexity of the data generating process. For instance, \citet{vanderlaan2024stabilized} suggest calibration on the full-sample, which avoids reserving data exclusively for calibration. This approach can mitigate instability without sacrificing theoretical guarantees, a hypothesis we test across multiple data-generating processes (DGP). Similarly, we reconcile the debates about stratified versus pooled calibration by demonstrating that efficient reuse of cross-fitted propensity scores obviates the need for group-specific adjustments in many settings. Further, beyond the work of \citet{ballinari2025improving}, we provide a theoretical extension of the double machine learning theory to allow for different sample-splitting schemes. By synthesizing these insights, our work clarifies when and how calibration improves ATE estimation, providing a bridge between theoretical calibration properties and practical implementation challenges. 

\textbf{Plan of the Paper.}
The rest of the paper is organized as follows. Section \ref{sec:prop_score} introduces propensity score calibration, highlighting different approaches and providing details on the properties of isotonic regression. Section \ref{sec:dml} proposes calibration algorithms for estimating the average treatment effect and establishes theoretical guarantees under double machine learning (DML), including convergence rates and asymptotic normality. Section \ref{sec:sim} demonstrates robustness through simulations across diverse and challenging data-generating processes. Details on the calibration of partially linear regression models, along with proofs of the theorems, are provided in Section 1 of the Supplementary Material. Implementation details for reproducibility, an analysis of how calibration affects the normalization of propensity score weights and covariate balance, as well as extended simulation results and sensitivity analyses, can be found in Sections 2 and 3 of the Supplement.

\section{Propensity Score Calibration}\label{sec:prop_score}
Let $D\in\{0,1\}$ be a binary treatment variable with covariates $X\in\mathcal{X}\subseteq\R^d$. The propensity score is defined as $m_0(x) := P(D = 1 \mid X = x) = \E[D \mid X = x].$ Since the propensity score represents a conditional expectation, it is calibrated such that $m_0(X) = \E[D \mid m_0(X)].$ The goal is to achieve a similar balancing property of an estimated version of $m_0(X)$. Given an estimate $\hat{m}(X)$ of $m_0(X)$, we consider popular calibration methods such as isotonic regression. Generally, we consider calibration procedures based on the pseudo-sample 
$((D_1, \hat{m}(X_1)), \dots,$ $(D_N, \hat{m}(X_N)))$. The calibration algorithm approximates $\E[D|\hat{m}(X)]$ which typically differs from $\E[D|m_0(X)]$. In the following, we denote the calibrated propensity score by $\tilde{m}: \mathcal{X} \to [0,1], \quad x\mapsto \tilde{m}(x).$
\subsection{Rate Comparison $\hat{m}(\cdot)$ and $\tilde{m}(\cdot)$}
For any estimate $\hat{m}(\cdot)$ of $m_0(\cdot)$ the mean-squared-error decomposes as
\begin{align}\label{eq:rmse_decomposition}
    \|\hat{m}(X) - m_0(X)\|_{P,2} &= \left(\mathbb{E}\left[\textrm{Var}\big(m_0(X)|\hat{m}(X)\big)\right] + \|\mathbb{E}[m_0(X)|\hat{m}(X)] - \hat{m}(X)\|_{P,2}^2 \right)^{1/2}.
\end{align}
The first term denotes the expected precision of $\hat{m}(\cdot)$, while the second term is the calibration error. Calibration procedures minimizing mean square error in the pseudo-sample approximate $\E[D|\hat{m}(X)]$, reducing this error. 
\begin{assumption}\label{assumption::calibration_rate}
Let $\tilde{m}(X)$ be an estimator of $\tilde{m}_0(X):=\E[D|\hat{m}(X)]$, with $\|\tilde{m}(X) - \tilde{m}_0(X)\|_{P,2}\le \tilde{\varepsilon}_N.$
\end{assumption}
\begin{lemma}\label{lemma::rmse_decomposition}
Under Assumption \ref{assumption::calibration_rate}:
\begin{align} \label{eq:calibration_rate}
    \|\tilde{m}(X) - m_0(X)\|_{P,2} &\le \left(\mathbb{E}\left[\textrm{Var}\big(m_0(X)|\hat{m}(X)\big)\right]\right)^{1/2} + \tilde{\varepsilon}_N.
\end{align}
\end{lemma}
Comparing \eqref{eq:rmse_decomposition} and \eqref{eq:calibration_rate} shows rate differences when the calibration error $\|\mathbb{E}[m_0(X)|\hat{m}(X)] - \hat{m}(X)\|_{P,2}^2$ dominates. Improvements occur if $\tilde{\varepsilon}_N = o(\|\mathbb{E}[m_0(X)|\hat{m}(X)] - \hat{m}(X)\|_{P,2})$. This is relevant in double ML settings requiring $\|\hat{m}(X) - m_0(X)\|_{P,2} = o(N^{-1/4})$: If $\tilde{\varepsilon}_N = o(N^{-1/4})$, $\tilde{m}(\cdot)$ satisfies the requirement with potential rate improvements from faster convergence.
\subsection{Calibration Methods}\label{sec:calib_methods}
\label{ch:isotonic_theory}
Model calibration has roots in meteorological forecasting to address reliability in probabilistic weather predictions \citep{toth2006weathercalib, dawid2014probforecast, gneiting2014weathercalib}.
Calibration is often achieved via post-hoc point calibrators that transform \(\hat{m}(X)\) into $f(\hat{m}(X))$, balancing two aims: (1) calibration validity, and (2) preservation of predictive sharpness (i.e., $f\circ\hat{m}(X)$ approximates \(\hat{m}\)’s discriminative power) \citep{gneiting2007calibrationsharpness, gupta2020binarycalibration}. 
Common approaches include parametric methods like Platt scaling, which fits a logistic sigmoid to \(f(X)\) \citep{platt1999plattscale, cox1958sigmoid}; non-parametric methods such as histogram binning, partitioning predictions into fixed intervals \citep{zadrozny2001calibiontrees, gupta2021binaryhistogram}, and isotonic regression, learning a monotonic transform via empirical risk minimization \citep{zadrozny2002multiclass, barlow1972isotonic}; as well as conformal methods like Venn-Abers predictors, refining calibration through cross-conformal inference \citep{vovk2014vennaberspredictors}.
Isotonic calibration, while distribution-free and tuning parameter-free, achieves asymptotic guarantees with \(O_P(N^{-1/3})\) convergence \citep{zhang2002isotonicbounds, van2023causal}. In contrast, histogram binning requires explicit bin specification and trades flexibility for finite-sample validity \citep{gupta2021binaryhistogram}. We focus on three common calibration methods, noting that calibration is an active area of research with potential for improvements and new proposals that require further simulation testing\footnote{The description of Platt scaling and Venn-Abers calibration can be found in Supplement \ref{S-appendix:calib_method}. As our theoretical results are built around isotonic regression, we introduce it here in more detail.}.

\subsubsection*{Isotonic Regression}
\begin{definition}
Given an estimate $\hat{m}(\cdot)$ of the propensity score, perform an isotonic regression as$f = \arg\min_{f\in\mathcal{F}_{\text{iso}}}\sum_{i=1}^N (D_i - (f\circ\hat{m})(X_i))^2$ with $\mathcal{F}_{\text{iso}}$ being the set of non-decreasing functions. The calibrated propensity score is then given by $\tilde{m} = f\circ\hat{m}.$
\end{definition}
Especially, the in-sample calibration property $\E_n[D|\tilde{m}(X_i)] = \tilde{m}(X_i)$
for $i\in\{1,\dots,N\},$ seems to be desirable (e.g. \citet{wuthrich2023isotonic}). We consider an estimated propensity score $\hat{m}(\cdot)$ based on a separate sample such that $\hat{m}(\cdot)$ can be considered a fixed function. Let $U:=\hat{m}(X)$ and define the pseudo sample as $Z:=(D,U)$, where $(Z_i)_{i=1}^N$ are iid. copies of $Z$. Consider the following assumption
\begin{assumption} \label{assumption::monotonicity}
    The regression function $\tilde{m}_0(u):= \E[D|U = u]$ is monotone.
\end{assumption}
\begin{remark}
Assumption \ref{assumption::monotonicity} is substantially weaker than requiring monotonicity in the original covariates $X$, intuitively only requiring the preliminary propensity score $\hat{m}(\cdot)$ 
to be monotone "on average". When $\hat{m}(\cdot)$ is not injective, distinct values of $X$ may map to the same $U$, allowing $\tilde{m}_0(U)$ to average over these $X$ values. This aggregation can smooth out non-monotonic behavior in $X$, making monotonicity in $U$ easier to satisfy. However, if $\hat{m}(\cdot)$ is bijective, each $U$ uniquely determines $X$, and monotonicity of $\tilde{m}_0(U)$ becomes equivalent to monotonicity of $\mathbb{E}[D|X = x]$ using the reparameterization $u = \hat{m}(x)$.
\end{remark}
Define $\tilde{m}$ as the estimator obtained by isotonic regression on the sample $(Z_i)_{i=1}^N$.
\begin{lemma}[Convergence of Isotonic Regression]
\label{lemma::convergence_isotonic_regression}
Under Assumption \ref{assumption::monotonicity}, the isotonic regression estimator $\tilde{m}$ satisfies: $\|\tilde{m}(U) - \tilde{m}_0(U)\|_{P,2} = O_P(N^{-1/3})$.
\end{lemma}

The rate follows from the bracketing entropy bound in Theorem 2.7.5 of \citet{vaart2023empirical} for monotonic functions. Relying on Lemma 3.4.3 of \citet{vaart2023empirical} gives the desired rate as mentioned in the corresponding Section 3.4.3.2. Earlier work by \citet{BirmanSolomjak1967} and \cite{geer2000empirical} established foundational approximation and entropy arguments, extended to additive isotonic regression in \citet{mammen2007additive}. For detailed convergence properties see e.g. \citet{zhang2002isotonicbounds}. 

Notably, similar convergence rates hold in settings where the data is not strictly i.i.d., such as when using sample splitting or cross-fitting to estimate \(\hat{m}(\cdot)\). Theorem 2 in \cite{vanderlaan2024stabilized} demonstrates that under cross-fitting regimes, where \(\hat{m}(\cdot)\) is trained on an independent sample and applied to the estimation sample, the pseudo-sample $Z$ is still sufficiently weakly dependent for the \(O_P(N^{-1/3})\) rate to hold. This aligns with the mean squared error (MSE) decomposition in \eqref{eq:rmse_decomposition}, where the \(O_P(N^{-1/3})\) \(L_2(P)\) convergence rate of isotonic regression (Lemma \ref{lemma::convergence_isotonic_regression}) ensures that the calibration error term \(\|\mathbb{E}[m_0(X)|\hat{m}(X)] - \hat{m}(X)\|_{P,2}^2\) decays as \(O_P(N^{-2/3})\). Theorem 1 in \cite{vanderlaan2024stabilized} further establishes distribution-free calibration guarantees, bounding the calibration error by \(O_P(N^{-2/3})\), irrespective of the smoothness of the inverse propensity score or the dimension of the covariates. Together, these results accommodate the calibration algorithms introduced in Section \ref{sec:dml}.
\section{Calibration for Double Machine Learning}\label{sec:dml}
The following section combines calibration and double machine learning. The first part focuses on high-level conditions 
for different double machine learning algorithms, whereas the second part states explicit conditions for particular double machine learning models with isotonic regression.
\subsection{Double Machine Learning Theory and Algorithms}

In this section, we state conditions which enable a re-estimation step for nuisance estimators in the double machine learning framework if the complexity of the re-estimation procedure is not too large.
As in \citet{Chernozhukov2018dml} we denote $\theta_0\in \Theta\subset \R$ the parameter of interest. The leading example is the average treatment effect (ATE) $\theta_0 = \E [ Y(1) - Y(0) ]$.
Further, we assume that $\theta_0$ satisfies the moment condition,
\begin{equation}\label{assumption::moment_condition}
    \E_P[\psi(W;\theta_0, \eta_0)]=0,
\end{equation}
where $\psi$ is a known score function, the  data $W$ is a random element in $(\mathcal{W}, \mathcal{A}_\mathcal{W})$ with probability measure $P\in\mathcal{P}_N$ and $\eta_0$ is the true value of the nuisance parameter $\eta\in T$, where $T$ is the convex subset of a normed vector space with norm $\|\cdot\|_T$.\\
The previous setting describes the standard double machine learning framework introduced in \cite{Chernozhukov2018dml}. For simplicity, we restrict ourselves to the case of linear score functions, that is
\begin{align}\label{assumption::linear_score}
\psi(w;\theta, \eta) = \psi^a(w;\eta)\theta + \psi^b(w;, \eta),
\end{align}
for all $w\in \mathcal{W}$, $\theta\in\Theta$ and $\eta\in T$.
Further, since we would like to consider a scenario with a re-estimation or calibration step, which might not affect all nuisance parameters, we define $\eta_0 = (\eta_0^{(1)}, \eta_0^{(2)})$,
where $\eta_0^{(2)}$ should be re-estimated as for example when $\eta_0^{(2)}$ is a propensity score to be calibrated. Correspondingly define $T = T^{(1)}\times T^{(2)}$.

Algorithm \ref{alg:dml2:standard} recaps the standard version of the double machine learning algorithm based on cross-fitting (cf. Definition 3.2 in \citet{Chernozhukov2018dml}).\\
Let $(W_i)_{i=1}^N$ be iid. copies of $W$ with probability measure $P$. To simplify notation, assume that $N$ is divisible by $K$.

\begin{algorithm}[!ht]
\caption{(uncalibrated) DML 2 Algorithm}
\setstretch{0.9}
\label{alg:dml2:standard}
{\footnotesize
\begin{algorithmic}[1]
\State \hspace*{0.02in} \textbf{Input:} Data $(W_i)_{i=1}^N$. A $K$-fold random partition $(I_k)_{k=1}^K$ of $[N]=\{1,\dots,N\}$ such that each fold $I_k$ is of size $n=N/K$. For each $k\in [K]=\{1,\dots,K\}$, define $I_k^c:=\{1,\dots,N\}\setminus I_k$.
\State For each $k\in[K]$, fit a machine learning estimator
$$\hat{\eta}_{0,k} = \hat{\eta}_{0}\big((W_i)_{i\in I_k^c}\big)$$
of $\eta_0$, where $\hat{\eta}_{0,k}$ is a random element in $T$, where the randomness only depends on the $(W_i)_{i\in I_k^c}$.
\State Construct the estimator $\hat{\theta}_0$ as the solution to
$$\frac{1}{K}\sum_{k=1}^K \E_{n,k}[\psi(W;\hat{\theta}_0, \hat{\eta}_{0,k})]=0,$$
where $\E_{n,k}[\psi(W)]=n^{-1}\sum_{i\in I_k} \psi(W_i)$ is the empirical expectation over $(W_i)_{i\in I_k}$.
\end{algorithmic}
}
\end{algorithm}

The standard DML 2 algorithm employs cross-fitting to handle the complexity of the estimated nuisance elements $\hat{\eta}_{0,k}$. 

\begin{remark}\label{theorem::dml_asympt_dist}
Theorem 3.1 in \citet{Chernozhukov2018dml} shows that the estimator according to Algorithm \ref{alg:dml2:standard} is asymptotically normally distributed. More specifically, it holds 
    \begin{align}\label{eq:dml_normal_dist}
        \sqrt{N}\sigma^{-1} (\hat{\theta}_0 - \theta_0) = \frac{1}{N}\sum_{i=1}^N \bar{\psi}(W_i) + O_P(\rho_N) 	\rightsquigarrow \N(0,1)
    \end{align}
uniformly over $P\in\mathcal{P}_N$, where the size of the remainder term obeys $\rho_N:=N^{-1/2} + r_N + r'_N + N^{1/2}\lambda_N + N^{1/2}\lambda'_N \lesssim \delta_N$, with $\delta_N \ge N^{-1/2}$. Here, $\bar{\psi}(\cdot):=\sigma^{-1}J_0^{-1} \psi(\cdot; \theta_0, \eta_0)$ is the influence function and the approximate variance is $\sigma^2:= J_0^{-2}\E_P[\psi(W; \theta_0, \eta_0)^2]$.
\end{remark}

In Remark \ref{theorem::dml_asympt_dist} it is assumed that $\hat{\eta}_{0,k}= (\hat{\eta}_{0,k}^{(1)}, \hat{\eta}_{0,k}^{(2)}) \in \T_N$ with probability $1-o(1)$, where $\T_N$ is a suitable nuisance realization set.
To enable the re-estimation of nuisance elements $\hat{\eta}_{0,k}^{(2)}$ the algorithm and the nuisance realization set $\T_N$ has to be slightly adapted. A simple and straightforward adaption is presented in Algorithm \ref{alg:dml2:naive}.

\begin{algorithm}[!htb]
\caption{(nested $K$-fold cross-fitting calibration) DML 2 Algorithm}
\label{alg:dml2:naive}
{\footnotesize
\begin{algorithmic}[1]
\State \textbf{Input:} Data $(W_i)_{i=1}^N$. A $K$-fold random partition $(I_k)_{k=1}^K$ of $[N]=\{1,\dots,N\}$ such that each fold $I_k$ is of size $n=N/K$. For each $k\in [K]=\{1,\dots,K\}$ define $I_k^c:=\{1,\dots,N\}\setminus I_k$.
\State For each $k\in[K]$, fit a machine learning estimator
$$\hat{\eta}^{(1)}_{0,k} = \hat{\eta}^{(1)}_{0}\big((W_i)_{i\in I_k^c}\big)$$
of $\eta_0^{(1)}$, where $\hat{\eta}^{(1)}_{0,k}$ is a random element in $T^{(1)}$, where the randomness only depends on the $(W_i)_{i\in I_k^c}$.
\State For each $k\in[K]$, split the training partition $I_k^c$ into two disjoint samples $I_{k,1}^c$ and $I_{k,2}^c$.
\State Use the first subset $I_{k,1}^c$ to fit a machine learning estimator
$$\hat{\eta}^{(2)}_{0,k} = \hat{\eta}^{(2)}_{0}\big((W_i)_{i\in I_{k,1}^c}\big)$$
of $\eta^{(2)}_0$, where $\hat{\eta}^{(2)}_{0,k}$ is a random element in $T^{(2)}$, where the randomness only depends on the $(W_i)_{i\in I_{k,1}^c}$.
\State Use the second subset $I_{k,2}^c$ and the estimated nuisance element $\hat{\eta}_{0,k}^{(2)}$ to fit a re-estimation procedure
$$\tilde{\eta}^{(2)}_{0,k} =\tilde{\eta}^{(2)}_{0}\big((W_i)_{i\in I_{k,2}^c}, \hat{\eta}^{(2)}_{0,k}\big)$$
of $\eta_0^{(2)}$, where $\tilde{\eta}^{(2)}_{0,k}$ is a random element in $T^{(2)}$, where the randomness only depends on the $(W_i)_{i\in I_{k}^c}$.
\State Construct the estimator $\tilde{\theta}_0$ as the solution to
$$\frac{1}{K}\sum_{k=1}^K \E_{n,k}\big[\psi\big(W;\tilde{\theta}_0, (\hat{\eta}^{(1)}_{0,k}, \tilde{\eta}^{(2)}_{0,k})\big)\big]=0,$$
where $\E_{n,k}[\psi(W)]=n^{-1}\sum_{i\in I_k} \psi(W_i)$ is the empirical expectation over $(W_i)_{i\in I_k}$.
\end{algorithmic}
}
\end{algorithm}

As already mentioned, the standard double machine learning procedure in Algorithm \ref{alg:dml2:standard} uses cross-fitting to handle the complexity of estimated nuisance elements $\hat{\eta}_{0,k}$. Algorithm \ref{alg:dml2:naive} is a straightforward extension, which leaves the cross-fitting unchanged. The approach just employs a nested sample splitting procedure, such that the calibrated nuisance elements $\tilde{\eta}^{(2)}_{0,k}$ still depend only on the observations of the training sample $(W_i)_{i\in I_{k}^c}$. As a consequence, only the predictive performance of the calibrated nuisance estimators must be ensured.

\begin{remark}\label{theorem::dml_asympt_dist_naive}
Let $\tilde{\theta}_0$ be the estimator according to Algorithm \ref{alg:dml2:naive}.
Under the assumptions in Theorem 3.1 in \citet{Chernozhukov2018dml}, Equation \eqref{eq:dml_normal_dist} in Remark \ref{theorem::dml_asympt_dist} holds analogously for $\tilde{\theta}_0$.
\end{remark}

Typically, Assumption 3.2 in Theorem 3.1 in \citet{Chernozhukov2018dml} requires high-quality nuisance estimators. In particular, the re-estimation procedure (calibration) has to converge sufficiently fast, i.e. $
    \|\tilde{\eta}^{(2)}_{0,k} -\eta^{(2)}_{0}\|_{P,2} \lesssim \varepsilon_N=o(N^{-1/4}).$
Since calibration properties are most important for small samples, further splitting of the training sample for calibration might not be desirable. 

Therefore, we state Assumptions 3.1 and 3.2 of \citet{Chernozhukov2018dml} for an adapted nuisance realization set $\widetilde{\T}_N$. It is worth noting that in the following assumption the calibrated nuisance elements $\tilde{\eta}_{0}^{(2)}((W_i)_{i\in [N]})) )$ may depend on the full data. 
This allows us to introduce new estimation algorithms that rely on more sophisticated splitting rules for calibration.

\begin{assumption}\label{assumption::dml}

Let \( c_0 > 0 \), \( c_1 > 0 \), and \( q \geq 2 \) be some finite constants such that \( c_0 \leq c_1\), and let \( \{\delta_N\}_{N \geq 1} \) and \( \{\Delta_N\}_{N \geq 1} \) be some sequences of positive constants converging to zero such that \( \delta_N \geq N^{-1/2} \). Also, let \( K \geq 2 \) be some fixed integer, and let \( \{\mathcal{P}_N\}_{N \geq 1} \) be some sequence of sets of probability distributions \( P \) of \(\mathcal{W}\) on \( W \).

\textbf{Assumption 3.1 (Linear scores with approximate Neyman orthogonality)} For all \( N \geq 3 \) and \( P \in \mathcal{P}_N \), the following conditions hold: \romaninline{1} The true parameter value \( \theta_0 \) obeys (\ref{assumption::moment_condition}). \romaninline{2} The score \( \psi \) is linear in the sense of (\ref{assumption::linear_score}). \romaninline{3} The map \( \eta \mapsto E_P[\psi(W;\theta,\eta)] \) is twice continuously Gateaux-differentiable on \( T \). \romaninline{4} The score \( \psi \) obeys the Neyman orthogonality or, more generally, the Neyman \( \lambda_N \) near-orthogonality condition at \( (\theta_0, \eta_0) \) with respect to the nuisance realization set \( \widetilde{\T}_N \subset T \) for $
    \lambda_N := \sup_{\eta \in \mathcal{T}_N} \left| \partial_\eta \E_P[\psi(W;\theta_0,\eta_0)][\eta - \eta_0]\right| \leq \delta_N N^{-1/2}.$
\romaninline{5} The identification condition holds; namely, the singular values of the matrix $J_0 := \E_P[\psi^a(W; \eta_0)]$ are between \( c_0 \) and \( c_1 \).
\bigskip

\textbf{Assumption 3.2 (Score regularity and quality of nuisance parameter estimators)} For all \( N \geq 3 \) and \( P \in \mathcal{P}_N \), the following conditions hold: 
\begin{itemize}
    \item[(a)] Given a random subset \( I \) of \( [N] \) of size \( n = N/K \), the nuisance parameter estimator \( \tilde{\eta}_0 := (\hat{\eta}_{0}^{(1)}((W_i)_{i\in I^c}), \tilde{\eta}_{0}^{(2)}((W_i)_{i\in [N]})) \) belongs to the realization set \( \widetilde{\T}_N \) with probability at least \( 1 - \Delta_N \), where \( \widetilde{\T}_N = \widetilde{\T}^{(1)}_N \times \widetilde{\T}^{(2)}_N \) contains \( \eta_0 \) and is constrained by the following conditions.
    \item[(b)] The moment conditions hold:
    \[
    m_N := \sup_{\eta \in \widetilde{\T}_N} \E_P[|\psi(W; \theta_0, \eta)|^{q}]^{1/q} \leq c_1;
    \quad
    m_N' := \sup_{\eta \in \widetilde{\T}_N} \E_P[|\psi^a(W; \eta)|^{q}]^{1/q} \leq c_1.
    \]
    \item[(c)] The following conditions on the statistical rates \( r_N \), \( r_N' \), and \( \lambda_N' \) hold:
    \[
    r_N := \sup_{\eta \in \widetilde{\T}_N} \left|\E_P[\psi^a(W; \eta)] - \E_P[\psi^a(W; \eta_0)]\right| \leq \delta_N,
    \]
    \[
    r_N' := \sup_{\eta^{(1)} \in \widetilde{\T}^{(1)}_N} \left(\E_P\left[\left(\psi\big(W; \theta_0, (\eta^{(1)}, \eta_0^{(2)})\big) - \psi\big(W; \theta_0, (\eta_0^{(1)}, \eta_0^{(2)})\big)\right)^2\right]\right)^{1/2} \leq \delta_N ,
    \]
    \[
    \lambda_N' := \sup_{r\in(0,1), \eta \in \widetilde{\T}_N} \left|\partial^2_r \E_P[\psi(W; \theta_0, \eta_0 + r (\eta - \eta_0))]\right| \leq \delta_N / \sqrt{N}.
    \]
    \item[(d)] The variance of the score \( \psi \) is non-degenerate: $c_0\le \E_P[\psi(W; \theta_0, \eta_0)^2].$
\end{itemize}

\end{assumption}

Indeed, if the calibration method is not too complex (see Assumption \ref{assumption::dml_calibration_complexity}), for example when isotonic regression is used for calibration, the additional sample split in algorithm \ref{alg:dml2:naive} can be avoided by calibrating the predictions on each ``test''-fold $I_k$ which are used to estimate the target parameter $\theta_0$. This procedure is described in Algorithm \ref{alg:dml2:calibration}. 

\begin{algorithm}[!htb]
\caption{($k$-fold cross-fitting calibration) DML 2 Algorithm}
\label{alg:dml2:calibration}
{\footnotesize
\begin{algorithmic}[1]
\State \textbf{Input:} Data $(W_i)_{i=1}^N$. A $K$-fold random partition $(I_k)_{k=1}^K$ of $[N]=\{1,\dots,N\}$ such that each fold $I_k$ is of size $n=N/K$. For each $k\in [K]=\{1,\dots,K\}$ define $I_k^c:=\{1,\dots,N\}\setminus I_k$.
\State For each $k\in[K]$, fit a machine learning estimator
$$\hat{\eta}_{0,k} = \hat{\eta}_{0}\big((W_i)_{i\in I_k^c}\big)$$
of $\eta_0$, where $\hat{\eta}_{0,k}$ is a random element in $T$, where the randomness only depends on the $(W_i)_{i\in I_k^c}$.
\State For each $k\in[K]$, rely on estimated nuisance element $\hat{\eta}_{0,k}^{(2)}$ to fit a re-estimation procedure
$$\tilde{\eta}^{(2)}_{0,k} =\tilde{\eta}^{(2)}_{0}\big((W_i)_{i\in I_{k}}, \hat{\eta}^{(2)}_{0,k}\big)$$
of $\eta_0^{(2)}$, where $\tilde{\eta}^{(2)}_{0,k}$ is a random element in $T^{(2)}$.
\State Construct the estimator $\tilde{\theta}_0$ as the solution to
$$\frac{1}{K}\sum_{k=1}^K \E_{n,k}\big[\psi\big(W;\tilde{\theta}_0, (\hat{\eta}^{(1)}_{0,k}, \tilde{\eta}^{(2)}_{0,k})\big)\big]=0,$$
where $\E_{n,k}[\psi(W)]=n^{-1}\sum_{i\in I_k} \psi(W_i)$ is the empirical expectation over $(W_i)_{i\in I_k}$.
\end{algorithmic}
}
\end{algorithm}

Although the calibrated nuisance estimator depends on the full data in this case, we will show an analog result as in Theorem 3.1 in \citet{Chernozhukov2018dml} under the Assumptions \ref{assumption::dml} and \ref{assumption::dml_calibration_complexity} in Theorem \ref{theorem::dml_asympt_dist_calibration}. As a slight modification of Algorithm \ref{alg:dml2:calibration} one can use different $K$-fold cross-fitting procedures for the estimated nuisance elements. For example, $2$-fold cross-fitting as described in Algorithm \ref{alg:dml2:calibration_split}, uses half of the data for nuisance estimation and the other half for calibration. Consequently, the calibration step might be more stable.

\begin{algorithm}[!htb]
\caption{(single split cross-fitting calibration) DML 2 Algorithm}
\label{alg:dml2:calibration_split}
{\footnotesize
\begin{algorithmic}[1]
\State \textbf{Input:} Data $(W_i)_{i=1}^N$. A $K$-fold random partition $(I_k)_{k=1}^K$ of $[N]=\{1,\dots,N\}$ such that each fold $I_k$ is of size $n=N/K$. A $2$-fold random partition $\tilde{I}_1\cup \tilde{I}_2 = \{1,\dots,N\}$ and $\tilde{I}_1\cap \tilde{I}_2 = \emptyset$. For each $k\in [K]=\{1,\dots,K\}$ define $I_k^c:=\{1,\dots,N\}\setminus I_k$ and for $j\in [2]=\{1,2\}$ define $\tilde{I}_j^c:=\{1,\dots,N\}\setminus \tilde{I}_j$.
\State For each $k\in[K]$, fit a machine learning estimator
$$\hat{\eta}^{(1)}_{0,k} = \hat{\eta}^{(1)}_{0}\big((W_i)_{i\in I_k^c}\big)$$
of $\eta_0^{(1)}$, where $\hat{\eta}^{(1)}_{0,k}$ is a random element in $T^{(1)}$, where the randomness only depends on the $(W_i)_{i\in I_k^c}$.
\State For each $j\in[2]$, fit a machine learning estimator
$$\hat{\eta}^{(2)}_{0,j} = \hat{\eta}^{(2)}_{0}\big((W_i)_{i\in \tilde{I}_j^c}\big)$$
of $\eta_0^{(2)}$, where $\hat{\eta}^{(2)}_{0,j}$ is a random element in $T^{(2)}$, where the randomness only depends on the $(W_i)_{i\in \tilde{I}_j^c}$. 
\State For each $j\in[2]$, rely on estimated nuisance element $\hat{\eta}_{0,j}^{(2)}$ to fit a re-estimation procedure
$$\tilde{\eta}^{(2)}_{0,j} =\tilde{\eta}^{(2)}_{0}\big((W_i)_{i\in \tilde{I}_{j}}, \hat{\eta}^{(2)}_{0,j}\big)$$
of $\eta_0^{(2)}$, where $\tilde{\eta}^{(2)}_{0,j}$ is a random element in $T^{(2)}$.
\State Construct the estimator $\tilde{\theta}_0$ as the solution to
$$\frac{1}{2K}\sum_{k=1}^K\sum_{j=1}^2 \E_{n,k,j}\big[\psi\big(W;\tilde{\theta}_0, (\hat{\eta}^{(1)}_{0,k}, \tilde{\eta}^{(2)}_{0,j})\big)\big]=0,$$
where $\E_{n,k,j}[\psi(W)]=n^{-1}\sum_{i\in I_k\cap \tilde{I}_j} \psi(W_i)$ is the empirical expectation over $(W_i)_{i\in I_k \cap \tilde{I}_j}$.
\end{algorithmic}
}
\end{algorithm}

Another option is to simultaneously calibrate all cross-fitted predictions $\hat{\eta}_0^{(2)}$ as described in Algorithm \ref{alg:dml2:calibration_all}. The main difference between Algorithm \ref{alg:dml2:calibration} and \ref{alg:dml2:calibration_all} is the dependency structure of the data used to calibrate the nuisance elements. In Algorithm \ref{alg:dml2:calibration} the recalibration is fitted on i.i.d. samples conditional on the corresponding ``training''-fold $I_k^c$, whereas in Algorithm \ref{alg:dml2:calibration_all} samples used for the calibration step have a complex dependency structure.
\begin{algorithm}[!htb]
\caption{(full-sample calibration) DML 2 Algorithm}
\label{alg:dml2:calibration_all}
{\footnotesize
\begin{algorithmic}[1]
\State \textbf{Input:} Data $(W_i)_{i=1}^N$. A $K$-fold random partition $(I_k)_{k=1}^K$ of $[N]=\{1,\dots,N\}$ such that each fold $I_k$ is of size $n=N/K$. For each $k\in [K]=\{1,\dots,K\}$ define $I_k^c:=\{1,\dots,N\}\setminus I_k$.
\State For each $k\in[K]$, fit a machine learning estimator
$$\hat{\eta}_{0,k} = \hat{\eta}_{0}\big((W_i)_{i\in I_k^c}\big)$$
of $\eta_0$, where $\hat{\eta}_{0,k}$ is a random element in $T$, where the randomness only depends on the $(W_i)_{i\in I_k^c}$.
\State Combine all estimated nuisance elements $\hat{\eta}_{0,k}^{(2)}$ to fit a re-estimation procedure
$$\tilde{\eta}^{(2)}_{0} =\tilde{\eta}^{(2)}_{0}\big((W_i)_{i=1}^N, (\hat{\eta}^{(2)}_{0,k})_{k\in [K]}\big)$$
of $\eta_0^{(2)}$, where $\tilde{\eta}^{(2)}_{0}$ is a random element in $T^{(2)}$.
\State Construct the estimator $\tilde{\theta}_0$ as the solution to
$$\frac{1}{K}\sum_{k=1}^K \E_{n,k}\big[\psi\big(W;\tilde{\theta}_0, (\hat{\eta}^{(1)}_{0,k}, \tilde{\eta}^{(2)}_{0})\big)\big]=0,$$
where $\E_{n,k}[\psi(W)]=n^{-1}\sum_{i\in I_k} \psi(W_i)$ is the empirical expectation over $(W_i)_{i\in I_k}$.
\end{algorithmic}
}
\end{algorithm}
\begin{assumption}[Calibration Complexity]\label{assumption::dml_calibration_complexity}
Let \( \{\tilde{r}_N\}_{N \geq 1} \) and \( \{\tilde{r}^{a}_N\}_{N \geq 1} \) be some sequences of positive constants converging to zero. The estimator $\tilde{\eta} = (\hat{\eta}^{(1)}, \tilde{\eta}^{(2)})$ is a random element in $T$ with nuisance realization set $\widetilde{\mathcal{T}}_N\subseteq \widetilde{\mathcal{T}}^{(1)}_N \times \widetilde{\mathcal{T}}^{(2)}_N$ such that
    \begin{enumerate}
        \item[(i)]
        Let $\eta^{(1)}$ is a fixed element in $\widetilde{\mathcal{T}}_N^{(1)}$ and define
        \begin{align*}
            \F_2(\eta^{(1)}):&=\left\{\psi\left(\cdot; \theta_0, (\eta^{(1)}, \eta^{(2)})\right)- \psi\left(\cdot; \theta_0, (\eta^{(1)}, \eta^{(2)}_0)\right)\big|(\eta^{(1)}, \eta^{(2)}) \in \widetilde{\mathcal{T}}_N\right\}.
        \end{align*}
        Further let $F_2(\eta^{(1)})(\cdot)$ be a measurable envelope for $\F_2(\eta^{(1)})$ such that $$\sup_{\eta^{(1)}\in \widetilde{\mathcal{T}}_N^{(1)}}\|F_2(\eta^{(1)})(W_i)\|_{P,q} \equiv \sup_{\eta^{(1)}\in \widetilde{\mathcal{T}}_N^{(1)}}\|F_2(\eta^{(1)})\|_{P,q} \le V_n$$
        for $q>2$. For each $\eta^{(1)}\in \widetilde{\mathcal{T}}_N^{(1)}$ define $\sigma^2_n$ as a sequence converging to zero such that
        \begin{align*}
            &\sup_{f\in\F_2(\eta^{(1)})} \E[f^2] \le \sigma^2_n\le \|F_2(\eta^{(1)})\|_{P,2}^2
        \end{align*}
        and $u_n$ such that
        \begin{align*}
            &\sup_{\eta^{(1)}\in \widetilde{\mathcal{T}}_N^{(1)}} J\left(\sigma_n /\|F_2(\eta^{(1)})\|_{P,2} , \F_2(\eta^{(1)}), F_2(\eta^{(1)})\right) \le u_n.
        \end{align*}
        Finally, assume the following growth condition is satisfied
        \begin{align*}
            &\sup_{\eta^{(1)}\in \widetilde{\mathcal{T}}_N^{(1)}} \Bigg(u_n \|F_2(\eta^{(1)})\|_{P,2} + \sigma_n \sqrt{\log(n)} \\
            \quad &+  n^{1/q-1/2}V_n \left(u_n^2\frac{\|F_2(\eta^{(1)})\|_{P,2} ^2}{\sigma_n^{2}}\vee\log(n)\right)\Bigg) \lesssim \tilde{r}_N
        \end{align*}

    \item[(ii)] It holds
        \begin{align*}
            \sup_{\eta^{(1)}\in \widetilde{\mathcal{T}}_N^{(1)}}\left|\E\left[\psi^a\left(W;(\eta^{(1)},\eta^{(2)}_{0})\right) - \psi^a\left(W;(\eta^{(1)}_0,\eta^{(2)}_{0})\right)\right]\right| \le \tilde{r}^a_N
        \end{align*}
        with probability converging to one.
        Further, the entropy conditions above also need to hold for $\psi^a$:\\
        Let $\eta^{(1)}$ is a fixed element in $\widetilde{\mathcal{T}}_N^{(1)}$ and define
        \begin{align*}
            \F^a_2(\eta^{(1)}):&=\left\{\psi^a\left(\cdot; (\eta^{(1)}, \eta^{(2)})\right)- \psi^a\left(\cdot; (\eta^{(1)}, \eta^{(2)}_0)\right)\big|(\eta^{(1)}, \eta^{(2)}) \in \widetilde{\mathcal{T}}_N\right\}.
        \end{align*}
        Further let $F^a_2(\eta^{(1)})(\cdot)$ be a measurable envelope for $\F^a_2(\eta^{(1)})$ such that $$\sup_{\eta^{(1)}\in \widetilde{\mathcal{T}}_N^{(1)}}\|F^a_2(\eta^{(1)})(W_i)\|_{P,q} \equiv \sup_{\eta^{(1)}\in \widetilde{\mathcal{T}}_N^{(1)}}\|F^a_2(\eta^{(1)})\|_{P,q} \le V_{n,a}$$
        for $q>2$. For each $\eta^{(1)}\in \widetilde{\mathcal{T}}_N^{(1)}$ define $\sigma^2_{n,a}$ as a sequence converging to zero such that
        \begin{align*}
           &0<\sup_{f\in\F^a_2(\eta^{(1)})} \E[f^2] \le \sigma^2_{n,a}\le \|F^a_2(\eta^{(1)})\|_{P,2}^2
        \end{align*}
        and $u_{n,a}$ such that
        \begin{align*}
            &\sup_{\eta^{(1)}\in \widetilde{\mathcal{T}}_N^{(1)}} J\left(\sigma_{n,a} /\|F^a_2(\eta^{(1)})\|_{P,2} , \F^a_2(\eta^{(1)}), F^a_2(\eta^{(1)})\right) \le u_{n,a}.
        \end{align*}
        Finally, assume the following growth condition is satisfied
        \begin{align*}
            &u_{n,a} \sup_{\eta^{(1)}\in \widetilde{\mathcal{T}}_N^{(1)}}\|F^a_2(\eta^{(1)})\|_{P,2} + \sigma_{n,a} \sqrt{\log(n)} \\
            \quad &+  n^{1/q-1/2}V_{n,a} \left(u_{n,a}^2\frac{\sup_{\eta^{(1)}\in \widetilde{\mathcal{T}}_N^{(1)}}\|F^a_2(\eta^{(1)})\|_{P,2} ^2}{\sigma_{n,a}^{2}}\vee\log(n)\right) \lesssim \tilde{r}^a_N.
        \end{align*}
    \end{enumerate}
\end{assumption}

Assumption \ref{assumption::dml_calibration_complexity} imposes high-level assumptions on the complexity of the calibration step. Assumption \ref{assumption::dml_calibration_complexity} (i) restricts the complexity of the class $\F_2(\eta^{(1)})$ via standard complexity measures.
If the function class is suitably measurable and the uniform entropy integral obeys
$ \log \sup_Q N\left(\epsilon \|F_2(\eta^{(1)})\|_{Q,2}, \F_2(\eta^{(1)}), \|\cdot\|_{Q,2}\right)\le C $,
    it holds
    \begin{align*}
        J\left(\sigma_n /\|F_2(\eta^{(1)})\|_{P,2} , \F_2(\eta^{(1)}), F_2(\eta^{(1)})\right) 
        &\lesssim \sigma_n^{1/2} \|F_2(\eta^{(1)})\|_{P,2}^{-1/2},
    \end{align*} 
    since
    \begin{align*}
        J\left(\delta, \F_2(\eta^{(1)}), F_2(\eta^{(1)})\right)
        := & \int_0^{\delta} \sup_Q \sqrt{1 + \log N(\epsilon\|\F_2(\eta^{(1)})\|_{Q,2}, \F_2(\eta^{(1)}), \|\cdot\|_{Q,2}})d\epsilon \\
        \le & \int_0^{\delta} \sqrt{1 + C \epsilon^{-1}}d\epsilon
        \le \delta + \sqrt{C}\int_0^{\delta} \epsilon^{-1/2}d\epsilon
        \lesssim  \sqrt{\delta}
    \end{align*}
    for any $\delta$ small enough and probability measure $Q$. The first part of Assumption \ref{assumption::dml_calibration_complexity} (ii) imposes a Lipschitz continuity condition on $\psi^{a}$ which is the first part of the linear score defined in Equation \eqref{assumption::linear_score}.
The second part of Assumption \ref{assumption::dml_calibration_complexity} (ii) provides similar complexity assumptions as for the function class $\F_2(\eta^{(1)})$ in Assumption \ref{assumption::dml_calibration_complexity} (i) and also the required growth rates. Remark that the conditions in Assumption \ref{assumption::dml_calibration_complexity} are quite similar to \citet{belloni2018uniformly}, but we build upon standard Donsker conditions. Again, it is worth noting that Assumption \ref{assumption::dml_calibration_complexity} (ii) depends only on the score $\psi^{a}$. In the case of a nonparametric causal model, also known as \textit{interactive regression model} (IRM), considered in Section \ref{ch:irm}, the first part of the linear score is given by $\psi^{a}=-1$, see Equation \eqref{score_IRM}, and therefore $\sigma^2_{n,a}=0$. Hence, in the interactive regression model, Theorem \ref{theorem::dml_asympt_dist_calibration} below holds with $\tilde{r}^a_N=0$ and only Assumption \ref{assumption::dml_calibration_complexity} (i) is required.

\begin{theorem}\label{theorem::dml_asympt_dist_calibration}
Let $\tilde{\theta}_0$ be the estimator according to Algorithm \ref{alg:dml2:calibration} to \ref{alg:dml2:calibration_all}. Assume $\delta_N \ge N^{-1/2}$ for all $N\ge 1$.
Under Assumption \ref{assumption::dml} and \ref{assumption::dml_calibration_complexity}, equation \eqref{eq:dml_normal_dist} in Remark \ref{theorem::dml_asympt_dist} holds with updated remainder, that is 
    \begin{align*}
        \sqrt{N}\sigma^{-1} (\hat{\theta}_0 - \theta_0) = \frac{1}{N}\sum_{i=1}^N \bar{\psi}(W_i) + O_P(\tilde{\rho}_N) 	\rightsquigarrow \N(0,1)
    \end{align*}
uniformly over $P\in\mathcal{P}_N$, where the size of the remainder term obeys
\begin{align*}
    \tilde{\rho}_N:= \rho_N + \tilde{r}_N + \tilde{r}^a_N = N^{-1/2} + r_N + r'_N + \tilde{r}_N + \tilde{r}^a_N + N^{1/2}\lambda_N + N^{1/2}\lambda'_N \lesssim \delta_N.
\end{align*}
\end{theorem}

The crucial difference between Algorithm \ref{alg:dml2:calibration} and \ref{alg:dml2:calibration_all} lies in the assumptions on the nuisance realization set $\mathcal{\tilde{T}}_N$, which require convergence rates of the recalibration procedure. In Algorithm \ref{alg:dml2:calibration}, estimation properties are well known, e.g. for isotonic regression see Section \ref{ch:isotonic_theory}. These proofs heavily rely on the i.i.d. assumption of the samples used for recalibration, which is violated for Algorithm \ref{alg:dml2:calibration_all}. Nevertheless, cross-fitting might result in only weak dependencies between different samples, such that the convergence rates might still be sufficient for the calibration with Algorithm \ref{alg:dml2:calibration_all}. This algorithm closely mirrors the IC-IPW approach described by \citet{vanderlaan2024stabilized}, where calibration is applied to cross-fitted propensity scores on the full sample, retaining the theoretical guarantees established in \citet{vanderlaan2024stabilized} (Theorems 1–2).

\subsection{Calibration for Double Machine Learning Models}\label{ch:calibration_dml_models}
The results of Section \ref{sec:dml} can be applied directly to different regression models and causal parameters of interest. 
In standard settings, a convergence rate of $\|\hat{m}_0 - m_0\|_{P,2}= o_P(N^{-1/4})$ is assumed. If isotonic regression is used for calibration, Lemma \ref{lemma::convergence_isotonic_regression} directly implies that the convergence rate of the calibrated propensity score will still satisfy the rate condition $\|\tilde{m}_0 - m_0\|_{P,2}= o_P(N^{-1/4})$. Furthermore, we state explicit assumptions to restrict the complexity of the nuisance calibration to avoid additional sample splitting.

\subsubsection*{Calibration in interactive regression models}\label{ch:irm}

Consider the fully heterogeneous or interactive regression model as in \citet{Chernozhukov2018dml}. This nonparametric regression model is often considered when augmented inverse probability weighting (AIPW) is used for estimation. Let $D\in\{0,1\}$ be a binary treatment variable and $W=(Y,D,X)$, where
\begin{align} 
    Y = g_0(D,X) + U,\; \E[U|D,X] &= 0, &
    D = m_0(X) + V,\; \E[V|X] &= 0. \label{IRM}
\end{align}
A common parameter of interest is the average treatment effect $\theta_0 := \E[g_0(1,X) - g_0(0,X)].$
Let the score function for the augmented inverse probability weighted estimator be
\begin{equation}\label{score_IRM}
\psi(W;\theta,\eta) := (g(1,X) - g(0,X)) + \frac{D}{m(X)}(Y - g(1,X)) - \frac{1-D}{1-m(X)}(Y - g(0,X)) -\theta
\end{equation}
where $\eta=(g,m)$ denotes the nuisance functions for the outcome regression $g_0(D,X) = E[Y|D,X]$ and propensity score $m_0(X)=E[D|X]$.

\begin{assumption}[cf. Assumption 5.1 in \citet{Chernozhukov2018dml}]\label{assumption::irm}
Let $\{\delta_N\},\{\Delta_N\}\searrow 0$; $c,\epsilon,C>0$, $q>4$, $K\geq 2$ fixed; $N/K\in\mathbb{N}$. For $\eta=(\eta_1,\dots,\eta_\ell)$, define $\|\eta\|_{p,q}:=\max_{1\leq j\leq\ell}\|\eta_j\|_{P,q}$. For all $P\in\mathcal{P}$, the following hold: (a) Equations \eqref{IRM} are satisfied, (b) $\|Y\|_{P,q}\leq C$, (c) $P(\epsilon\leq m_0(X)\leq1-\epsilon)=1$, (d) $\|U\|_{P,2}\geq c$, (e) $\|\E_P[U^2|X]\|_{P,\infty}\leq C$ and (f) for a random subset $I\subset[N]$ of size $n=N/K$, the nuisance parameter estimator $\hat{\eta}_0=\hat{\eta}_0((W_i)_{i\in I^c})$ satisfies, with $P$-probability $\geq1-\Delta_N$:
\begin{itemize}[leftmargin=2cm,noitemsep]
            \item[\romaninline{1}] $\|\hat{\eta}_0-\eta_0\|_{P,2}\leq\delta_N$, $\|\hat{\eta}_0-\eta_0\|_{P,q}\leq C$ 
             \item[\romaninline{2}] $\|\hat{m}_0-m_0\|_{P,2}\times\|\hat{g}_0-g_0\|_{P,2}\leq\delta_N N^{-1/2}$ with $\|\hat{g}_0-g_0\|_{P,\infty}\leq C$, $\|\hat{m}_0-1/2\|_{P,\infty}\leq1/2-\epsilon$
\end{itemize}
\end{assumption}

Under Assumption \ref{assumption::irm}, Remark \ref{theorem::dml_asympt_dist} holds (\citealt[Theorem 5.1]{Chernozhukov2018dml}). Our assumption strengthens the original by requiring $q>4$ rather than $q>2$. For Theorem \ref{theorem::dml_asympt_dist_calibration}, we add:

\begin{assumption}[Calibration rate/complexity]\label{assumption::irm_calibration_rate}
The following assumptions hold: \romaninline{1}  With $P$-probability $\geq 1-\Delta_N$, we have: $\|\tilde{m}(X)-m_0(X)\|_{P,2}\lesssim\varepsilon_N\leq\log^{-1/2}N$, $\varepsilon_N(\cdot)\|\hat{g}_0-g_0\|_{P,2}\leq\delta_N N^{-1/2}$. Further, the predictions are well separated from zero and one, $\|\tilde{m}(X)-1/2\|_{P,\infty}\leq1/2-\epsilon$. \romaninline{2} Let $\tilde{m}(\cdot)\in\mathcal{M}$, such that the covering numbers obey $\sup_Q N(\epsilon,\mathcal{M},L_2(Q))\leq C\epsilon^{-1}$.
\end{assumption}

Assumption \ref{assumption::irm_calibration_rate} imposes mild conditions on the calibration procedure. Assumption \ref{assumption::irm_calibration_rate}(i) ensures that the convergence rate of the calibrated propensity score $\tilde{m}(\cdot)$ is still sufficiently fast, while Assumption \ref{assumption::irm_calibration_rate}(ii) restricts the complexity of calibration, so that additional cross-fitting can be avoided.

\begin{theorem}\label{theorem::irm}
    Under Assumptions \ref{assumption::irm} and \ref{assumption::irm_calibration_rate}(i) Remark \ref{theorem::dml_asympt_dist_naive} is valid. If additionally Assumption \ref{assumption::irm_calibration_rate}(ii) is satisfied, Theorem \ref{theorem::dml_asympt_dist_calibration} holds.
\end{theorem}

The proof of Theorem \ref{theorem::irm} is given in the Supplementary Material.
As mentioned in the beginning of Section \ref{ch:calibration_dml_models}, often convergence rates of $\|\hat{m}_0 - m_0\|_{P,2}= o_P(N^{-1/4})$ and $\|\hat{g}_0 - g_0\|_{P,2}= o_P(N^{-1/4})$ are assumed implying the conditions of Assumption \ref{assumption::irm}(f)(ii). Considering Lemma \ref{lemma::convergence_isotonic_regression} this immediately implies Assumption \ref{assumption::irm_calibration_rate} if isotonic regression is used for calibration as the convergence rate of the calibrated propensity score is given by $\|\tilde{m}_0 - m_0\|_{P,2}= o_P(N^{-1/4})$ and the complexity of monotone functions satisfies Assumption \ref{assumption::irm_calibration_rate}(ii).

\section{Simulation Study}\label{sec:sim}
In this section, we investigate the introduced calibrated propensity score models from Section \ref{ch:calibration_dml_models} through an extensive simulation study\footnote{The code for the simulation study is available at the following link: \url{https://github.com/JanRabenseifner/Causal-Propensity-Calibration.git}. The simulation is executed on an HPC cluster in parallel, using different seeds for the DGPs.}. We evaluate the impact of calibration methods (Venn-ABERS, Platt scaling, isotonic regression) on the performance of causal estimators (IPW, DML), supplemented by analyses of weight normalization and a comparison to covariate-balancing reweighting estimators (e.g., entropy balancing). Performance is assessed using calibration diagnostics (e.g., calibration plots, expected calibration error) and causal estimation metrics (RMSE, MAE, and variance) to unravel the interplay between robustness, forecast accuracy, and covariate balance.

To assess the contribution of potentially miss-calibrated propensity scores, we briefly introduce the causal estimators considered. The inverse probability weighting (IPW) estimator uses estimates of the propensity scores $\hat{m}(D=1|X)$ directly. Here, an estimate $\hat{\theta}$ of the ATE is computed as $\frac{1}{n}\sum_{i=1}^n \bigg( \frac{D^{(i)}Y^{(i)}}{\hat{m}(D=1|X^{(i)})} - \frac{(1-D^{(i)})Y^{(i)}}{1-\hat{m}(D=1|X^{(i)})}\bigg)$.
Especially treated units with low propensity scores and non-treated units with high propensity scores have extreme contributions. This can be critical if the underlying propensity score model is misspecified or overconfident.

The interactive regression model (IRM, Section~\ref{ch:irm}) allows for heterogeneous treatment effects without strong form assumptions. Contrary, the partially linear regression model (PLR, Supplement \ref{S-ch:plm}) imposes an additive structure\footnote{Both, the IRM and PLR model, are implemented via the \texttt{DoubleML} package \citep{DoubleML2022Python, doubleml2024R}.}. One obstacle in evaluating causal ATE models is that the true value of the causal parameter $\theta_0$ is not observed in observational studies.  For a fair evaluation, we have selected four external sources for the data generating processes (DGPs), each proposing different challenges for the models. An overview and detailed descriptions of the DGPs are provided in the Supplement \ref{S-appendix:extradgp}. The DGPs are characterized by varying levels of noise (DGP 1, \citet{belloni2017program}), different dimensionality of observed covariates (DGP 1), and underlying nonlinearities (DGP 2, \citet{deshpande2023calibrated}; DGP 3, \citet{van2023causal}), overlap violations (DGP 2), or unbalancedness (DGP 4, \citet{ballinari2024calibrating, nie2020unbalanced}). The DGPs satisfy the unconfoundedness assumption, $Y(d)\perp D \mid X$, with $Y(d)$ indicating the potential outcome under treatment $D=d$. Hence, these settings allow for identification of the average treatment effect, $\theta_0 = \mathbb{E}[Y(d=1)-Y(d=0)]$. 

\subsection{Learners and Calibration Methods}

We test different learners for the outcome regression and the propensity score estimation. For the outcome regression, we consider a simple linear regression along with the tree-based Machine Learning algorithms LightGBM (LGBM) \citep{ke2017lightgbm} and random forest. Both are flexible machine learning algorithms that perform well across a wide variety of datasets.
For the propensity score estimation, we consider logistic regression, LGBM classifier, and random forest classifier. All models are employed within their default settings. Employing different fine-tuning schemes could benefit either approach and distort the comparison.

For propensity calibration, we consider the three approaches introduced in \ref{sec:calib_methods}.  First, we utilize \texttt{IsotonicRegression} from the \texttt{scikit-learn} package \citep{scikit-learn}. In addition, we employ the Inductive Venn–ABERS predictor (VAP) introduced by \citet{vovk2015vennabers} and available at \citet{Petej2024vap}. VAP builds on the groupings in the outcome space made by isotonic regression. It utilizes potential labels to fit separate isotonic regressions. Thus, simple isotonic regression receives a coarser partitioning of the outcome space. Lastly, we incorporate Platt scaling, implemented via the \texttt{CalibratedClassifierCV} in the \texttt{scikit-learn} package.

\subsection{Calibration Metrics} 

For binary classification, the $\ell_p$ Expected Calibration Error (ECE) \citep{naeini2015ece, sun2024l2ece}, for $p \ge 1$, is defined as:
$\label{ece} \mathrm{ECE}_{p} := \mathbb{E} \left[ \mathbb{E} \left[\|D - m(X)\|^p \mid m(X) \right] \right]^{\frac{1}{p}}.$ To approximate the expected calibration error (ECE), the estimated propensity scores $\hat{m}(X)$ are divided across the probabilistic output range $[0, 1]$ into equally spaced intervals \citep{naeini2015ece} $\{I_0, I_1, ..., I_M\}$ or quantiles \citep{nguyen2015posterior} of $\hat{m}(x)$. This allows us to generate buckets $\{B_i\}_{i=1}^M$, where $B_i = \{(X, D) \mid m(D=1 \mid X) \in I_i\}$. Each predicted probability is assigned to the appropriate bin. The calibration error is then defined as the difference between the fraction of correct predictions (accuracy) and the mean predicted probability (confidence) within each bin: $\operatorname{ECE}_p = \sum_{i=1}^{M} \frac{n_i}{N} \left\| \operatorname{acc}_i(B_i) - \operatorname{conf}_i(B_i) \right\|_p,$ where $\operatorname{acc}_i(B_i) = \frac{1}{|B_i|} \sum_{j=1}^{|B_i|} D_j$ and
$\operatorname{conf}_i(B_i) = \frac{1}{|B_i|} \sum_{j=1}^{|B_i|} m(D=1 \mid X_j)$. 
\vspace{-0.5em}    
\begin{wrapfigure}{!r}{0.45\textwidth} 
\centering 
\includegraphics[width=0.43\textwidth]{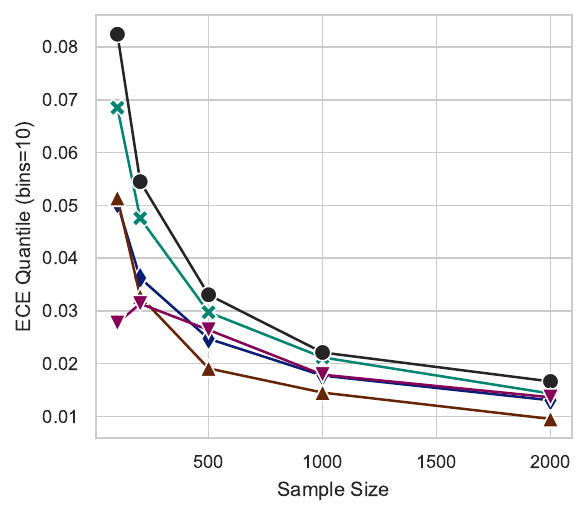} 
\fbox{%
    \scriptsize
    \begin{tabular}{@{}l@{\hspace{2.0em}}l@{\hspace{4.0em}}l@{\hspace{2.0em}}l@{}}
        \begin{tikzpicture}[baseline=0.3ex]
            \useasboundingbox (0,0.3) rectangle (2em,0.9ex);
            \draw[color=cbgreen2, line width=1.5pt] (0,0.9ex) -- (2em,0.9ex);
            \node[color=cbgreen2] at (1em,0.9ex) {\Large$\times$};
        \end{tikzpicture} 
        & \raggedright\arraybackslash Alg-1-uncalib 

        & \begin{tikzpicture}[baseline=0.3ex]
            \useasboundingbox (0,0.3) rectangle (2em,0.9ex);
            \draw[color=cbblack, line width=1.5pt] (0,0.9ex) -- (2em,0.9ex);
            \node[draw=cbblack, circle, line width=0.8pt, fill=cbblack, minimum size=3pt] at (1em,0.9ex) {};
          \end{tikzpicture} 
        & \raggedright\arraybackslash Alg-2-nested-cf \\

        \begin{tikzpicture}[baseline=0.3ex]
            \useasboundingbox (0,0.3) rectangle (2em,0.9ex);
            \draw[color=cbdarkblue, line width=1.5pt] (0,0.9ex) -- (2em,0.9ex);
            \node[draw=cbdarkblue, diamond, line width=0.8pt, fill=cbdarkblue, minimum size=3pt] at (1em,0.9ex) {};
        \end{tikzpicture} 
        & \raggedright\arraybackslash Alg-3-cf 
        & \begin{tikzpicture}[baseline=0.3ex]
            \useasboundingbox (0,0.3) rectangle (2em,0.9ex);
            \draw[color=cbbrown, line width=1.5pt] (0,0.9ex) -- (2em,0.9ex);
            \node[draw=cbbrown, regular polygon, regular polygon sides=3, fill=cbbrown, line width=0.8pt, minimum size=3pt] at (1em,0.9ex) {};
          \end{tikzpicture} 
        & \raggedright\arraybackslash Alg-4-single-split \\

        \begin{tikzpicture}[baseline=0.3ex]
            \useasboundingbox (0,0.3) rectangle (2em,0.9ex);
            \draw[color=cbmagenta, line width=1.5pt] (0,0.9ex) -- (2em,0.9ex);
            \node[draw=cbmagenta, regular polygon, regular polygon sides=3, shape border rotate=180, line width=0.8pt, fill=cbmagenta, minimum size=3pt] at (1em,0.9ex) {};
        \end{tikzpicture} 
        & \multicolumn{3}{@{}l@{}}{\raggedright\arraybackslash Alg-5-full-sample} \\
    \end{tabular}%
}
\vspace{-0.2em}    
\caption{Quantile ECE, DGP 1, \\m = LGBM, n  =2000, p = 20} 
\label{fig:ece_n_obs}
\vspace{-1em}
\end{wrapfigure}
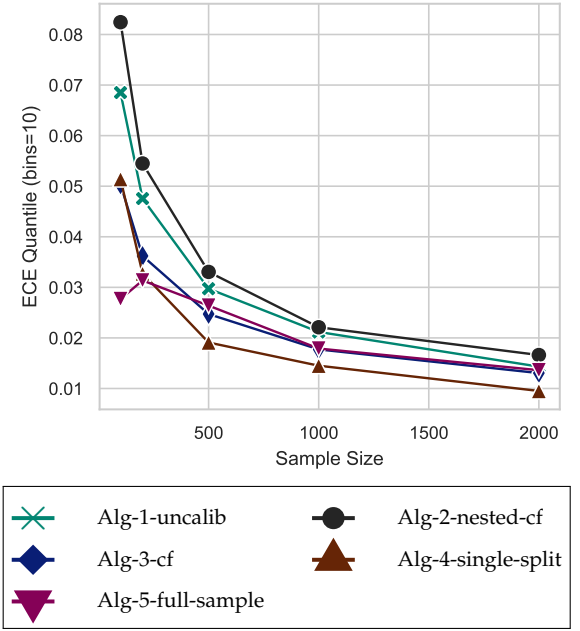

In Figure \ref{fig:ece_n_obs}, we can observe that the uncalibrated Algorithm \ref{alg:dml2:standard}, as well as the nested k-fold cross-fit Algorithm \ref{alg:dml2:naive} are poorly calibrated for small sample sizes. Additionally, a version of Algorithm \ref{alg:dml2:standard} clipped at the one percent level is included. This helps neglect some of the miscalibration, but still performs worse than Algorithms \ref{alg:dml2:calibration_split} and \ref{alg:dml2:calibration_all} for all sample sizes. In propensity weighting, severe deviations in the middle of the propensity score distribution are not particularly critical. However, deviations at the boundaries are crucial because they can lead to exploding weights. Therefore, it is generally advisable to include visualizations to assess both the overlap in propensity scores and their calibration properties. The overlap ratio plot, based on the reliability diagram, splits the probability space into equal parts. For each propensity bin, the plot displays the actual proportion of treated and untreated units separately. 
\begin{figure}[!h]
  \centering
\includegraphics[width=0.9\textwidth]{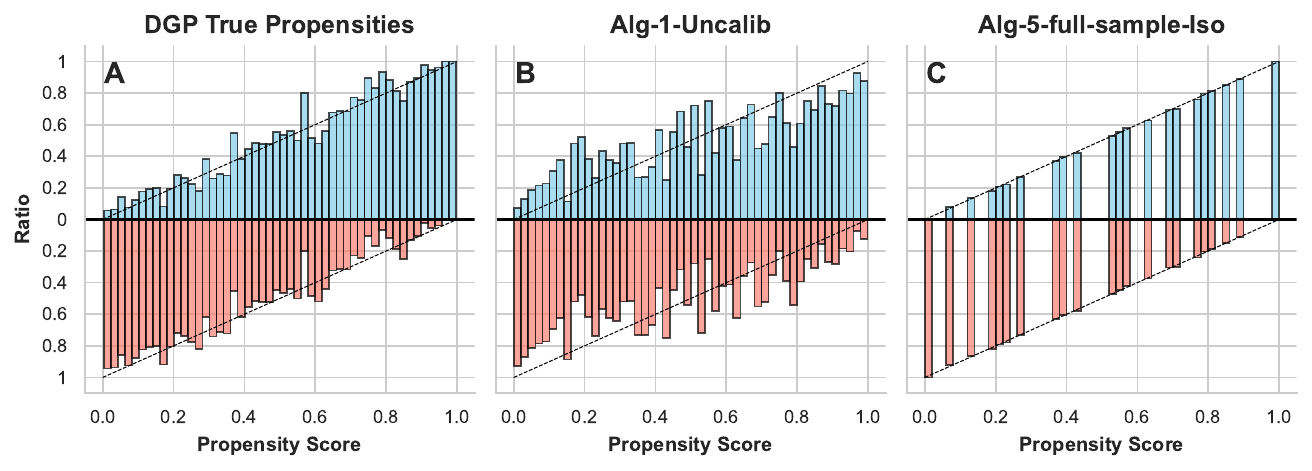}
  \vspace{-0.5em}    
    \fbox{%
        \scriptsize 
        \begin{tabular}{@{}l@{\hspace{1em}}llrr}
            \textcolor{cbblue3}{\rule[+0.4ex]{2em}{3.5pt}} & Treated & 
            \textcolor{cborange2}{\rule[+0.4ex]{2em}{3.5pt}} & Untreated
        \end{tabular}
    }
 \vspace{-0.2em}    
  \caption{Overlap Ratios, DGP 1, n = 2000, p = 20, m = LGBM}
  \label{fig:overlap_ratio}
\end{figure}
\vspace{-1em}
No clear violation of the overlap assumption can be seen for the true underlying propensities displayed in the left panel of Figure \ref{fig:overlap_ratio}. The black dotted lines represent perfect calibration. The ratios illustrate the deviations for both the treated and untreated groups in the uncalibrated Algorithm \ref{alg:dml2:standard} in the middle panel. Notably critical are the substantial proportions of treated observations with estimated propensity scores near zero and untreated observations with propensities close to one. In contrast, the perfect calibration property of Algorithm \ref{alg:dml2:calibration_all} is displayed in the right panel.

\subsection{General Findings}

As expected, the nested cross-fitting Algorithm \ref{alg:dml2:naive} faces stability challenges in small sample size settings. Venn-Abers calibration relies on isotonic regression combined with sample splitting. Consequently, Algorithm \ref{alg:dml2:calibration}, when used with Venn-Abers, encountered similar instability. Additionally, the two-fold calibration Algorithm \ref{alg:dml2:calibration_split}, combined with isotonic regression, required clipping at the 1-percent level to maintain stability. This instability arises from the known limitation of isotonic regression, which is prone to overfitting, especially with small calibration sets \citep{vanderlaan2024selfcalibrating}. To ensure a fair comparison, the uncalibrated Algorithm \ref{alg:dml2:standard} is presented both unclipped and with a restriction at the 1-percent threshold (Alg-1-Clipped). Algorithms \ref{alg:dml2:calibration} and \ref{alg:dml2:calibration_all} were only clipped at a threshold of $10^{-12}$. Given the sample sizes, such violations seem unlikely. This serves more as a general recommendation, as the added clipping bias is negligible. Table \ref{tab:results_overview} provides a summary of the results across all treatment models and the algorithms discussed, in combination with isotonic regression.
\begin{table}[H]
\begin{center}
    \begin{footnotesize}
    \begin{singlespace}
    \captionsetup{justification=centering}
\caption{Results Overview}
 \label{tab:results_overview}
        \scalebox{0.85}{
\centering
\begin{tabular}{lllccccccccc}
\toprule
DGP & Model & Method & \multicolumn{3}{c}{m = Logit} & \multicolumn{3}{c}{m = Random Forest} & \multicolumn{3}{c}{m = LGBM} \\
 &  & & MAE & RMSE & Std. dev. & MAE & RMSE & Std. dev. & MAE & RMSE & Std. dev. \\
\midrule
\multirow{12}{*}{1} & IRM & Alg-1-Clipped & 0.07 & 0.10 & 0.10 & 0.06 & 0.08 & 0.07 & 0.22 & 0.27 & 0.24 \\
& IRM & Alg-1-Uncalib & 0.08 & 0.13 & 0.13 & 1.85e+06 & 1.85e+07 & 1.84e+07 & 0.48 & 0.60 & 0.54 \\
& IRM & Alg-2-nested-cf-Iso & 0.12 & 0.15 & 0.14 & 0.10 & 0.12 & 0.12 & 0.10 & 0.12 & 0.12 \\
& IRM & Alg-3-cf-Iso & 0.06 & 0.08 & 0.07 & 0.06 & 0.08 & 0.08 & 0.06 & 0.08 & 0.07 \\
& IRM & Alg-4-single-split-Iso & 0.10 & 0.13 & 0.12 & 0.07 & 0.08 & 0.06 & 0.07 & 0.09 & 0.06 \\
& IRM & Alg-5-full-sample-Iso & 0.06 & 0.08 & 0.08 & 0.06 & 0.08 & 0.07 & 0.07 & 0.08 & 0.08 \\
\cline{2-12}
& IPW & Alg-1-Clipped & 0.08 & 0.11 & 0.11 & 0.18 & 0.19 & 0.06 & 0.94 & 0.98 & 0.31 \\
& IPW & Alg-1-Uncalib & 0.10 & 0.16 & 0.16 & 1.09e+06 & 1.09e+07 & 1.09e+07 & 1.57 & 1.77 & 0.83 \\
& IPW & Alg-5-full-sample & 0.07 & 0.09 & 0.08 & 0.12 & 0.14 & 0.06 & 0.14 & 0.16 & 0.07 \\
\cline{2-12}
& PLR & Alg-1-Clipped & 0.05 & 0.06 & 0.06 & 0.05 & 0.06 & 0.06 & 0.05 & 0.06 & 0.06 \\
& PLR & Alg-1-Uncalib & 0.05 & 0.06 & 0.06 & 0.05 & 0.06 & 0.06 & 0.05 & 0.06 & 0.06 \\
& PLR & Alg-5-full-sample & 0.05 & 0.06 & 0.06 & 0.05 & 0.06 & 0.06 & 0.05 & 0.07 & 0.07 \\
\midrule
\multirow{12}{*}{2} & IRM & Alg-1-Clipped & 0.09 & 0.11 & 0.11 & 0.31 & 0.39 & 0.39 & 0.20 & 0.26 & 0.26 \\
& IRM & Alg-1-Uncalib & 0.09 & 0.11 & 0.11 & 2.22e+09 & 2.89e+09 & 2.86e+09 & 0.24 & 0.32 & 0.32 \\
& IRM & Alg-2-nested-cf-Iso & 0.18 & 0.23 & 0.22 & 0.16 & 0.22 & 0.22 & 0.19 & 0.24 & 0.24 \\
& IRM & Alg-3-cf-Iso & 0.09 & 0.11 & 0.11 & 0.09 & 0.12 & 0.12 & 0.10 & 0.12 & 0.12 \\
& IRM & Alg-4-single-split-Iso & 0.15 & 0.20 & 0.20 & 0.09 & 0.11 & 0.11 & 0.09 & 0.11 & 0.11 \\
& IRM & Alg-5-full-sample-Iso & 0.09 & 0.11 & 0.11 & 0.09 & 0.11 & 0.11 & 0.09 & 0.12 & 0.12 \\
\cline{2-12}
& IPW & Alg-1-Clipped & 0.09 & 0.12 & 0.11 & 4.45 & 4.59 & 1.11 & 2.67 & 2.74 & 0.61\\
& IPW & Alg-1-Uncalib & 0.09 & 0.12 & 0.11 & 1.17e+10 & 1.37e+10 & 7.23e+09 & 2.86 & 2.98 & 0.81 \\
& IPW & Alg-5-full-sample-Iso & 0.09 & 0.11 & 0.11 & 0.20 & 0.23 & 0.10 & 0.18 & 0.20 & 0.10\\
\cline{2-12}
& PLR & Alg-1-Clipped & 0.09 & 0.11 & 0.10 & 0.12 & 0.14 & 0.10 & 0.10 & 0.11 & 0.10 \\
& PLR & Alg-1-Uncalib & 0.09 & 0.11 & 0.10 & 0.12 & 0.14 & 0.10 & 0.10 & 0.11 & 0.10 \\
& PLR & Alg-5-full-sample-Iso & 0.09 & 0.11 & 0.10 & 0.09 & 0.11 & 0.10 & 0.09 & 0.10 & 0.10 \\

\hline
\multirow{12}{*}{3} & IRM & Alg-1-Clipped & 0.05 & 0.07 & 0.07 & 0.08 & 0.10 & 0.10 & 0.11 & 0.13 & 0.13 \\
& IRM & Alg-1-Uncalib & 0.05 & 0.07 & 0.07 & 9.97e+07 & 2.74e+08 & 2.73e+08 & 0.11 & 0.14 & 0.14 \\
& IRM & Alg-2-nested-cf-Iso & 0.10 & 0.13 & 0.12 & 0.08 & 0.10 & 0.10 & 0.10 & 0.13 & 0.13 \\
& IRM & Alg-3-cf-Iso & 0.05 & 0.07 & 0.07 & 0.06 & 0.07 & 0.07 & 0.05 & 0.07 & 0.06 \\
& IRM & Alg-4-single-split-Iso & 0.10 & 0.12 & 0.11 & 0.05 & 0.07 & 0.06 & 0.05 & 0.07 & 0.06 \\
& IRM & Alg-5-full-sample-Iso & 0.05 & 0.07 & 0.07 & 0.06 & 0.07 & 0.07 & 0.05 & 0.07 & 0.07 \\
\cline{2-12}
& IPW & Alg-1-Clipped & 0.06 & 0.08 & 0.08 & 0.54 & 0.58 & 0.21 & 2.00 & 2.02 & 0.31\\
& IPW & Alg-1-Uncalib & 0.06 & 0.08 & 0.08 & 3.07e+08 & 8.49e+08 & 7.91e+08 & 2.05 & 2.07 & 0.34 \\
& IPW & Alg-5-full-sample-Iso & 0.06 & 0.08 & 0.08 & 0.37 & 0.37 & 0.07 & 0.42 & 0.43 & 0.07\\
\cline{2-12}
& PLR & Alg-1-Clipped & 0.06 & 0.08 & 0.08 & 0.06 & 0.07 & 0.07 & 0.08 & 0.10 & 0.07 \\
& PLR & Alg-1-Uncalib & 0.06 & 0.08 & 0.08 & 0.06 & 0.07 & 0.07 & 0.08 & 0.10 & 0.07 \\
& PLR & Alg-5-full-sample-Iso & 0.07 & 0.09 & 0.08 & 0.06 & 0.07 & 0.07 & 0.06 & 0.07 & 0.07 \\
\hline
\multirow{12}{*}{4} & IRM & Alg-1-Clipped & 0.04 & 0.06 & 0.06 & 0.07 & 0.09 & 0.09 & 0.17 & 0.21 & 0.20 \\
& IRM & Alg-1-Uncalib & 0.04 & 0.06 & 0.06 & 1.74e+08 & 2.70e+08 & 2.68e+08 & 0.36 & 0.46 & 0.43 \\
& IRM & Alg-2-nested-cf-Iso & 0.05 & 0.07 & 0.07 & 0.05 & 0.07 & 0.07 & 0.06 & 0.07 & 0.07 \\
& IRM & Alg-3-cf-Iso & 0.04 & 0.06 & 0.06 & 0.05 & 0.06 & 0.06 & 0.05 & 0.06 & 0.06 \\
& IRM & Alg-4-single-split-Iso & 0.05 & 0.06 & 0.06 & 0.05 & 0.06 & 0.06 & 0.04 & 0.06 & 0.06 \\
& IRM & Alg-5-full-sample-Iso & 0.04 & 0.06 & 0.06 & 0.05 & 0.06 & 0.06 & 0.04 & 0.06 & 0.06 \\
\cline{2-12}
& IPW & Alg-1-Clipped & 0.13 & 0.15 & 0.06 & 0.54 & 0.55 & 0.13 & 6.14 & 6.17 & 0.52 \\
& IPW & Alg-1-Uncalib & 0.13 & 0.15 & 0.06 & 4.64e+08 & 6.89e+08 & 5.11e+08 & 8.37 & 8.45 & 1.21 \\
& IPW & Alg-5-full-sample-Iso & 0.07 & 0.08 & 0.05 & 0.11 & 0.12 & 0.05 & 0.10 & 0.11 & 0.06 \\
\cline{2-12}
& PLR & Alg-1-Clipped & 0.08 & 0.09 & 0.06 & 0.05 & 0.06 & 0.05 & 0.05 & 0.06 & 0.05 \\
& PLR & Alg-1-Uncalib & 0.08 & 0.09 & 0.06 & 0.05 & 0.06 & 0.05 & 0.05 & 0.06 & 0.05 \\
& PLR & Alg-5-full-sample-Iso & 0.08 & 0.09 & 0.06 & 0.07 & 0.09 & 0.05 & 0.08 & 0.09 & 0.05 \\
\bottomrule
\end{tabular}
}
\caption*{\scriptsize For all DGPs: g = LGBM, and for Algorithms 2 - 5: Calibration = Isotonic Regression; DGP 1: n = 2000, p = 20, R2\_d = 0.5; DGP 2: n = 2000, p = 3, overlap = 0.5; DGP 3: n = 2000, p = 4; DGP 4: n = 4000, p = 20, share treated = 0.1}
    \end{singlespace}
    \end{footnotesize}
\end{center}
\end{table}
\vspace{-1.5cm}

\begin{figure}[H]
  \centering
    \includegraphics[width=\textwidth]{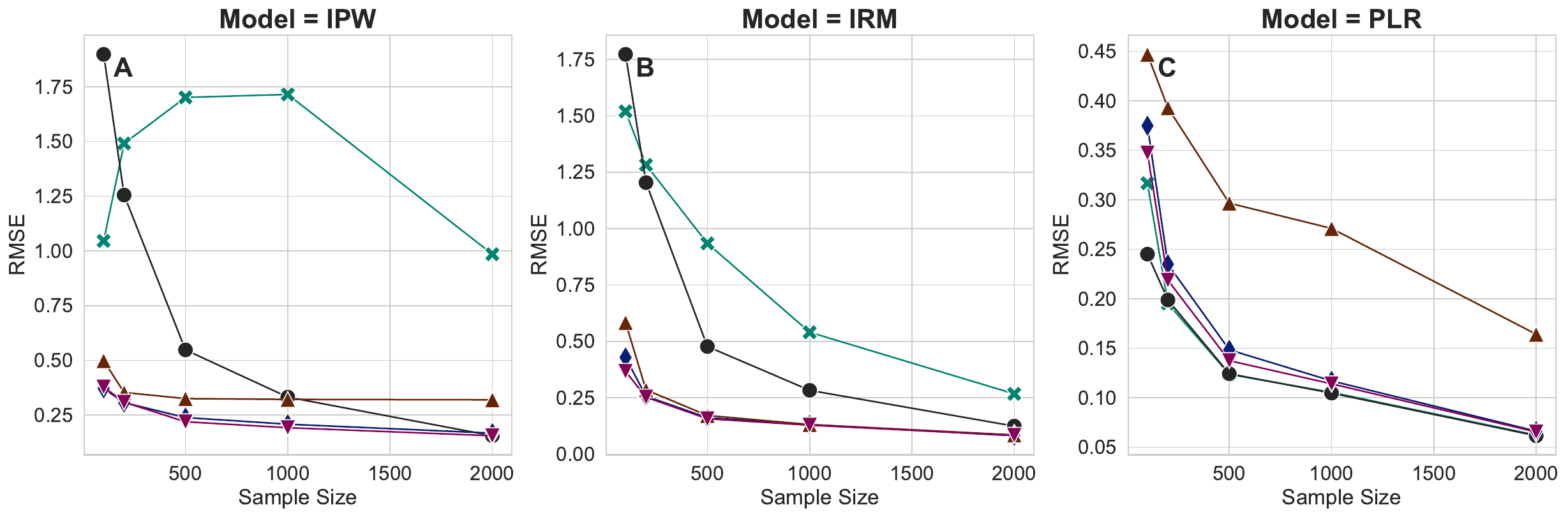}
    \vspace{-0.5em}  
    \fbox{%
        \scriptsize
        \begin{tabular}{@{}l@{\hspace{0.8em}}llllllllll}
            \raisebox{0.5ex}{\begin{tikzpicture}[baseline=-0.5ex]
                \draw[color=cbgreen2, line width=1.5pt] (0,0) -- (2em,0);
                \node[color=cbgreen2] at (1em,0) {\Large$\times$};
            \end{tikzpicture}} & Alg-1-uncalib &
            
            \raisebox{0.5ex}{\begin{tikzpicture}[baseline=-0.5ex]
                \draw[color=cbblack, line width=1.5pt] (0,0) -- (2em,0);
                \node[draw=cbblack, circle, line width=0.8pt, fill=cbblack, minimum size=3pt] at (1em,0) {};
            \end{tikzpicture}} & Alg-2-nested-cf &
            
            \raisebox{0.5ex}{\begin{tikzpicture}[baseline=-0.5ex]
                \draw[color=cbdarkblue, line width=1.5pt] (0,0) -- (2em,0);
                \node[draw=cbdarkblue, diamond, line width=0.8pt, fill=cbdarkblue, minimum size=3pt] at (1em,0) {};
            \end{tikzpicture}} & Alg-3-cf &
            
            \raisebox{0.5ex}{\begin{tikzpicture}[baseline=-0.5ex]
                \draw[color=cbbrown, line width=1.5pt] (0,0) -- (2em,0);
                \node[draw=cbbrown, regular polygon, regular polygon sides=3, fill=cbbrown, line width=0.8pt, minimum size=3pt] at (1em,0) {};
            \end{tikzpicture}} &  Alg-4-single-split &
            
            \raisebox{0.5ex}{\begin{tikzpicture}[baseline=-0.5ex]
                \draw[color=cbmagenta, line width=1.5pt] (0,0) -- (2em,0);
                \node[draw=cbmagenta, regular polygon, regular polygon sides=3, shape border rotate=180, line width=0.8pt,fill=cbmagenta, minimum size=3pt] at (1em,0) {};
            \end{tikzpicture}} & Alg-5-full-sample \\
        \end{tabular}%
    }
 \vspace{-0.2em}    
  \caption{DGP 1, n  =2000, p = 20, R2D = 0.5, m = LGBM, g = LGBM}
  \label{fig:ate_errors_n}
\end{figure}
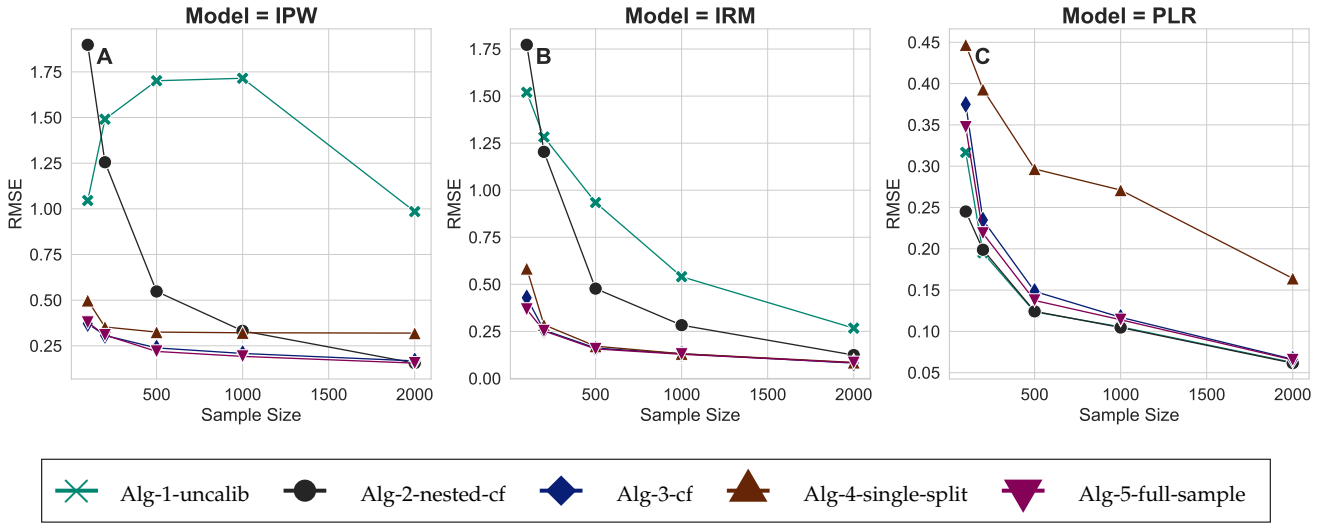

\begin{figure}[H]
  \centering
    \includegraphics[width=\textwidth]{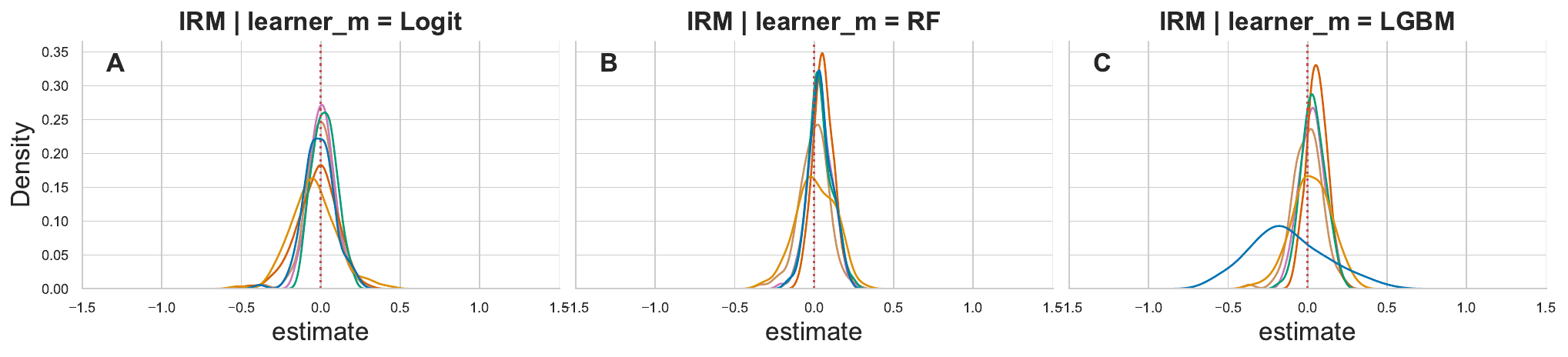}
  \vspace{-0.7em}    
    \fbox{%
        \scriptsize 
        \begin{tabular}{@{}l@{\hspace{1em}}llllll}
            \textcolor{cbblue}{\rule[+0.5ex]{2em}{1.5pt}} & Alg-1-uncalib &
            \textcolor{cborange}{\rule[+0.5ex]{2em}{1.5pt}} & Alg-2-nested-cf &
            \textcolor{cbgreen}{\rule[+0.5ex]{2em}{1.5pt}} & Alg-3-cf \\
            \textcolor{cbred}{\rule[+0.5ex]{2em}{1.5pt}} & Alg-4-single-split &
            \textcolor{cbpurple}{\rule[+0.5ex]{2em}{1.5pt}} & Alg-5-full-sample &
            \textcolor{cbyellow}{\rule[+0.5ex]{2em}{1.5pt}} & Oracle \\
        \end{tabular}
    }
  \caption{DGP 1, n  =2000, p = 20, R2D = 0.5, m = LGBM, g = LGBM}
\label{fig:learner_m}
\end{figure}

Across all DGPs and settings, we can observe that calibration improves the inverse propensity-based IPW and IRM especially in combination with the tree-based propensity learners. The PLR model produces stable results across all settings, with minimal improvement from clipping or calibration for most DGPs. In general, the doubly-robust calibrated IRM and the PLR outperform the IPW. Algorithm \ref{alg:dml2:calibration_split} is biased in the PLR model for the tree-based methods random forest and LGBM in combination with VAP or isotonic regression\footnote{For more details on the influence of sample size on the proposed Algorithms, we refer to the Supplement \ref{S-appendix:extrasim}, Figures \ref{S-fig:ate_errors_n_irm}, \ref{S-fig:ate_errors_n_drug}, \ref{S-fig:ate_errors_n_nl}, \ref{S-fig:ate_errors_n_unb}.}. This bias appears to persist regardless of the sample size, as demonstrated on the right-hand side of Figure \ref{fig:ate_errors_n}. 
    \vspace{-0.5em}  
   \vspace{-1.5em}  
The impact of calibration is strongly dependent on the underlying propensity score learner. In contrast to \citet{blasiok2023doesoptimizingproperloss}, we generally observe good calibration properties for logistic regression and the least improvements through calibration.
\begin{figure}[H]
  \centering
    \includegraphics[width=0.9\textwidth]{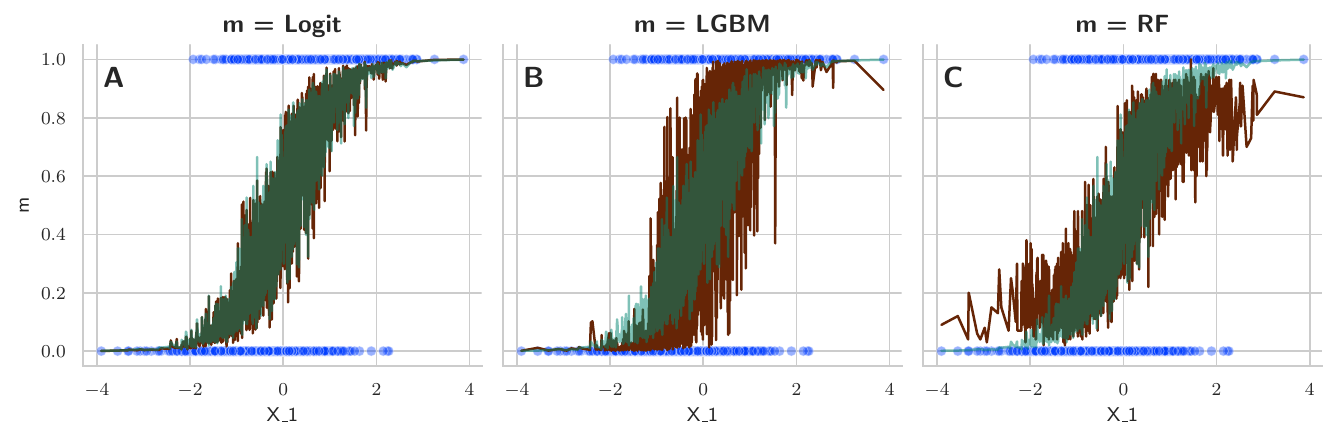}
  \vspace{-0.5em}    
  \fbox{%
        \scriptsize 
        \begin{tabular}{@{}l@{\hspace{1em}}ccccc}
            \textcolor{cbblue2}{\tikz[baseline=-0.7ex]{\fill (0,0) circle (0.4em);}} & Observed treatment &
            \textcolor{cbgreen2}{\rule[+0.5ex]{2em}{1.5pt}} & True $m_0$ & 
            \textcolor{cbbrown}{\rule[+0.5ex]{2em}{1.5pt}} & Predicted $\hat{m}$ 
        \end{tabular}
    }
  \vspace{-0.2em}
  \caption{DGP 1, n  =2000, p = 20, R2D = 0.5}
 \label{fig:ps_calib_iso_2_irm}
 \end{figure}
\vspace{-1.5em} Random forest can be both under-confident, as seen in Figure \ref{fig:ps_calib_iso_2_irm} for DGP 1, and over-confident for DGPs 2 and 3\footnote{The corresponding figures are located in the Supplementary Material, Figures \ref{S-fig:ps_calib_iso_2_drug} and \ref{S-fig:ps_calib_iso_2_nl}.}. As shown by \citet{johansson2023wellcalibrated}, random forest tends to be under-confident for the minority class in unbalanced settings (DGP 4, Figure \ref{S-fig:ps_calib_iso_2_drug} in the Supplementary Material). In general, the combination of random forest with calibration performs well across various settings. On the other hand, boosting-based methods, such as LGBM, tend to be over-confident.
Figure \ref{fig:learner_m} displays the distribution of ATE estimates for different propensity score learners under the IRM across 100 repetitions with different seeds in DGP 1. Calibration can correct for both under-confident and over-confident learners. However, the impact appears strongest for over-confident learners. For more details, we refer to the Supplementary Material, where the robustness of our algorithms is tested with respect to the propensity and outcome learners, different clipping thresholds, and various levels of signal-to-noise ratio (DGP 1), overlap (DGP 2), and share of treated units (DGP 4).
\vspace{-0.5em}  
\vspace{-1.5em}    

\section{Discussion}

We extended and adapted some of the simulation settings to gain further insights from the studies implemented by \citet{deshpande2023calibrated}, \citet{Gutman2024platt}, \citet{ballinari2024calibrating}, and \citet{vanderlaan2024stabilized}.
In our simulation study, our objective was to determine whether calibration works and to explore the best methods to achieve effective calibration.

\citet{deshpande2023calibrated} employ a non-cross-fitted version of Algorithm \ref{alg:dml2:calibration_split}, where the data is split for calibration only. They achieve significant improvements, even in logistic regression, through calibration. This result seems counterintuitive. However, given their deterministic propensity scores in the drug effectiveness DGP, it is unsurprising that logistic regression is miscalibrated. In our adapted version, DGP 2, with non-deterministic propensity scores, such large improvements are not observed. In particular, Algorithm \ref{alg:dml2:calibration_split} without clipping, or in combination with PLR, is not recommended based on our findings.

\citet{ballinari2024calibrating} implements the nested cross-fitting Algorithm \ref{alg:dml2:naive}. As demonstrated, nested cross-fitting combined with isotonic regression is effective only when clipping is applied. Generally, Algorithm \ref{alg:dml2:naive} shows poor calibration properties  and is not recommended for small sample sizes, where calibration has the most significant impact. The unbalanced and nonlinear settings labeled "difficult" and "extreme" for DGPs 4 and 5 by \citet{ballinari2024calibrating} were tested under DGP 4 in this study. We show that either using a larger share of observations for calibration or adding 1-percent clipping allows isotonic regression to lower RMSEs for boosting-based methods and remain stable for other learners. The instability observed in the results of \citet{ballinari2024calibrating} was also addressed by \citet{vanderlaan2024stabilized}. The authors concluded that, in a discretized version, it is important to calibrate the treated and untreated observations separately. However, as we demonstrate, the instability of isotonic regression is more likely due to small sample size issues. Our calibration algorithms, \ref{alg:dml2:calibration} and \ref{alg:dml2:calibration_all}, are both stable without clipping or separate calibration for treated and untreated units. The latter follows \citet{vanderlaan2024stabilized}'s recommendation to calibrate on the full sample using cross-fitted propensity scores.
In summary, our findings emphasize the critical role of method selection and sample size in calibration procedures. While (nested) cross-fitting for the calibration step is not always necessary, it may require supplemental clipping in settings prone to overfitting. Crucially, propensity score calibration enhances the robustness of inverse propensity-weighted ATE estimates.

\section{Acknowledgment}
This work was partly funded by the Bavarian Joint Research Program (BayVFP) – Digitization
(Funding reference: DIK0294/01). Economic AI kindly thanks the the VDI/VDE-IT Munich for the organization and the Free State of Bavaria for the financial support.

\bibliographystyle{abbrvnat}
\bibliography{paper/paper-ref}

\begin{thebibliography}{47}
\providecommand{\natexlab}[1]{#1}
\providecommand{\url}[1]{\texttt{#1}}
\expandafter\ifx\csname urlstyle\endcsname\relax
  \providecommand{\doi}[1]{doi: #1}\else
  \providecommand{\doi}{doi: \begingroup \urlstyle{rm}\Url}\fi

\bibitem[Bach et~al.(2022)Bach, Chernozhukov, Kurz, and Spindler]{DoubleML2022Python}
P.~Bach, V.~Chernozhukov, M.~S. Kurz, and M.~Spindler.
\newblock {DoubleML} -- {A}n object-oriented implementation of double machine learning in {P}ython.
\newblock \emph{Journal of Machine Learning Research}, 23\penalty0 (53):\penalty0 1--6, 2022.

\bibitem[Bach et~al.(2024)Bach, Kurz, Chernozhukov, Spindler, and Klaassen]{doubleml2024R}
P.~Bach, M.~S. Kurz, V.~Chernozhukov, M.~Spindler, and S.~Klaassen.
\newblock {DoubleML}: {A}n object-oriented implementation of double machine learning in {R}.
\newblock \emph{Journal of Statistical Software}, 108\penalty0 (3):\penalty0 1--56, 2024.
\newblock \doi{10.18637/jss.v108.i03}.

\bibitem[Ballinari(2024)]{ballinari2024calibrating}
D.~Ballinari.
\newblock Calibrating doubly-robust estimators with unbalanced treatment assignment, 2024.

\bibitem[Ballinari and Bearth(2025)]{ballinari2025improving}
D.~Ballinari and N.~Bearth.
\newblock Improving the finite sample estimation of average treatment effects using double/debiased machine learning with propensity score calibration, 2025.

\bibitem[Barlow and Brunk(1972)]{barlow1972isotonic}
R.~E. Barlow and H.~D. Brunk.
\newblock The isotonic regression problem and its dual.
\newblock \emph{Journal of the American Statistical Association}, 67\penalty0 (337):\penalty0 140--147, 1972.
\newblock ISSN 01621459, 1537274X.

\bibitem[Belloni et~al.(2017)Belloni, Chernozhukov, Fern{\'a}ndez-Val, and Hansen]{belloni2017program}
A.~Belloni, V.~Chernozhukov, I.~Fern{\'a}ndez-Val, and C.~Hansen.
\newblock Program evaluation and causal inference with high-dimensional data.
\newblock \emph{Econometrica}, 85\penalty0 (1):\penalty0 233--298, 2017.

\bibitem[Belloni et~al.(2018)Belloni, Chernozhukov, Chetverikov, and Wei]{belloni2018uniformly}
A.~Belloni, V.~Chernozhukov, D.~Chetverikov, and Y.~Wei.
\newblock Uniformly valid post-regularization confidence regions for many functional parameters in z-estimation framework.
\newblock \emph{Annals of statistics}, 46\penalty0 (6B):\penalty0 3643, 2018.

\bibitem[Birman and Solomjak(1967)]{BirmanSolomjak1967}
M.~S. Birman and M.~Z. Solomjak.
\newblock Piecewise-polynomial approximations of functions of the classes $w_p^\alpha$.
\newblock \emph{Matematicheskii Sbornik}, 73 (115)\penalty0 (3):\penalty0 295--317, 1967.

\bibitem[Błasiok et~al.(2023)Błasiok, Gopalan, Hu, and Nakkiran]{blasiok2023doesoptimizingproperloss}
J.~Błasiok, P.~Gopalan, L.~Hu, and P.~Nakkiran.
\newblock When does optimizing a proper loss yield calibration?, 2023.

\bibitem[Chernozhukov et~al.(2018)Chernozhukov, Chetverikov, Demirer, Duflo, Hansen, Newey, and Robins]{Chernozhukov2018dml}
V.~Chernozhukov, D.~Chetverikov, M.~Demirer, E.~Duflo, C.~Hansen, W.~Newey, and J.~Robins.
\newblock Double/debiased machine learning for treatment and structural parameters.
\newblock \emph{The Econometrics Journal}, 21\penalty0 (1):\penalty0 C1--C68, 2018.
\newblock \doi{https://doi.org/10.1111/ectj.12097}.

\bibitem[Chernozhukov et~al.(2022)Chernozhukov, Newey, and Singh]{chernozhukov2022autoDML}
V.~Chernozhukov, W.~K. Newey, and R.~Singh.
\newblock {Automatic Debiased Machine Learning of Causal and Structural Effects}.
\newblock \emph{Econometrica}, 90\penalty0 (3):\penalty0 967--1027, May 2022.
\newblock \doi{10.3982/ECTA18515}.

\bibitem[Chernozhukov et~al.(2024)Chernozhukov, Newey, Quintas-Martinez, and Syrgkanis]{chernozhukov2024automaticdebiasedmachinelearning}
V.~Chernozhukov, W.~K. Newey, V.~Quintas-Martinez, and V.~Syrgkanis.
\newblock Automatic debiased machine learning via riesz regression, 2024.

\bibitem[Cox(1958)]{cox1958sigmoid}
D.~R. Cox.
\newblock Two further applications of a model for binary regression.
\newblock \emph{Biometrika}, 45\penalty0 (3-4):\penalty0 562--565, 12 1958.
\newblock ISSN 0006-3444.
\newblock \doi{10.1093/biomet/45.3-4.562}.

\bibitem[Dawid(2014)]{dawid2014probforecast}
A.~P. Dawid.
\newblock \emph{Probability Forecasting}.
\newblock John Wiley \& Sons, Ltd, 2014.
\newblock ISBN 9781118445112.
\newblock \doi{https://doi.org/10.1002/9781118445112.stat02978}.

\bibitem[Deshpande and Kuleshov(2023)]{deshpande2023calibrated}
S.~Deshpande and V.~Kuleshov.
\newblock Calibrated propensity scores for causal effect estimation.
\newblock \emph{arXiv preprint arXiv:2306.00382}, 2023.

\bibitem[Gamarnik(1998)]{gamarnik1998efficient}
D.~Gamarnik.
\newblock Efficient learning of monotone concepts via quadratic optimization.
\newblock In \emph{Proceedings of the eleventh annual conference on computational learning theory}, pages 134--143, 1998.

\bibitem[Gneiting(2014)]{gneiting2014weathercalib}
T.~Gneiting.
\newblock Calibration of medium-range weather forecasts, 03/2014 2014.

\bibitem[Gneiting and Ranjan(2011)]{Gneiting2011probcalib}
T.~Gneiting and R.~Ranjan.
\newblock Combining predictive distributions.
\newblock \emph{Electronic Journal of Statistics}, 7, 06 2011.
\newblock \doi{10.1214/13-EJS823}.

\bibitem[Gneiting et~al.(2007)Gneiting, Balabdaoui, and Raftery]{gneiting2007calibrationsharpness}
T.~Gneiting, F.~Balabdaoui, and A.~E. Raftery.
\newblock Probabilistic forecasts, calibration and sharpness.
\newblock \emph{Journal of the Royal Statistical Society: Series B (Statistical Methodology)}, 69\penalty0 (2):\penalty0 243--268, 2007.
\newblock \doi{https://doi.org/10.1111/j.1467-9868.2007.00587.x}.

\bibitem[Gupta and Ramdas(2021)]{gupta2021binaryhistogram}
C.~Gupta and A.~K. Ramdas.
\newblock Distribution-free calibration guarantees for histogram binning without sample splitting, 2021.

\bibitem[Gupta et~al.(2020)Gupta, Podkopaev, and Ramdas]{gupta2020binarycalibration}
C.~Gupta, A.~Podkopaev, and A.~Ramdas.
\newblock Distribution-free binary classification: prediction sets, confidence intervals and calibration.
\newblock In H.~Larochelle, M.~Ranzato, R.~Hadsell, M.~Balcan, and H.~Lin, editors, \emph{Advances in Neural Information Processing Systems}, volume~33, pages 3711--3723. Curran Associates, Inc., 2020.

\bibitem[Gutman et~al.(2024)Gutman, Karavani, and Shimoni]{Gutman2024platt}
R.~Gutman, E.~Karavani, and Y.~Shimoni.
\newblock Improving inverse probability weighting by post-calibrating its propensity scores.
\newblock \emph{Epidemiology}, 35\penalty0 (4):\penalty0 473–480, Apr. 2024.
\newblock ISSN 1044-3983.
\newblock \doi{10.1097/ede.0000000000001733}.

\bibitem[Johansson et~al.(2023)Johansson, Löfström, and Sönströd]{johansson2023wellcalibrated}
U.~Johansson, T.~Löfström, and C.~Sönströd.
\newblock Well-calibrated probabilistic predictive maintenance using venn-abers, 2023.

\bibitem[Ke et~al.(2017)Ke, Meng, Finley, Wang, Chen, Ma, Ye, and Liu]{ke2017lightgbm}
G.~Ke, Q.~Meng, T.~Finley, T.~Wang, W.~Chen, W.~Ma, Q.~Ye, and T.-Y. Liu.
\newblock Lightgbm: A highly efficient gradient boosting decision tree.
\newblock \emph{Advances in neural information processing systems}, 30:\penalty0 3146--3154, 2017.

\bibitem[Mammen and Yu(2007)]{mammen2007additive}
E.~Mammen and K.~Yu.
\newblock Additive isotone regression.
\newblock \emph{Lecture Notes-Monograph Series}, pages 179--195, 2007.

\bibitem[Naeini et~al.(2015)Naeini, Cooper, and Hauskrecht]{naeini2015ece}
P.~M. Naeini, G.~Cooper, and M.~Hauskrecht.
\newblock Obtaining well calibrated probabilities using bayesian binning.
\newblock \emph{Proceedings of the ... AAAI Conference on Artificial Intelligence. AAAI Conference on Artificial Intelligence}, 2015:\penalty0 2901--2907, 04 2015.
\newblock \doi{10.1609/aaai.v29i1.9602}.

\bibitem[Nguyen and O'Connor(2015)]{nguyen2015posterior}
K.~Nguyen and B.~O'Connor.
\newblock Posterior calibration and exploratory analysis for natural language processing models.
\newblock \emph{arXiv preprint arXiv:1508.05154}, 2015.

\bibitem[Nie and Wager(2020)]{nie2020unbalanced}
X.~Nie and S.~Wager.
\newblock {Quasi-oracle estimation of heterogeneous treatment effects}.
\newblock \emph{Biometrika}, 108\penalty0 (2):\penalty0 299--319, 2020.
\newblock ISSN 0006-3444.

\bibitem[Pedregosa et~al.(2011)Pedregosa, Varoquaux, Gramfort, Michel, Thirion, Grisel, Blondel, Prettenhofer, Weiss, Dubourg, Vanderplas, Passos, Cournapeau, Brucher, Perrot, and Duchesnay]{scikit-learn}
F.~Pedregosa, G.~Varoquaux, A.~Gramfort, V.~Michel, B.~Thirion, O.~Grisel, M.~Blondel, P.~Prettenhofer, R.~Weiss, V.~Dubourg, J.~Vanderplas, A.~Passos, D.~Cournapeau, M.~Brucher, M.~Perrot, and E.~Duchesnay.
\newblock Scikit-learn: Machine learning in {P}ython.
\newblock \emph{Journal of Machine Learning Research}, 12:\penalty0 2825--2830, 2011.

\bibitem[Petej(2024)]{Petej2024vap}
I.~Petej.
\newblock venn-abers.
\newblock \url{https://github.com/ip200/venn-abers}, 2024.

\bibitem[Platt(1999)]{platt1999plattscale}
J.~Platt.
\newblock Probabilistic outputs for support vector machines and comparisons to regularized likelihood methods.
\newblock \emph{Advances in large margin classifiers}, 10\penalty0 (3):\penalty0 61--74, 1999.

\bibitem[Rosenbaum and Rubin(1983)]{rosenbaum1983central}
P.~R. Rosenbaum and D.~B. Rubin.
\newblock The central role of the propensity score in observational studies for causal effects.
\newblock \emph{Biometrika}, 70\penalty0 (1):\penalty0 41--55, 1983.

\bibitem[Sun et~al.(2024)Sun, Chaudhari, Barnett, and Dobriban]{sun2024l2ece}
Y.~Sun, P.~Chaudhari, I.~J. Barnett, and E.~Dobriban.
\newblock A confidence interval for the $\ell_2$ expected calibration error, 2024.

\bibitem[Toth et~al.(2006)Toth, Talagrand, and Zhu]{toth2006weathercalib}
Z.~Toth, O.~Talagrand, and Y.~Zhu.
\newblock \emph{The attributes of forecast systems: a general framework for the evaluation and calibration of weather forecasts}, page 584–595.
\newblock Cambridge University Press, 2006.

\bibitem[Vaart and Wellner(2023)]{vaart2023empirical}
A.~v.~d. Vaart and J.~A. Wellner.
\newblock Empirical processes.
\newblock In \emph{Weak Convergence and Empirical Processes: With Applications to Statistics}, pages 127--384. Springer, 2023.

\bibitem[{van de Geer}(2000)]{geer2000empirical}
S.~A. {van de Geer}.
\newblock \emph{Empirical Processes in M-estimation}, volume~6.
\newblock Cambridge university press, 2000.

\bibitem[van~der Laan and Alaa(2024)]{vanderlaan2024selfcalibrating}
L.~van~der Laan and A.~M. Alaa.
\newblock Self-calibrating conformal prediction, 2024.

\bibitem[van~der Laan et~al.(2023)van~der Laan, Ulloa-Perez, Carone, and Luedtke]{van2023causal}
L.~van~der Laan, E.~Ulloa-Perez, M.~Carone, and A.~Luedtke.
\newblock Causal isotonic calibration for heterogeneous treatment effects.
\newblock In A.~Krause, E.~Brunskill, K.~Cho, B.~Engelhardt, S.~Sabato, and J.~Scarlett, editors, \emph{Proceedings of the 40th International Conference on Machine Learning}, volume 202 of \emph{Proceedings of Machine Learning Research}, pages 34831--34854. PMLR, 23--29 Jul 2023.

\bibitem[van~der Laan et~al.(2024{\natexlab{a}})van~der Laan, Lin, Carone, and Luedtke]{vanderlaan2024stabilized}
L.~van~der Laan, Z.~Lin, M.~Carone, and A.~Luedtke.
\newblock Stabilized inverse probability weighting via isotonic calibration, 2024{\natexlab{a}}.

\bibitem[van~der Laan et~al.(2024{\natexlab{b}})van~der Laan, Luedtke, and Carone]{vanderlaan2024automaticdoublyrobustinference}
L.~van~der Laan, A.~Luedtke, and M.~Carone.
\newblock Automatic doubly robust inference for linear functionals via calibrated debiased machine learning, 2024{\natexlab{b}}.

\bibitem[Vovk and Petej(2014)]{vovk2014vennaberspredictors}
V.~Vovk and I.~Petej.
\newblock Venn-abers predictors, 2014.

\bibitem[Vovk et~al.(2015)Vovk, Petej, and Fedorova]{vovk2015vennabers}
V.~Vovk, I.~Petej, and V.~Fedorova.
\newblock Large-scale probabilistic predictors with and without guarantees of validity.
\newblock \emph{CoRR}, abs/1511.00213, 2015.

\bibitem[W{\"u}thrich and Ziegel(2023)]{wuthrich2023isotonic}
M.~V. W{\"u}thrich and J.~Ziegel.
\newblock Isotonic recalibration under a low signal-to-noise ratio.
\newblock \emph{arXiv preprint arXiv:2301.02692}, 2023.

\bibitem[Yang and Barber(2019)]{yang2019contraction}
F.~Yang and R.~F. Barber.
\newblock Contraction and uniform convergence of isotonic regression, 2019.

\bibitem[Zadrozny and Elkan(2001)]{zadrozny2001calibiontrees}
B.~Zadrozny and C.~Elkan.
\newblock Obtaining calibrated probability estimates from decision trees and naive bayesian classifiers.
\newblock \emph{ICML}, 1, 05 2001.

\bibitem[Zadrozny and Elkan(2002)]{zadrozny2002multiclass}
B.~Zadrozny and C.~Elkan.
\newblock Transforming classifier scores into accurate multiclass probability estimates.
\newblock In \emph{Proceedings of the Eighth ACM SIGKDD International Conference on Knowledge Discovery and Data Mining}, KDD '02, page 694–699, New York, NY, USA, 2002. Association for Computing Machinery.
\newblock ISBN 158113567X.
\newblock \doi{10.1145/775047.775151}.

\bibitem[Zhang(2002)]{zhang2002isotonicbounds}
C.-H. Zhang.
\newblock {Risk bounds in isotonic regression}.
\newblock \emph{The Annals of Statistics}, 30\penalty0 (2):\penalty0 528 -- 555, 2002.
\newblock \doi{10.1214/aos/1021379864}.

\end{thebibliography}


\begin{thebibliography}{27}
\providecommand{\natexlab}[1]{#1}
\providecommand{\url}[1]{\texttt{#1}}
\expandafter\ifx\csname urlstyle\endcsname\relax
  \providecommand{\doi}[1]{doi: #1}\else
  \providecommand{\doi}{doi: \begingroup \urlstyle{rm}\Url}\fi

\bibitem[Austin and Stuart(2015)]{austin2015covbalance}
P.~C. Austin and E.~A. Stuart.
\newblock Moving towards best practice when using inverse probability of treatment weighting (iptw) using the propensity score to estimate causal treatment effects in observational studies.
\newblock \emph{Statistics in Medicine}, 34\penalty0 (28):\penalty0 3661--3679, 2015.
\newblock \doi{https://doi.org/10.1002/sim.6607}.

\bibitem[Ballinari(2024)]{ballinari2024calibrating}
D.~Ballinari.
\newblock Calibrating doubly-robust estimators with unbalanced treatment assignment, 2024.

\bibitem[Belloni et~al.(2017)Belloni, Chernozhukov, Fern{\'a}ndez-Val, and Hansen]{belloni2017program}
A.~Belloni, V.~Chernozhukov, I.~Fern{\'a}ndez-Val, and C.~Hansen.
\newblock Program evaluation and causal inference with high-dimensional data.
\newblock \emph{Econometrica}, 85\penalty0 (1):\penalty0 233--298, 2017.

\bibitem[Belloni et~al.(2018)Belloni, Chernozhukov, Chetverikov, and Wei]{belloni2018uniformly}
A.~Belloni, V.~Chernozhukov, D.~Chetverikov, and Y.~Wei.
\newblock Uniformly valid post-regularization confidence regions for many functional parameters in z-estimation framework.
\newblock \emph{Annals of statistics}, 46\penalty0 (6B):\penalty0 3643, 2018.

\bibitem[Busso et~al.(2014)Busso, DiNardo, and McCrary]{busso2014normalizedweights}
M.~Busso, J.~DiNardo, and J.~McCrary.
\newblock New evidence on the finite sample properties of propensity score reweighting and matching estimators.
\newblock \emph{The Review of Economics and Statistics}, 96\penalty0 (5):\penalty0 885--897, 2014.
\newblock ISSN 00346535, 15309142.

\bibitem[Chernozhukov et~al.(2014)Chernozhukov, Chetverikov, and Kato]{10.1214/14-AOS1230}
V.~Chernozhukov, D.~Chetverikov, and K.~Kato.
\newblock {Gaussian approximation of suprema of empirical processes}.
\newblock \emph{The Annals of Statistics}, 42\penalty0 (4):\penalty0 1564 -- 1597, 2014.
\newblock \doi{10.1214/14-AOS1230}.

\bibitem[Chernozhukov et~al.(2018)Chernozhukov, Chetverikov, Demirer, Duflo, Hansen, Newey, and Robins]{Chernozhukov2018dml}
V.~Chernozhukov, D.~Chetverikov, M.~Demirer, E.~Duflo, C.~Hansen, W.~Newey, and J.~Robins.
\newblock Double/debiased machine learning for treatment and structural parameters.
\newblock \emph{The Econometrics Journal}, 21\penalty0 (1):\penalty0 C1--C68, 2018.
\newblock \doi{https://doi.org/10.1111/ectj.12097}.

\bibitem[Deshpande and Kuleshov(2023)]{deshpande2023calibrated}
S.~Deshpande and V.~Kuleshov.
\newblock Calibrated propensity scores for causal effect estimation.
\newblock \emph{arXiv preprint arXiv:2306.00382}, 2023.

\bibitem[Friedman(1991)]{friedman1991multivariate}
J.~H. Friedman.
\newblock Multivariate adaptive regression splines.
\newblock \emph{The Annals of Statistics}, 19\penalty0 (1):\penalty0 1--67, 1991.

\bibitem[Graham et~al.(2012)Graham, De~Xavier~Pinto, and Egel]{graham2012ipt}
B.~S. Graham, C.~C. De~Xavier~Pinto, and D.~Egel.
\newblock Inverse probability tilting for moment condition models with missing data.
\newblock \emph{The Review of Economic Studies}, 79\penalty0 (3):\penalty0 1053--1079, 04 2012.
\newblock ISSN 0034-6527.
\newblock \doi{10.1093/restud/rdr047}.

\bibitem[Greifer(2025)]{greifer2025weightit}
N.~Greifer.
\newblock \emph{WeightIt: Weighting for Covariate Balance in Observational Studies}, 2025.
\newblock URL \url{https://ngreifer.github.io/WeightIt/}.
\newblock R package version 1.4.0, https://github.com/ngreifer/WeightIt.

\bibitem[Hainmueller(2012)]{hainmueller2012ebalatt}
J.~Hainmueller.
\newblock Entropy balancing for causal effects: A multivariate reweighting method to produce balanced samples in observational studies.
\newblock \emph{Political Analysis}, 20\penalty0 (1):\penalty0 25–46, 2012.
\newblock \doi{10.1093/pan/mpr025}.

\bibitem[Imai and Ratkovic(2014)]{imai2014cbps}
K.~Imai and M.~Ratkovic.
\newblock Covariate balancing propensity score.
\newblock \emph{Journal of the Royal Statistical Society Series B}, 76\penalty0 (1):\penalty0 243--263, 2014.

\bibitem[Imbens(2004)]{imbens2004trim}
G.~Imbens.
\newblock Nonparametric estimation of average treatment effects under exogeneity: A review.
\newblock \emph{The Review of Economics and Statistics}, 86:\penalty0 4--29, 02 2004.
\newblock \doi{10.1162/003465304323023651}.

\bibitem[Kumar et~al.(2019)Kumar, Liang, and Ma]{kumar2019plattcaliberror}
A.~Kumar, P.~S. Liang, and T.~Ma.
\newblock Verified uncertainty calibration.
\newblock In H.~Wallach, H.~Larochelle, A.~Beygelzimer, F.~d\textquotesingle Alch\'{e}-Buc, E.~Fox, and R.~Garnett, editors, \emph{Advances in Neural Information Processing Systems}, volume~32. Curran Associates, Inc., 2019.

\bibitem[Künzel et~al.(2019)Künzel, Sekhon, Bickel, and Yu]{Kunzel2019meta}
S.~R. Künzel, J.~S. Sekhon, P.~J. Bickel, and B.~Yu.
\newblock Metalearners for estimating heterogeneous treatment effects using machine learning.
\newblock \emph{Proceedings of the National Academy of Sciences}, 116\penalty0 (10):\penalty0 4156–4165, Feb. 2019.
\newblock ISSN 1091-6490.
\newblock \doi{10.1073/pnas.1804597116}.

\bibitem[Lambrou et~al.(2012)Lambrou, Papadopoulos, Nouretdinov, and Gammerman]{lambrou2012ivap}
A.~Lambrou, H.~Papadopoulos, I.~Nouretdinov, and A.~Gammerman.
\newblock Reliable probability estimates based on support vector machines for large multiclass datasets.
\newblock \emph{IFIP Advances in Information and Communication Technology}, 382:\penalty0 182--191, 09 2012.
\newblock \doi{10.1007/978-3-642-33412-2_19}.

\bibitem[Li and Sur(2025)]{li2025plattangular}
Y.~Li and P.~Sur.
\newblock Optimal and provable calibration in high-dimensional binary classification: Angular calibration and platt scaling, 2025.

\bibitem[Naeini et~al.(2015)Naeini, Cooper, and Hauskrecht]{naeini2015ece}
P.~M. Naeini, G.~Cooper, and M.~Hauskrecht.
\newblock Obtaining well calibrated probabilities using bayesian binning.
\newblock \emph{Proceedings of the ... AAAI Conference on Artificial Intelligence. AAAI Conference on Artificial Intelligence}, 2015:\penalty0 2901--2907, 04 2015.
\newblock \doi{10.1609/aaai.v29i1.9602}.

\bibitem[Nie and Wager(2020)]{nie2020unbalanced}
X.~Nie and S.~Wager.
\newblock {Quasi-oracle estimation of heterogeneous treatment effects}.
\newblock \emph{Biometrika}, 108\penalty0 (2):\penalty0 299--319, 2020.
\newblock ISSN 0006-3444.

\bibitem[Nixon et~al.(2020)Nixon, Dusenberry, Jerfel, Nguyen, Liu, Zhang, and Tran]{nixon2020adaptivece}
J.~Nixon, M.~Dusenberry, G.~Jerfel, T.~Nguyen, J.~Liu, L.~Zhang, and D.~Tran.
\newblock Measuring calibration in deep learning, 2020.

\bibitem[Nouretdinov et~al.(2018)Nouretdinov, Volkhonskiy, Lim, Toccaceli, and Gammerman]{nouretdinov2018ivap}
I.~Nouretdinov, D.~Volkhonskiy, P.~Lim, P.~Toccaceli, and A.~Gammerman.
\newblock Inductive {V}enn-{A}bers predictive distribution.
\newblock In A.~Gammerman, V.~Vovk, Z.~Luo, E.~Smirnov, and R.~Peeters, editors, \emph{Proceedings of the Seventh Workshop on Conformal and Probabilistic Prediction and Applications}, volume~91 of \emph{Proceedings of Machine Learning Research}, pages 15--36. PMLR, 11--13 Jun 2018.

\bibitem[Platt(1999)]{platt1999plattscale}
J.~Platt.
\newblock Probabilistic outputs for support vector machines and comparisons to regularized likelihood methods.
\newblock \emph{Advances in large margin classifiers}, 10\penalty0 (3):\penalty0 61--74, 1999.

\bibitem[van~der Laan et~al.(2023)van~der Laan, Ulloa-Perez, Carone, and Luedtke]{van2023causal}
L.~van~der Laan, E.~Ulloa-Perez, M.~Carone, and A.~Luedtke.
\newblock Causal isotonic calibration for heterogeneous treatment effects.
\newblock In A.~Krause, E.~Brunskill, K.~Cho, B.~Engelhardt, S.~Sabato, and J.~Scarlett, editors, \emph{Proceedings of the 40th International Conference on Machine Learning}, volume 202 of \emph{Proceedings of Machine Learning Research}, pages 34831--34854. PMLR, 23--29 Jul 2023.

\bibitem[Vovk and Petej(2014)]{vovk2014vennaberspredictors}
V.~Vovk and I.~Petej.
\newblock Venn-abers predictors, 2014.

\bibitem[Vovk et~al.(2004)Vovk, Shafer, and Nouretdinov]{vovk2004calibration}
V.~Vovk, G.~Shafer, and I.~Nouretdinov.
\newblock Self-calibrating probability forecasting.
\newblock In \emph{Advances in Neural Information Processing Systems 16 - Proceedings of the 2003 Conference, NIPS 2003}, Advances in Neural Information Processing Systems. Neural information processing systems foundation, 2004.
\newblock ISBN 0262201526.
\newblock 17th Annual Conference on Neural Information Processing Systems, NIPS 2003 ; Conference date: 08-12-2003 Through 13-12-2003.

\bibitem[Zubizarreta(2015)]{zubizarreta2015optweight}
J.~R. Zubizarreta.
\newblock Stable weights that balance covariates for estimation with incomplete outcome data.
\newblock \emph{Journal of the American Statistical Association}, 110\penalty0 (511):\penalty0 910--922, 2015.
\newblock \doi{10.1080/01621459.2015.1023805}.

\end{thebibliography}

\clearpage

\ifarXiv
    

\section*{Supplementary material}

# 1   Details on Calibration

## 1.1   Details on Calibration Methods

### 1.1.1   Venn-Abers Calibration

Venn-Abers predictors (VAPs) (Vovk et al., 2004) provide calibrated probability estimates by considering both possible labels for a test instance and fitting separate isotonic regressions for each case. For binary outcomes (Vovk and Petej, 2014) $D \in \{0,1\}$, VAPs assign each test unit $X_{l+1}$ two probabilities: $\tilde{m}_0(W)$ and $\tilde{m}_1(W)$, derived from isotonic regression under the assumed labels $D_{l+1} = 0$ and $D_{l+1} = 1$. These estimates are oracle calibrated—the true label's corresponding probability is valid. While aligning one probability distribution with the true outcome ensures accurate calibration, the "oracle" selector $S = D$ is not known in practice. This means we often need to use heuristic methods, like averaging, to combine information, even though these methods do not come with formal guarantees. Computational expense also arises from refitting isotonic models per test instance.

Inductive Venn-Abers predictors (IVAPs) (Lambrou et al., 2012; Nouretdinov et al., 2018) address these issues by splitting data into a proper training set (to fit a propensity model $\hat{m}(X)$) and a calibration set (to fit isotonic maps $\tilde{m}_0, \tilde{m}_1$). For square loss, IVAPs combine $p_0 = \tilde{m}_0(\hat{m}(X))$ and $p_1 = \tilde{m}_1(\hat{m}(X))$ into a single probability: $p = p_1 + \frac{p_0^2}{2} - \frac{p_1^2}{2}$. As Vovk and Petej (2014) point out, this can be rewritten as: $p = \bar{p} + (p_1 - p_0)\left(\frac{1}{2} - \bar{p}\right), \quad \bar{p} = \frac{p_0 + p_1}{2}$. Thus, $p$ is a regularized version of $\bar{p}$ moving the prediction towards $\frac{1}{2}$. This characteristic is particularly advantageous in mitigating the risk of inflated weights during treatment effect estimation.

### 1.1.2 Platt Scaling

Platt scaling (Platt, 1999) leverages the robust calibration properties of log-loss. It applies logistic regression, expressed by the function $f(\hat{m}(X)) := \frac{1}{1+\exp(A\hat{m}(X)+B)}$, to the scores produced by an estimator, using treatment assignment labels as targets for calibration, where $A < 0$ and $B$ are parameters. The parameters $A$ and $B$ are estimated using a maximum likelihood method on the same training set as the original classifier $m(x)$. To avoid overfitting, a held-out calibration set or cross-validation can be used. Platt also recommends clipping the outputs $d$ to target probabilities in the range $\left(\frac{1}{N_0+2}, \frac{N_1+1}{N_1+2}\right)$, where $N_0$ is the number of control units and $N_1$ is the number of treated units in the calibration set. This motivates the testing of additional clipping in small sample sizes, a setting where the parametric form assumption is especially appropriate. Unlike isotonic regression, for which van der Laan et al. (2023) established distribution-free calibration guarantees, Platt scaling lacks universal theoretical guarantees. This limitation is critical in propensity calibration, where treatment assignments can be unbalanced, or propensity scores deviate from logistic normality (e.g., highly skewed or multimodal distributions). Empirical studies further challenge its reliability: Kumar et al. (2019) show that Platt scaling's apparent calibration is often inflated due to the systematic underestimation of errors in continuous output spaces, where true calibration cannot be verified without uncheckable smoothness assumptions. Recent work by Li and Sur (2025) identifies specific conditions - notably Gaussian-like or light-tailed feature distributions - under which Platt scaling achieves Bregman optimality, minimizing divergences such as log loss or squared error, even in high-dimensional settings.

## 1.2 Alternative Calibration Metrics

Commonly employed are the L1-Norm and L2-Norm. (Nixon et al., 2020) suggests using either the L2-norm or adaptive intervals, as the L1-norm is highly susceptible to the decision on the number of intervals. To assess subpopulations with extreme propensity scores, the maximum calibration error (MCE) (Naeini et al., 2015) can be used. MCE is defined as the maximum deviation across all bins, given by: $\max_{i \in \{1,...,M\}} |\text{acc}_i(B_i) - \text{conf}_i(B_i)|$. This error reflects the

worst-case deviation between predicted and actual values across all intervals. All calibration metrics are visualized with respect to the observation size and the underlying propensity learner in Supplement 4.

## 2  Details on Calibration for Double Machine Learning Models

### 2.1  Calibration in Partially Linear Regression Models

Consider the partially linear regression model as in Chernozhukov et al. (2018). Let $D \in \{0, 1\}$ be a binary treatment variable and $W = (Y, D, X)$, where

$$Y = \theta_0 D + g_0(X) + U, \ \mathbb{E}[U|D, X] = 0, \qquad D = m_0(X) + V, \ \mathbb{E}[V|X] = 0, \qquad (1)$$

with $\theta_0$ being the parameter of interest. Let

$$\psi(W; \theta, \eta) := (Y - l(X) - \theta(D - m(X)))(D - m(X))$$

be the "partialling-out" score function, where $\eta = (l, m)$ denotes the nuisance functions for the outcome regression $l_0(X) = \mathbb{E}[Y|X]$ and propensity score $m_0(X) = \mathbb{E}[D|X]$.

**Assumption S1** (cf. Assumption 4.1 in Chernozhukov et al. (2018)). *Let $(\delta_N)_{n=1}^\infty$ and $(\Delta_N)_{n=1}^\infty$ be sequences of positive constants approaching 0 as before. Also, let $c, C$ and $q$ be fixed strictly positive constants such that $q > 4$, and let $K \geq 2$ be a fixed integer. Moreover, for any $\eta = (\eta_1, \eta_2)$, where $\eta_1$ and $\eta_2$ are functions mapping the support of X to $\mathbb{R}$, denote $\|\eta\|_{p,q} = \|\eta_1\|_{p,q} \vee \|\eta_2\|_{p,q}$. For simplicity, assume that $N/K$ is an integer. For all probability laws $P \in \mathcal{P}$ for the triple $(Y, D, X)$, the following conditions hold:*

*(a) Equations (1) hold*

*(b) $\|Y\|_{P,q} + \|D\|_{P,q} \leq C$*

*(c) $\|UV\|_{P,2} \geq c^2$ and $\mathbb{E}_P[V^2] \geq c$*

*(d) $\|\mathbb{E}_P[U^2 \mid X]\|_{P,\infty} \leq C$ and $\|\mathbb{E}_P[V^2 \mid X]\|_{P,\infty} \leq C$*

(e) *Given a random subset $I$ of $[N]$ of size $n = N/K$, the nuisance parameter estimator $\hat{\eta}_0 = ((W_i)_{i \in I^c})$ obeys the following conditions for all $N \geq 1$. With P-probability no less than $1 - \Delta_N$,*

(i) $\|\hat{\eta}_0 - \eta_0\|_{P,q} \leq C$ and $\|\hat{\eta}_0 - \eta_0\|_{P,2} \leq \delta_N$

(ii) $\|\hat{m}_0 - m_0\|_{P,2} \times \left( \|\hat{m}_0 - m_0\|_{P,2} + \|\hat{\ell}_0 - \ell_0\|_{P,2} \right) \leq \delta_N N^{-1/2}$.

Under Assumption S1, Remark 2 holds (cf. Theorem 4.1 in Chernozhukov et al. (2018)). To apply Theorem 1 using a calibrated propensity score estimator, Assumption 4 has to be satisfied. To this end, we introduce the following Assumption S2.

**Assumption S2** (Calibration rate and complexity). *Let $\tilde{m}(X)$ be an estimator of $\tilde{m}_0(X) := \mathbb{E}[D|\hat{m}(X)]$. We assume*

(i) *The following convergence rates hold with P-probability no less than $1 - \Delta_N$,*

$$\|\tilde{m}(X) - \tilde{m}_0(X)\|_{P,2} \lesssim \varepsilon_N,$$

such that

$$\varepsilon_N \cdot \left( \varepsilon_N + \|\hat{\ell}_0 - \ell_0\|_{P,2} \right) \leq \delta_N N^{-1/2}.$$

(ii) *Let $\tilde{m}(\cdot) \in \mathcal{M}$, such that the covering numbers obey*

$$\sup_Q N(\epsilon, \mathcal{M}, L_2(Q)) \leq C\epsilon^{-1}.$$

It is worth noting that this calibration assumption does not necessarily require $D$ to be binary.

**Theorem S1.** *Under Assumptions S1 and S2(i) Remark 3 is valid. If additionally Assumption S2(ii) is satisfied, Theorem 1 holds.*

The proof of Theorem S1 is similar to the proof of Theorem 2 and is therefore left out. It just needs an additional complexity argument to verify Assumption 4 (ii) based on Assumption S2(ii).

## 2.2 Proofs

### 2.2.1 Proofs for Section 2

It holds

$$\|\hat{m}(X) - m_0(X)\|_{P,2}^2$$
$$= \mathbb{E}\left[\left(m_0(X) - \mathbb{E}[m_0(X)|\hat{m}(X)] + \mathbb{E}[m_0(X)|\hat{m}(X)] - \hat{m}(X)\right)^2\right]$$
$$= \mathbb{E}\left[\mathbb{E}\left[\left(m_0(X) - \mathbb{E}[m_0(X)|\hat{m}(X)] + \mathbb{E}[m_0(X)|\hat{m}(X)] - \hat{m}(X)\right)^2 |\hat{m}(X)\right]\right]$$
$$= \mathbb{E}\left[\text{Var}(m_0(X)|\hat{m}(X))\right] + \|\mathbb{E}[m_0(X)|\hat{m}(X)] - \hat{m}(X)\|_{P,2}^2$$
$$+ 2\mathbb{E}\left[\left(\mathbb{E}[m_0(X)|\hat{m}(X)] - \hat{m}(X)\right)\underbrace{\mathbb{E}\left[\left(m_0(X) - \mathbb{E}[m_0(X)|\hat{m}(X)]\right)|\hat{m}(X)\right]}_{=0}\right]$$

such that

$$\|\hat{m}(X) - m_0(X)\|_{P,2} = \left(\mathbb{E}\left[\text{Var}(m_0(X)|\hat{m}(X))\right] + \|\mathbb{E}[m_0(X)|\hat{m}(X)] - \hat{m}(X)\|_{P,2}^2\right)^{1/2}.$$

**Proof of Lemma 1.**

Under Assumption 1 we can decompose the root-mean-squared-error as follows

$$\|\tilde{m}(X) - m_0(X)\|_{P,2} \leq \underbrace{\|\tilde{m}(X) - \mathbb{E}[D|\hat{m}(X)]\|_{P,2}}_{\leq \tilde{\varepsilon}_N} + \|\mathbb{E}[D|\hat{m}(X)] - m_0(X)\|_{P,2}$$
$$\leq \left(\mathbb{E}\left[\text{Var}(m_0(X)|\hat{m}(X))\right]\right)^{1/2} + \tilde{\varepsilon}_N$$

due to

$$\|\mathbb{E}[D|\hat{m}(X)] - m_0(X)\|_{P,2}^2 = \|\mathbb{E}[m_0(X)|\hat{m}(X)] - m_0(X)\|_{P,2}^2$$
$$= \mathbb{E}\left[\mathbb{E}\left[\left(m_0(X) - \mathbb{E}[m_0(X)|\hat{m}(X)]\right)^2\right]|\hat{m}(X)\right]$$
$$= \mathbb{E}\left[\text{Var}(m_0(X)|\hat{m}(X))\right].$$

Here, we used that

$$\sigma(\hat{m}(X)) \subseteq \sigma(X),$$

such that

$$\mathbb{E}[m_0(X)|\hat{m}(X)] = \mathbb{E}[\mathbb{E}[D|X]|\hat{m}(X)] = \mathbb{E}[D|\hat{m}(X)].$$

□

### 2.2.2 Proofs for Section 3

For any $k \in \{1, \ldots, K\}$, we use the following empirical process notation

$$\mathbb{G}_{n,k}[\phi(W)] = \frac{1}{\sqrt{n}} \sum_{i \in I_k} \left( \phi(W_i) - \int \phi(w) dP_N \right)$$

for any $P_N$-integrable function $\phi$ on $\mathcal{W}$.

**Proof of Theorem 1.**

This proof adjusts the proof of Theorem 3.1 in Chernozhukov et al. (2018). Upon inspecting the Step 1 (DML2 case) of the proof of Theorem 3.1 in Chernozhukov et al. (2018), it is worth noting that the only terms which are affected by the calibration are

$$\tilde{R}_{N,1} := \frac{1}{K} \sum_{k=1}^{K} \mathbb{E}_{n,k} \left[ \psi^a \left( W; (\hat{\eta}_{0,k}^{(1)}, \tilde{\eta}_{0,k}^{(2)}) \right) \right] - \mathbb{E} \left[ \psi^a \left( W_i; (\eta_{0,k}^{(1)}, \eta_{0,k}^{(2)}) \right) \right]$$

$$\tilde{R}_{N,2} := \frac{1}{K} \sum_{k=1}^{K} \mathbb{E}_{n,k} \left[ \psi \left( W; \theta_0, (\hat{\eta}_{0,k}^{(1)}, \tilde{\eta}_{0,k}^{(2)}) \right) \right] - \frac{1}{N} \sum_{i=1}^{N} \psi \left( W_i; \theta_0, (\eta_{0,k}^{(1)}, \eta_{0,k}^{(2)}) \right)$$

as the other terms do not depend on the calibrated element $\tilde{\eta}_{0,k}^{(2)}$. First, we focus on the term $\tilde{R}_{N,2}$. Following Step 3 of the proof of Theorem 3.1, we obtain by triangle inequality

$$\left| \mathbb{E}_{n,k} \left[ \psi \left( W; \theta_0, (\hat{\eta}_{0,k}^{(1)}, \tilde{\eta}_{0,k}^{(2)}) \right) \right] - \frac{1}{n} \sum_{i \in I_k} \psi \left( W_i; \theta_0, (\eta_{0,k}^{(1)}, \eta_{0,k}^{(2)}) \right) \right| \leq \frac{\mathcal{I}_{3,k} + \mathcal{I}_{4,k}}{\sqrt{n}},$$

where

$$\mathcal{I}_{3,k} := \left| \mathbb{G}_{n,k} \left[ \psi \left( W; \theta_0, (\hat{\eta}_{0,k}^{(1)}, \tilde{\eta}_{0,k}^{(2)}) \right) \right] - \mathbb{G}_{n,k} \left[ \psi \left( W; \theta_0, (\eta_{0,k}^{(1)}, \eta_{0,k}^{(2)}) \right) \right] \right|$$

$$\mathcal{I}_{4,k} := \sqrt{n} \left| \mathbb{E} \left[ \psi \left( W; \theta_0, (\hat{\eta}_{0,k}^{(1)}, \tilde{\eta}_{0,k}^{(2)}) \right) \mid (W_i)_{i \in I_k^c} \right] - \mathbb{E} \left[ \psi \left( W; \theta_0, (\eta_{0,k}^{(1)}, \eta_{0,k}^{(2)}) \right) \right] \right|.$$

Due to Assumption 3 it holds $(\hat{\eta}_{0,k}^{(1)}, \tilde{\eta}_{0,k}^{(2)}) \in \widetilde{\mathcal{T}}_N$ with probability $\geq 1 - \Delta_N$. Define $\mathcal{E}_N$ as the event that $(\hat{\eta}_{0,k}^{(1)}, \tilde{\eta}_{0,k}^{(2)}) \in \widetilde{\mathcal{T}}_N$ for all $k \in [K]$. Therefore, $P(\mathcal{E}_N) \geq 1 - K\Delta_N$. Consequently, the whole argument of Chernozhukov et al. (2018) is still valid which implies

$$\mathcal{I}_{4,k} = O_{P_N}(\sqrt{n}(\lambda_N + \lambda'_N)).$$

Next, due to the triangle inequality

$$\mathcal{I}_{3,k} \leq \left| \mathbb{G}_{n,k} \left[ \psi \left( W; \theta_0, (\hat{\eta}_{0,k}^{(1)}, \tilde{\eta}_{0,k}^{(2)}) \right) \right] - \mathbb{G}_{n,k} \left[ \psi \left( W; \theta_0, (\hat{\eta}_{0,k}^{(1)}, \eta_{0,k}^{(2)}) \right) \right] \right|$$
$$+ \left| \mathbb{G}_{n,k} \left[ \psi \left( W; \theta_0, (\hat{\eta}_{0,k}^{(1)}, \eta_{0,k}^{(2)}) \right) \right] - \mathbb{G}_{n,k} \left[ \psi \left( W; \theta_0, (\eta_{0,k}^{(1)}, \eta_{0,k}^{(2)}) \right) \right] \right|$$
$$=: \mathcal{I}_{3,k}^{(1)} + \mathcal{I}_{3,k}^{(2)},$$

where $\mathcal{I}_{3,k}^{(2)} = O_{P_N}(r'_N)$ by Assumption 3.2, see Chernozhukov et al. (2018). To bound $\mathcal{I}_{3,k}^{(1)}$, we rely on classical empirical process theory. Since conditionally on $(W_i)_{i \in I_k^c}$ the $\hat{\eta}_{0,k}^{(1)}$ is non-stochastic, remark that

$$\mathbb{E}\left[ \mathcal{I}_{3,k}^{(1)} \mid (W_i)_{i \in I_k^c} \right]$$
$$= \mathbb{E}\left[ \left| \mathbb{G}_{n,k} \left[ \psi \left( W; \theta_0, (\hat{\eta}_{0,k}^{(1)}, \tilde{\eta}_{0,k}^{(2)}) \right) \right] - \mathbb{G}_{n,k} \left[ \psi \left( W; \theta_0, (\hat{\eta}_{0,k}^{(1)}, \eta_{0,k}^{(2)}) \right) \right] \right| \,\bigg|\, (W_i)_{i \in I_k^c} \right]$$
$$\leq \sup_{\eta^{(1)} \in \widetilde{\mathcal{T}}_N^{(1)}} \mathbb{E}\left[ \sup_{\eta^{(2)}:(\eta^{(1)},\eta^{(2)}) \in \widetilde{\mathcal{T}}_N} \left| \mathbb{G}_{n,k}\left[ \psi\left(W;\theta_0,(\eta^{(1)},\eta^{(2)})\right) \right] \right. \right.$$
$$\left. \left. - \mathbb{G}_{n,k}\left[ \psi\left(W;\theta_0,(\eta^{(1)},\eta_{0,k}^{(2)})\right) \right] \right| \,\bigg|\, (W_i)_{i \in I_k^c} \right]$$
$$\leq \sup_{\eta^{(1)} \in \tilde{\mathcal{T}}_N^{(1)}} \mathbb{E}\left[ \sup_{f \in \mathcal{F}_2(\eta^{(1)})} |\mathbb{G}_{n,k}(f)| \,\bigg|\, (W_i)_{i \in I_k^c} \right]$$

8

$$\leq \sup_{\eta^{(1)} \in \widetilde{\mathcal{T}}_N^{(1)}} \mathbb{E}[\sup_{f \in \mathcal{F}_2(\eta^{(1)})} |\mathbb{G}_{n,k}(f)|]$$

$$= \sup_{\eta^{(1)} \in \widetilde{\mathcal{T}}_N^{(1)}} \|\mathbb{G}_{n,k}\|_{\mathcal{F}_2(\eta^{(1)})}$$

where $\mathcal{F}_2$ is defined in Assumption 4. Let $\eta^{(1)}$ be a any element of $\widetilde{\mathcal{T}}_N^{(1)}$. Relying on Theorem 5.2 of Chernozhukov et al. (2014), it holds

$$\mathbb{E}\left[\|\mathbb{G}_{n,k}\|_{\mathcal{F}_2(\eta^{(1)})}\right] \lesssim J\left(\delta_n, \mathcal{F}_2(\eta^{(1)}), F_2(\eta^{(1)})\right) \|F_2(\eta^{(1)})\|_{P,2}$$
$$+ \frac{\|M\|_{P,2} J^2\left(\delta_n, \mathcal{F}_2(\eta^{(1)}), F_2(\eta^{(1)})\right)}{\delta_n^2 \sqrt{n}}$$

with $\delta_n = \sigma_n / \|F_2(\eta^{(1)})\|_{P,2}$, $M = \max_{1 \leq i \leq n} F_2(\eta^{(1)})(W_i)$, where $\sup_{f \in \mathcal{F}_2(\eta^{(1)})} \mathbb{E}[f^2] \leq \sigma_n^2 \leq \|F_2(\eta^{(1)})\|_{P,2}^2$. Under Assumption 4(i) and using $\|M\|_{P,q} \leq n^{1/q} \|F_2(\eta^{(1)})\|_{P,q}$ this implies

$$\mathbb{E}\left[\|\mathbb{G}_{n,k}\|_{\mathcal{F}_2(\eta^{(1)})}\right] \lesssim u_n \|F_2(\eta^{(1)})\|_{P,2} + n^{1/q-1/2} V_n u_n^2 \frac{\|F_2(\eta^{(1)})\|_{P,2}^2}{\sigma_n^2}.$$

This enables the use of Theorem 5.1 of Chernozhukov et al. (2014) with $t = \log(n)$ such that for any fixed $\eta^{(1)} \in \widetilde{\mathcal{T}}_N^{(1)}$

$$\|\mathbb{G}_{n,k}\|_{\mathcal{F}_2(\eta^{(1)})} \leq (1+\alpha) \mathbb{E}\left[\|\mathbb{G}_{n,k}\|_{\mathcal{F}_2(\eta^{(1)})}\right]$$
$$+ C(q) \left[\left(\sigma_n + \frac{\|M\|_{P,q}}{\sqrt{n}}\right) \sqrt{\log(n)} + \frac{\|M\|_{P,2} \log(n)}{\alpha \sqrt{n}}\right]$$
$$\lesssim (1+\alpha) \left(u_n \|F_2(\eta^{(1)})\|_{P,2} + n^{1/q-1/2} V_n u_n^2 \frac{\|F_2(\eta^{(1)})\|_{P,2}^2}{\sigma_n^2}\right)$$
$$+ C(q) \left((\sigma_n + n^{1/q-1/2} V_n) \sqrt{\log(n)} + n^{1/q-1/2} V_n \frac{\log(n)}{\alpha}\right)$$

with probability $> 1 - \log(n)^{-q/2}$ for all $\alpha > 0$ and $C(q) > 0$ is a constant only depending on $q$. Consequently

$$\|\mathbb{G}_{n,k}\|_{\mathcal{F}_2(\eta^{(1)})} \lesssim u_n \|F_2(\eta^{(1)})\|_{P,2} + \sigma_n \sqrt{\log(n)}$$

9

$$+ n^{1/q-1/2} V_n \left( u_n^2 \frac{\|F_2(\eta^{(1)})\|_{P,2}^2}{\sigma_n^2} \vee \log(n) \right)$$

$$\leq \tilde{r}_N$$

with probability $> 1 - c \log(n)^{-1}$, where $\tilde{r}_N$ does not depend on $\eta^{(1)}$ due to growth condition 4 (iii). Hence, by Lemma 6.1 in Chernozhukov et al. (2018)

$$\mathcal{I}_{3,k}^{(1)} = O_{P_N}(\tilde{r}_N).$$

This implies that

$$|\tilde{R}_{N,2}| = O_{P_N}(N^{-1/2}(r'_N + \tilde{r}_N) + \lambda_N + \lambda'_N)$$

Next, we show $|\tilde{R}_{N,1}| = |\hat{J}_0 - J_0| = O_{P_N}\left(N^{-1/2} + r_N + \tilde{r}_N^a\right)$. As in Chernozhukov et al. (2018) is suffices to show that for any $k \in [K]$,

$$\left| \mathbb{E}_{n,k} \left[ \psi^a \left( W; (\hat{\eta}_{0,k}^{(1)}, \tilde{\eta}_{0,k}^{(2)}) \right) \right] - \mathbb{E} \left[ \psi^a \left( W; (\eta_0^{(1)}, \eta_{0,k}^{(2)}) \right) \right] \right|$$
$$= O_{P_N}(N^{-1/2} + r_N + \tilde{r}_N).$$

It holds

$$\left| \mathbb{E}_{n,k} \left[ \psi^a \left( W; (\hat{\eta}_{0,k}^{(1)}, \tilde{\eta}_{0,k}^{(2)}) \right) \right] - \mathbb{E} \left[ \psi^a \left( W; (\eta_0^{(1)}, \eta_{0,k}^{(2)}) \right) \right] \right|$$
$$\leq \left| \mathbb{E}_{n,k} \left[ \psi^a \left( W; (\hat{\eta}_{0,k}^{(1)}, \tilde{\eta}_{0,k}^{(2)}) \right) \right] - \mathbb{E}_{n,k} \left[ \psi^a \left( W; (\hat{\eta}_{0,k}^{(1)}, \eta_0^{(2)}) \right) \right] \right|$$
$$+ \left| \mathbb{E}_{n,k} \left[ \psi^a \left( W; (\hat{\eta}_{0,k}^{(1)}, \eta_0^{(2)}) \right) \right] - \mathbb{E} \left[ \psi^a \left( W; (\hat{\eta}_{0,k}^{(1)}, \eta_0^{(2)}) \right) | (W_i)_{i \in I_k^c} \right] \right|$$
$$+ \left| \mathbb{E} \left[ \psi^a \left( W; (\hat{\eta}_{0,k}^{(1)}, \eta_0^{(2)}) \right) | (W_i)_{i \in I_k^c} \right] - \mathbb{E} \left[ \psi^a \left( W; (\eta_0^{(1)}, \eta_{0,k}^{(2)}) \right) \right] \right|$$
$$= \mathcal{I}_{1,k}^{(1)} + \mathcal{I}_{1,k}^{(2)} + \mathcal{I}_{2,k},$$

where $\mathcal{I}_{1,k}^{(2)} = O_{P_N}(N^{-1/2})$ and $\mathcal{I}_{2,k} = O_{P_N}(r_N)$ by the same terms as in Step 2 of the proof of

Theorem 3.1 in Chernozhukov et al. (2018). Next, we show $\mathcal{I}_{1,k}^{(1)} = O_{P_N}(r_N + \tilde{r}_N)$.

$$\mathcal{I}_{1,k}^{(1)} = \left| \mathbb{E}_{n,k} \left[ \psi^a \left( W; (\hat{\eta}_{0,k}^{(1)}, \tilde{\eta}_{0,k}^{(2)}) \right) \right] - \mathbb{E}_{n,k} \left[ \psi^a \left( W; (\hat{\eta}_{0,k}^{(1)}, \eta_0^{(2)}) \right) \right] \right|$$

$$\leq \left| \mathbb{G}_{n,k} \left[ \psi^a \left( W; (\hat{\eta}_{0,k}^{(1)}, \tilde{\eta}_{0,k}^{(2)}) \right) \right] - \mathbb{G}_{n,k} \left[ \psi^a \left( W; (\hat{\eta}_{0,k}^{(1)}, \eta_0^{(2)}) \right) \right] \right|$$

$$+ \left| \mathbb{E} \left[ \psi^a \left( W; (\hat{\eta}_{0,k}^{(1)}, \tilde{\eta}_{0,k}^{(2)}) \right) \right] - \mathbb{E} \left[ \psi^a \left( W; (\hat{\eta}_{0,k}^{(1)}, \eta_0^{(2)}) \right) \right] \right|$$

At first, remark

$$\left| \mathbb{E} \left[ \psi^a \left( W; (\hat{\eta}_{0,k}^{(1)}, \tilde{\eta}_{0,k}^{(2)}) \right) \right] - \mathbb{E} \left[ \psi^a \left( W; (\hat{\eta}_{0,k}^{(1)}, \eta_0^{(2)}) \right) \right] \right|$$

$$\leq \left| \mathbb{E} \left[ \psi^a \left( W; (\hat{\eta}_{0,k}^{(1)}, \tilde{\eta}_{0,k}^{(2)}) \right) \right] - \mathbb{E} \left[ \psi^a \left( W; (\eta_0^{(1)}, \eta_0^{(2)}) \right) \right] \right|$$

$$+ \left| \mathbb{E} \left[ \psi^a \left( W; (\eta_0^{(1)}, \eta_0^{(2)}) \right) \right] - \mathbb{E} \left[ \psi^a \left( W; (\hat{\eta}_{0,k}^{(1)}, \eta_0^{(2)}) \right) \right] \right|$$

$$\lesssim r_N + \tilde{r}_N^a$$

by Assumption 4 (ii). Further, with the same argument as above

$$\mathbb{E} \left[ \left| \mathbb{G}_{n,k} \left[ \psi^a \left( W; (\hat{\eta}_{0,k}^{(1)}, \tilde{\eta}_{0,k}^{(2)}) \right) - \psi^a \left( W; (\hat{\eta}_{0,k}^{(1)}, \eta_0^{(2)}) \right) \right] \right| \Big| (W_i)_{i \in I_k^c} \right]$$

$$\leq \sup_{\eta^{(1)} \in \tilde{\mathcal{T}}_N^{(1)}} \| \mathbb{G}_{n,k} \|_{\mathcal{F}_2^a(\eta^{(1)})}$$

with

$$\mathcal{F}_2^a(\eta^{(1)}) := \left\{ \psi^a \left( \cdot; (\eta^{(1)}, \eta^{(2)}) \right) - \psi^a \left( \cdot; (\eta^{(1)}, \eta_0^{(2)}) \right) \mid (\eta^{(1)}, \eta^{(2)}) \in \tilde{\mathcal{T}}_N \right\}.$$

Following the same arguments as above, combined with Assumption 4 (ii), we obtain

$$\| \mathbb{G}_{n,k} \|_{\mathcal{F}_2^a(\eta^{(1)})} \lesssim u_{n,a} \| F_2^a(\eta^{(1)}) \|_{P,2} + \sigma_{n,a} \sqrt{\log(n)}$$

$$+ n^{1/q-1/2} V_{n,a} \left( u_{n,a}^2 \frac{\| F_2^a(\eta^{(1)}) \|_{P,2}^2}{\sigma_{n,a}^2} \vee \log(n) \right)$$

$$\leq \tilde{r}_N^a.$$

Again, by Lemma 6.1 in Chernozhukov et al. (2018)

$$\mathcal{I}_{1,k}^{(1)} = O_{P_N}(r_N + \tilde{r}_N^a)$$

and therefore

$$|\tilde{R}_{N,1}| = O_{P_N}\left(N^{-1/2} + r_N + \tilde{r}_N^a\right).$$

$\square$

**Proof of Theorem 2.**

The first part is an application of Theorem 5.1 Chernozhukov et al. (2018) with nuisance realization set $\widetilde{\mathcal{T}}_N$.

The second part follows by verifying Assumption 4. At first remark that Assumption 4(ii) holds since $\psi^a(W; \eta) = -1$ for all $\eta \in T$. Further, it holds

$$\mathbb{E}\left[\left(\psi\left(W; \theta_0, (\eta^{(1)}, \eta^{(2)})\right) - \psi\left(W; \theta_0, (\eta^{(1)}, \eta_0^{(2)})\right)\right)^2\right]^{1/2}$$
$$= \left\|\psi\left(W; \theta_0, (\eta^{(1)}, \eta^{(2)})\right) - \psi\left(W; \theta_0, (\eta^{(1)}, \eta_0^{(2)})\right)\right\|_{P,2}$$
$$\leq \mathcal{I}_1 + \mathcal{I}_2$$

with

$$\mathcal{I}_1 := \left\|\left(\tilde{m}(X)^{-1} - m_0(X)^{-1}\right) D(Y - g(1, X))\right\|_{P,2}$$
$$\mathcal{I}_2 := \left\|\left((1 - \tilde{m}(X))^{-1} - (1 - m_0(X))^{-1}\right) (1 - D)(Y - g(0, X))\right\|_{P,2}.$$

Remark that $\epsilon \leq m_0(X) \leq 1 - \epsilon$ and $\epsilon \leq \tilde{m}(X) \leq 1 - \epsilon$, such that

$$\mathcal{I}_1 \leq \epsilon^{-2} \|(m_0(X) - \tilde{m}(X))D(Y - g(1, X))\|_{P,2}$$
$$\leq \epsilon^{-2} \|(m_0(X) - \tilde{m}(X))(g_0(1, X) + U - g(1, X))\|_{P,2}$$

12

$$\leq \epsilon^{-2} \left\| (m_0(X) - \tilde{m}(X))(g_0(1,X) - g(1,X)) \right\|_{P,2} + \epsilon^{-2} \left\| (m_0(X) - \tilde{m}(X))U \right\|_{P,2}$$

$$\lesssim \|m_0(X) - \tilde{m}(X)\|_{P,2}$$

since $\mathbb{E}[U^2|X] \leq C$ and $\|g - g_0\|_{P,\infty} \leq C$. With a analogous argument it holds

$$\mathcal{I}_2 \lesssim \|m_0(X) - \tilde{m}(X)\|_{P,2}.$$

This implies

$$\sup_{f \in \mathcal{F}_2(\eta^{(1)})} \|f\|_{P,2}^2$$
$$\leq \sup_{(\eta^{(1)}, \eta^{(2)}) \in \widetilde{\mathcal{T}}_N} \mathbb{E}\left[\left(\psi\left(W; \theta_0, (\eta^{(1)}, \eta^{(2)})\right) - \psi\left(W; \theta_0, (\eta^{(1)}, \eta_0^{(2)})\right)\right)^2\right]$$
$$\leq C \|m_0(X) - \tilde{m}(X)\|_{P,2}^2$$
$$\lesssim \epsilon_N^2.$$

Further, remark that using the same argument as above

$$\left|\psi\left(W; \theta_0, (\eta^{(1)}, \eta^{(2)})\right) - \psi\left(W; \theta_0, (\eta^{(1)}, \eta_0^{(2)})\right)\right|$$
$$\lesssim |m_0(X) - \tilde{m}(X)||D(Y - g(1,X)) + (1-D)(Y - g(0,X))|$$
$$\leq L(W)|m_0(X) - \tilde{m}(X)|$$

where

$$\sup_{\eta^{(1)} \in \widetilde{\mathcal{T}}_N^{(1)}} \|L(W)\|_{P,q} \lesssim \|U\|_{P,q} + \sup_{\eta^{(1)} \in \widetilde{\mathcal{T}}_N^{(1)}} \|g_0(D,X) - g(D,X)\|_{P,q} \lesssim 1.$$

Applying Lemma O.1 from Belloni et al. (2018) implies

$$\log \sup_Q N(\epsilon \|L(W)\|_{Q,2}, \mathcal{F}_2(\eta^{(1)}), \|\cdot\|_{Q,2})$$

$$\leq \log \sup_Q N(\epsilon, \mathcal{M}, \|\cdot\|_{Q,2})$$

$$\leq C\epsilon^{-1}$$

by Assumption 6 (ii), where we can use 1 as an envelope for $\mathcal{M}$. According to the discussion of Assumption 4, it holds

$$u_n \lesssim \sigma_n^{1/2} \|F_2(\eta^{(1)})\|_{P,2}^{-1/2}.$$

Further,

$$\|F_2(\eta^{(1)})\|_{P,q} \leq \left\| \sup_{\eta^{(2)} \in \widetilde{\mathcal{T}}_N^{(2)}} \left( \psi\left(W; \theta_0, (\eta^{(1)}, \eta^{(2)})\right) - \psi\left(W; \theta_0, (\eta^{(1)}, \eta_0^{(2)})\right) \right) \right\|_{P,q}$$

$$= \left\| L(W) \sup_{\eta^{(2)} \in \widetilde{\mathcal{T}}_N^{(2)}} |\eta^{(2)}(X) - \eta_0^{(2)}(X)| \right\|_{P,q} \lesssim 1.$$

Choose $\sigma_n = \epsilon_N \vee \log^{1/2}(n) n^{1/q-1/2}$, such that

$$\sup_{f \in \mathcal{F}_2(\eta^{(1)})} \|f\|_{P,2}^2 \leq \sigma_n^2 \leq \|F_2(\eta^{(1)})\|_{P,2}^2$$

for each $\eta^{(1)}$ and $n$ large enough. Consequently,

$$\sigma_n^{1/2} \sup_{\eta^{(1)} \in \widetilde{\mathcal{T}}_N^{(1)}} \|F_2(\eta^{(1)})\|_{P,2}^{1/2} + \sigma_n \sqrt{\log(n)}$$

$$+ n^{1/q-1/2} V_n \left( \frac{\sup_{\eta^{(1)} \in \widetilde{\mathcal{T}}_N^{(1)}} \|F_2(\eta^{(1)})\|_{P,2}}{\sigma_n} \vee \log(n) \right)$$

$$\leq C\sigma_n + \sigma_n \sqrt{\log(n)} + Cn^{1/q-1/2} \left( \sigma_n^{-1} \vee \log(n) \right) = o(1).$$

$\square$

# 3 Details on the DGPs

Table 1: Overview of the DGPs

| DGP | Covariates | Description of $D \mid X$ | Description of $g_0(D, X)$ |
|---|---|---|---|
| 1 | $X_i \sim N(0, \Sigma)$ with $\Sigma_{kj} = (0.5)^{\|j-k\|}$ | Sigmoid function combined with Uniform indicator. $D$ is exogenous conditional on $X$. | Heterogeneous treatment effect, linear with simple interaction, $\epsilon_i \stackrel{iid}{\sim} \mathcal{N}(0,1)$ |
| 2 | $X_1 \sim$ Bin$(1, 0.5)$, $X_2 \mid X_1 \sim \Gamma$, $X_3 \mid X_2 \sim$ Beta | Highly nonlinear, tree-based via conditions, $D \sim$ Bin$(1, m_0(X))$ | Homogeneous treatment effect, Poisson with simple linear combination of covariates |
| 3 | $X \sim$ Unif$[-1,1]^4$ | Expit function combined with $D \sim$ Bin$(1, m_0(X))$ | Heterogeneous treatment effect, nonlinear transformations of the covariates, $\epsilon_i \stackrel{iid}{\sim} \mathcal{N}(0,1)$ |
| 4 | $X \sim$ Unif$[0,1]^{20}$ | Unbalanced and nonlinear $m_0(X) = \alpha \left(1 + \beta_{2,4}(\min(X_1, X_2))\right)$, $D \sim$ Bin$(1, m_0(X))$ | Heterogeneous treatment effect, scaled Friedman function for $g_0(X)$ and simple interaction of $D$ and $X$, $\epsilon_i \stackrel{iid}{\sim} \mathcal{N}(0,1)$ |

A detailed description of DGP 2-4 is given in the Supplementary Material.

In our first DGP, data units were generated following Belloni et al. (2017)[1].

DGP 2 is adapted from Deshpande and Kuleshov (2023). Three covariates are simulated: gender ($X_1$), age ($X_2$), and disease severity ($X_3$), while treatment ($D$) corresponds to the administration of a drug. The outcome $Y$ is the time taken for the recovery of a patient. In addition to the original setting, we introduced non-deterministic treatment assignments to ensure positivity. The function for the treatment assignment resembles the approach of tree-based models, favoring the IRM and PLR in combination with LGBM and random forest.

For a complicated nonlinear outcome regression, we implemented DGP 3 (van der Laan et al., 2023). The included transformations make it difficult for linear models as well as tree-based classifiers through the local linearity. Given the nonlinear and heterogeneous influence of the treatment on the outcome, we expect the IRM model to outperform the PLR. However, the treatment assignment itself can be modeled consistently, favoring the IPW.

The unbalanced treatment assignment DGP 4 follows Ballinari (2024) and is adapted from Nie and Wager (2020). Three settings of the share of treated $\mathbb{E}[D]$ are tested: $\mathbb{E}[D] \in$

---
[1]This DGP is available at DoubleMLIRMData.

$\{0.05, 0.1, 0.2\}$. To address positivity violations, we tested three clipping thresholds for the propensity scores: $1e - 12$, $0.01$, and $0.1$. Given the majority non-treatment class, a lower threshold is sufficient to limit the contribution of a single observation.

**DGP 1 IRM, Detailed Description**

$$d_i = 1\left\{\frac{\exp(c_D X_i'\beta)}{1 + \exp(c_D X_i'\beta)} > V_i\right\}, \qquad V_i \sim \mathcal{U}(0,1),$$

$$y_i = \theta D_i + c_Y X_i'\beta D_i + \zeta_i, \qquad \zeta_i \sim \mathcal{N}(0,1),$$

where $v_i \sim U(0,1)$, $\zeta_i \sim N(0,1)$, $V_i$ and $\zeta_i$ are independent, $p = \dim(X_i)$, the covariates $X_i \sim N(0, \Sigma)$ with $\Sigma_{kj} = (0.5)^{|j-k|}$. $\beta$ is a $p \times 1$ vector with elements set as $\beta_{1,j} = (1/j)^2$ for $j = 1, ..., p$. $c_D$ and $c_Y$ are scalars given by

$$c_Y = \sqrt{\frac{R_Y^2}{(1-R_Y^2)\beta'\Sigma\beta}}, \qquad c_D = \sqrt{\frac{(\pi^2/3)R_D^2}{(1-R_D^2)\beta'\Sigma\beta}}.$$

that control the strength of the relationship between the controls, the outcome, and the treatment variable. The underlying treatment effect is heterogeneous. Hence, to accurately model the interaction effects between treatment and covariates, the IRM should be considered. Simple propensity score weighting in the IPW, as well as the additive structural form assumptions for the treatment effect of the PLR, should be inconsistent for this DGP.

**DGP 2 Drug Effectiveness, Detailed Description**

In our second DGP, data units were generated as follows:

- Gender indicator $X_1$ and age $X_2$ are simulated as:

$$X_1 \sim \text{Bin}(0, 0.5), \quad X_2 \sim \Gamma(\kappa_{\text{age}}, \psi_{\text{age}}).$$

The distribution of age is gender dependent:

$$\kappa_{\text{age}} = \left(\frac{\mu_{\text{age}}}{\sigma_{\text{age}}}\right)^2, \quad \psi_{\text{age}} = \frac{\sigma_{\text{age}}^2}{\mu_{\text{age}}}, \text{ and}$$

$$\mu_{\text{age}} = \begin{cases} 49 & \text{if } X_1 = 1, \\ 51 & \text{if } X_1 = 0, \end{cases} \quad \sigma_{\text{age}} = \begin{cases} 7 & \text{if } X_1 = 1, \\ 8 & \text{if } X_1 = 0. \end{cases}$$

- Disease severity $X_3$ is simulated based on age:

$$\mu_{\text{disease}} = \text{clip}\left(\frac{X_2 - 20}{20}, 0.1, 5\right), \quad X_3 \sim \text{Beta}(\mu_{\text{disease}}, 2).$$

- Propensity scores are generated using linear predictors, incorporating adjustable levels of overlap through the parameter $\beta$:

$$\text{coefficients} = \begin{bmatrix} (-0.4, 0.2, 0.8), \\ (-0.4 + 2(1 - \beta), 0.2, 0.8), \\ (-0.4 + 2(1 - \beta), 0.3, 1), \\ (-0.4 + 2(1 - \beta), 0.1, 1.2), \\ (-0.4 + 2(1 - \beta), 0.1, 1.2) \end{bmatrix},$$

$$\text{linear\_predictors} = \Big(b_0 + b_1 \cdot \text{normalize}(X_2) + b_2 \cdot \text{normalize}(X_3)$$
$$+ \epsilon \mid (b_0, b_1, b_2) \in \text{coefficients}\Big),$$
$$\epsilon \sim \mathcal{N}(0, 0.5).$$

Propensity scores are computed via logistic transformation:

$$\text{prop\_scores} = \left[\frac{1}{1 + \exp(-\text{lp})} \mid \text{lp} \in \text{linear\_predictors}\right].$$

- Treatment assignment $m_0$ is initialized with prop_scores[0] and updated based on conditions:

$$\text{conditions} = \begin{cases} ((X_1 = 0) \text{ and } (X_2 > 55) \text{ and } (X_3 \leq 0.55), \text{prop\_scores}[1]), \\ ((X_1 = 1) \text{ and } (X_2 > 55) \text{ and } (X_3 \leq 0.55), \text{prop\_scores}[2]), \\ ((X_1 = 0) \text{ and } (X_3 > 0.55), \text{prop\_scores}[3]), \\ ((X_1 = 1) \text{ and } (X_3 > 0.55), \text{prop\_scores}[4]) \end{cases}$$

for (condition, score) $\in$ conditions : $m_0$[condition] = score[condition].

Final treatment assignment $D_i$ is:

$$D_i \sim \text{Bin}(1, m_0).$$

- The outcome $Y_i$ is simulated as:

$$Y_{di} = \text{Pois}(2 + 0.5X_1 + 0.03X_2 + 2X_3 - d), \qquad Y_i = D_i \cdot Y_{1i} + (1 - D_i) \cdot Y_{0i}.$$

This DGP captures complex relationships between gender, age, disease severity, and treatment assignment, providing a realistic framework for evaluating causal inference methods. The treatment assignments are highly nonlinear and resemble decision trees with underlying local linearities.

**DGP 3 Nonlinear, Detailed Description**

For the nonlinear outcome regression, the detailed data generation process is as follows:

$$X \sim \text{Unif}[-1, 1]^4, \quad D|X \sim \text{Bin}(1, p(X)), \quad Y \sim \text{Bin}(1, \mu(X, D))$$

with the treatment assignment probability given by:

$$p(X) = \text{expit}\{-0.25 - X_1 + 0.5X_2 - X_3 + 0.5X_4\}.$$

The outcome function is based on nonlinear transformations of the covariates, with added treatment effect interactions.

$$\mu(X, D) = 1.5 + 1.5D + 2D|X_1||X_2| - 2.5(1 - D)|X_2|X_3 + 2.5X_3$$
$$+ 2.5(1 - D)\sqrt{|X_4|} - 1.5D \cdot I(X_2 < 0.5) + 1.5(1 - D)I(X_4 < 0).$$

The outcome is then generated by:

$$Y_d \sim \text{N}(\mu(X, D), 1), \quad Y = DY_1 + (1 - D)Y_0.$$

**DGP 4 Unbalanced, Detailed Description**

The unbalanced treatment assignment setting follows Ballinari (2024) and is adapted from Nie and Wager (2020).

$$X_i \sim \text{Unif}[0, 1]^{20}, \quad D|X \sim \text{Bin}(1, p(X)), \quad \epsilon_i \stackrel{\text{iid}}{\sim} \mathcal{N}(0, 1),$$
$$Y = b(X) + (D - 0.5)(X_1 + X_2) + \epsilon_i.$$

The baseline main effect is the scaled Friedman (1991) function:

$$b(X_i) = \sin(\pi X_1 X_2) + 2(X_3 - 0.5)^2 + X_4 + 0.5X_5.$$

For the propensity score, we follow Künzel et al. (2019) and set:

$$m_0(X) = \alpha \left(1 + \beta_{2,4}(\min(X_1, X_2))\right),$$

where $\beta_{2,4}(\cdot)$ is the beta cumulative distribution function with shape parameters 2 and 4. This unbalanced setting entails difficult nuisance components and an easy treatment effect function, favoring doubly robust causal models compared to IPW.

The share of treated is $\mathbb{E}[D] = (31/21)\alpha$. Three settings of the share of treated $\mathbb{E}[D]$ are tested: $\mathbb{E}[D] \in \{0.05, 0.1, 0.2\}$. For unbalanced applications, several adjustments for ATE estimators exist to address positivity violations. The most commonly used method is clipping the propensity scores to preset levels. Throughout our simulations, we tested three different levels of clipping thresholds.

As Imbens (2004) points out, for the IPTW with $n = 4000$ observations, choosing a clipping cut-off at $0.01^2$ limits the contribution of a single observation to 0.025. We also tested more severe clipping as employed in Nie and Wager (2020), with a clipping level at 0.1, and alternatively, without clipping [3]. Particularly, for the setting with only 5% treated, we expect that a clipping threshold of 0.1 introduces a strong bias.

---

[2]Given the majority non-treatment class, a lower threshold is sufficient.
[3]To ensure computational stability we set the threshold at $1e - 12$.

# 4 Detailed Simulation Results

## 4.1 Desirable Properties

Modern causal inference methods using weighting estimators address two core challenges: (1) achieving covariate balance, where the weighted covariate distributions satisfy $\mathbb{E}[w\mathbf{X} \mid D = 1] = \mathbb{E}[w\mathbf{X} \mid D = 0]$ for treatment $D \in \{0,1\}$ and covariates $\mathbf{X}$, and (2) ensuring normalization $\sum_{i:D_i=1} w_i = 1$ and $\sum_{i:D_i=0} w_i = 1$ to stabilize weights (Busso et al., 2014). Covariate balance is quantified via standardized mean differences (SMD):

$$\text{SMD}_k = \frac{\bar{X}^w_{k,D=1} - \bar{X}^w_{k,D=0}}{\sqrt{(s^w_{k,D=1})^2 + (s^w_{k,D=0})^2/2}},$$

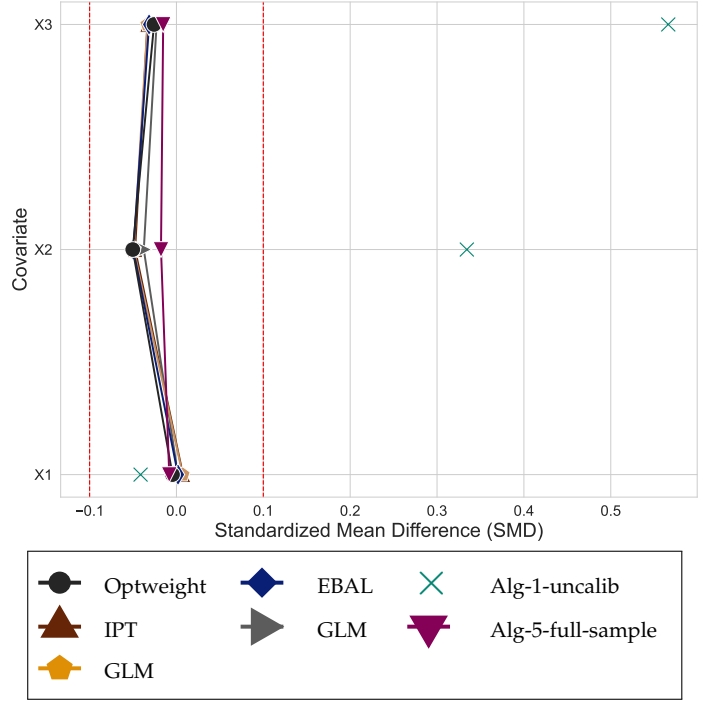

Figure S1: SMD across covariates for DGP 2. Dashed line indicates $|\text{SMD}| = 0.1$.

where $\bar{X}^w_{k,D}$ and $s^w_{k,D}$, with $D \in \{0,1\}$, are the weighted means and standard deviations of $X_k$ in the treated and control groups, respectively. The SMD expresses an imbalance in standard deviation units, with $|\text{SMD}_k| < 0.1$ indicating an adequate balance (Austin and Stuart, 2015). Figure S1 demonstrates SMD reduction across methods. Entropy balancing (EBAL) minimizes KL/ Rényi divergence from uniform weights $q_i = 1/n$ under balance constraints $\sum_{i:D_i=0} w_i \phi(\mathbf{X}_i) = \frac{1}{n_1} \sum_{i:D_i=1} \phi(\mathbf{X}_i)$ (Hainmueller, 2012). OptimWeight solves $\min_w \sum_{i:D_i=0}(w_i - 1/n_0)^2$ with $\ell_\infty$-norm balance constraints $\|\frac{1}{n_1} \sum_{i:D_i=1} \mathbf{X}_i - \sum_{i:D_i=0} w_i \mathbf{X}_i\|_\infty \leq \delta$ (Zubizarreta, 2015). Covariate balancing propensity score (CBPS) estimates the propensity score $m(\mathbf{X}_i; \beta) = \text{expit}(\mathbf{X}_i^T \beta)$ via GMM, combining score equations $\sum_{i=1}^n [D_i - m(\mathbf{X}_i; \beta)]\mathbf{X}_i = 0$ with balancing moments Imai and Ratkovic (2014). Inverse probability tilting (IPT) solves

dual-moment conditions for normalization $\sum D_i/m(\mathbf{X}_i;\theta) = n$ and balance $\sum D_i\mathbf{X}_i/m(\mathbf{X}_i;\theta)$ $= \sum(1-D_i)\mathbf{X}_i/(1-m(\mathbf{X}_i;\theta))$ (Graham et al., 2012). GLM uses $m(\mathbf{X}_i) = \text{expit}(\mathbf{X}_i^T\hat{\beta}_{\text{MLE}})$ with normalized IPW weights. To ensure a fair comparison, only first-order moments were rebalanced in the setup of the outcome equation for the `weightit` approaches. Similarly, Alg-3-cf and Alg-5-full-sample used a linear outcome function without further fine-tuning, as excessive fine-tuning could bias the comparison. This means the outcome equation is misspecified for all methods. Figure S1 shows that methods from the R package `weightit` (Greifer, 2025) (EBAL, OptWeight, CBPS, GLM, IPT) as well as the full-sample calibrated IRM model achieve covariate balance, unlike the unadjusted IRM model. Table 2 compares performance metrics and normalization.

Table 2: Comparison under covariate balance

| Method | | m | Coverage | CI Length | Norm $D=1$ | Norm $D=0$ | RMSE | Std. dev. | MAE |
|---|---|---|---|---|---|---|---|---|---|
| Alg-3-cf | isotonic | Logit | 0.960 | 0.413 | 0.996 | 0.987 | 0.098 | 0.097 | 0.077 |
| Alg-3-cf | isotonic | RF | 0.960 | 0.389 | 0.997 | 0.994 | 0.096 | 0.094 | 0.075 |
| Alg-3-cf | platt | Logit | 0.960 | 0.388 | 1.003 | 0.984 | 0.094 | 0.093 | 0.075 |
| Alg-3-cf | platt | RF | 0.940 | 0.358 | 0.984 | 0.961 | 0.092 | 0.091 | 0.072 |
| Alg-5-full-sample | isotonic | Logit | 0.970 | 0.404 | 0.999 | 0.998 | 0.097 | 0.095 | 0.077 |
| Alg-5-full-sample | isotonic | RF | 0.940 | 0.386 | 1.000 | 1.000 | 0.095 | 0.094 | 0.075 |
| Alg-5-full-sample | platt | Logit | 0.960 | 0.397 | 1.003 | 0.996 | 0.094 | 0.093 | 0.075 |
| Alg-5-full-sample | platt | RF | 0.950 | 0.373 | 0.998 | 0.988 | 0.094 | 0.093 | 0.074 |
| Cbps | weighted | Logit | 0.950 | 0.371 | 1.000 | 1.000 | 0.094 | 0.093 | 0.074 |
| Ebal | weighted | - | 0.950 | 0.376 | 1.000 | 1.000 | 0.094 | 0.093 | 0.074 |
| Glm | weighted | Logit | 0.950 | 0.371 | 1.000 | 1.000 | 0.094 | 0.093 | 0.075 |
| Ipt | weighted | Logit | 0.950 | 0.371 | 1.000 | 1.000 | 0.094 | 0.093 | 0.075 |
| Optweight | weighted | - | 0.960 | 0.376 | 1.000 | 1.000 | 0.093 | 0.092 | 0.074 |

DGP 2: n = 2000, p = 3, overlap = 0.5, Clip = 1e-12 g = Linear

All `weightit` methods enforce exact normalization via $\mathbb{E}[D/m(\mathbf{X})] = 1$ and $\mathbb{E}[(1-D)/(1-m(\mathbf{X}))] = 1$ (columns 'Norm D = 1' and 'Norm D = 0'), ensuring weight stability. Full-sample calibration achieves higher levels of normalization compared to their cross-fitted counterparts. However, the remaining small deviations from optimally normalized weights do not deteriorate performance, as seen in Table 2.

## 4.2 Extended Results Overview

Table 3: IPW Results 2

| DGP | Method | m = Logit | | | m = Random Forest | | | m = LGBM | | |
|---|---|---|---|---|---|---|---|---|---|---|
| | | MAE | RMSE | Std. dev. | MAE | RMSE | Std. dev. | MAE | RMSE | Std. dev. |
| 1 | Alg-1-Clipped | 0.08 | 0.11 | 0.11 | 0.18 | 0.19 | 0.06 | 0.94 | 0.98 | 0.31 |
| | Alg-1-Uncalib | 0.10 | 0.16 | 0.16 | 1.09e+06 | 1.09e+07 | 1.09e+07 | 1.57 | 1.77 | 0.83 |
| | Alg-2-nested-cf-IVAP | 0.10 | 0.12 | 0.09 | 0.14 | 0.15 | 0.07 | 0.17 | 0.19 | 0.08 |
| | Alg-2-nested-cf-Iso | 0.20 | 0.25 | 0.19 | 0.13 | 0.17 | 0.16 | 0.13 | **0.16** | 0.16 |
| | Alg-2-nested-cf-Platt | 0.10 | 0.17 | 0.17 | 0.14 | 0.15 | 0.06 | 0.22 | 0.23 | 0.06 |
| | Alg-3-cf-IVAP | 0.11 | 0.13 | 0.07 | 0.14 | 0.16 | 0.07 | 0.17 | 0.18 | 0.07 |
| | Alg-3-cf-Iso | 0.09 | 0.11 | 0.07 | 0.13 | 0.15 | 0.06 | 0.15 | 0.17 | 0.06 |
| | Alg-3-cf-Platt | 0.13 | 0.14 | 0.06 | 0.35 | 0.36 | 0.05 | 0.31 | 0.31 | 0.05 |
| | Alg-4-single-split-IVAP | 0.11 | 0.12 | 0.07 | 0.31 | 0.31 | 0.03 | 0.33 | 0.33 | 0.04 |
| | Alg-4-single-split-Iso | 0.13 | 0.18 | 0.16 | 0.30 | 0.30 | 0.03 | 0.32 | 0.32 | 0.04 |
| | Alg-4-single-split-Platt | 0.10 | 0.12 | 0.09 | 0.16 | 0.17 | 0.06 | 0.23 | 0.24 | 0.06 |
| | Alg-5-full-sample-IVAP | 0.08 | 0.10 | 0.08 | 0.12 | **0.14** | 0.07 | 0.14 | **0.16** | 0.07 |
| | Alg-5-full-sample-Iso | 0.07 | **0.09** | 0.08 | 0.12 | **0.14** | 0.06 | 0.14 | **0.16** | 0.07 |
| | Alg-5-full-sample-Platt | 0.09 | 0.11 | 0.10 | 0.14 | 0.16 | 0.06 | 0.22 | 0.22 | 0.05 |
| 2 | Alg-1-Clipped | 0.09 | 0.12 | 0.11 | 4.45 | 4.59 | 1.11 | 2.67 | 2.74 | 0.61 |
| | Alg-1-Uncalib | 0.09 | 0.12 | 0.11 | 1.17e+10 | 1.37e+10 | 7.23e+09 | 2.86 | 2.98 | 0.81 |
| | Alg-2-nested-cf-IVAP | 0.21 | 0.26 | 0.26 | 0.28 | 0.33 | 0.26 | 0.29 | 0.33 | 0.25 |
| | Alg-2-nested-cf-Iso | 0.60 | 0.81 | 0.66 | 0.47 | 0.61 | 0.61 | 0.48 | 0.64 | 0.62 |
| | Alg-2-nested-cf-Platt | 0.20 | 0.25 | 0.25 | 0.29 | 0.34 | 0.24 | 0.29 | 0.34 | 0.23 |
| | Alg-3-cf-IVAP | 0.10 | 0.12 | 0.12 | 0.21 | 0.24 | 0.12 | 0.20 | 0.23 | 0.11 |
| | Alg-3-cf-Iso | 0.09 | **0.11** | 0.11 | 0.21 | 0.24 | 0.11 | 0.20 | 0.23 | 0.11 |
| | Alg-3-cf-Platt | 0.13 | 0.16 | 0.11 | 0.44 | 0.45 | 0.11 | 0.50 | 0.51 | 0.10 |
| | Alg-4-single-split-IVAP | 0.11 | 0.14 | 0.14 | 0.96 | 0.97 | 0.14 | 0.90 | 0.91 | 0.11 |
| | Alg-4-single-split-Iso | 0.61 | 0.77 | 0.53 | 0.95 | 0.96 | 0.14 | 0.87 | 0.88 | 0.11 |
| | Alg-4-single-split-Platt | 0.10 | 0.12 | 0.11 | 0.26 | 0.28 | 0.12 | 0.27 | 0.29 | 0.11 |
| | Alg-5-full-sample-IVAP | 0.09 | **0.11** | 0.11 | 0.20 | **0.22** | 0.11 | 0.18 | **0.20** | 0.11 |
| | Alg-5-full-sample-Iso | 0.09 | **0.11** | 0.11 | 0.20 | 0.23 | 0.10 | 0.18 | **0.20** | 0.10 |
| | Alg-5-full-sample-Platt | 0.10 | 0.12 | 0.11 | 0.27 | 0.29 | 0.10 | 0.27 | 0.29 | 0.10 |
| 3 | Alg-1-Clipped | 0.06 | **0.08** | 0.08 | 0.54 | 0.58 | 0.21 | 2.00 | 2.02 | 0.31 |
| | Alg-1-Uncalib | 0.06 | **0.08** | 0.08 | 3.07e+08 | 8.49e+08 | 7.91e+08 | 2.05 | 2.07 | 0.34 |
| | Alg-2-nested-cf-IVAP | 0.12 | 0.15 | 0.14 | 0.36 | 0.38 | 0.11 | 0.46 | 0.48 | 0.11 |
| | Alg-2-nested-cf-Iso | 0.94 | 1.03 | 0.41 | 0.26 | **0.32** | 0.31 | 0.27 | **0.34** | 0.34 |
| | Alg-2-nested-cf-Platt | 0.11 | 0.14 | 0.14 | 0.40 | 0.42 | 0.10 | 0.53 | 0.54 | 0.10 |
| | Alg-3-cf-IVAP | 0.07 | **0.08** | 0.08 | 0.32 | 0.33 | 0.08 | 0.38 | 0.38 | 0.07 |
| | Alg-3-cf-Iso | 0.07 | 0.09 | 0.08 | 0.36 | 0.37 | 0.07 | 0.42 | 0.42 | 0.07 |
| | Alg-3-cf-Platt | 0.12 | 0.14 | 0.07 | 0.87 | 0.87 | 0.08 | 0.82 | 0.82 | 0.08 |
| | Alg-4-single-split-IVAP | 0.07 | 0.09 | 0.09 | 1.13 | 1.13 | 0.07 | 1.03 | 1.03 | 0.08 |
| | Alg-4-single-split-Iso | 0.73 | 0.80 | 0.34 | 1.13 | 1.13 | 0.07 | 1.03 | 1.03 | 0.08 |
| | Alg-4-single-split-Platt | 0.08 | 0.09 | 0.08 | 0.44 | 0.45 | 0.08 | 0.57 | 0.57 | 0.08 |
| | Alg-5-full-sample-IVAP | 0.06 | **0.08** | 0.08 | 0.35 | 0.36 | 0.07 | 0.40 | 0.41 | 0.07 |
| | Alg-5-full-sample-Iso | 0.06 | **0.08** | 0.08 | 0.37 | 0.37 | 0.07 | 0.42 | 0.43 | 0.07 |
| | Alg-5-full-sample-Platt | 0.07 | 0.09 | 0.08 | 0.42 | 0.43 | 0.07 | 0.52 | 0.53 | 0.07 |
| 4 | Alg-1-Clipped | 0.13 | 0.15 | 0.06 | 0.54 | 0.55 | 0.13 | 6.14 | 6.17 | 0.52 |
| | Alg-1-Uncalib | 0.13 | 0.15 | 0.06 | 4.64e+08 | 6.89e+08 | 5.11e+08 | 8.37 | 8.45 | 1.21 |
| | Alg-2-nested-cf-IVAP | 0.10 | 0.13 | 0.11 | 0.12 | 0.15 | 0.11 | 0.11 | 0.14 | 0.11 |
| | Alg-2-nested-cf-Iso | 0.21 | 0.24 | 0.13 | 0.19 | 0.23 | 0.13 | 0.25 | 0.29 | 0.15 |
| | Alg-2-nested-cf-Platt | 0.13 | 0.16 | 0.11 | 0.15 | 0.18 | 0.11 | 0.15 | 0.18 | 0.12 |
| | Alg-3-cf-IVAP | 0.05 | **0.05** | 0.05 | 0.06 | **0.08** | 0.05 | 0.06 | **0.07** | 0.06 |
| | Alg-3-cf-Iso | 0.05 | 0.06 | 0.05 | 0.09 | 0.10 | 0.05 | 0.07 | 0.09 | 0.06 |
| | Alg-3-cf-Platt | 0.07 | 0.09 | 0.06 | 0.13 | 0.16 | 0.10 | 0.18 | 0.23 | 0.16 |
| | Alg-4-single-split-IVAP | 0.05 | 0.06 | 0.06 | 1.40 | 1.41 | 0.09 | 1.43 | 1.43 | 0.07 |
| | Alg-4-single-split-Iso | 0.07 | 0.09 | 0.07 | 0.88 | 0.98 | 0.43 | 0.96 | 1.02 | 0.37 |
| | Alg-4-single-split-Platt | 0.10 | 0.12 | 0.06 | 0.12 | 0.13 | 0.06 | 0.12 | 0.13 | 0.05 |
| | Alg-5-full-sample-IVAP | 0.06 | 0.08 | 0.06 | 0.10 | 0.11 | 0.05 | 0.09 | 0.10 | 0.06 |
| | Alg-5-full-sample-Iso | 0.07 | 0.08 | 0.05 | 0.11 | 0.12 | 0.05 | 0.10 | 0.11 | 0.06 |
| | Alg-5-full-sample-Platt | 0.09 | 0.11 | 0.06 | 0.12 | 0.13 | 0.06 | 0.12 | 0.13 | 0.06 |

DGP 1: n = 2000, p = 20, R2_d = 0.5; DGP 2: n = 2000, p = 3, overlap = 0.5;
DGP 3: n = 2000, p = 4; DGP 4: n = 4000, p = 20, share treated = 0.1; g = LGBM,
lowest RMSEs per DGP and propensity learner are highlighted

Table 4: IRM Results

| DGP | Method | m = Logit | | | m = Random Forest | | | m = LGBM | | |
|---|---|---|---|---|---|---|---|---|---|---|
| | | MAE | RMSE | Std. dev. | MAE | RMSE | Std. dev. | MAE | RMSE | Std. dev. |
| 1 | Alg-1-Clipped | 0.07 | 0.10 | 0.10 | 0.06 | 0.08 | 0.07 | 0.22 | 0.27 | 0.24 |
| | Alg-1-Uncalib | 0.08 | 0.13 | 0.13 | 1.85e+06 | 1.85e+07 | 1.84e+07 | 0.48 | 0.60 | 0.54 |
| | Alg-2-nested-cf-IVAP | 0.07 | 0.09 | 0.08 | 0.06 | **0.08** | 0.07 | 0.07 | **0.08** | 0.08 |
| | Alg-2-nested-cf-Iso | 0.12 | 0.15 | 0.14 | 0.10 | 0.12 | 0.12 | 0.10 | 0.12 | 0.12 |
| | Alg-2-nested-cf-Platt | 0.08 | 0.14 | 0.14 | 0.06 | **0.08** | 0.07 | 0.07 | **0.08** | 0.07 |
| | Alg-3-cf-IVAP | 0.06 | 0.08 | 0.07 | 0.06 | **0.08** | 0.08 | 0.07 | **0.08** | 0.07 |
| | Alg-3-cf-Iso | 0.06 | 0.08 | 0.07 | 0.06 | **0.08** | 0.08 | 0.06 | **0.08** | 0.07 |
| | Alg-3-cf-Platt | 0.06 | **0.07** | 0.07 | 0.07 | 0.08 | 0.06 | 0.07 | **0.08** | 0.07 |
| | Alg-4-single-split-IVAP | 0.06 | 0.08 | 0.07 | 0.07 | **0.08** | 0.06 | 0.07 | 0.09 | 0.06 |
| | Alg-4-single-split-Iso | 0.10 | 0.13 | 0.12 | 0.07 | **0.08** | 0.06 | 0.07 | 0.09 | 0.06 |
| | Alg-4-single-split-Platt | 0.07 | 0.08 | 0.08 | 0.06 | **0.08** | 0.07 | 0.07 | **0.08** | 0.07 |
| | Alg-5-full-sample-IVAP | 0.07 | 0.08 | 0.08 | 0.07 | 0.09 | 0.08 | 0.07 | **0.08** | 0.08 |
| | Alg-5-full-sample-Iso | 0.06 | 0.08 | 0.08 | 0.06 | **0.08** | 0.07 | 0.07 | **0.08** | 0.08 |
| | Alg-5-full-sample-Platt | 0.07 | 0.09 | 0.09 | 0.06 | **0.08** | 0.07 | 0.07 | **0.08** | 0.07 |
| 2 | Alg-1-Clipped | 0.09 | **0.11** | 0.11 | 0.31 | 0.39 | 0.39 | 0.20 | 0.26 | 0.26 |
| | Alg-1-Uncalib | 0.09 | **0.11** | 0.11 | 2.22e+09 | 2.89e+09 | 2.86e+09 | 0.24 | 0.32 | 0.32 |
| | Alg-2-nested-cf-IVAP | 0.09 | **0.11** | 0.11 | 0.09 | **0.11** | 0.11 | 0.09 | 0.12 | 0.12 |
| | Alg-2-nested-cf-Iso | 0.18 | 0.23 | 0.22 | 0.16 | 0.22 | 0.22 | 0.19 | 0.24 | 0.24 |
| | Alg-2-nested-cf-Platt | 0.09 | **0.11** | 0.11 | 0.09 | **0.11** | 0.11 | 0.09 | **0.11** | 0.11 |
| | Alg-3-cf-IVAP | 0.09 | **0.11** | 0.11 | 0.10 | 0.12 | 0.12 | 0.10 | 0.12 | 0.12 |
| | Alg-3-cf-Iso | 0.09 | **0.11** | 0.11 | 0.09 | 0.12 | 0.12 | 0.10 | 0.12 | 0.12 |
| | Alg-3-cf-Platt | 0.09 | **0.11** | 0.11 | 0.09 | **0.11** | 0.11 | 0.09 | **0.11** | 0.11 |
| | Alg-4-single-split-IVAP | 0.09 | **0.11** | 0.11 | 0.09 | **0.11** | 0.11 | 0.09 | **0.11** | 0.11 |
| | Alg-4-single-split-Iso | 0.15 | 0.20 | 0.20 | 0.09 | **0.11** | 0.11 | 0.09 | **0.11** | 0.11 |
| | Alg-4-single-split-Platt | 0.09 | **0.11** | 0.11 | 0.09 | **0.11** | 0.11 | 0.09 | **0.11** | 0.11 |
| | Alg-5-full-sample-IVAP | 0.09 | **0.11** | 0.11 | 0.09 | **0.11** | 0.11 | 0.09 | 0.12 | 0.12 |
| | Alg-5-full-sample-Iso | 0.09 | **0.11** | 0.11 | 0.09 | **0.11** | 0.11 | 0.09 | 0.12 | 0.12 |
| | Alg-5-full-sample-Platt | 0.09 | **0.11** | 0.11 | 0.09 | **0.11** | 0.11 | 0.09 | **0.11** | 0.11 |
| 3 | Alg-1-Clipped | 0.05 | 0.07 | 0.07 | 0.08 | 0.10 | 0.10 | 0.11 | 0.13 | 0.13 |
| | Alg-1-Uncalib | 0.05 | 0.07 | 0.07 | 9.97e+07 | 2.74e+08 | 2.73e+08 | 0.11 | 0.14 | 0.14 |
| | Alg-2-nested-cf-IVAP | 0.05 | 0.07 | 0.07 | 0.05 | **0.07** | 0.06 | 0.05 | **0.07** | 0.07 |
| | Alg-2-nested-cf-Iso | 0.10 | 0.13 | 0.12 | 0.08 | 0.10 | 0.10 | 0.10 | 0.13 | 0.13 |
| | Alg-2-nested-cf-Platt | 0.05 | 0.07 | 0.07 | 0.05 | **0.07** | 0.06 | 0.05 | **0.07** | 0.06 |
| | Alg-3-cf-IVAP | 0.05 | 0.07 | 0.06 | 0.06 | **0.07** | 0.07 | 0.05 | **0.07** | 0.07 |
| | Alg-3-cf-Iso | 0.05 | 0.07 | 0.07 | 0.06 | **0.07** | 0.07 | 0.05 | **0.07** | 0.06 |
| | Alg-3-cf-Platt | 0.05 | **0.06** | 0.06 | 0.05 | **0.07** | 0.06 | 0.05 | **0.07** | 0.06 |
| | Alg-4-single-split-IVAP | 0.05 | 0.07 | 0.07 | 0.05 | **0.07** | 0.06 | 0.05 | **0.07** | 0.06 |
| | Alg-4-single-split-Iso | 0.10 | 0.12 | 0.11 | 0.05 | **0.07** | 0.06 | 0.05 | **0.07** | 0.06 |
| | Alg-4-single-split-Platt | 0.05 | 0.07 | 0.06 | 0.06 | **0.07** | 0.07 | 0.05 | **0.07** | 0.06 |
| | Alg-5-full-sample-IVAP | 0.05 | 0.07 | 0.07 | 0.06 | **0.07** | 0.07 | 0.05 | **0.07** | 0.07 |
| | Alg-5-full-sample-Iso | 0.05 | 0.07 | 0.07 | 0.06 | **0.07** | 0.07 | 0.05 | **0.07** | 0.07 |
| | Alg-5-full-sample-Platt | 0.05 | 0.07 | 0.07 | 0.06 | **0.07** | 0.07 | 0.05 | **0.07** | 0.06 |
| 4 | Alg-1-Clipped | 0.04 | 0.06 | 0.06 | 0.07 | 0.09 | 0.09 | 0.17 | 0.21 | 0.20 |
| | Alg-1-Uncalib | 0.04 | 0.06 | 0.06 | 1.74e+08 | 2.70e+08 | 2.68e+08 | 0.36 | 0.46 | 0.43 |
| | Alg-2-nested-cf-IVAP | 0.04 | **0.05** | 0.05 | 0.05 | **0.06** | 0.06 | 0.04 | **0.06** | 0.06 |
| | Alg-2-nested-cf-Iso | 0.05 | 0.07 | 0.07 | 0.05 | 0.07 | 0.07 | 0.06 | 0.07 | 0.07 |
| | Alg-2-nested-cf-Platt | 0.04 | 0.06 | 0.05 | 0.05 | **0.06** | 0.06 | 0.04 | **0.06** | 0.06 |
| | Alg-3-cf-IVAP | 0.04 | 0.06 | 0.06 | 0.05 | **0.06** | 0.06 | 0.05 | **0.06** | 0.06 |
| | Alg-3-cf-Iso | 0.04 | 0.06 | 0.06 | 0.05 | **0.06** | 0.06 | 0.05 | **0.06** | 0.06 |
| | Alg-3-cf-Platt | 0.04 | 0.06 | 0.06 | 0.05 | **0.06** | 0.06 | 0.05 | **0.06** | 0.06 |
| | Alg-4-single-split-IVAP | 0.04 | 0.06 | 0.06 | 0.04 | **0.06** | 0.06 | 0.04 | **0.06** | 0.06 |
| | Alg-4-single-split-Iso | 0.05 | 0.06 | 0.06 | 0.05 | **0.06** | 0.06 | 0.04 | **0.06** | 0.06 |
| | Alg-4-single-split-Platt | 0.04 | 0.06 | 0.06 | 0.05 | **0.06** | 0.06 | 0.04 | **0.06** | 0.06 |
| | Alg-5-full-sample-IVAP | 0.04 | 0.06 | 0.06 | 0.05 | **0.06** | 0.06 | 0.04 | **0.06** | 0.06 |
| | Alg-5-full-sample-Iso | 0.04 | 0.06 | 0.06 | 0.05 | **0.06** | 0.06 | 0.04 | **0.06** | 0.06 |
| | Alg-5-full-sample-Platt | 0.04 | 0.06 | 0.06 | 0.05 | **0.06** | 0.06 | 0.04 | **0.06** | 0.05 |

DGP 1: n = 2000, p = 20, R2_d = 0.5; DGP 2: n = 2000, p = 3, overlap = 0.5;
DGP 3: n = 2000, p = 4; DGP 4: n = 4000, p = 20, share treated = 0.1; g = LGBM,
lowest RMSEs per DGP and propensity learner are highlighted

Table 5: PLR Results

| DGP | Method | m = Logit | | | m = Random Forest | | | m = LGBM | | |
|---|---|---|---|---|---|---|---|---|---|---|
| | | MAE | RMSE | Std. dev. | MAE | RMSE | Std. dev. | MAE | RMSE | Std. dev. |
| 1 | Alg-1-Clipped | 0.05 | **0.06** | 0.06 | 0.05 | **0.06** | 0.06 | 0.05 | **0.06** | 0.06 |
| | Alg-1-Uncalib | 0.05 | **0.06** | 0.06 | 0.05 | **0.06** | 0.06 | 0.05 | **0.06** | 0.06 |
| | Alg-2-nested-cf-IVAP | 0.05 | **0.06** | 0.06 | 0.05 | **0.06** | 0.06 | 0.05 | **0.06** | 0.06 |
| | Alg-2-nested-cf-Iso | 0.05 | **0.06** | 0.06 | 0.05 | **0.06** | 0.06 | 0.05 | **0.06** | 0.06 |
| | Alg-2-nested-cf-Platt | 0.05 | **0.06** | 0.06 | 0.05 | **0.06** | 0.06 | 0.05 | **0.06** | 0.06 |
| | Alg-3-cf-IVAP | 0.05 | **0.06** | 0.06 | 0.05 | **0.06** | 0.06 | 0.05 | **0.06** | 0.06 |
| | Alg-3-cf-Iso | 0.05 | **0.06** | 0.06 | 0.05 | **0.06** | 0.06 | 0.05 | 0.07 | 0.07 |
| | Alg-3-cf-Platt | 0.05 | **0.06** | 0.06 | 0.05 | **0.06** | 0.06 | 0.05 | 0.07 | 0.07 |
| | Alg-4-single-split-IVAP | 0.05 | **0.06** | 0.06 | 0.14 | 0.18 | 0.18 | 0.12 | 0.15 | 0.15 |
| | Alg-4-single-split-Iso | 0.05 | **0.06** | 0.06 | 0.15 | 0.20 | 0.20 | 0.13 | 0.16 | 0.16 |
| | Alg-4-single-split-Platt | 0.05 | **0.06** | 0.06 | 0.05 | **0.06** | 0.06 | 0.05 | **0.06** | 0.06 |
| | Alg-5-full-sample-IVAP | 0.05 | **0.06** | 0.06 | 0.05 | **0.06** | 0.06 | 0.05 | **0.06** | 0.06 |
| | Alg-5-full-sample-Iso | 0.05 | **0.06** | 0.06 | 0.05 | **0.06** | 0.06 | 0.05 | 0.07 | 0.07 |
| | Alg-5-full-sample-Platt | 0.05 | **0.06** | 0.06 | 0.05 | **0.06** | 0.06 | 0.05 | **0.06** | 0.06 |
| 2 | Alg-1-Clipped | 0.09 | **0.11** | 0.10 | 0.12 | 0.14 | 0.10 | 0.10 | 0.11 | 0.10 |
| | Alg-1-Uncalib | 0.09 | **0.11** | 0.10 | 0.12 | 0.14 | 0.10 | 0.10 | 0.11 | 0.10 |
| | Alg-2-nested-cf-IVAP | 0.09 | **0.11** | 0.10 | 0.10 | 0.12 | 0.10 | 0.09 | 0.11 | 0.10 |
| | Alg-2-nested-cf-Iso | 0.09 | **0.11** | 0.10 | 0.10 | 0.12 | 0.10 | 0.09 | 0.11 | 0.10 |
| | Alg-2-nested-cf-Platt | 0.09 | **0.11** | 0.10 | 0.10 | 0.12 | 0.10 | 0.09 | 0.11 | 0.10 |
| | Alg-3-cf-IVAP | 0.09 | **0.11** | 0.10 | 0.10 | 0.12 | 0.10 | 0.09 | 0.11 | 0.10 |
| | Alg-3-cf-Iso | 0.09 | **0.11** | 0.11 | 0.09 | **0.11** | 0.10 | 0.09 | **0.10** | 0.10 |
| | Alg-3-cf-Platt | 0.09 | **0.11** | 0.10 | 0.11 | 0.13 | 0.10 | 0.09 | 0.11 | 0.10 |
| | Alg-4-single-split-IVAP | 0.09 | **0.11** | 0.10 | 0.23 | 0.26 | 0.14 | 0.13 | 0.16 | 0.12 |
| | Alg-4-single-split-Iso | 0.09 | **0.11** | 0.10 | 0.24 | 0.27 | 0.15 | 0.13 | 0.16 | 0.12 |
| | Alg-4-single-split-Platt | 0.09 | **0.11** | 0.10 | 0.09 | **0.11** | 0.10 | 0.09 | 0.11 | 0.10 |
| | Alg-5-full-sample-IVAP | 0.09 | **0.11** | 0.10 | 0.09 | **0.11** | 0.10 | 0.09 | **0.10** | 0.10 |
| | Alg-5-full-sample-Iso | 0.09 | **0.11** | 0.10 | 0.09 | **0.11** | 0.10 | 0.09 | **0.10** | 0.10 |
| | Alg-5-full-sample-Platt | 0.09 | **0.11** | 0.10 | 0.09 | **0.11** | 0.10 | 0.09 | 0.11 | 0.10 |
| 3 | Alg-1-Clipped | 0.06 | **0.08** | 0.08 | 0.06 | **0.07** | 0.07 | 0.08 | 0.10 | 0.07 |
| | Alg-1-Uncalib | 0.06 | **0.08** | 0.08 | 0.06 | **0.07** | 0.07 | 0.08 | 0.10 | 0.07 |
| | Alg-2-nested-cf-IVAP | 0.06 | **0.08** | 0.08 | 0.07 | 0.09 | 0.07 | 0.09 | 0.11 | 0.07 |
| | Alg-2-nested-cf-Iso | 0.06 | **0.08** | 0.08 | 0.07 | 0.09 | 0.07 | 0.09 | 0.11 | 0.07 |
| | Alg-2-nested-cf-Platt | 0.06 | **0.08** | 0.08 | 0.07 | 0.08 | 0.07 | 0.09 | 0.11 | 0.07 |
| | Alg-3-cf-IVAP | 0.06 | **0.08** | 0.08 | 0.06 | 0.08 | 0.07 | 0.06 | 0.08 | 0.07 |
| | Alg-3-cf-Iso | 0.07 | 0.09 | 0.08 | 0.06 | 0.08 | 0.08 | 0.06 | 0.08 | 0.08 |
| | Alg-3-cf-Platt | 0.07 | **0.08** | 0.08 | 0.10 | 0.12 | 0.08 | 0.07 | 0.09 | 0.08 |
| | Alg-4-single-split-IVAP | 0.06 | **0.08** | 0.08 | 1.01 | 1.02 | 0.16 | 0.43 | 0.44 | 0.11 |
| | Alg-4-single-split-Iso | 0.06 | **0.08** | 0.08 | 1.10 | 1.11 | 0.18 | 0.46 | 0.47 | 0.12 |
| | Alg-4-single-split-Platt | 0.06 | **0.08** | 0.08 | 0.06 | 0.08 | 0.08 | 0.08 | 0.09 | 0.07 |
| | Alg-5-full-sample-IVAP | 0.06 | **0.08** | 0.08 | 0.06 | 0.08 | 0.07 | 0.06 | **0.08** | 0.07 |
| | Alg-5-full-sample-Iso | 0.07 | 0.09 | 0.08 | 0.06 | **0.07** | 0.07 | 0.06 | **0.07** | 0.07 |
| | Alg-5-full-sample-Platt | 0.06 | **0.08** | 0.08 | 0.06 | 0.08 | 0.08 | 0.06 | **0.07** | 0.07 |
| 4 | Alg-1-Clipped | 0.08 | **0.09** | 0.06 | 0.05 | **0.06** | 0.05 | 0.05 | **0.06** | 0.05 |
| | Alg-1-Uncalib | 0.08 | **0.09** | 0.06 | 0.05 | **0.06** | 0.05 | 0.05 | **0.06** | 0.05 |
| | Alg-2-nested-cf-IVAP | 0.08 | **0.09** | 0.05 | 0.07 | 0.09 | 0.05 | 0.07 | 0.09 | 0.05 |
| | Alg-2-nested-cf-Iso | 0.08 | **0.09** | 0.06 | 0.07 | 0.09 | 0.05 | 0.07 | 0.09 | 0.05 |
| | Alg-2-nested-cf-Platt | 0.08 | **0.09** | 0.06 | 0.07 | 0.09 | 0.05 | 0.07 | 0.09 | 0.05 |
| | Alg-3-cf-IVAP | 0.07 | **0.09** | 0.06 | 0.07 | 0.08 | 0.05 | 0.07 | 0.09 | 0.05 |
| | Alg-3-cf-Iso | 0.08 | 0.10 | 0.06 | 0.08 | 0.09 | 0.05 | 0.08 | 0.10 | 0.05 |
| | Alg-3-cf-Platt | 0.08 | **0.09** | 0.05 | 0.07 | 0.09 | 0.05 | 0.08 | 0.09 | 0.06 |
| | Alg-4-single-split-IVAP | 0.08 | 0.10 | 0.06 | 0.56 | 0.59 | 0.19 | 0.56 | 0.58 | 0.14 |
| | Alg-4-single-split-Iso | 0.08 | **0.09** | 0.06 | 0.26 | 0.31 | 0.17 | 0.26 | 0.30 | 0.16 |
| | Alg-4-single-split-Platt | 0.08 | **0.09** | 0.06 | 0.07 | 0.09 | 0.05 | 0.08 | 0.09 | 0.05 |
| | Alg-5-full-sample-IVAP | 0.08 | **0.09** | 0.06 | 0.07 | 0.09 | 0.05 | 0.08 | 0.09 | 0.05 |
| | Alg-5-full-sample-Iso | 0.08 | **0.09** | 0.06 | 0.07 | 0.09 | 0.05 | 0.08 | 0.09 | 0.05 |
| | Alg-5-full-sample-Platt | 0.08 | **0.09** | 0.06 | 0.07 | 0.09 | 0.05 | 0.08 | 0.09 | 0.05 |

DGP 1: n = 2000, p = 20, R2_d = 0.5; DGP 2: n = 2000, p = 3, overlap = 0.5;
DGP 3: n = 2000, p = 4; DGP 4: n = 4000, p = 20, share treated = 0.1; g = LGBM,
lowest RMSEs per DGP and propensity learner are highlighted

## 4.3 Sensitivity Analysis

The following figures display the results over 100 repetitions for the ATE estimators. Additionally, *Oracle* estimates are provided, utilizing true propensity scores from the DGPs. Furthermore, the figures with respect to the sample size and number of covariates include the estimators with clipped propensity scores. The combination of clipping and calibration is also displayed in histogram plots. If no clipping is employed, the clipping threshold is set to $1^{-12}$ to ensure stability.

Table 6: Overview of Simulated Settings per DGP

| DGP | Parameter | Values |
|---|---|---|
| 1 | $n$ | [100, 200, 500, 1000, 2000] |
|   | $p$ | [5, 20, 50, 100, 200] |
|   | $R_d^2$ | [0.2, 0.5, 0.8] |
|   | Clip | [1e-12, 0.01, 0.1] |
| 2 | $n$ | [200, 500, 1000, 2000, 4000] |
|   | $p$ | 3 |
|   | Overlap | [0.1, 0.5, 0.9] |
|   | Clip | [1e-12, 0.01, 0.1] |
| 3 | $n$ | [200, 500, 1000, 2000, 4000] |
|   | $p$ | 4 |
|   | Clip | [1e-12, 0.01, 0.1] |
| 4 | $n$ | [2000, 4000, 6000, 8000] |
|   | $p$ | 20 |
|   | $\mathbb{E}[D]$ | [0.05, 0.1, 0.2] |
|   | Clip | [1e-12, 0.01, 0.1] |

## Propensity Scores by DGP

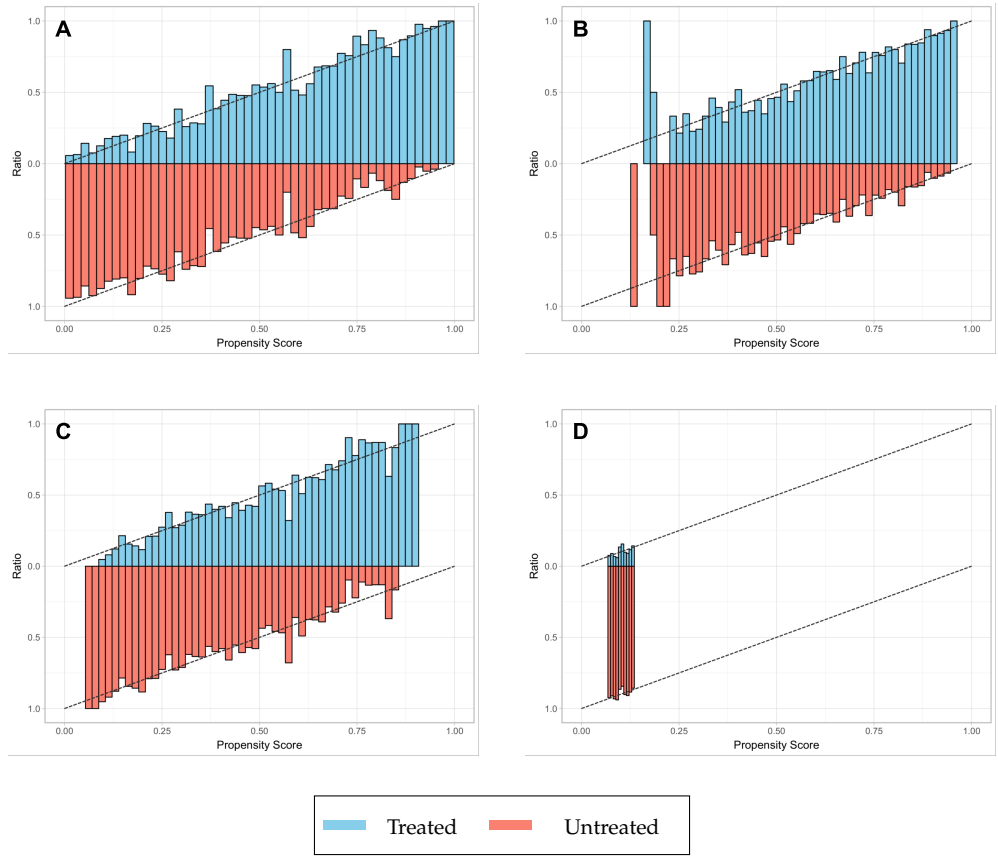

Figure S2: Underlying true propensity scores, divided by treatment allocation. The DGPs presented are: Panel A - DGP 1 (IRM), Panel B - DGP 2 (Drug), Panel C - DGP 3 (Nonlinear), and Panel D - DGP 4 (Unbalanced).

## Propensity Learner

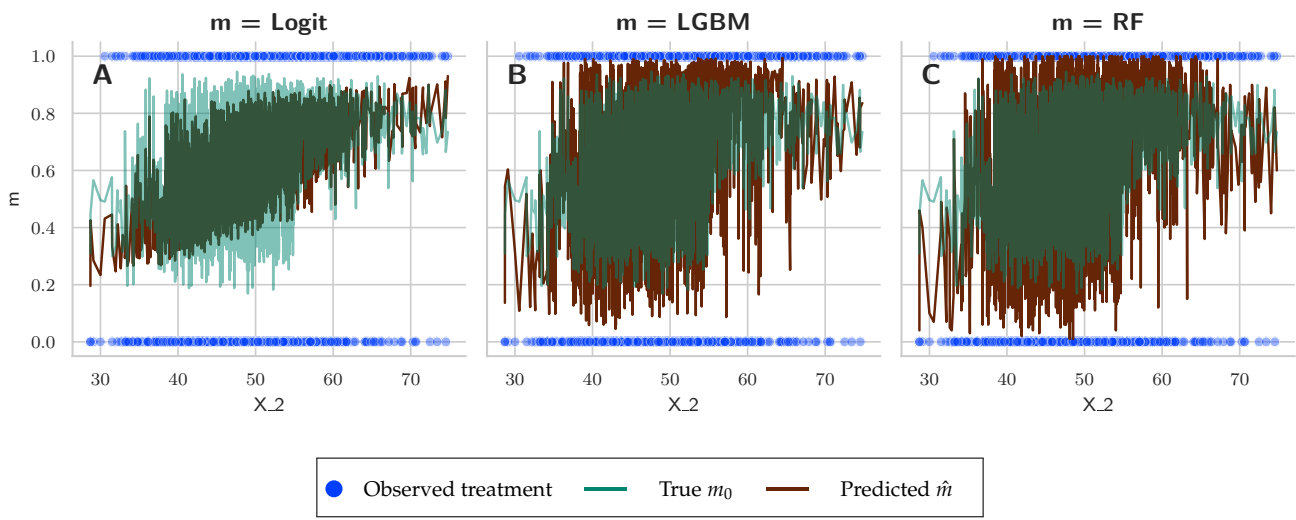

Figure S3: DGP 2 Drug, Overlap = 0.5, n = 2000, p = 3

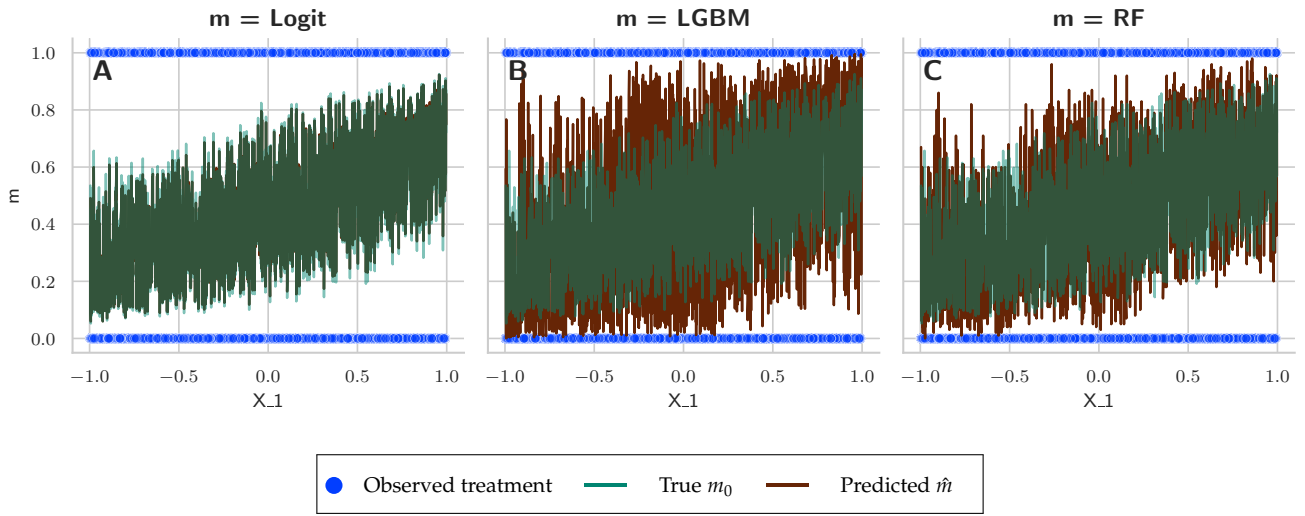

Figure S4: DGP 3 Nonlinear, n = 2000, p = 4

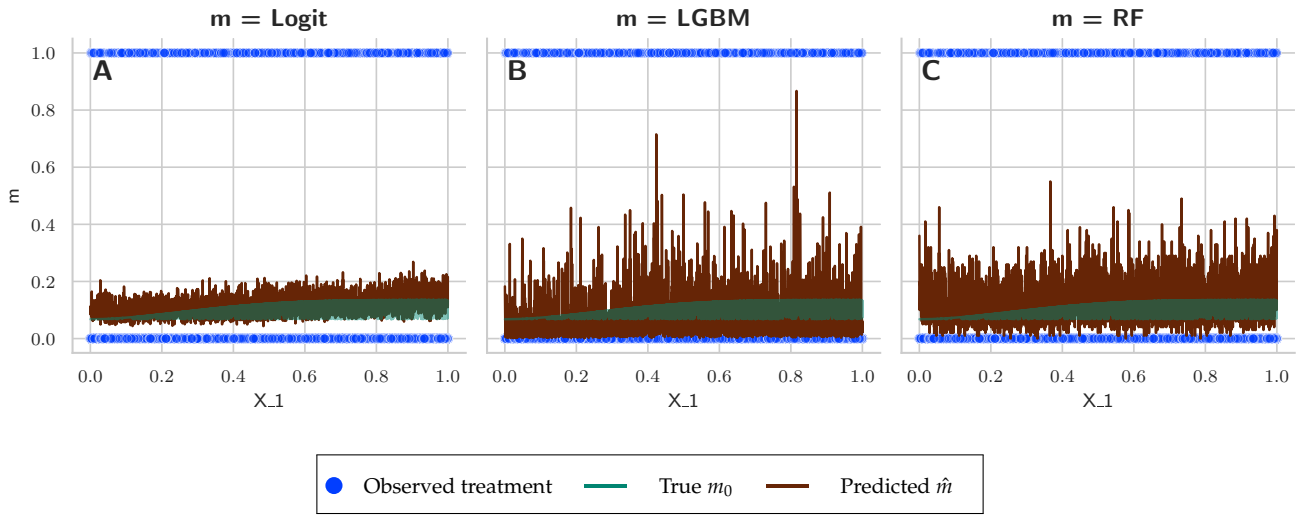

Figure S5: DGP 4 Unbalanced, $\alpha = 0.1$, p = 20

**Calibrated Propensity Scores**

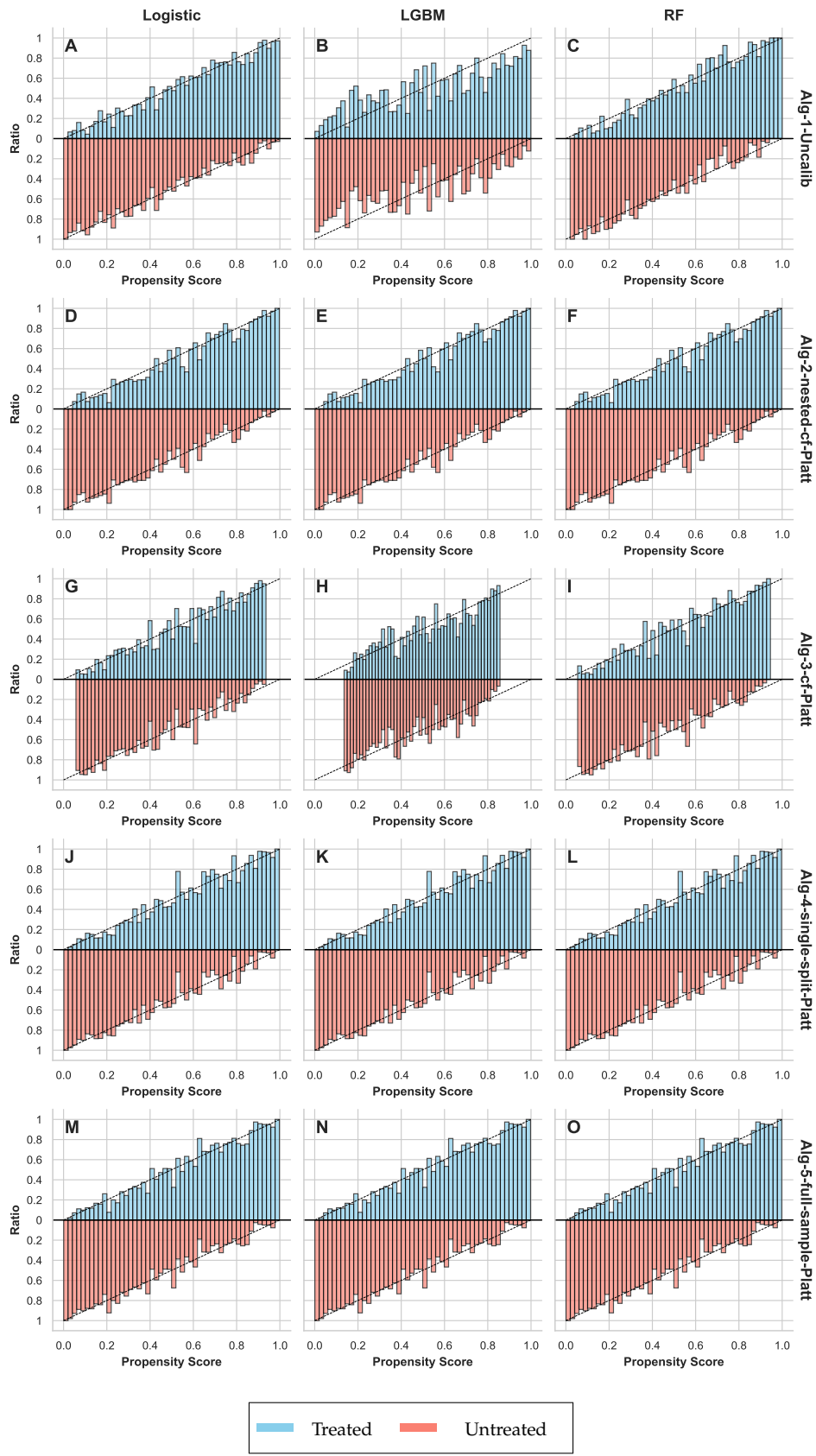

Figure S6: DGP 1 IRM, R2D = 0.5, m = LGBM, g = LGBM, n = 2000, p = 20, Calibration method for Algorithms 2-4: Platt Scaling

29

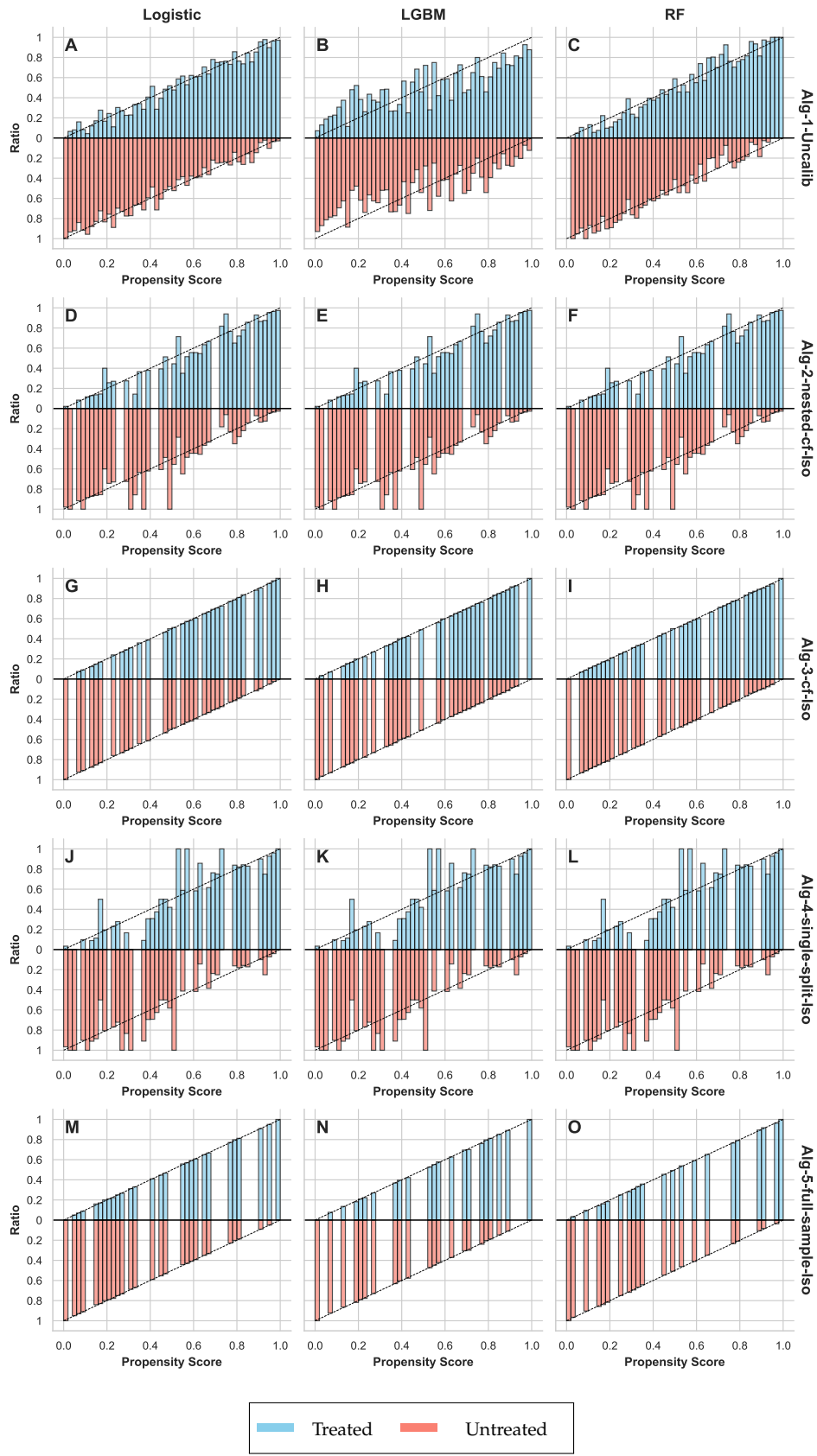

Figure S7: DGP 1 IRM, R2D = 0.5, m = LGBM, g = LGBM, n = 2000, p = 20, Calibration method for Algorithms 2-5: isotonic regression

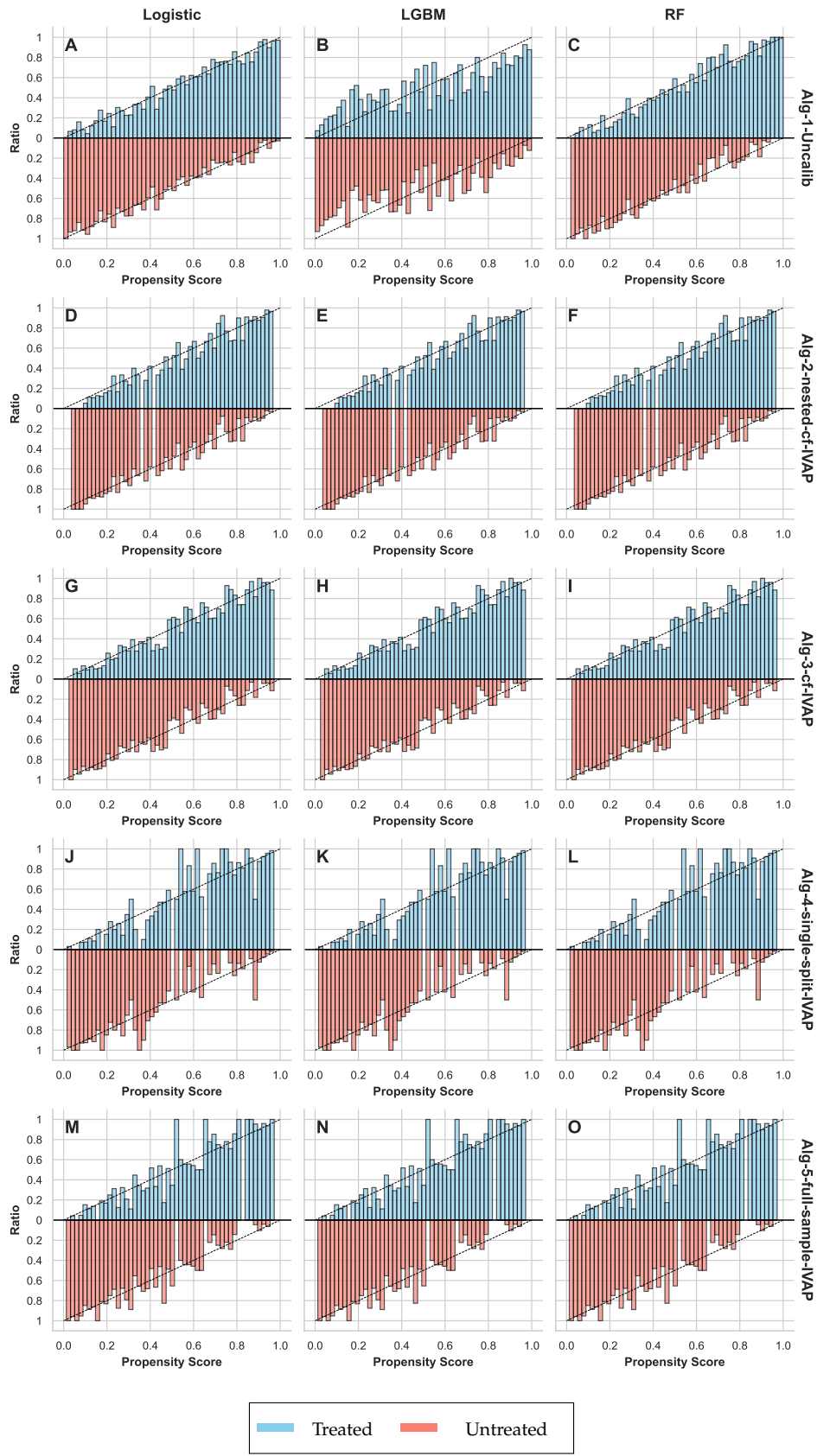

Figure S8: DGP 1 IRM, R2D = 0.5, m = LGBM, g = LGBM, n = 2000, p = 20, Calibration method for Algorithms 2-4: Venn-ABERS

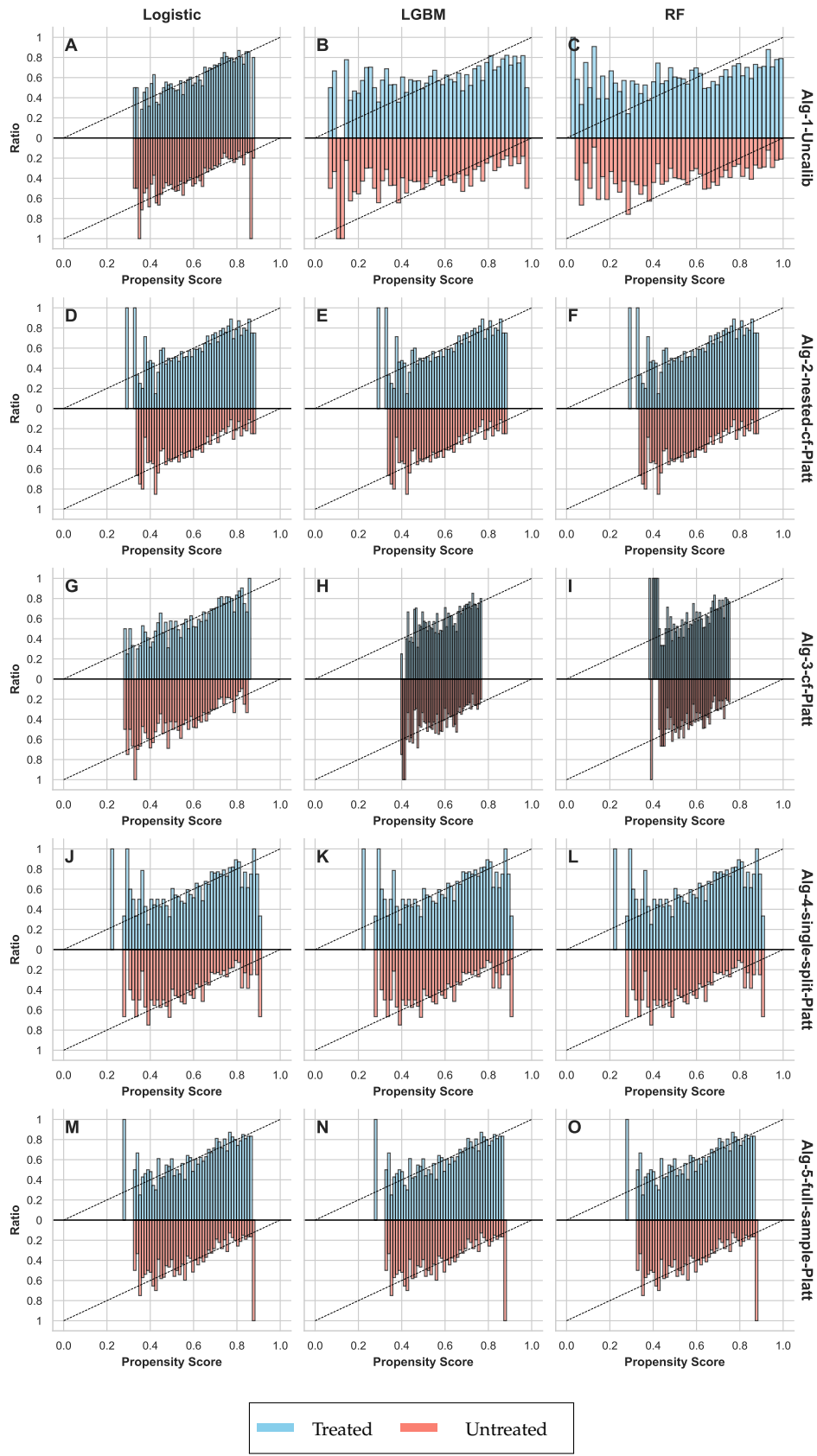

Figure S9: DGP 2 Drug, Overlap = 0.5, m = LGBM, g = LGBM, n = 2000, p = 3, Calibration method for Algorithms 2-4: Platt Scaling

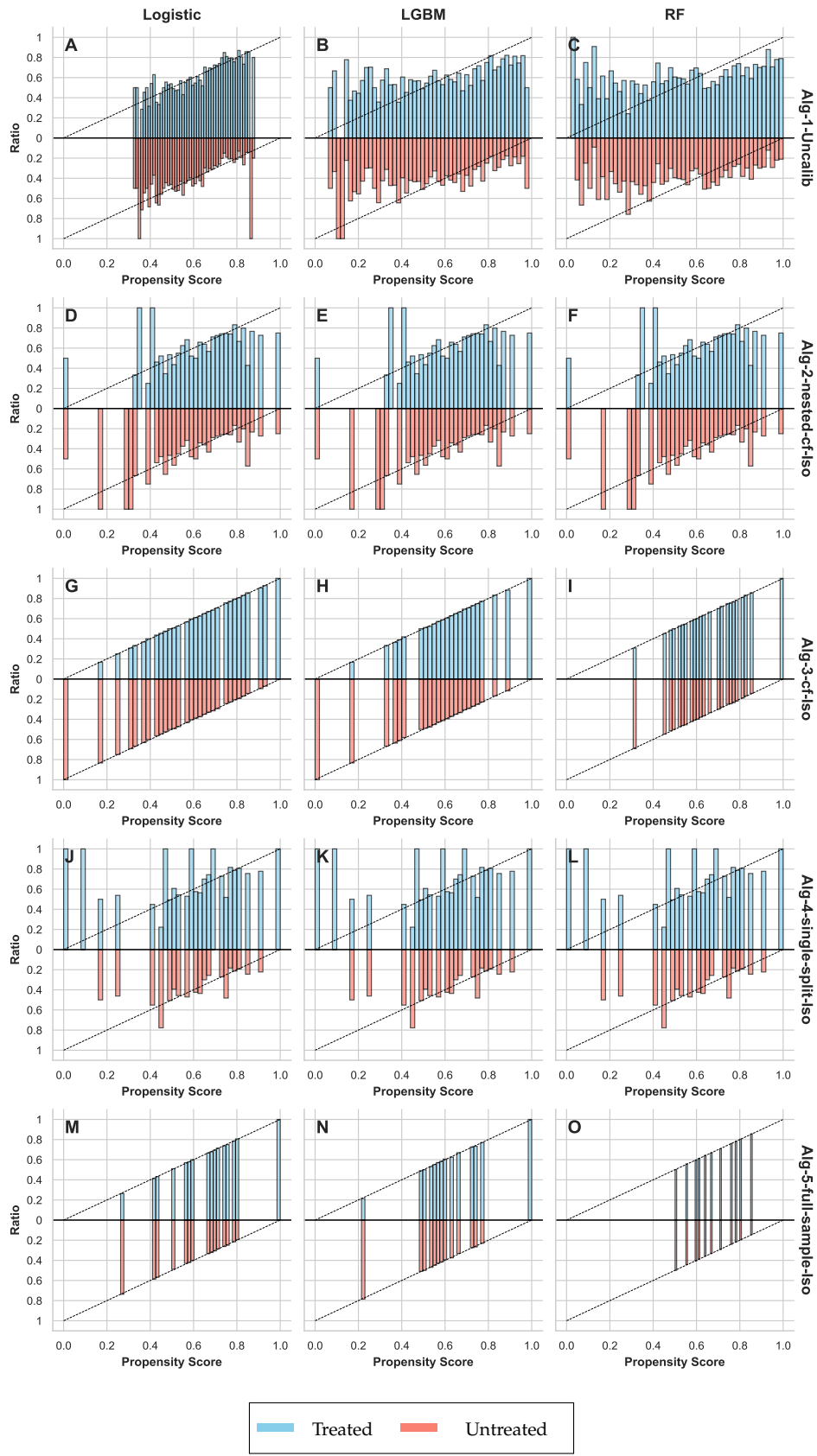

Figure S10: DGP 2 Drug, Overlap = 0.5, m = LGBM, g = LGBM, n = 2000, p = 3, Calibration method for Algorithms 2-5: isotonic regression

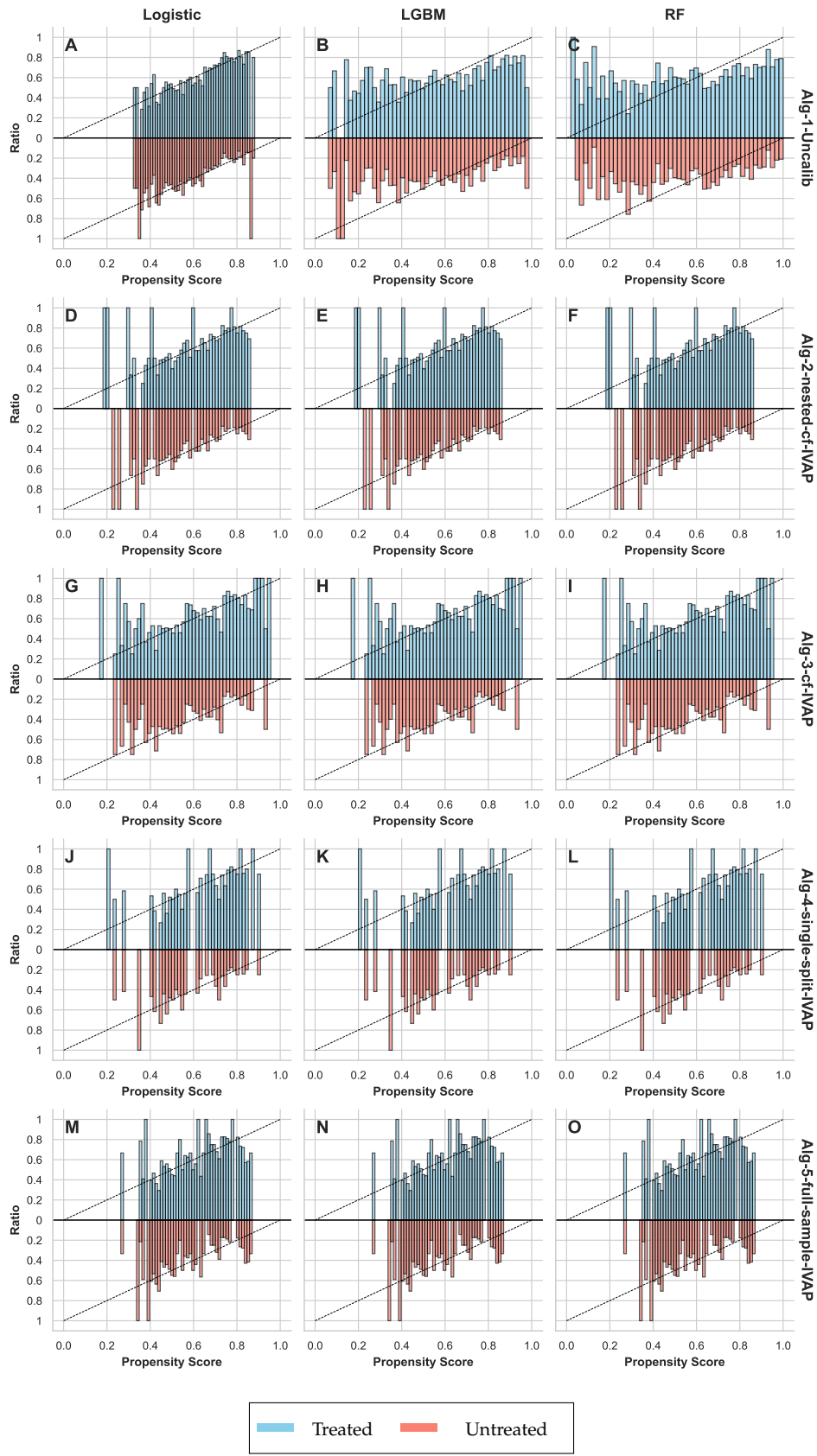

Figure S11: DGP 2 Drug, Overlap = 0.5, m = LGBM, g = LGBM, n = 2000, p = 3, Calibration method for Algorithms 2-4: Venn-ABERS

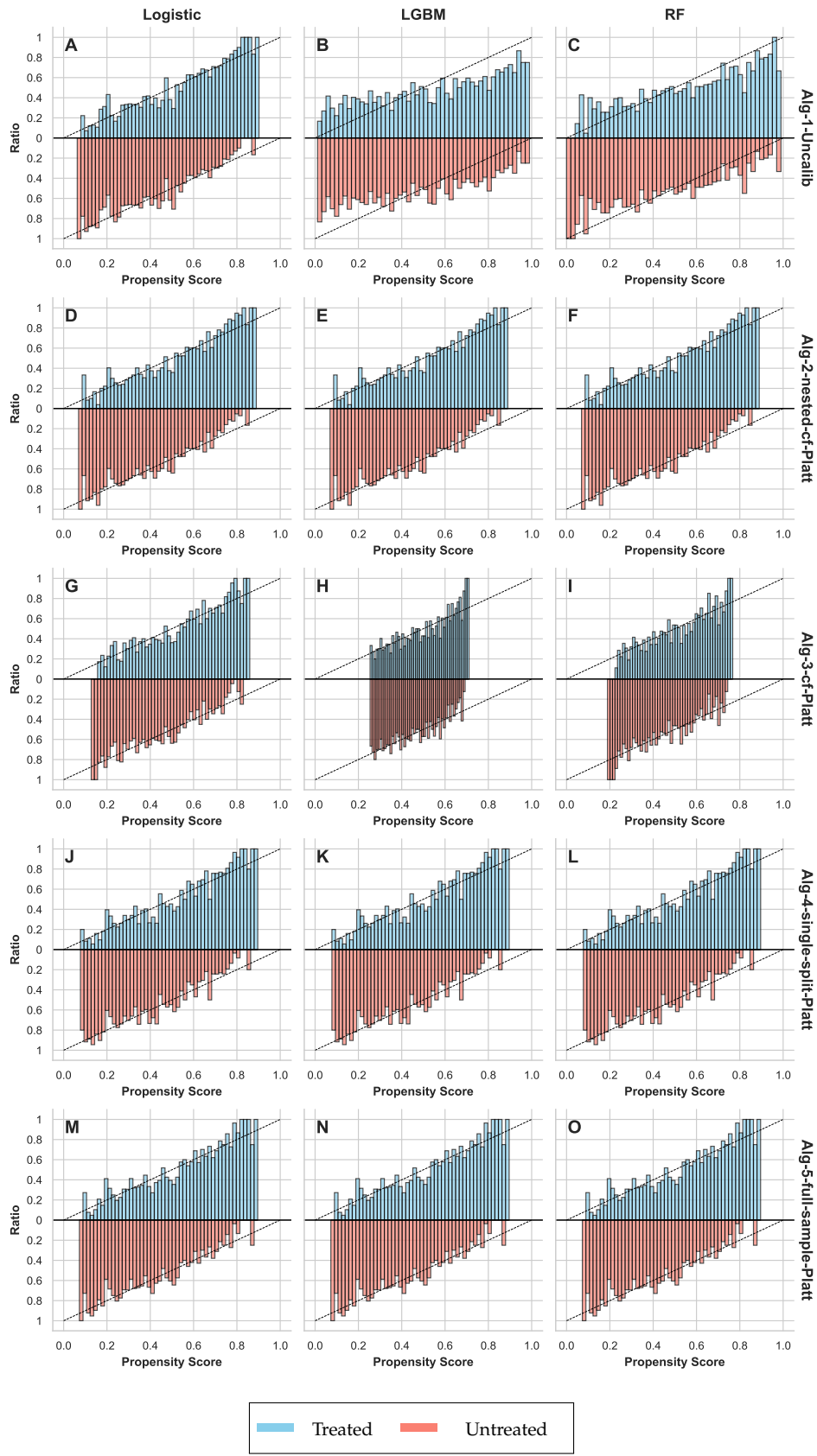

Figure S12: DGP 3 Nonlinear, m = LGBM, g = LGBM, n = 2000, p = 4, Calibration method for Algorithms 2-4: Platt Scaling

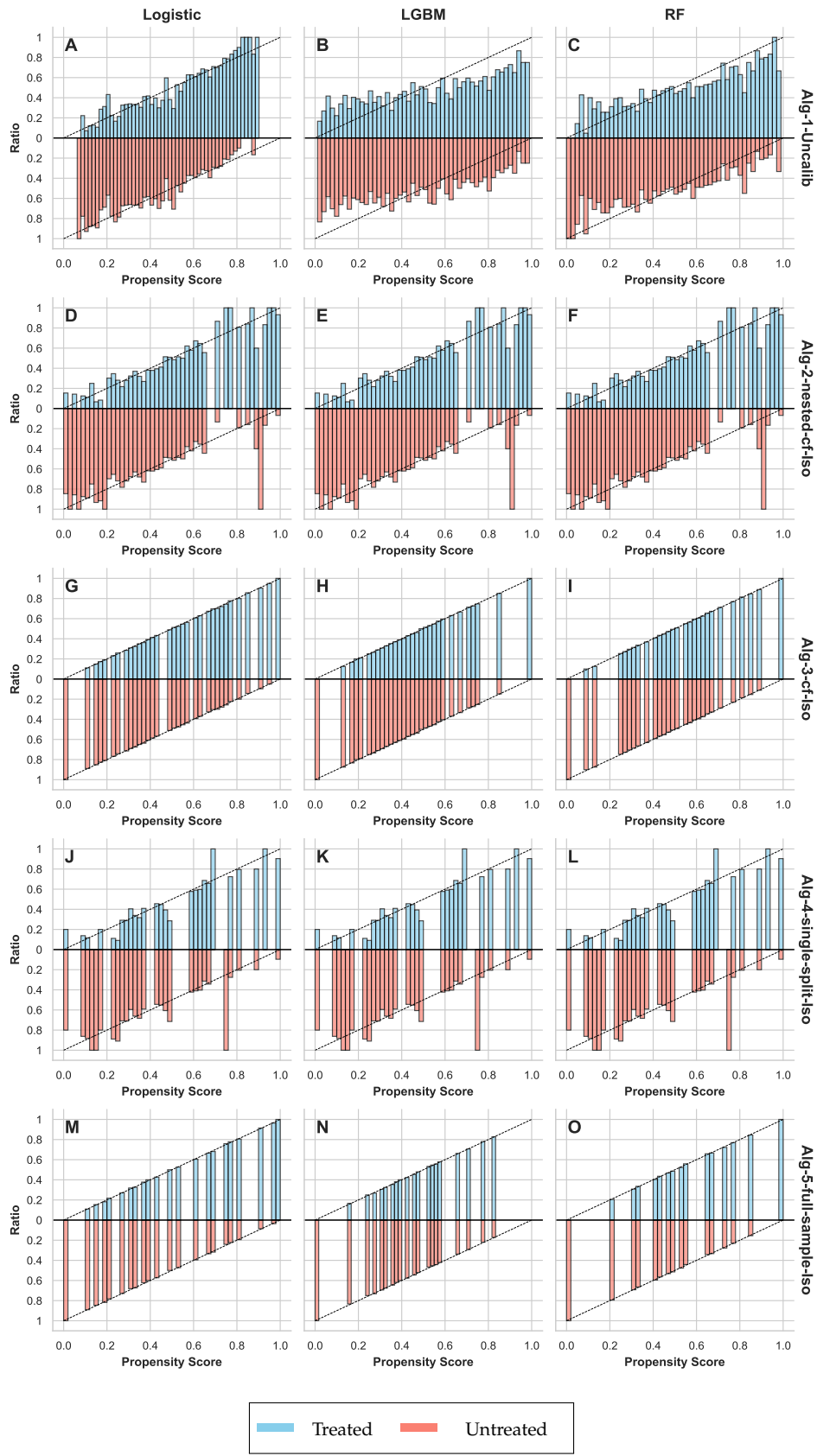

Figure S13: DGP 3 Nonlinear, m = LGBM, g = LGBM, n = 2000, p = 4, Calibration method for Algorithms 2-5: isotonic regression

36

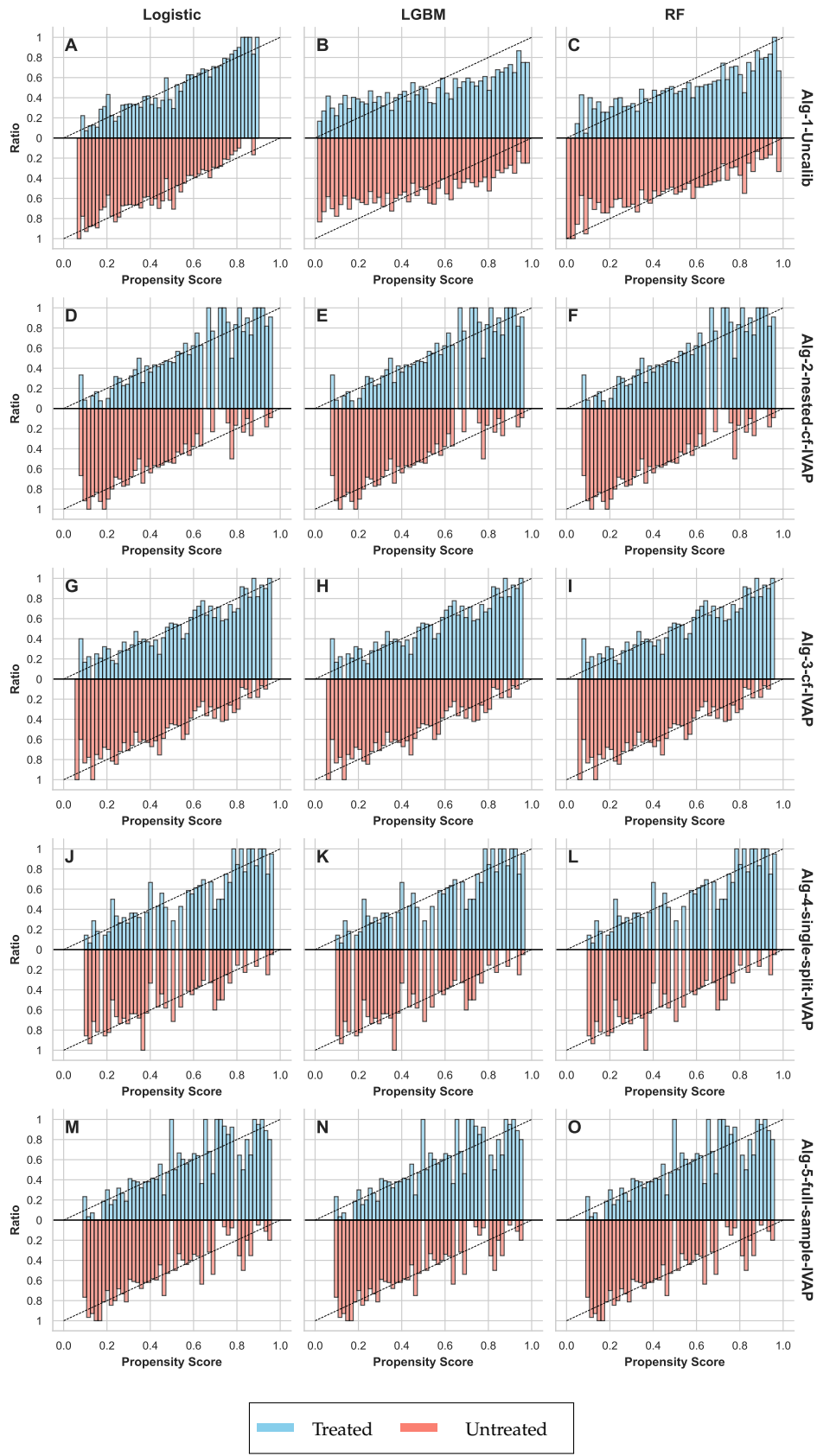

Figure S14: DGP 3 Nonlinear, m = LGBM, g = LGBM, n = 2000, p = 4, Calibration method for Algorithms 2-4: Venn-ABERS

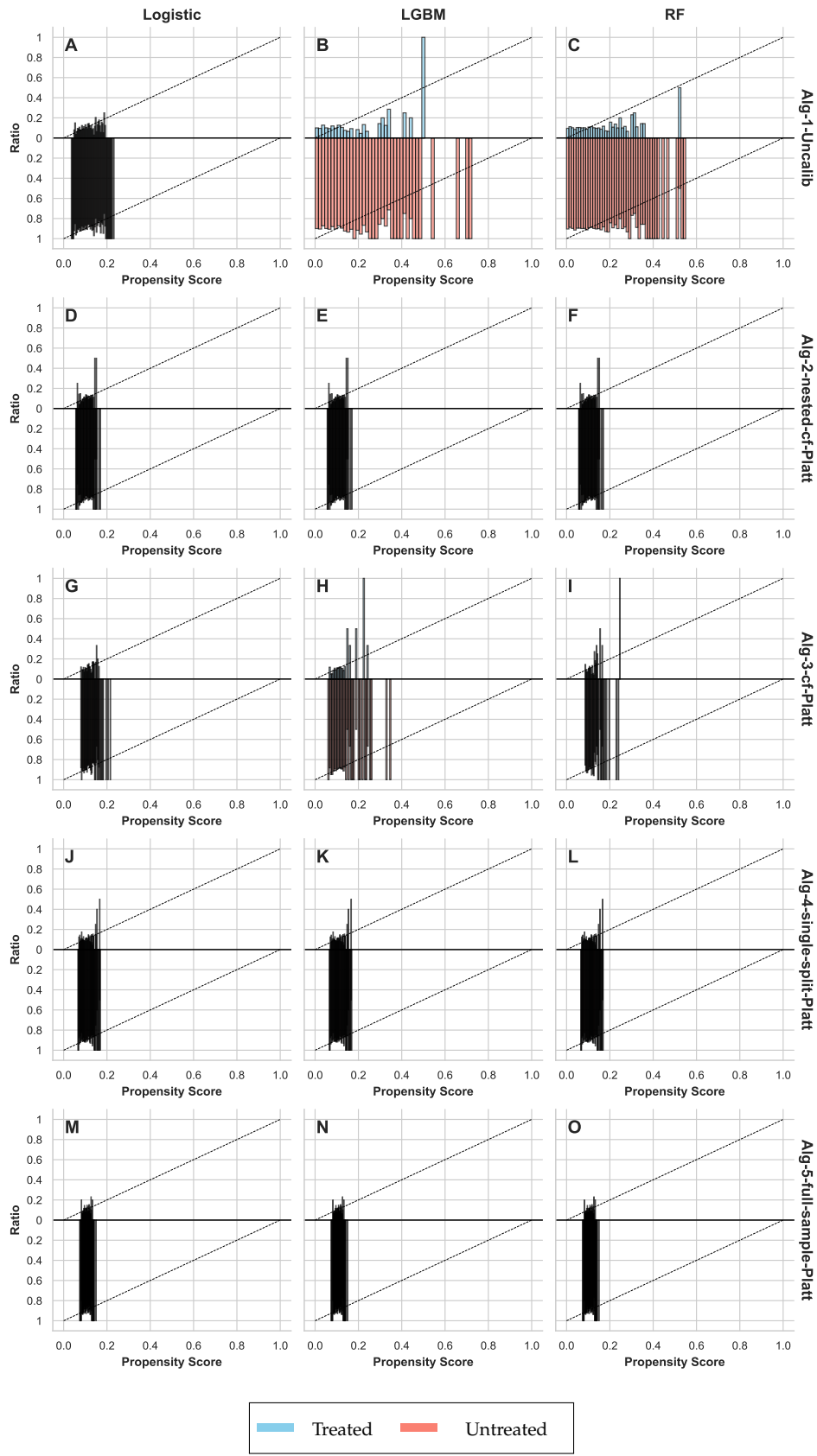

Figure S15: DGP 4 Unbalanced, $\alpha = 0.1$, m = LGBM, g = LGBM, n = 4000, p = 20, Calibration method for Algorithms 2-4: Platt Scaling

38

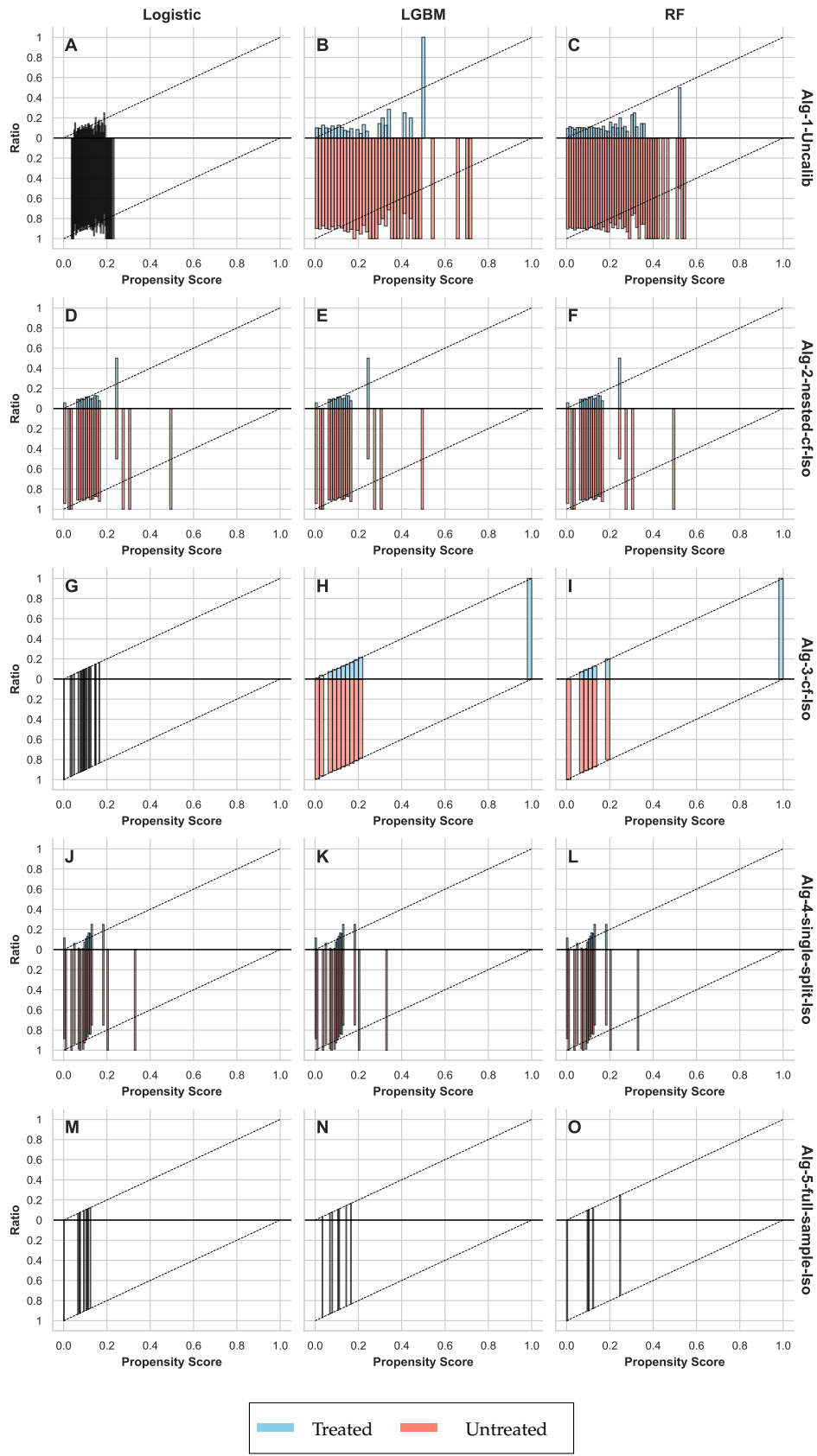

Figure S16: DGP 4 Unbalanced, $\alpha = 0.1$, m = LGBM, g = LGBM, n = 4000, p = 20, Calibration method for Algorithms 2-5: isotonic regression

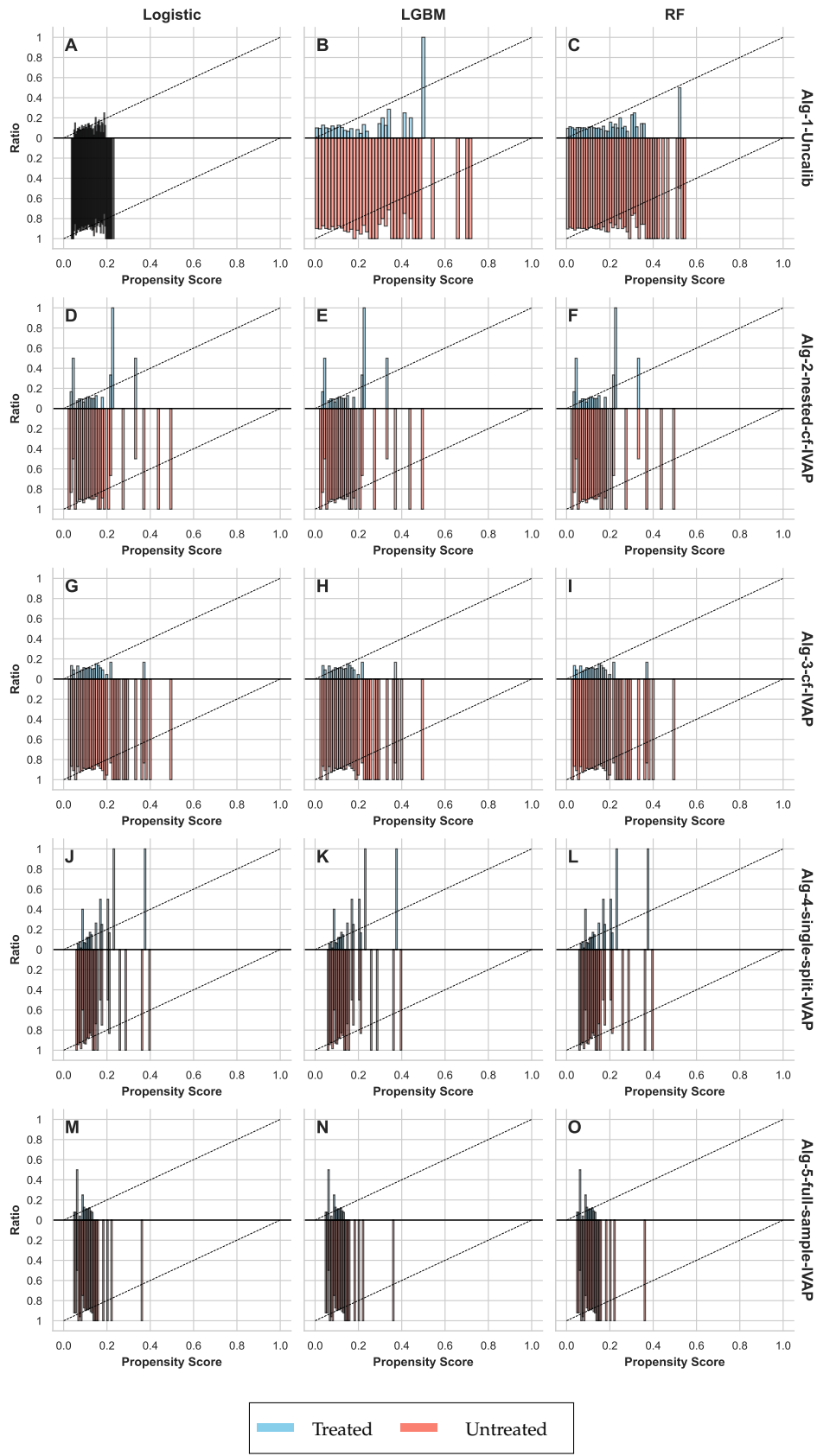

Figure S17: DGP 4 Unbalanced, $\alpha = 0.1$, m = LGBM, g = LGBM, n = 4000, p = 20, Calibration method for Algorithms 2-4: Venn-ABERS

# Calibration Errors

## DGP 1 IRM

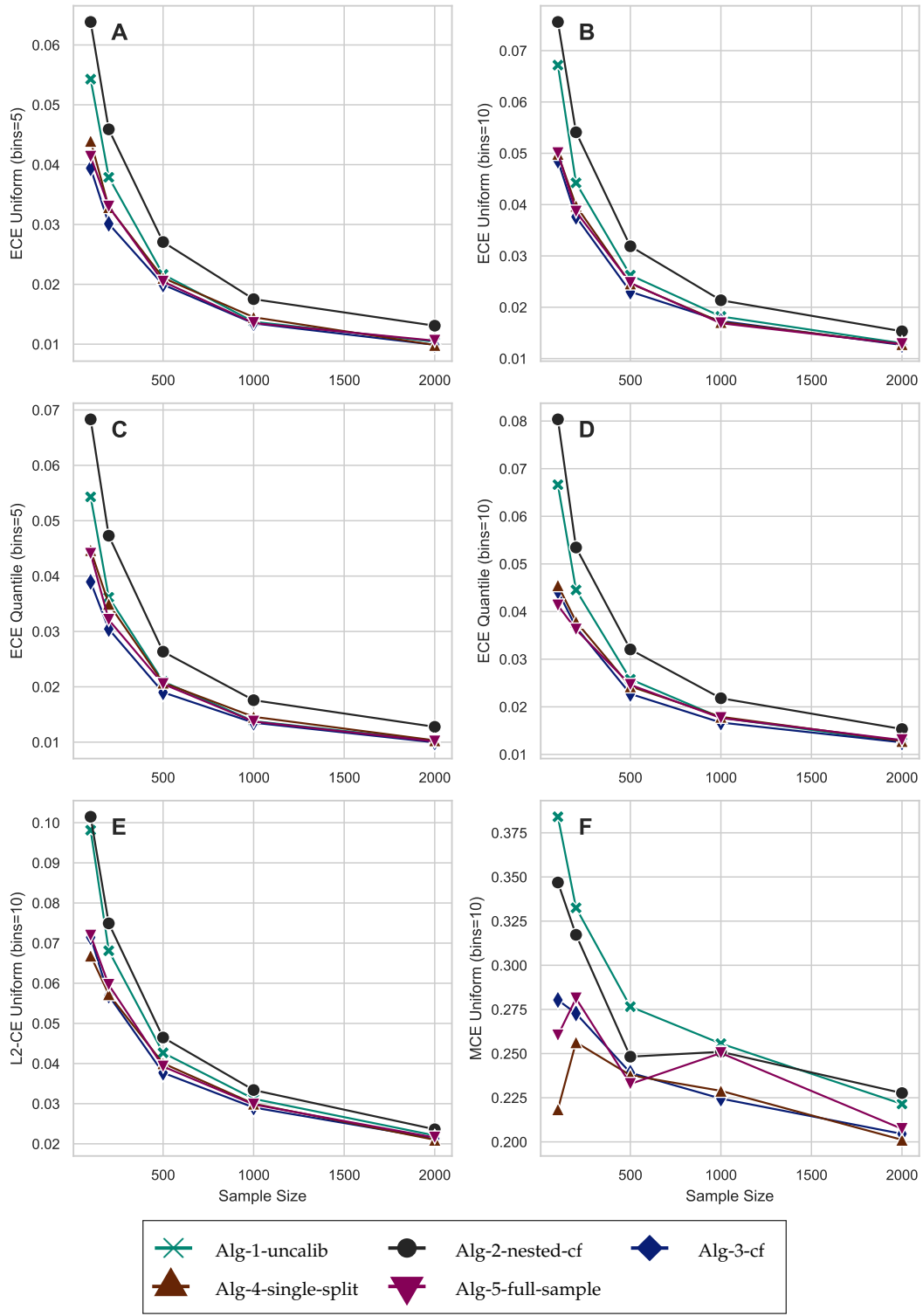

Figure S18: DGP 1 IRM, R2D = 0.5, m = Logit, p = 20, Clipping threshold for Algorithms 1,2,4 = 0.01, Calibration method for Algorithms 2-4: isotonic regression

41

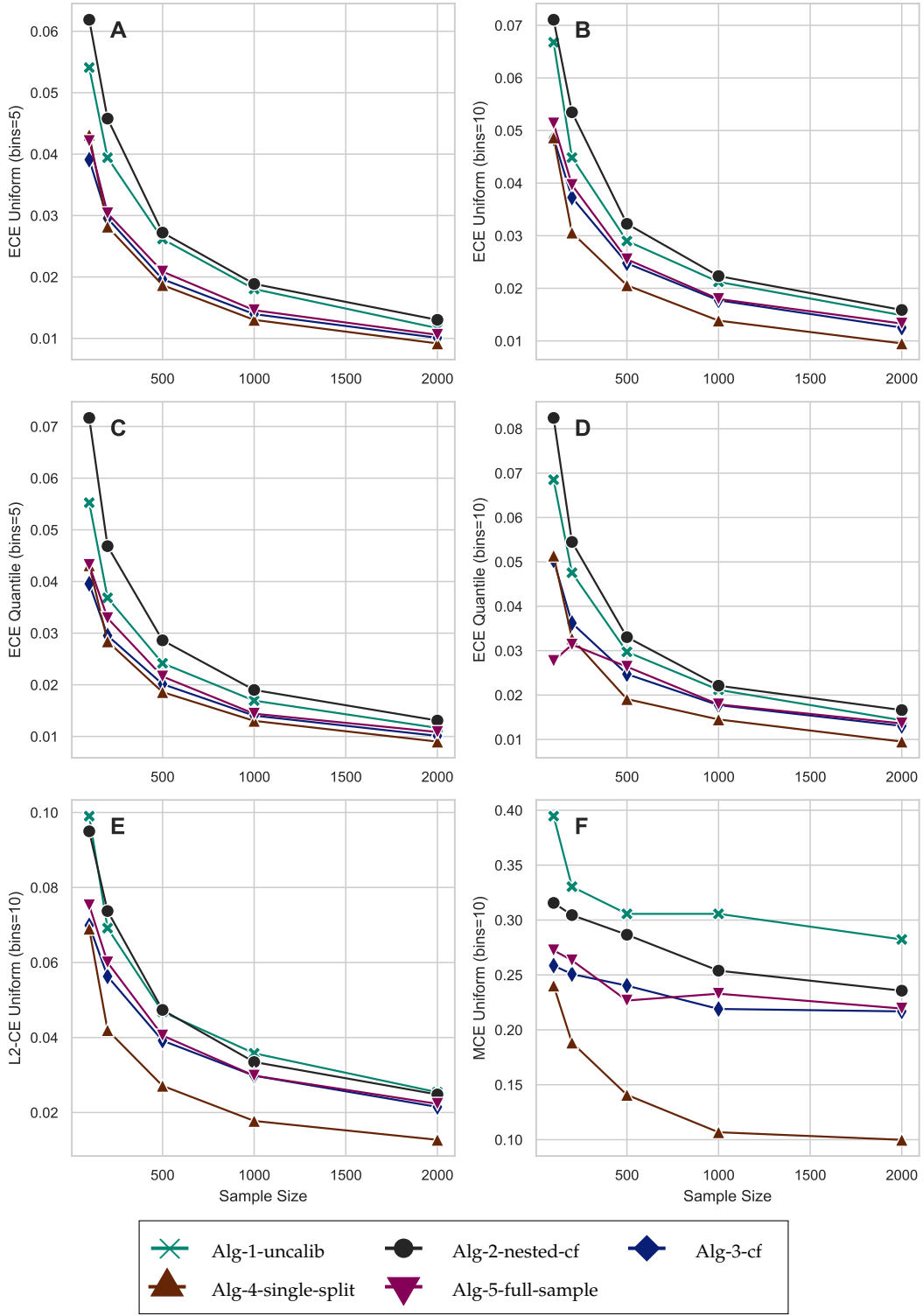

Figure S19: DGP 1 IRM, R2D = 0.5, m = LGBM, p = 20, Calibration method for Algorithms 2-5: isotonic regression

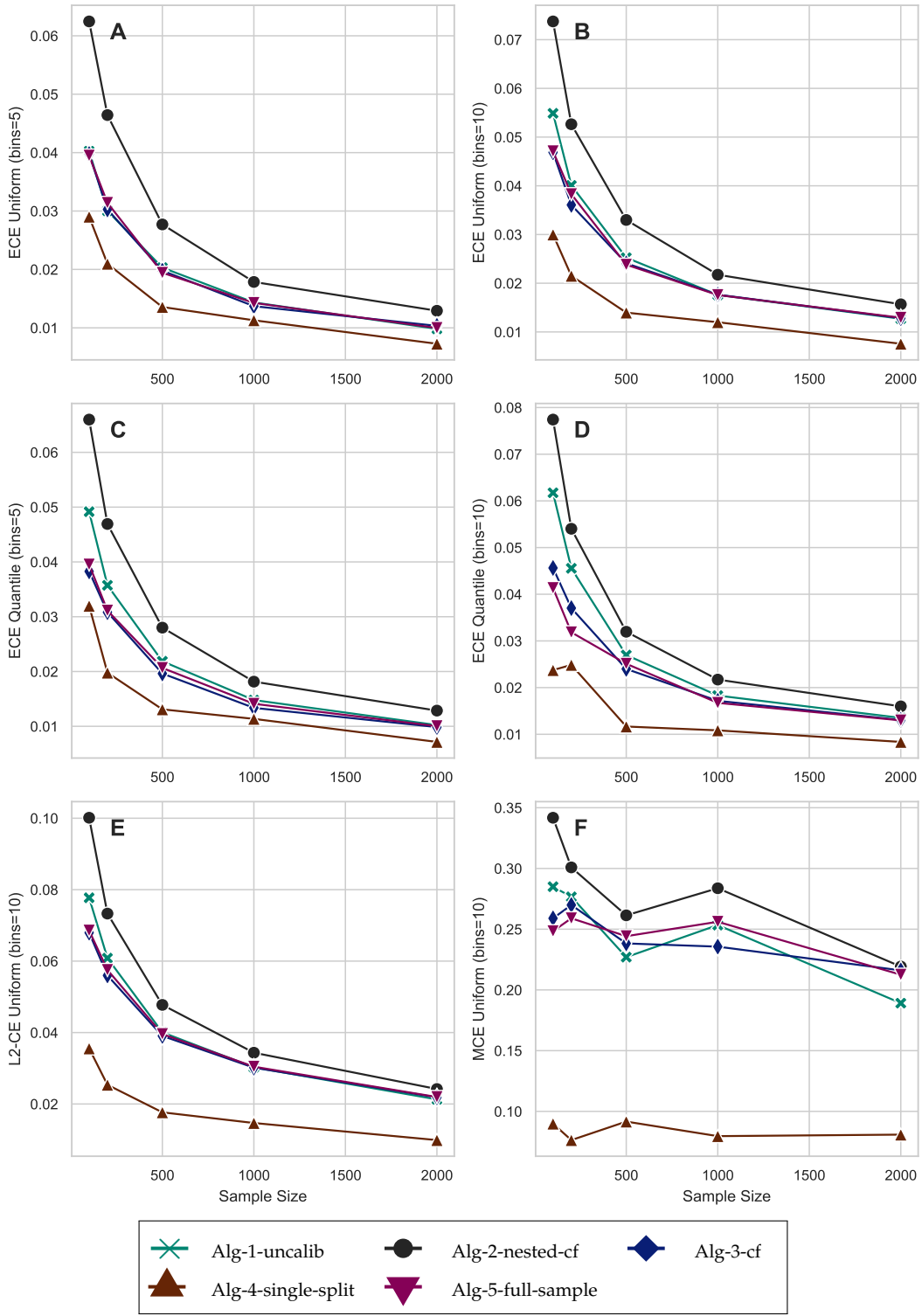

Figure S20: DGP 1 IRM, R2D = 0.5, m = RF, p = 20, Calibration method for Algorithms 2-5: isotonic regression

**DGP 2 Drug**

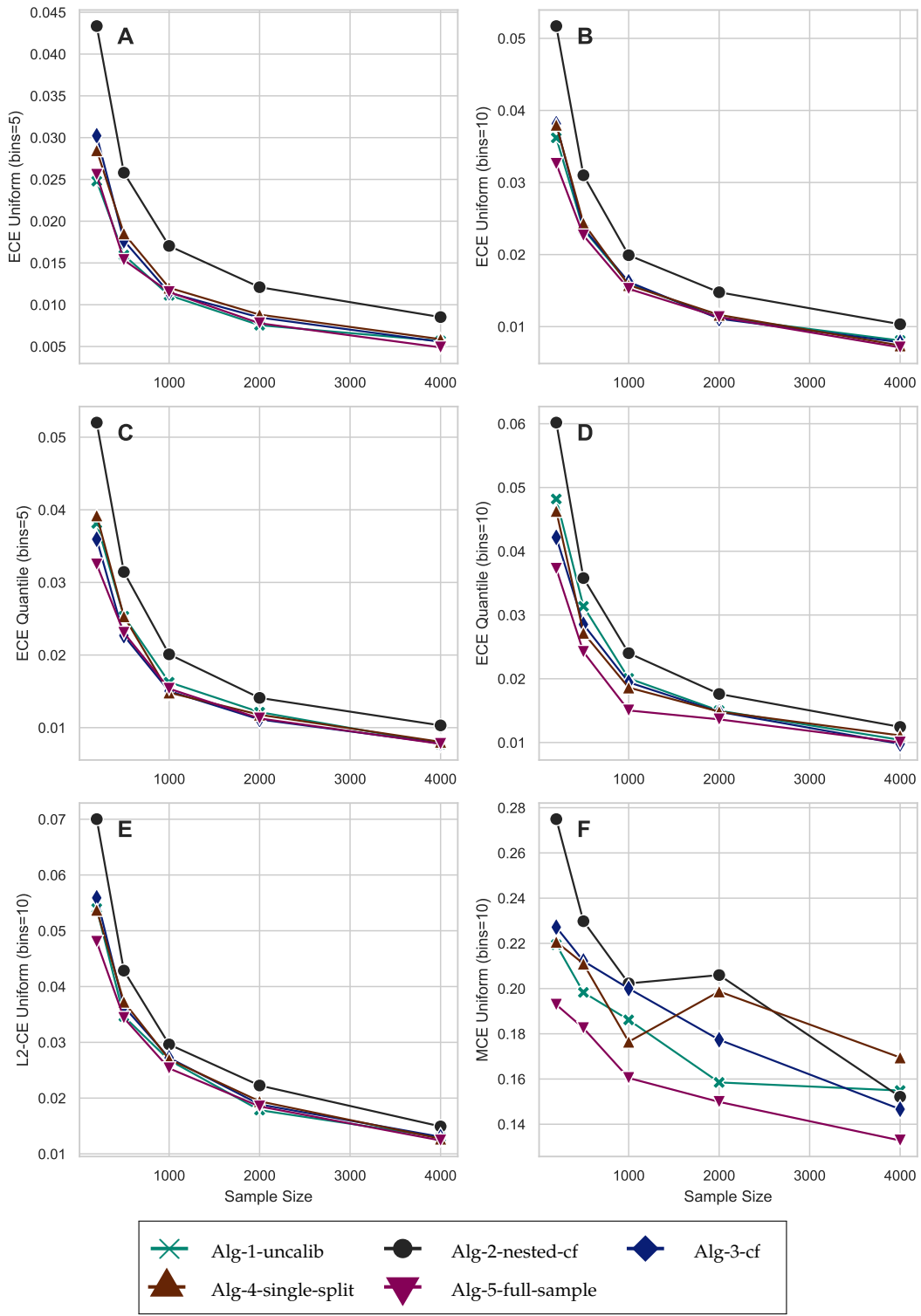

Figure S21: DGP 2 Drug, Overlap = 0.5, m = Logit, n = 2000, p = 3, Calibration method for Algorithms 2-4: isotonic regression

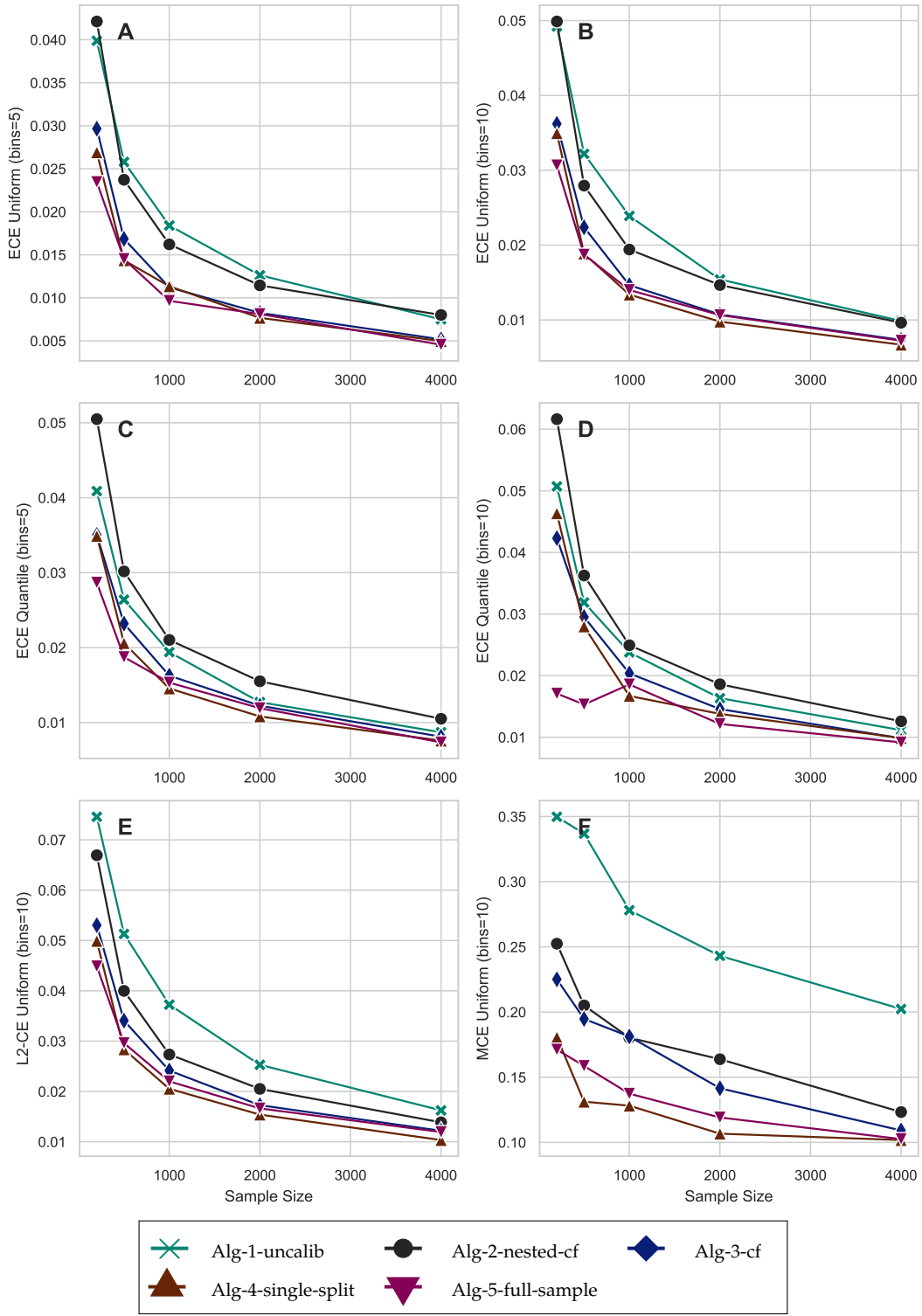

Figure S22: DGP 2 Drug, Overlap = 0.5, m = LGBM, p = 3, Calibration method for Algorithms 2-5: isotonic regression

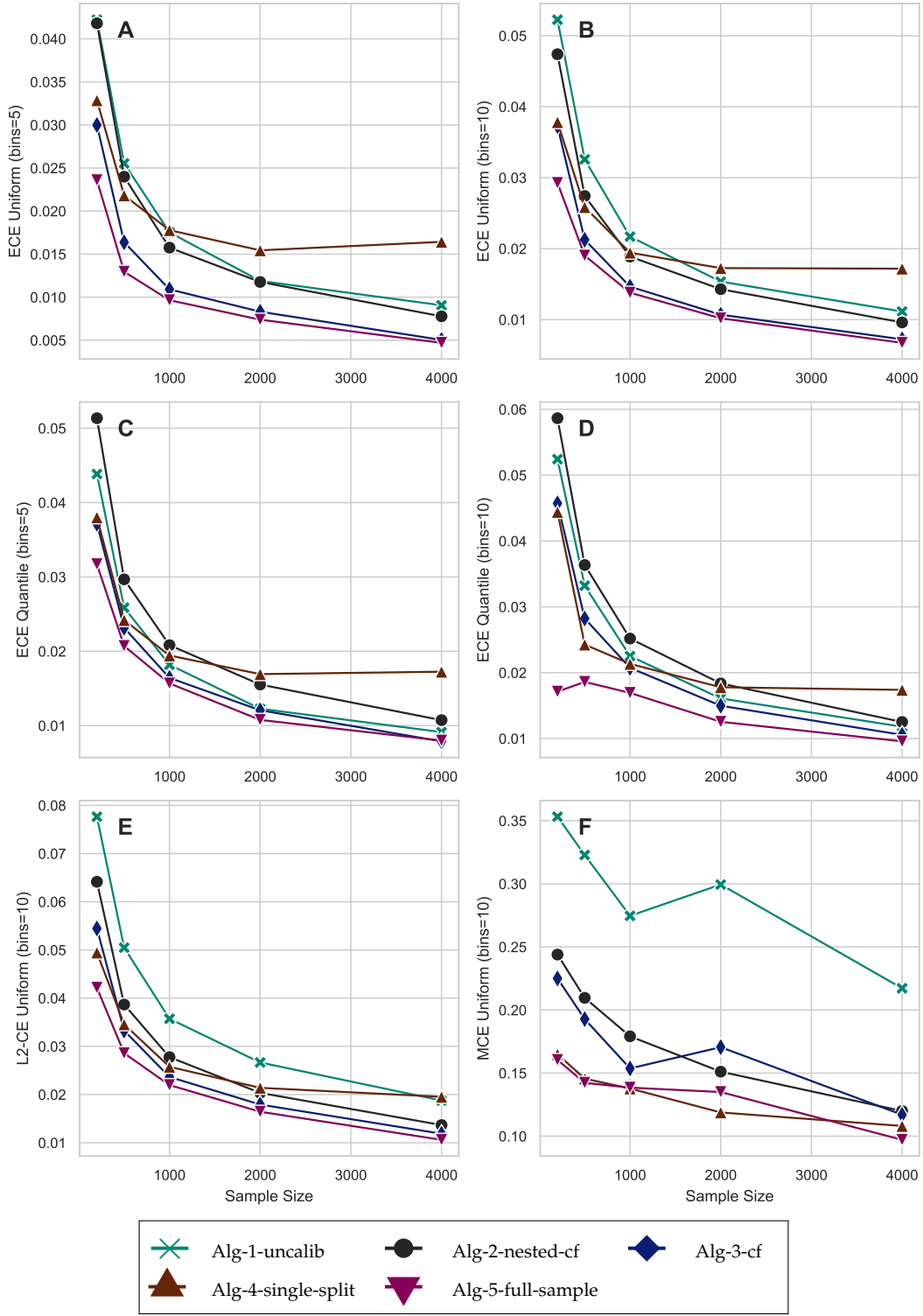

Figure S23: DGP 2 Drug, Overlap = 0.5, m = RF, p = 3, Calibration method for Algorithms 2-5: isotonic regression

**Nonlinear**

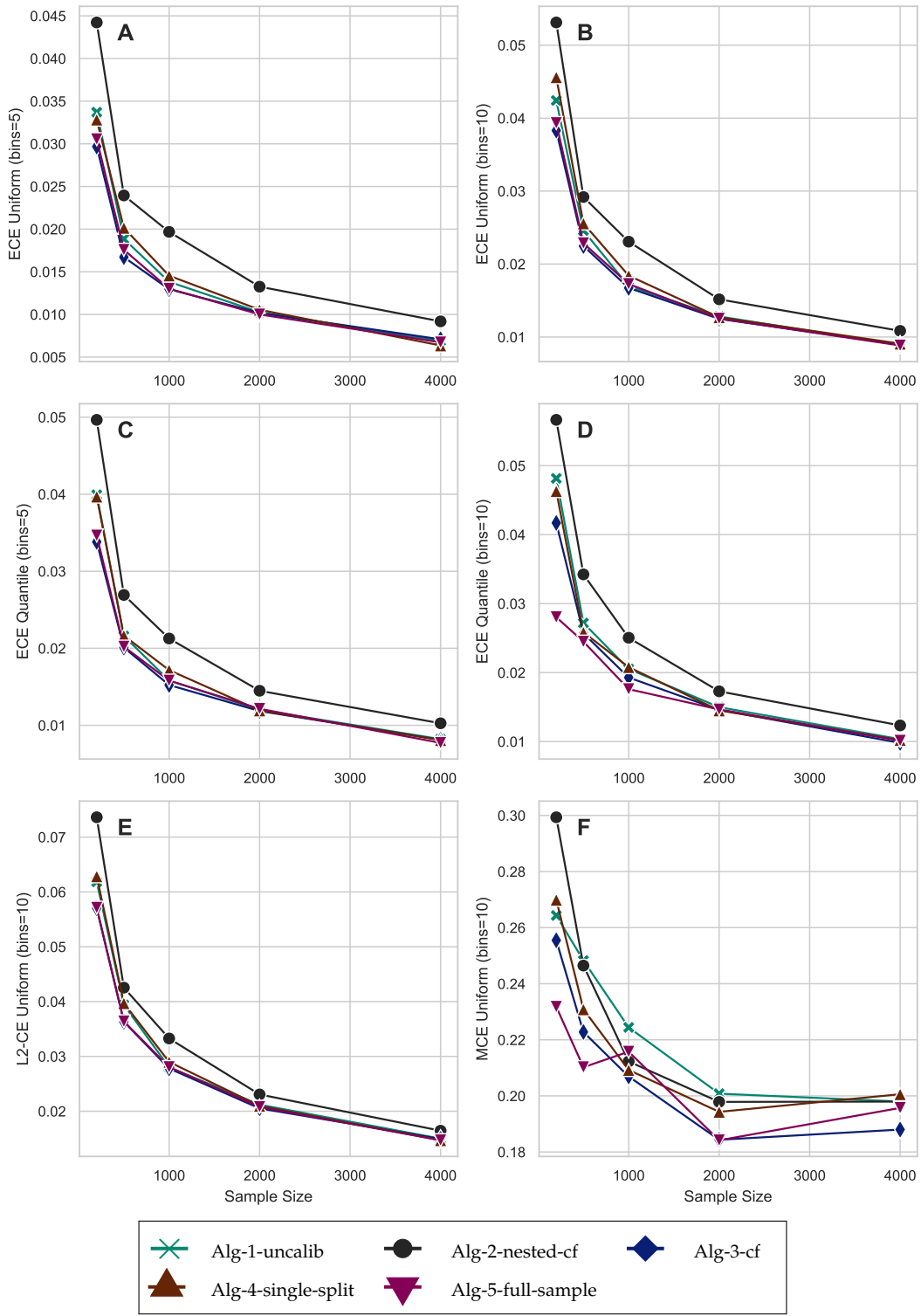

Figure S24: DGP 3 Nonlinear, m = Logit, p = 4, Calibration method for Algorithms 2-4: isotonic regression

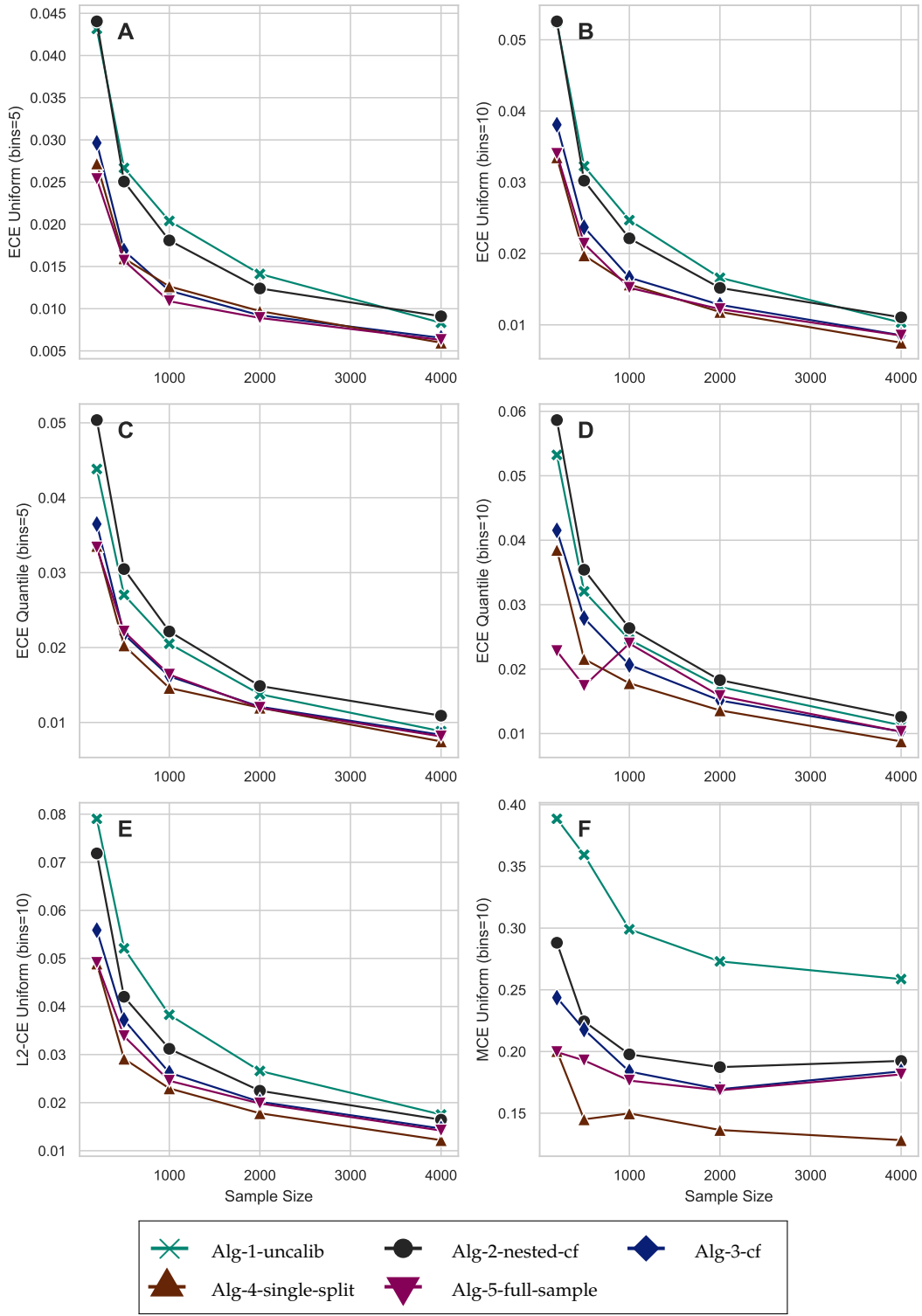

Figure S25: DGP 3 Nonlinear, m = LGBM, p = 4, Calibration method for Algorithms 2-5: isotonic regression

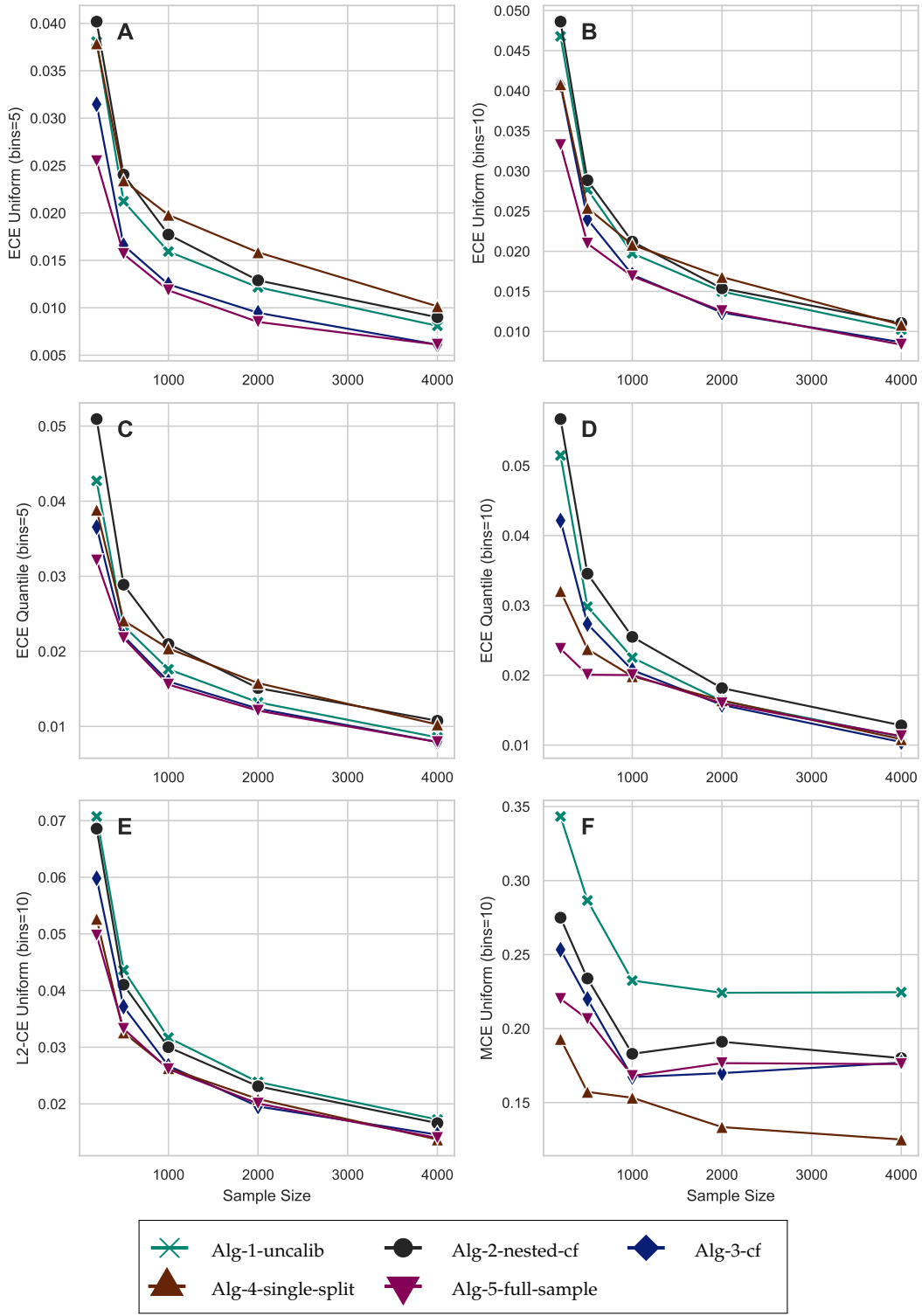

Figure S26: DGP 3 Nonlinear, m = RF, p = 4, Calibration method for Algorithms 2-5: isotonic regression

49

**DGP 4 Unbalanced**

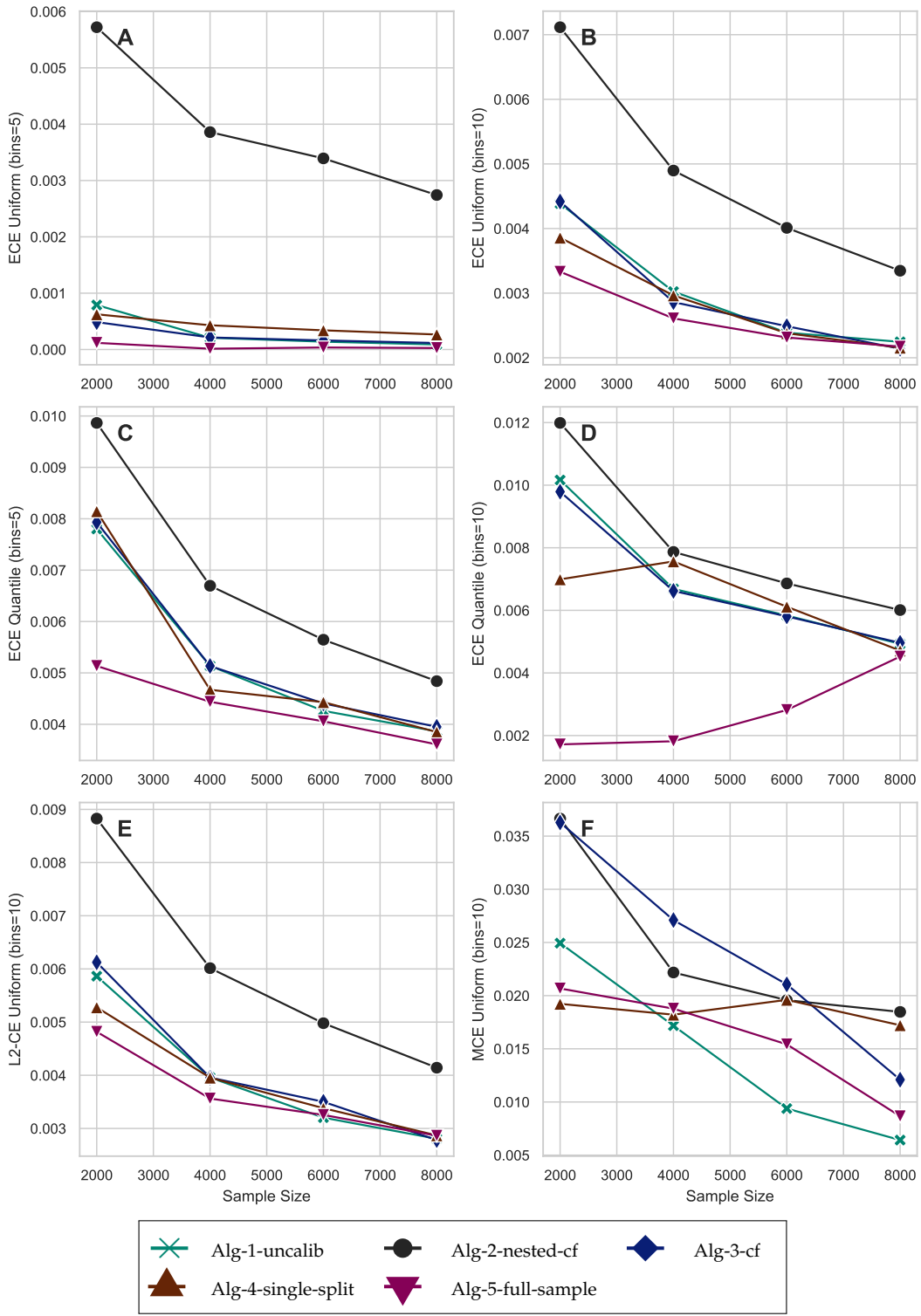

Figure S27: DGP 4 Unbalanced, Share_treated = 0.1, m = Logit, p = 20, Calibration method for Algorithms 2-4: isotonic regression

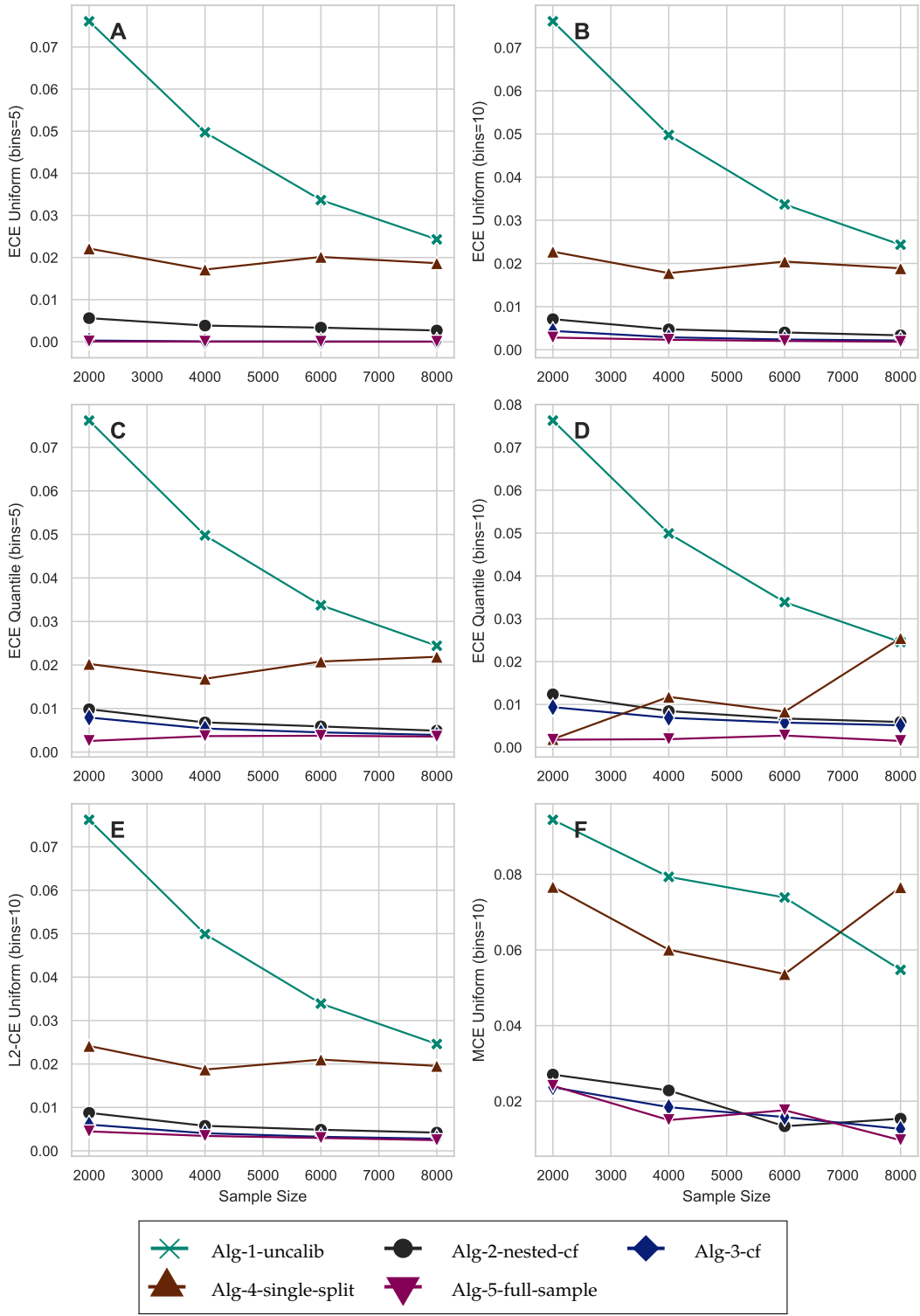

Figure S28: DGP 4 Unbalanced, Share_treated = 0.1, m = LGBM, p = 20, Calibration method for Algorithms 2-5: isotonic regression

51

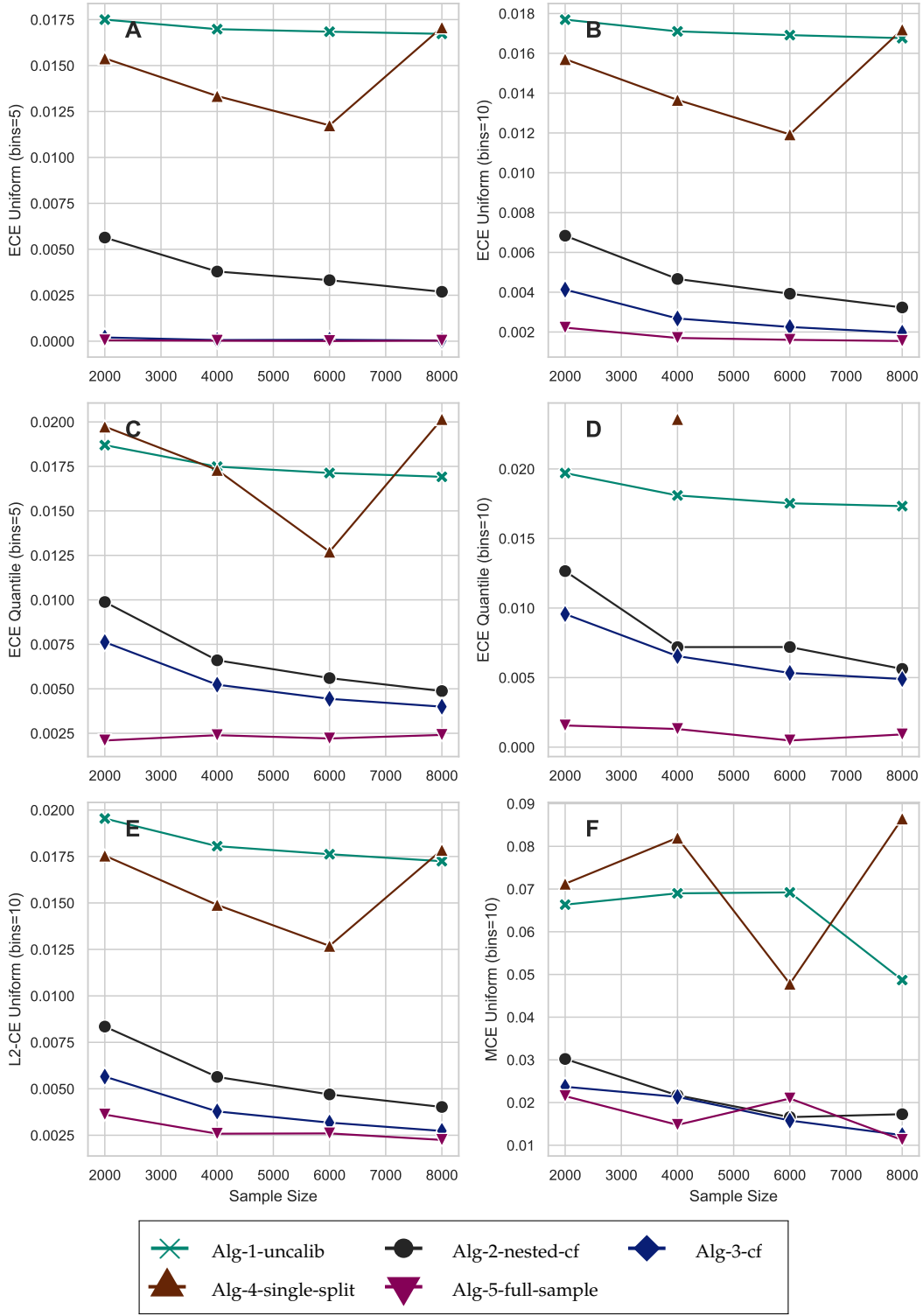

Figure S29: DGP 4 Unbalanced, Share_treated = 0.1, m = RF, p = 20, Calibration method for Algorithms 2-5: isotonic regression

# ATE Errors on Sample Size

## DGP 1 IRM

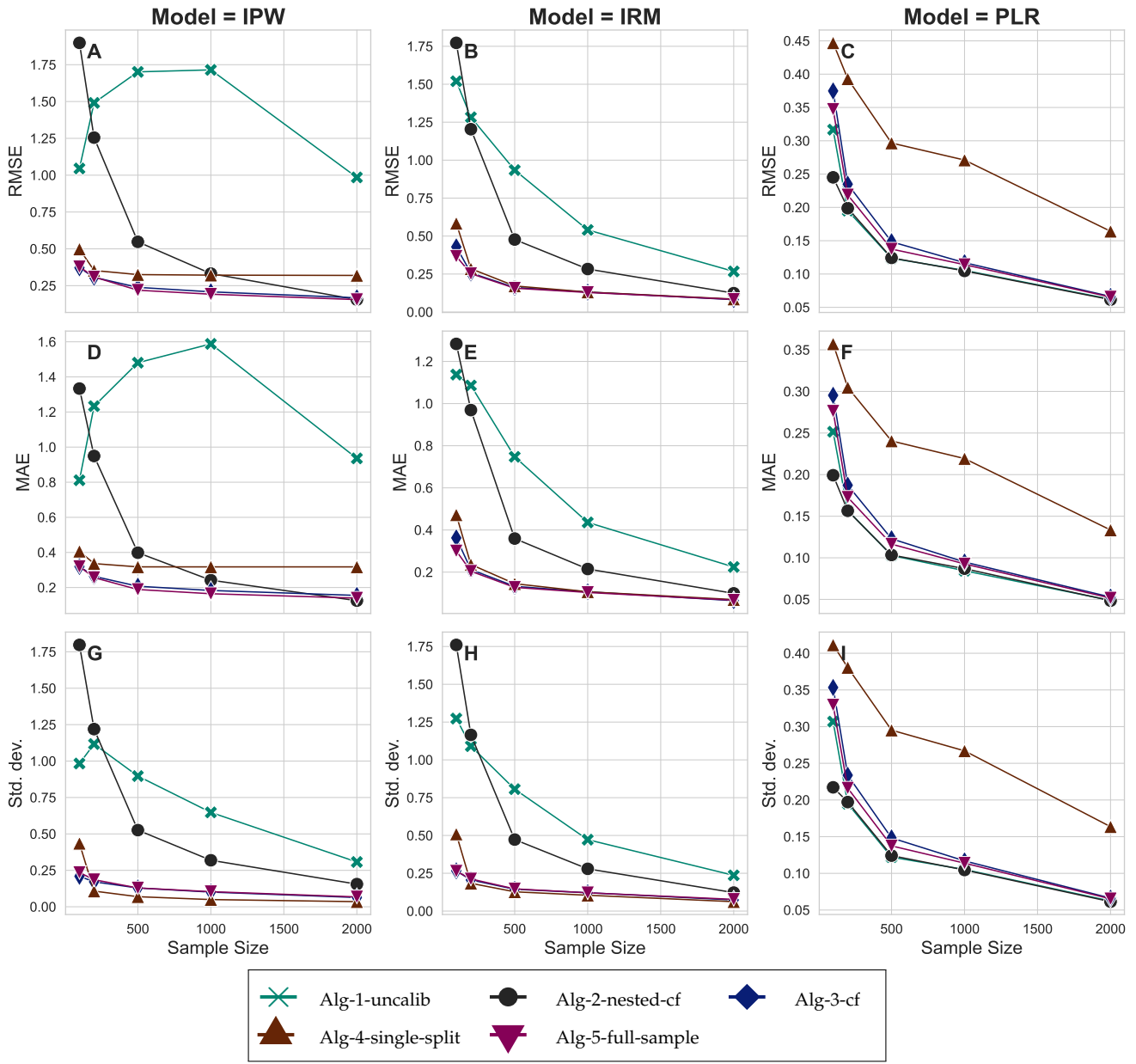

Figure S30: DGP 1 IRM, R2D = 0.5, m = LGBM, g = LGBM, p = 20, Calibration method for Algorithms 2-5: isotonic regression, Clip = 0.01

53

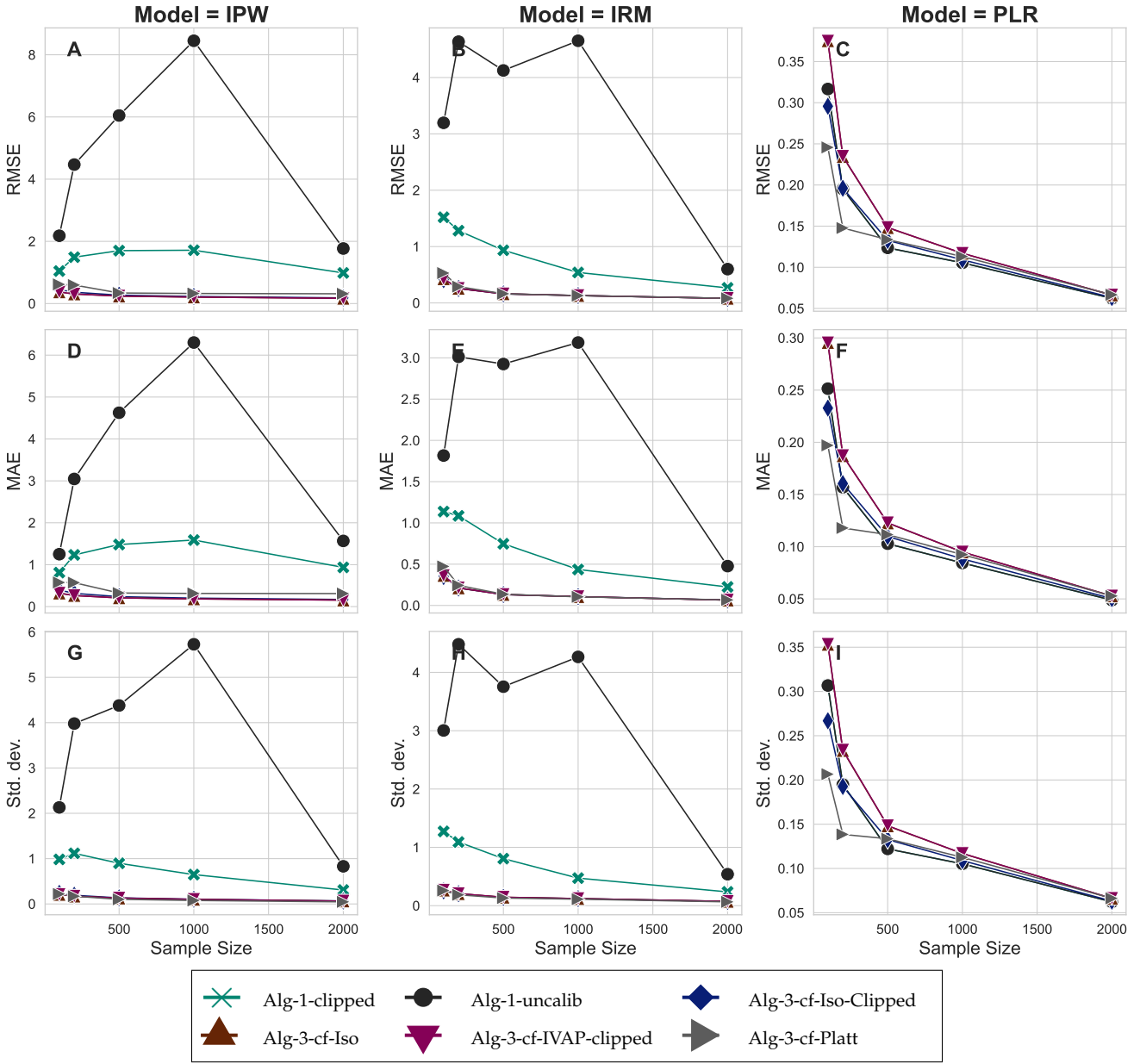

Figure S31: DGP 1 IRM, different calibration methods for Algorithm 3, R2D = 0.5, m = LGBM, g = LGBM, p = 20, , Clip = 0.01

**DGP 2 Drug**

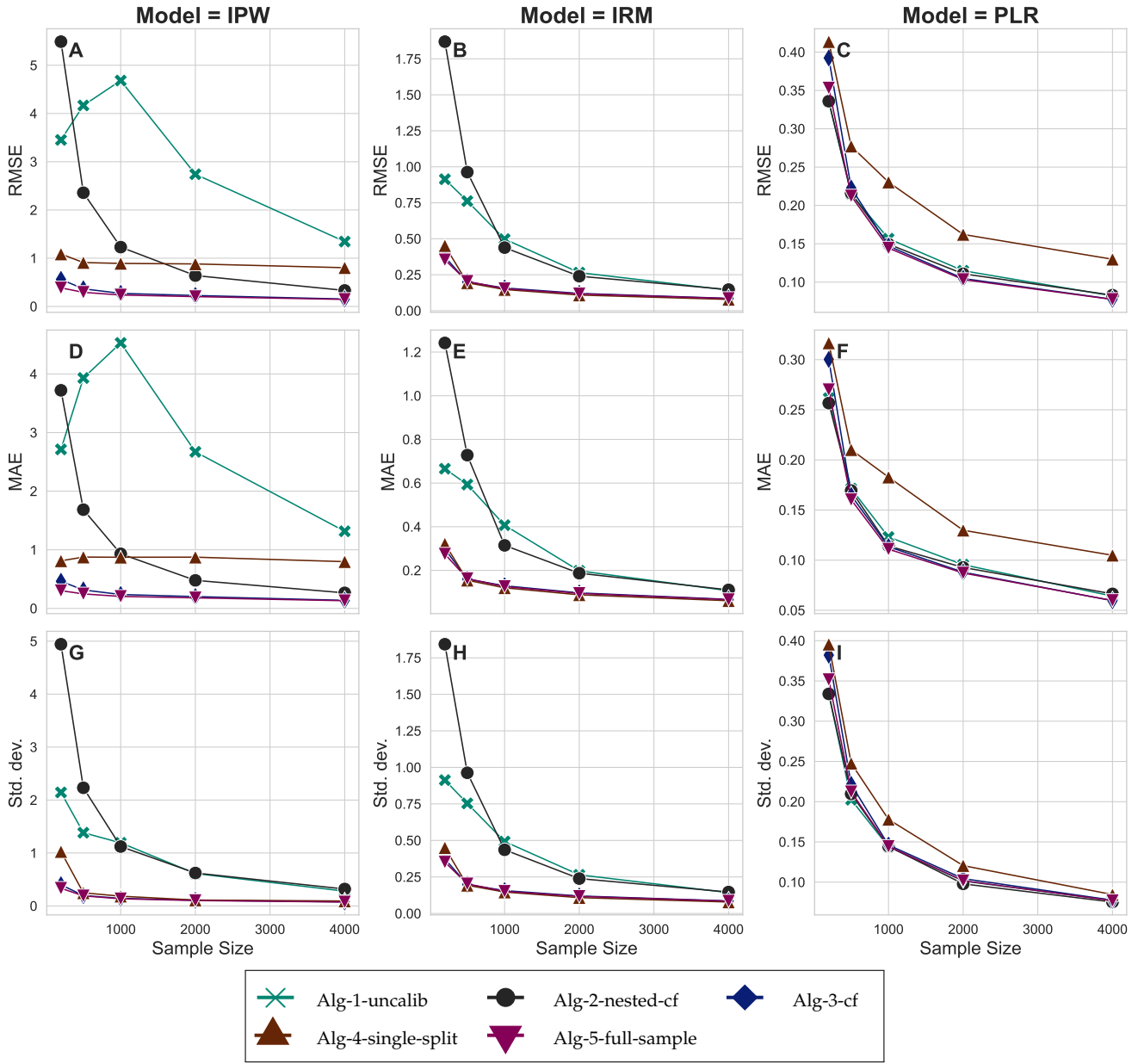

Figure S32: DGP 2 Drug, Overlap = 0.5, m = LGBM, g = LGBM, p = 3, Calibration method for Algorithms 2-5: isotonic regression, Clip = 0.01

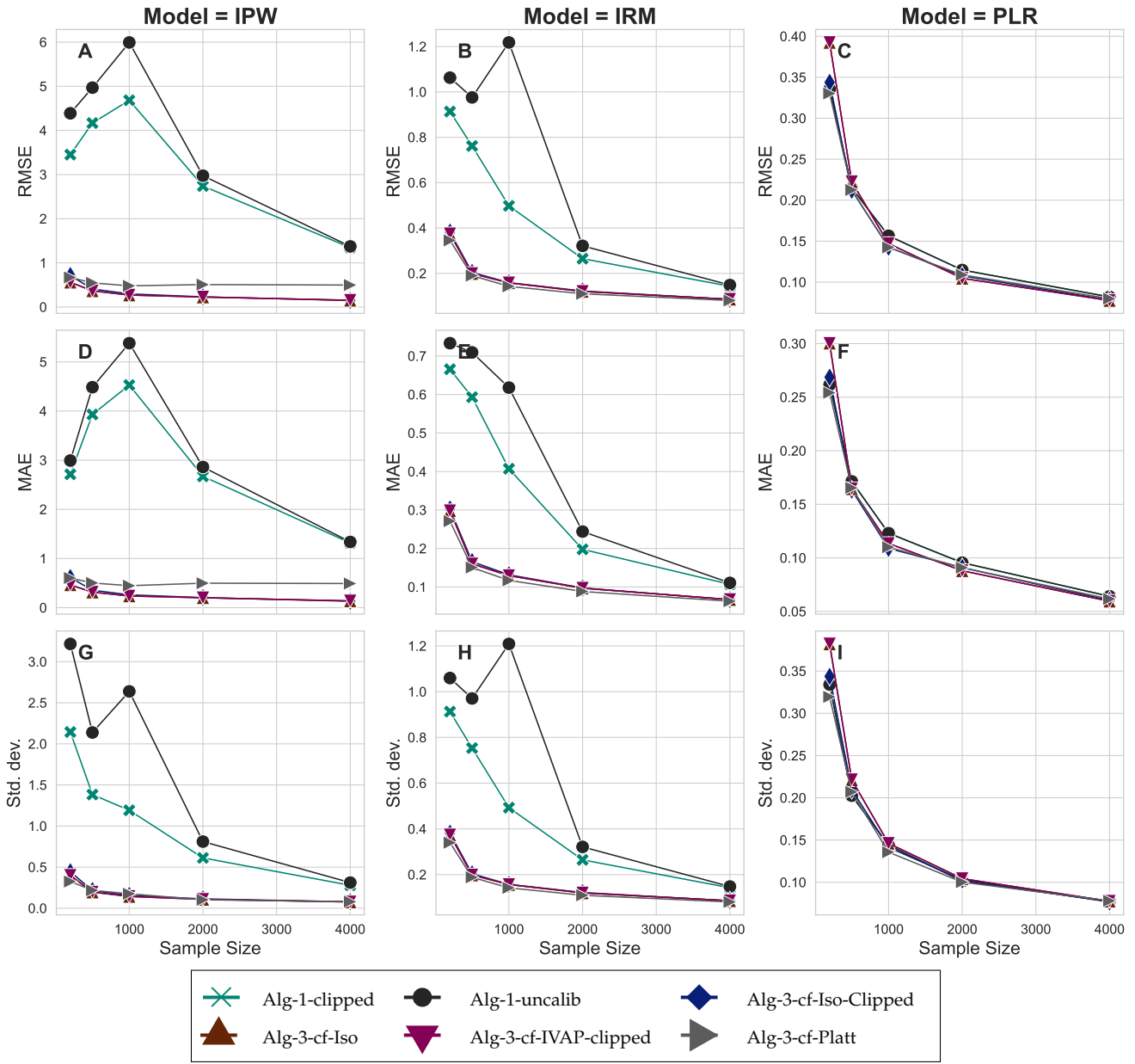

Figure S33: DGP 2 Drug, different calibration methods for Algorithm 3, Overlap = 0.5, m = LGBM, g = LGBM, p = 3, Clip = 0.01

**DGP 3 Nonlinear**

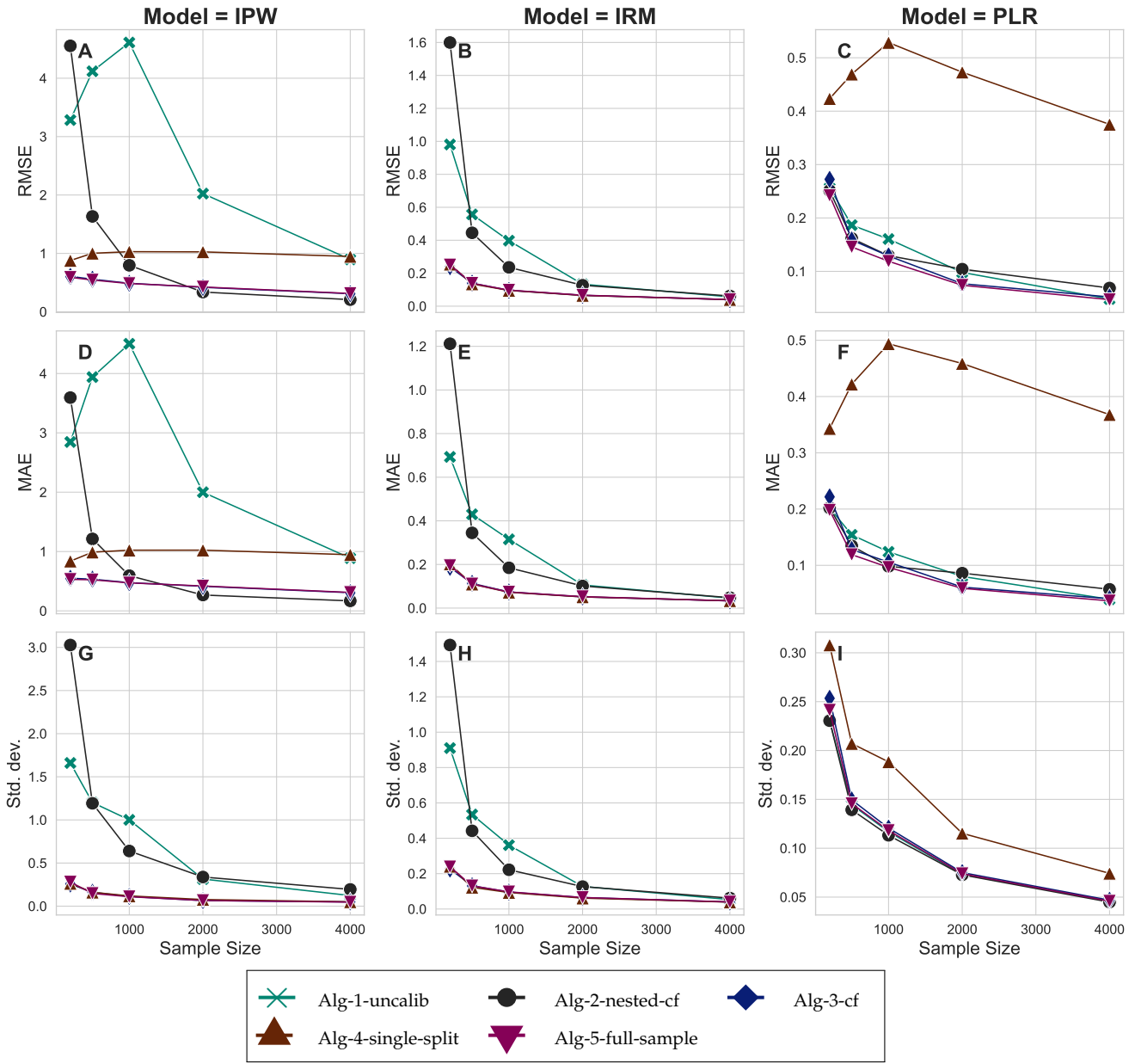

Figure S34: DGP 3 Nonlinear, m = LGBM, g = LGBM, p = 4, Calibration method for Algorithms 2-5: isotonic regression, Clip = 0.01

57

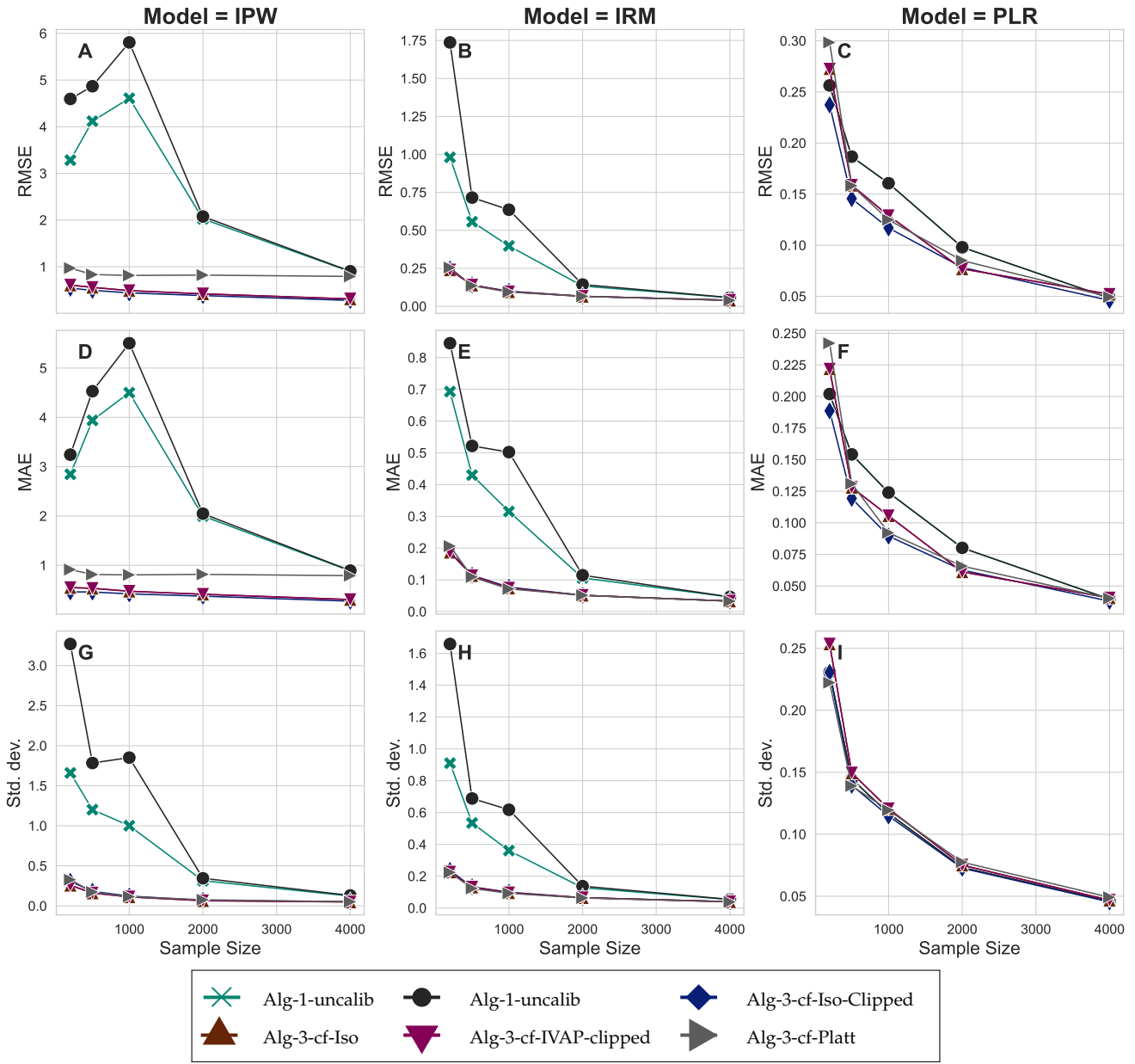

Figure S35: DGP 3 Nonlinear, different calibration methods for Algorithm 3, m = LGBM, g = LGBM, p = 4, Clip = 0.01

**DGP 4 Unbalanced**

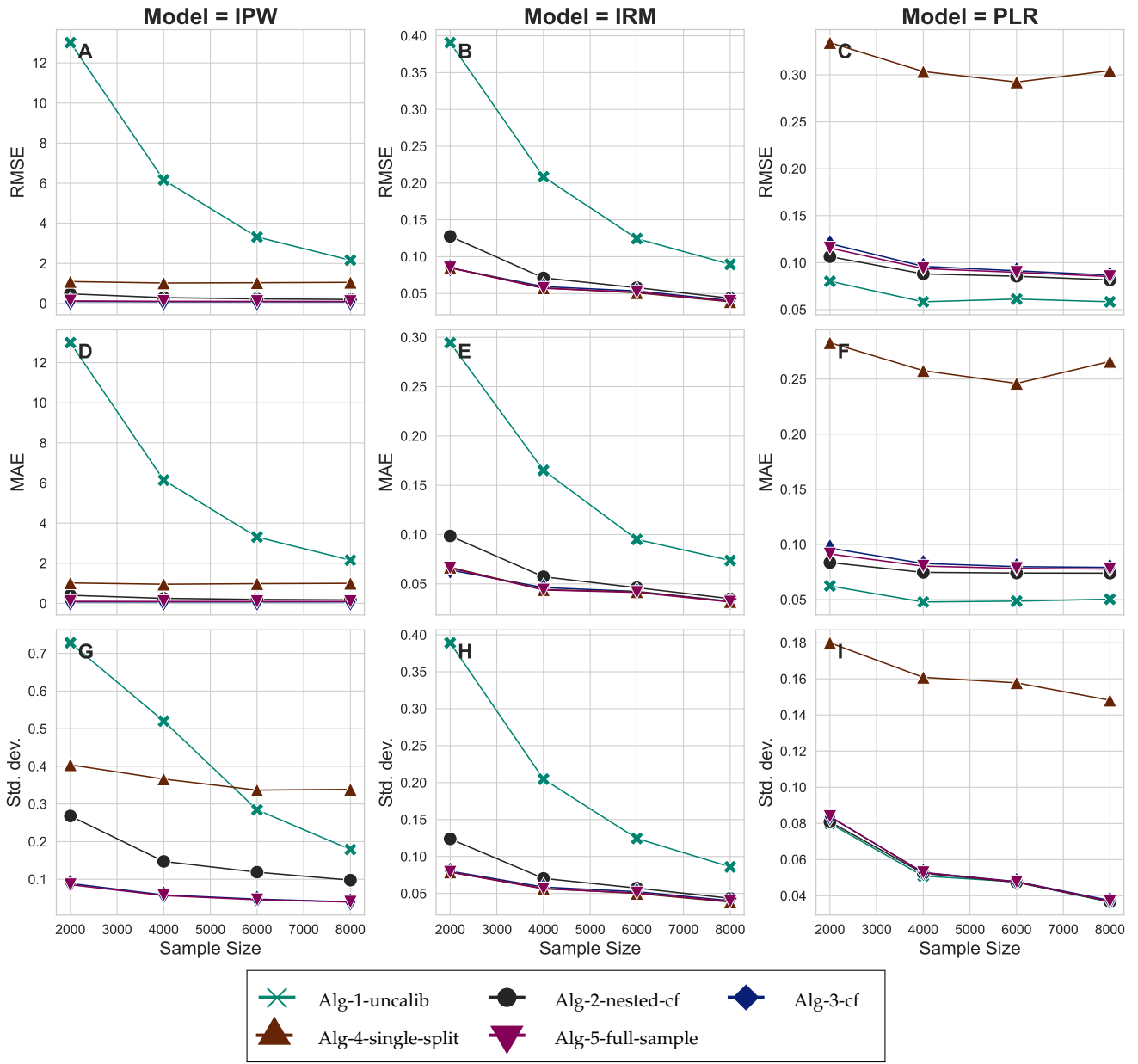

Figure S36: DGP 4 Unbalanced, Share_treated = 0.1, m = LGBM, g = LGBM, p = 20, Calibration method for Algorithms 2-5: isotonic regression, Clip = 0.01

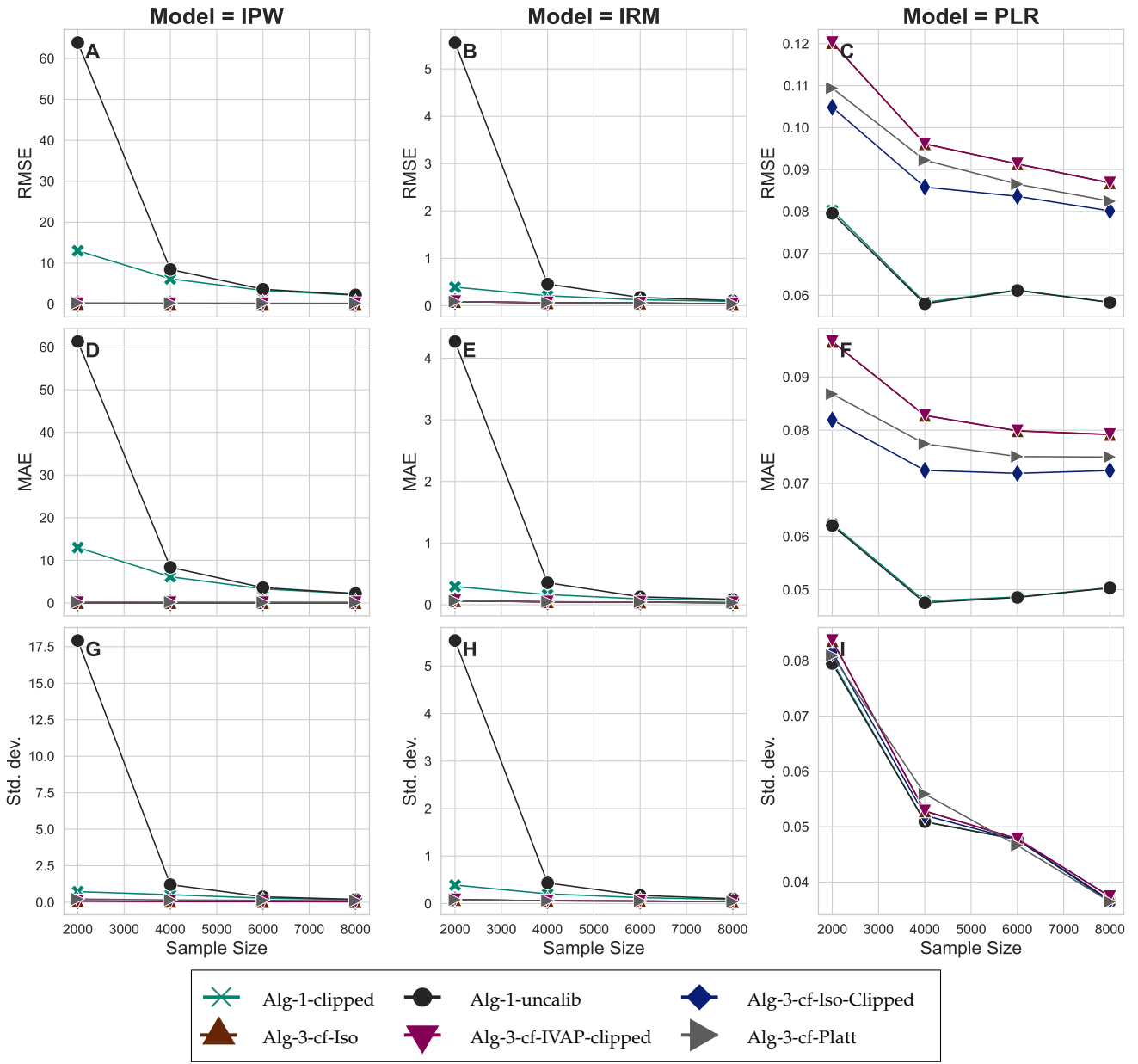

Figure S37: DGP 4 Unbalanced, different calibration methods for Algorithm 3, Share_treated = 0.1, m = LGBM, g = LGBM, p = 20, Clip = 0.01

**Propensity Learner on ATE Distribution**

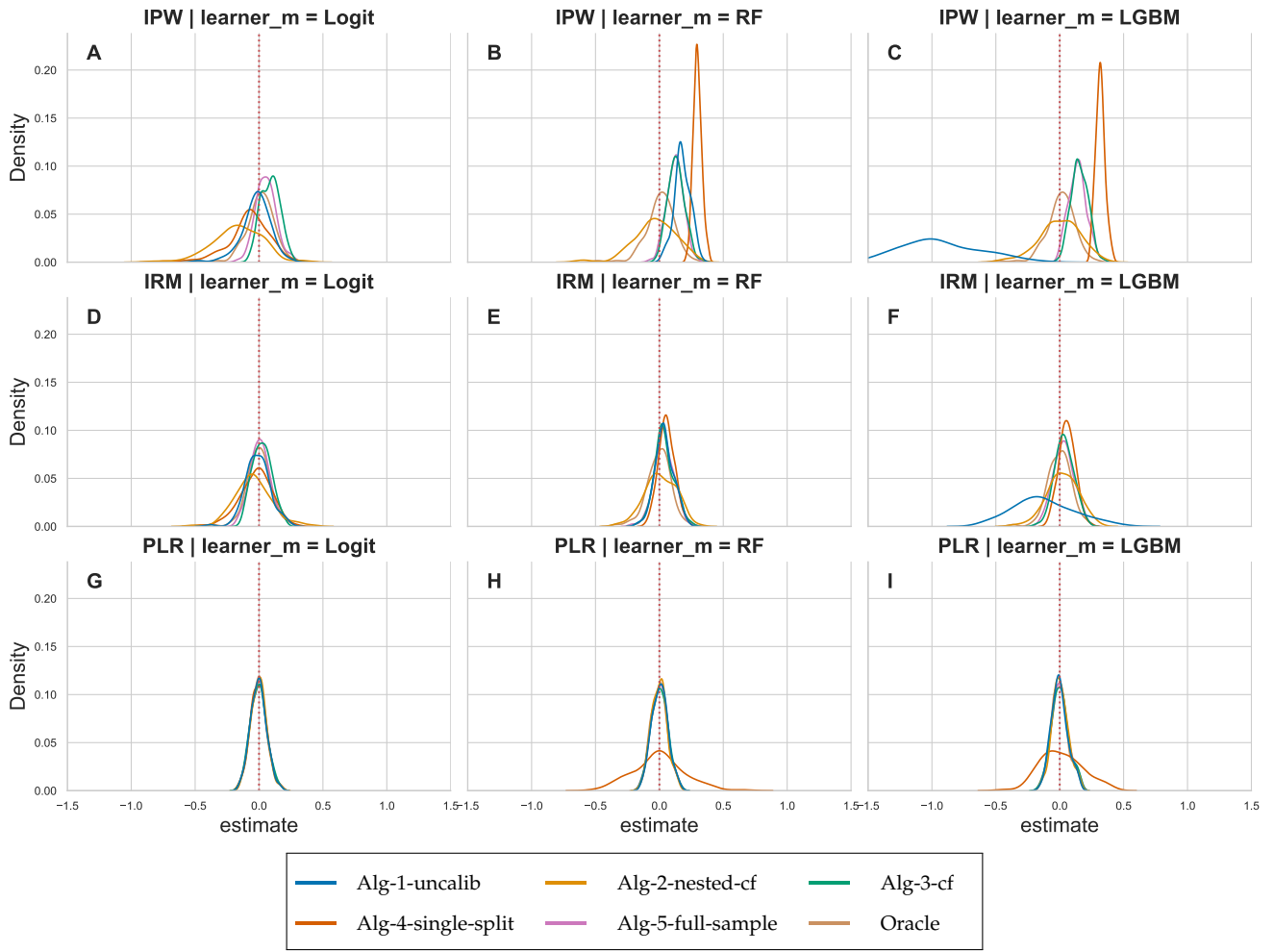

Figure S38: DGP 1 IRM, R2D = 0.5, m = LGBM, g = LGBM, n = 2000, p = 20, Calibration method for Algorithms 2-5: isotonic regression, Clip = 0.01

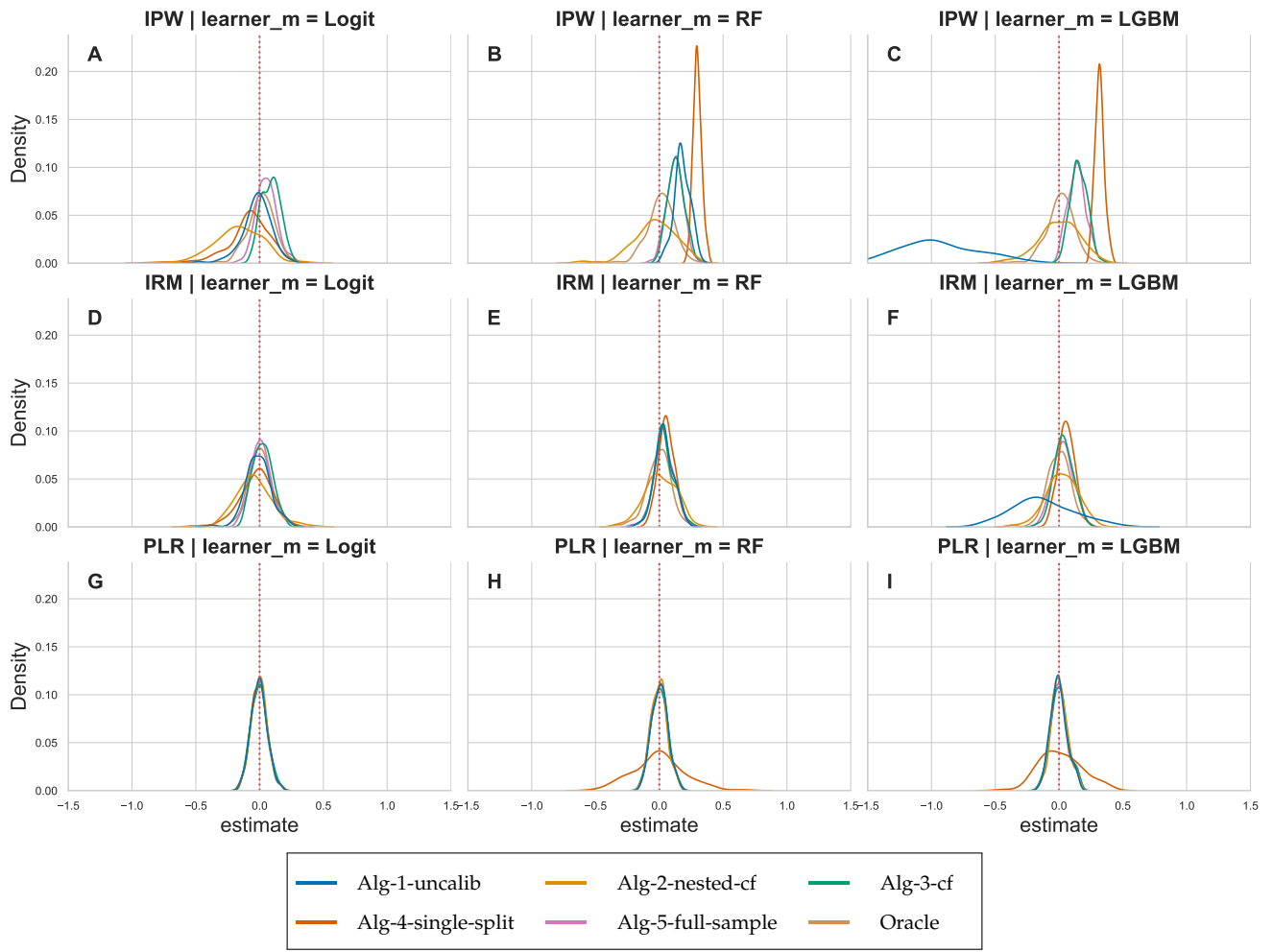

Figure S39: DGP 1 IRM, R2D = 0.5, m = LGBM, g = LGBM, n = 2000, p = 20, Calibration method for Algorithms 2-5: isotonic regression, Clip = 0.01

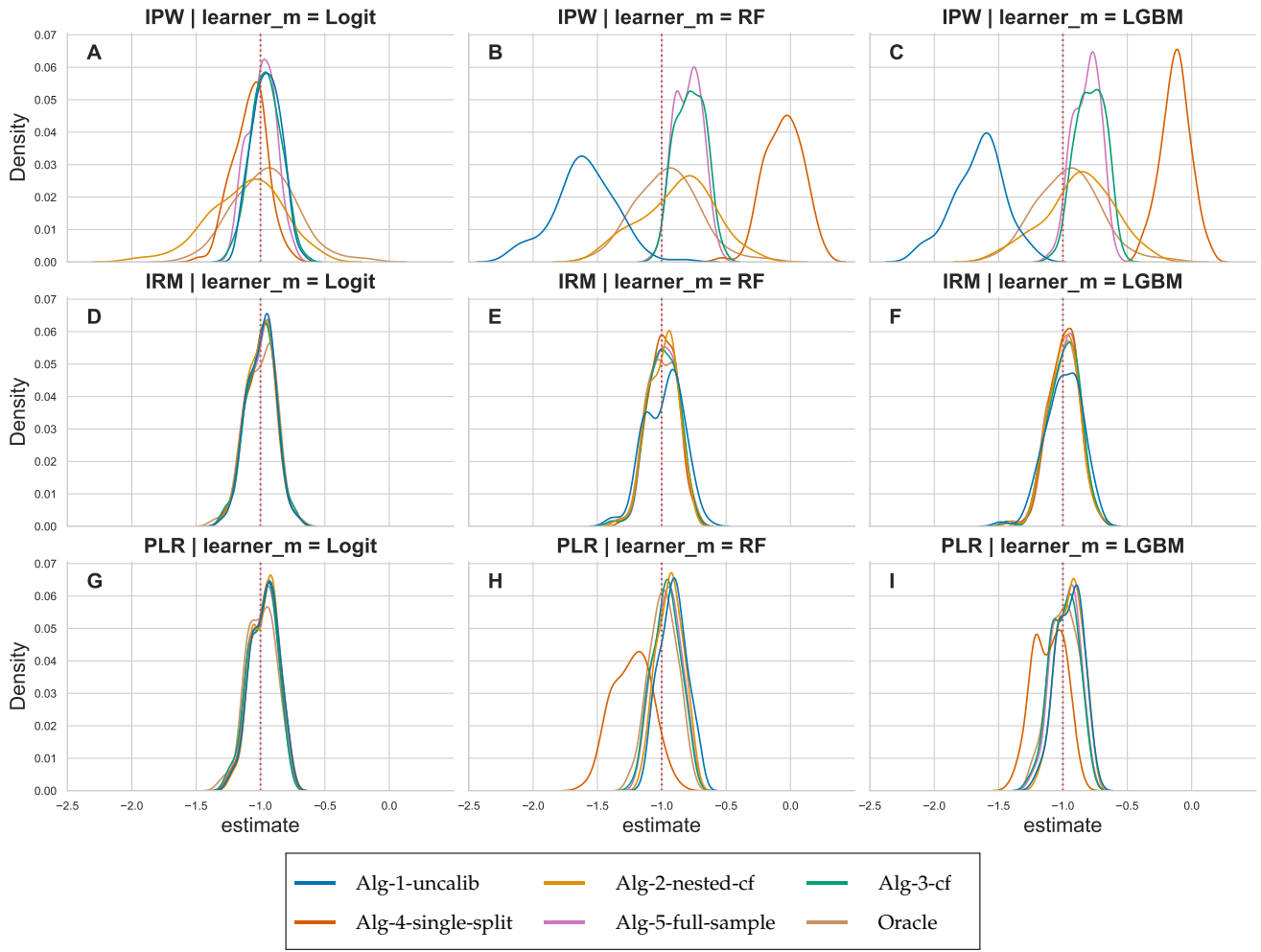

Figure S40: DGP 2 Drug, Overlap = 0.5, m = LGBM, g = LGBM, n = 2000, p = 3, Calibration method for Algorithms 2-5: isotonic regression, Clip = 0.01

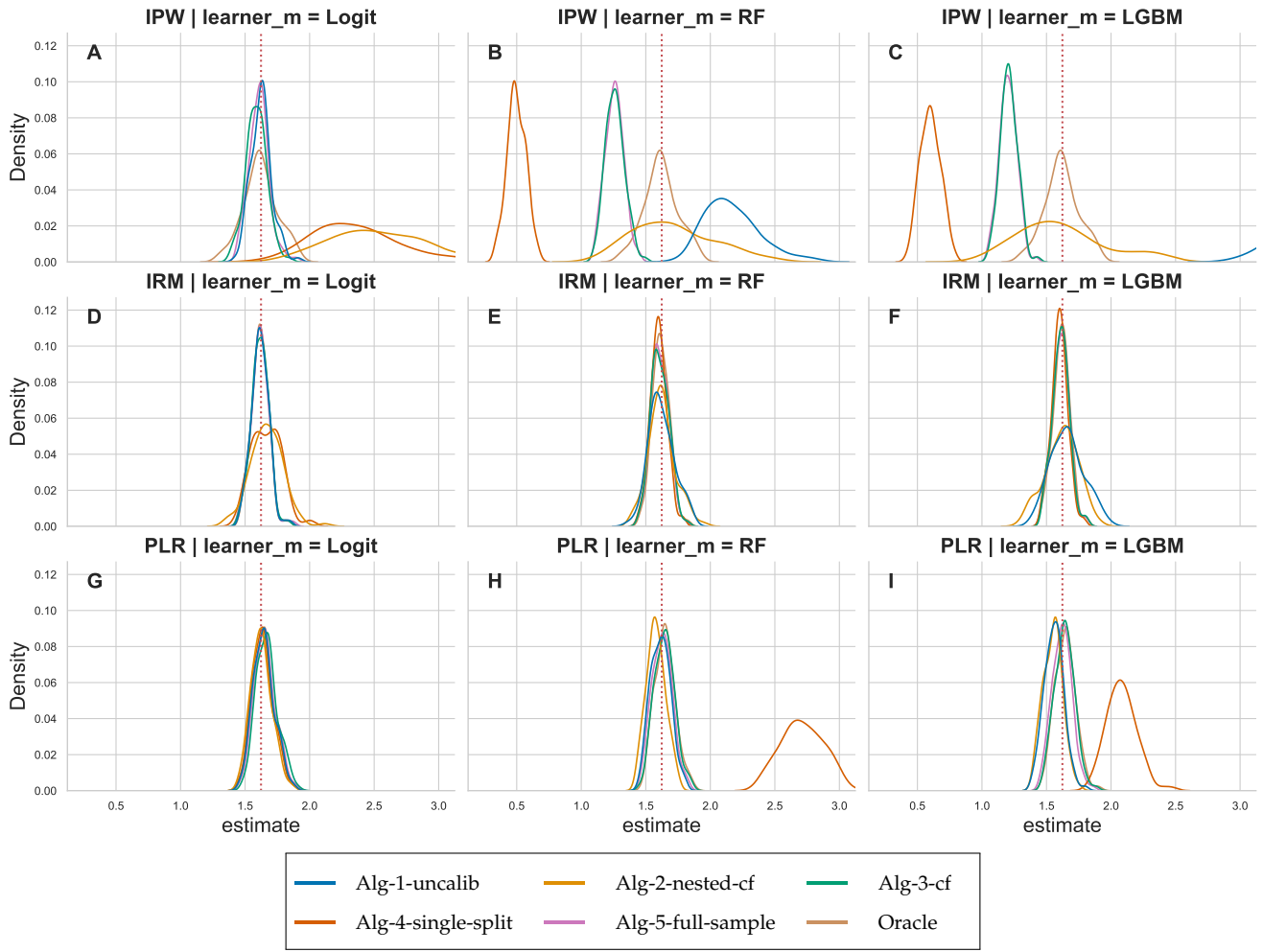

Figure S41: DGP 3 Nonlinear, m = LGBM, g = LGBM, n = 2000, p = 4, Calibration method for Algorithms 2-5: isotonic regression, Clip = 0.01

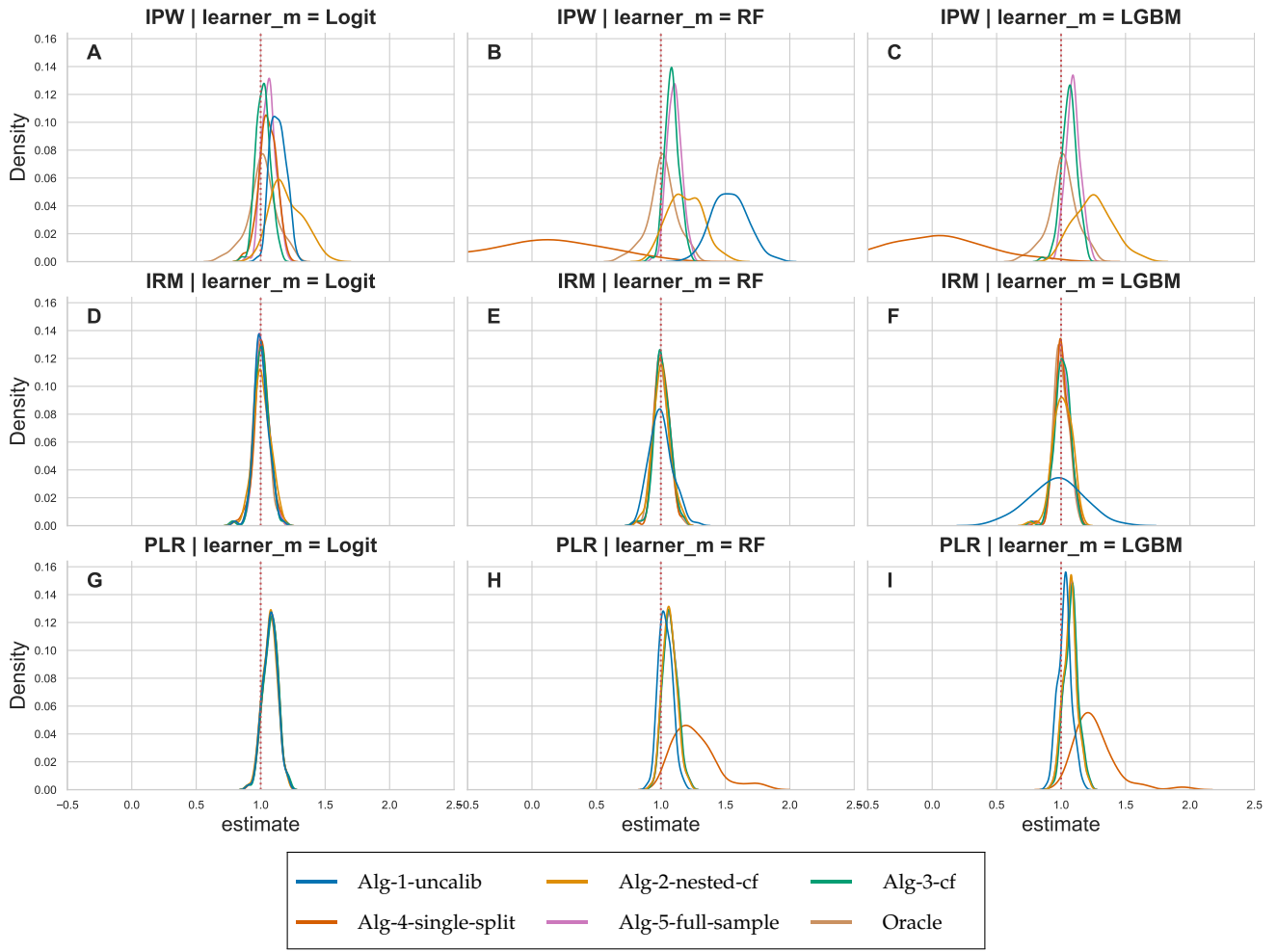

Figure S42: DGP 4 Unbalanced, Share_treated = 0.1, m = LGBM, g = LGBM, n = 4000, p = 20, Calibration method for Algorithms 2-5: isotonic regression, Clip = 0.01

**Outcome Learner on ATE Distribution**

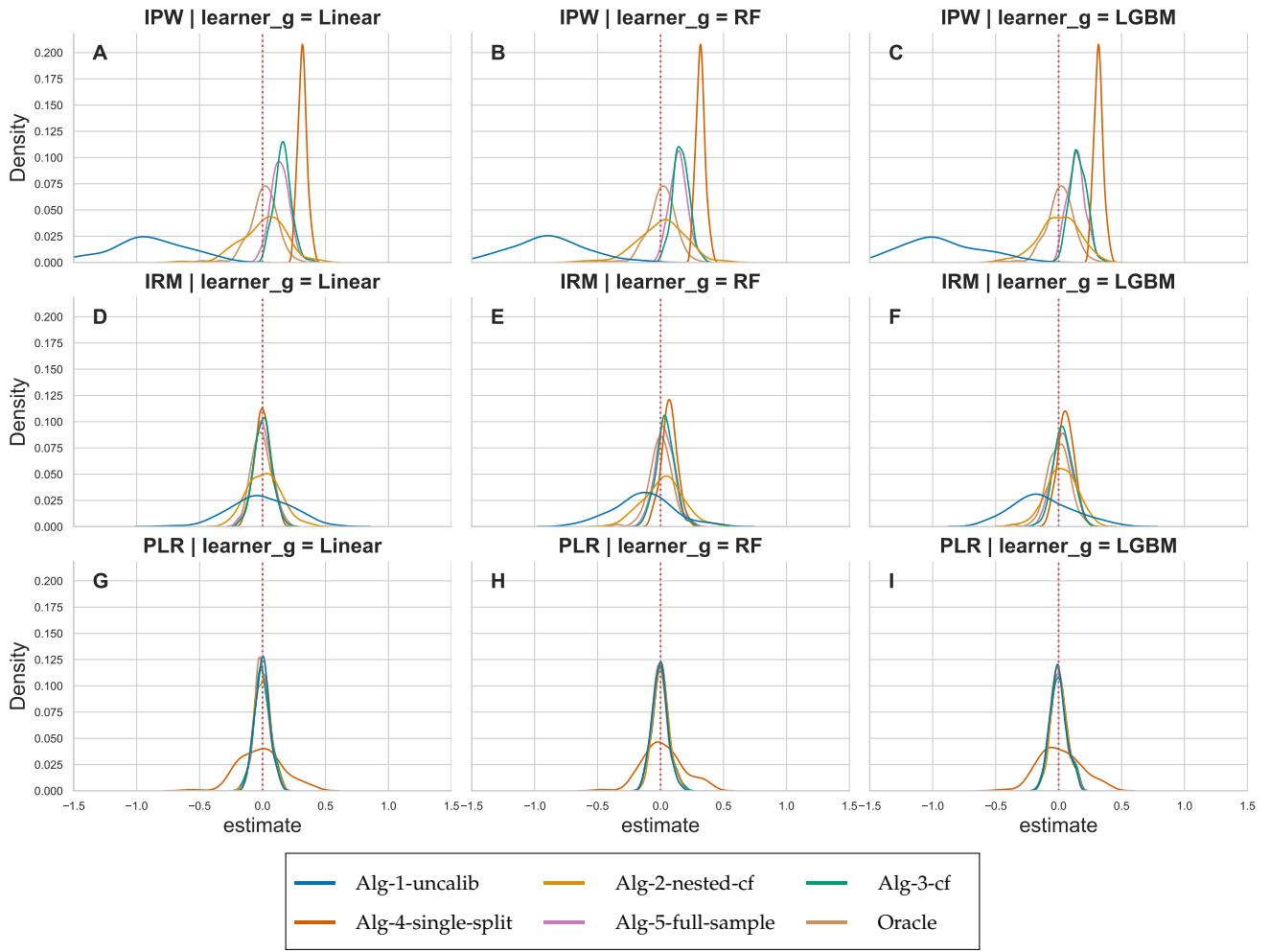

Figure S43: DGP 1 IRM, R2D = 0.5, m = LGBM, g = LGBM, n = 2000, p = 20, Calibration method for Algorithms 2-5: isotonic regression, Clip = 0.01

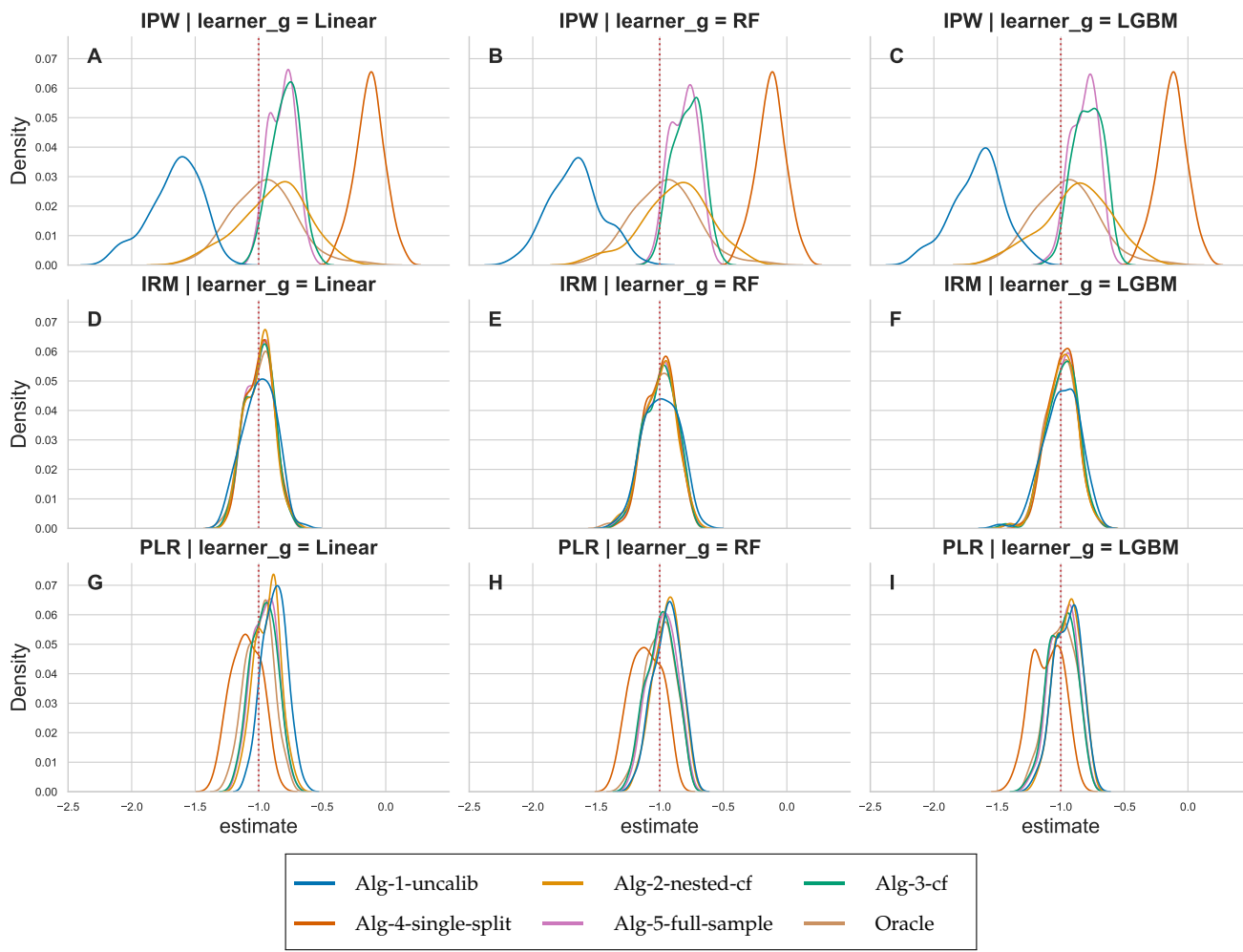

Figure S44: DGP 2 Drug, Overlap = 0.5, m = LGBM, g = LGBM, n = 2000, p = 3, Calibration method for Algorithms 2-5: isotonic regression, Clip = 0.01

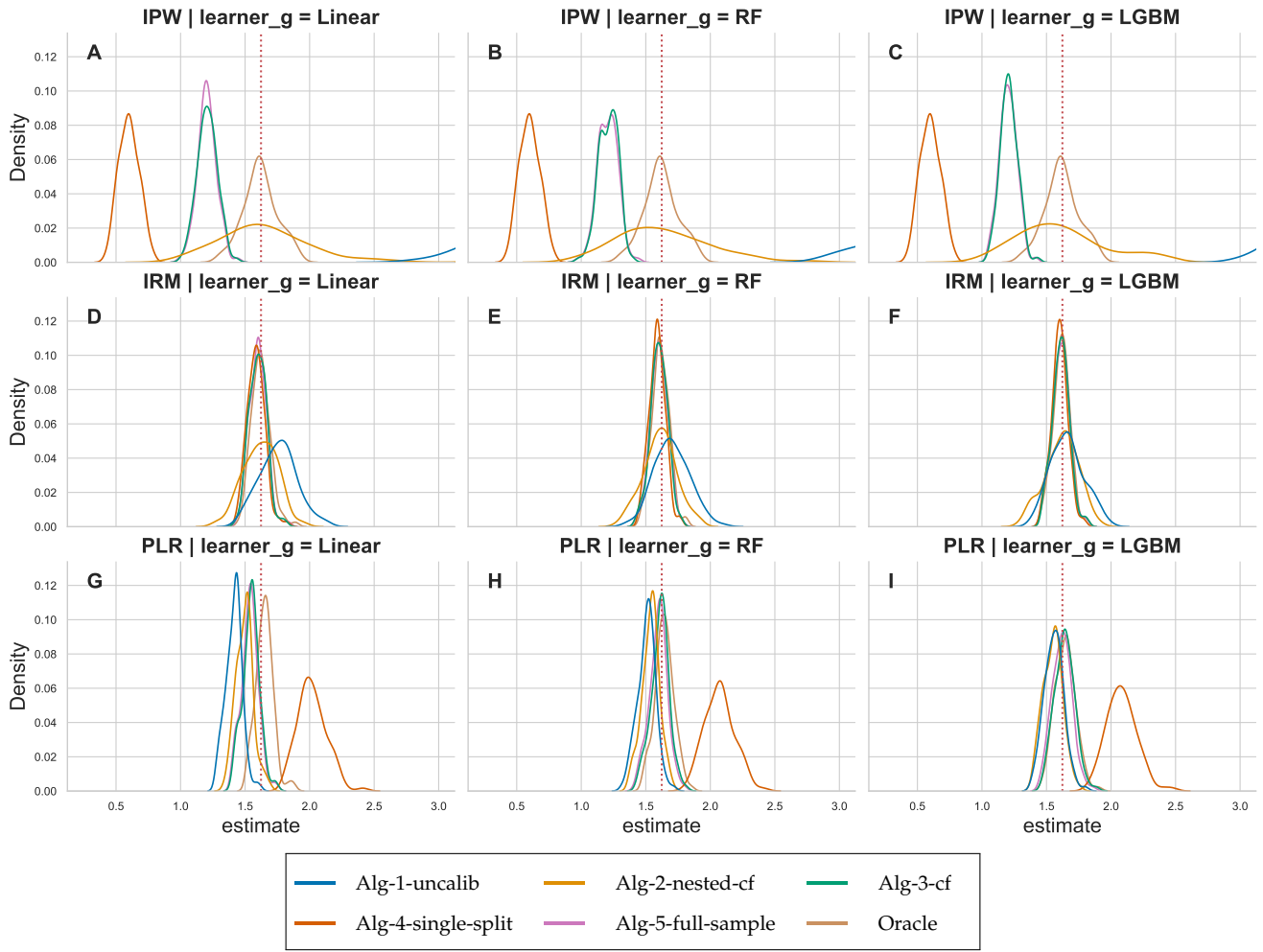

Figure S45: DGP 3 Nonlinear, m = LGBM, g = LGBM, n = 2000, p = 4, Calibration method for Algorithms 2-5: isotonic regression, Clip = 0.01

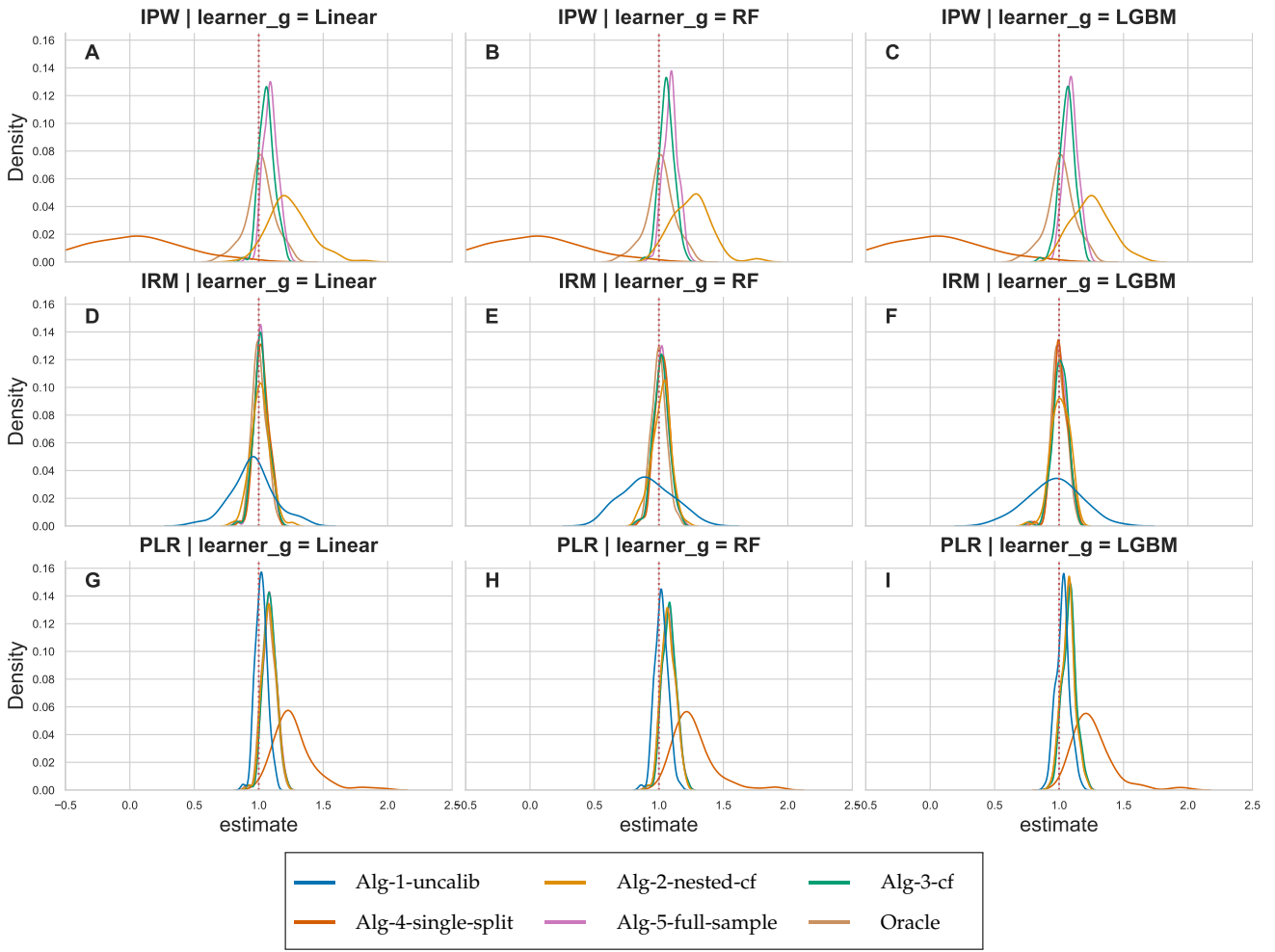

Figure S46: DGP 4 Unbalanced, Share_treated = 0.1, m = LGBM, g = LGBM, n = 4000, p = 20, Calibration method for Algorithms 2-5: isotonic regression, Clip = 0.01

**Number of Covariates on ATE Distribution**

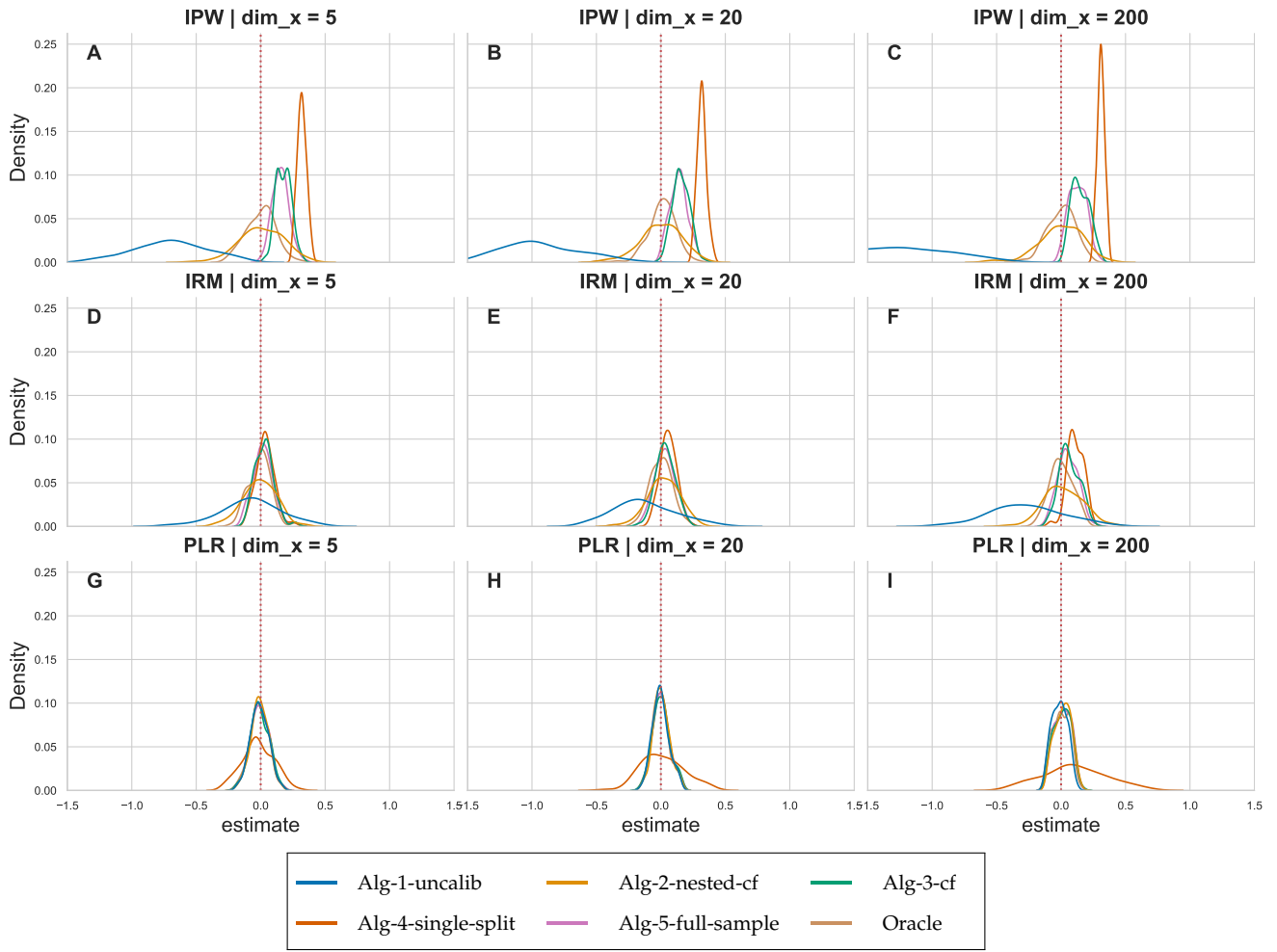

Figure S47: DGP 1 IRM, R2D = 0.5, m = LGBM, g = LGBM, n = 2000, Calibration method for Algorithms 2-5: isotonic regression, Clip = 0.01

**Clipping Threshold on ATE Distribution**

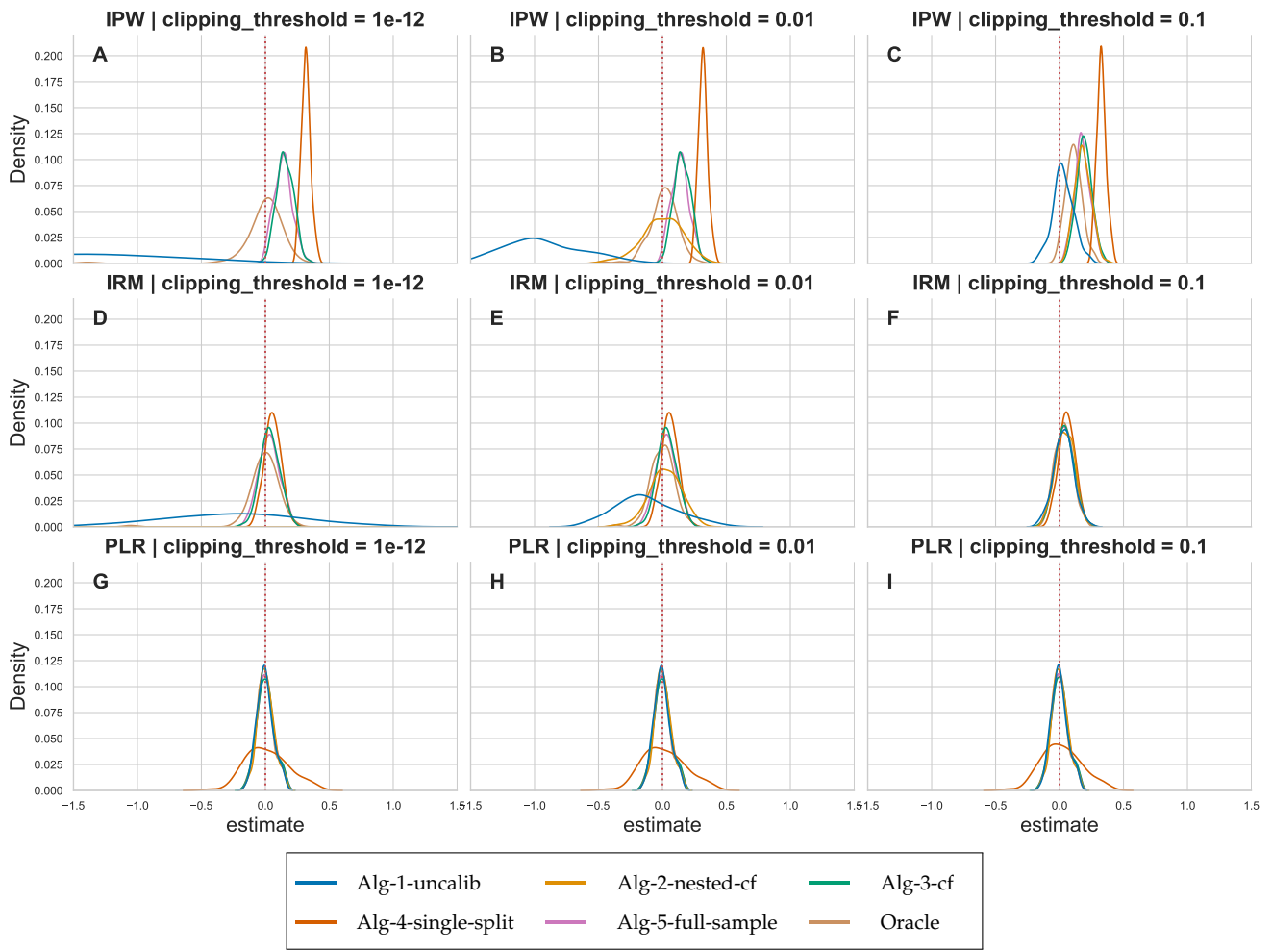

Figure S48: DGP 1 IRM, R2D = 0.5, m = LGBM, g = LGBM, n = 2000, p = 20, Calibration method for Algorithms 2-5: isotonic regression

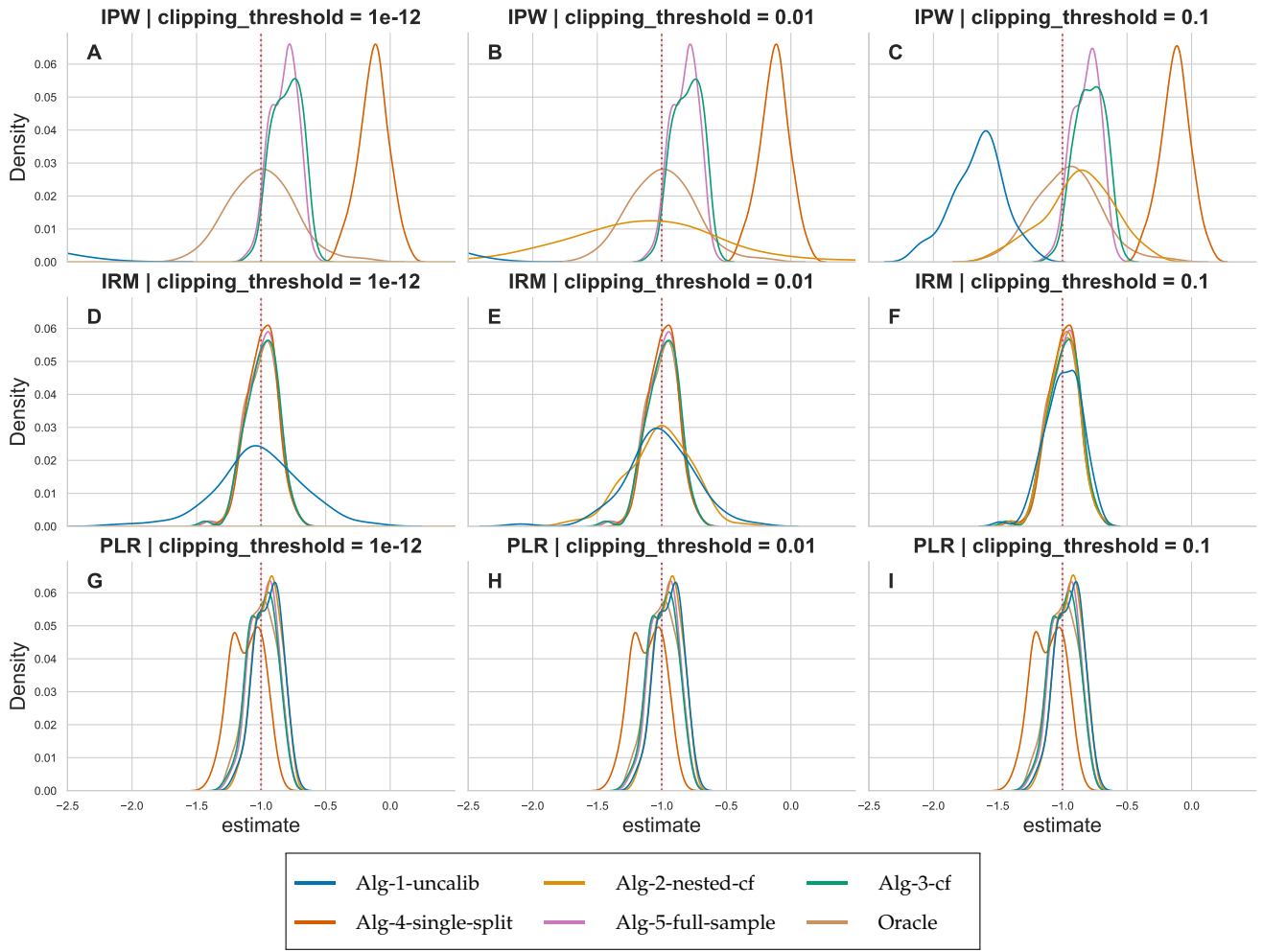

Figure S49: DGP 2 Drug, Overlap = 0.5, m = LGBM, g = LGBM, n = 2000, p = 3, Calibration method for Algorithms 2-5: isotonic regression

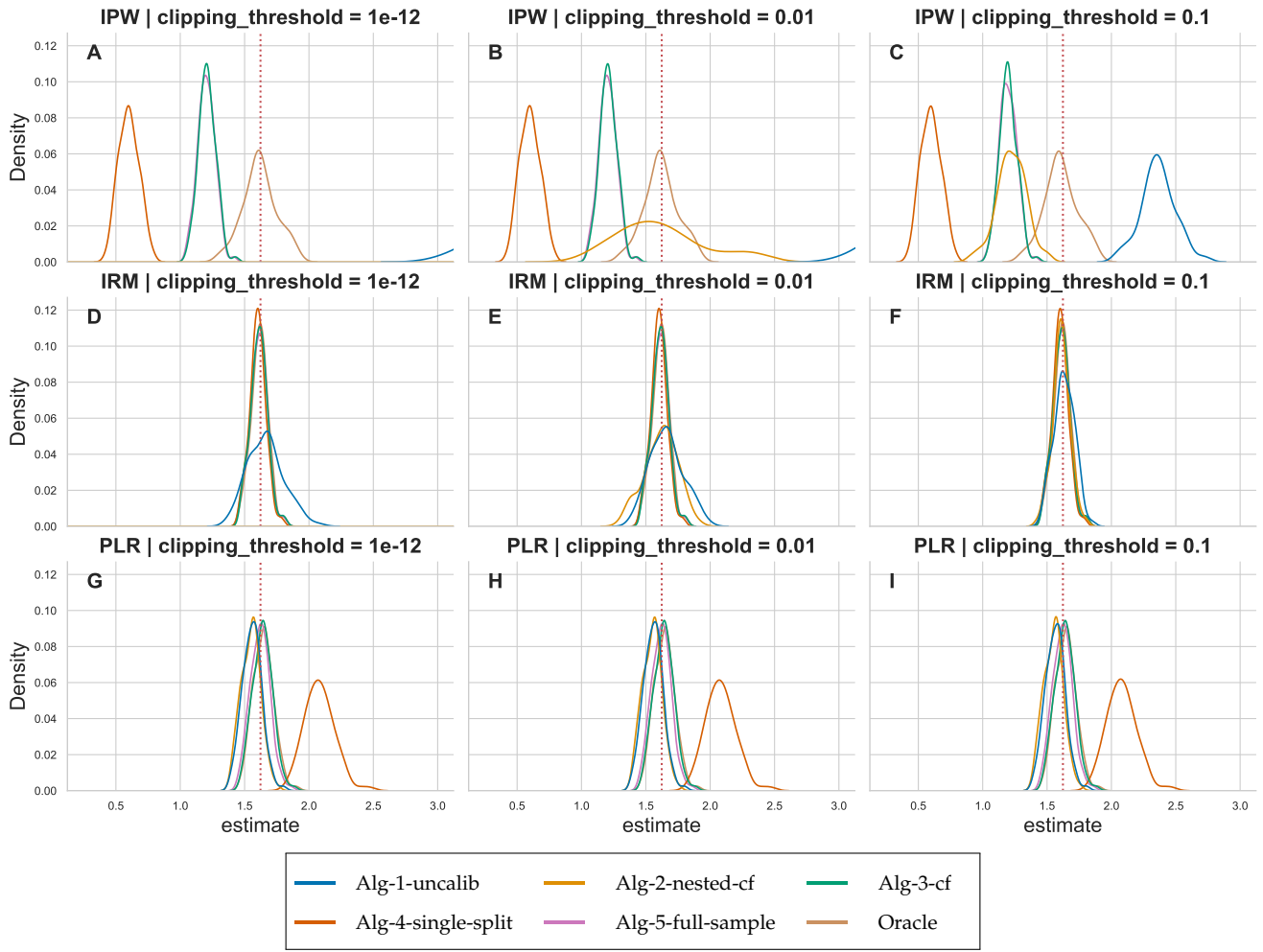

Figure S50: DGP 3 Nonlinear, m = LGBM, g = LGBM, n = 2000, p = 4, Calibration method for Algorithms 2-5: isotonic regression, Clip = 0.01

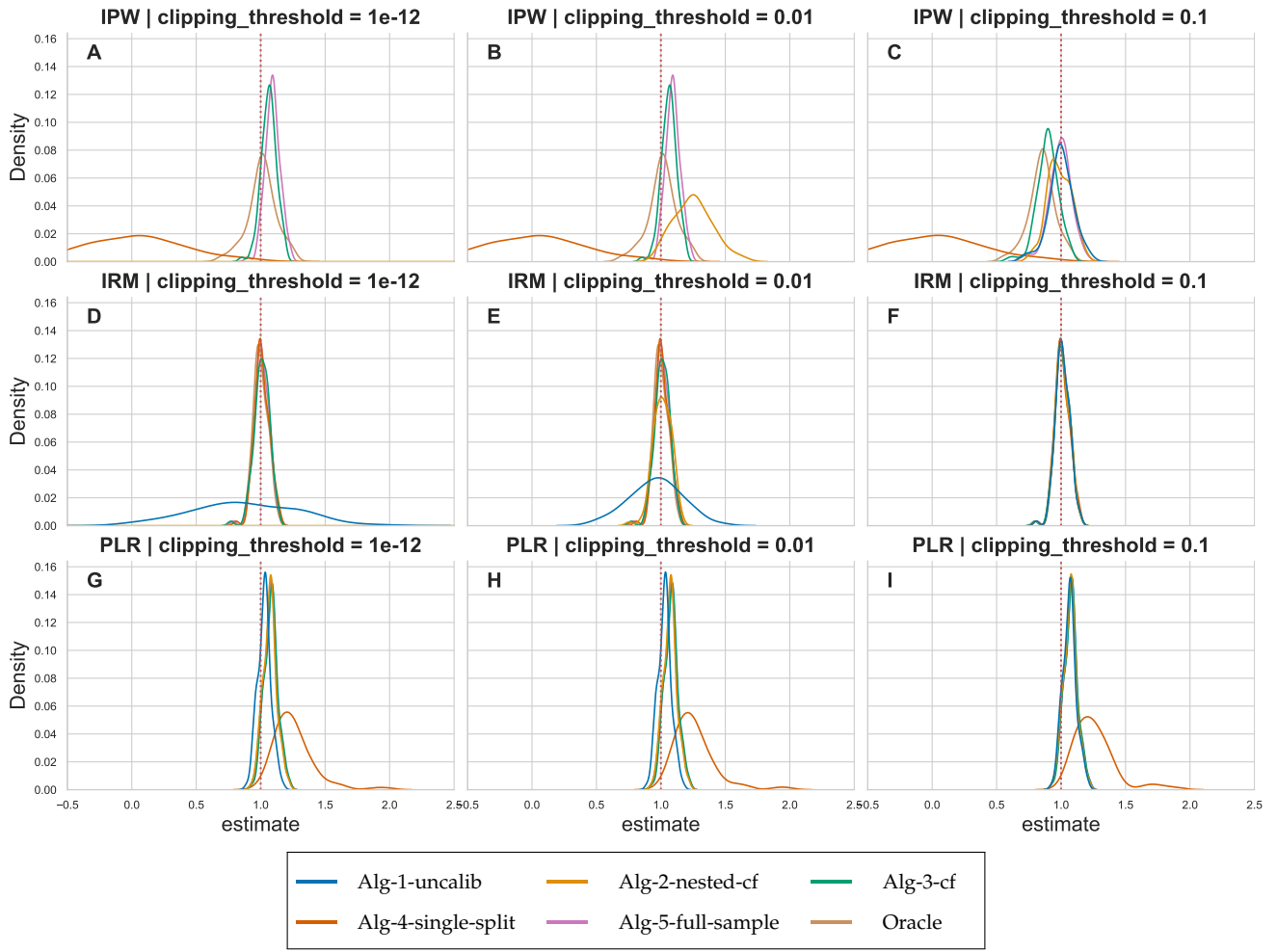

Figure S51: DGP 4 Unbalanced, Share_treated = 0.1, m = LGBM, g = LGBM, n = 4000, p = 20, Calibration method for Algorithms 2-5: isotonic regression

**DGP Setting on ATE Distribution**

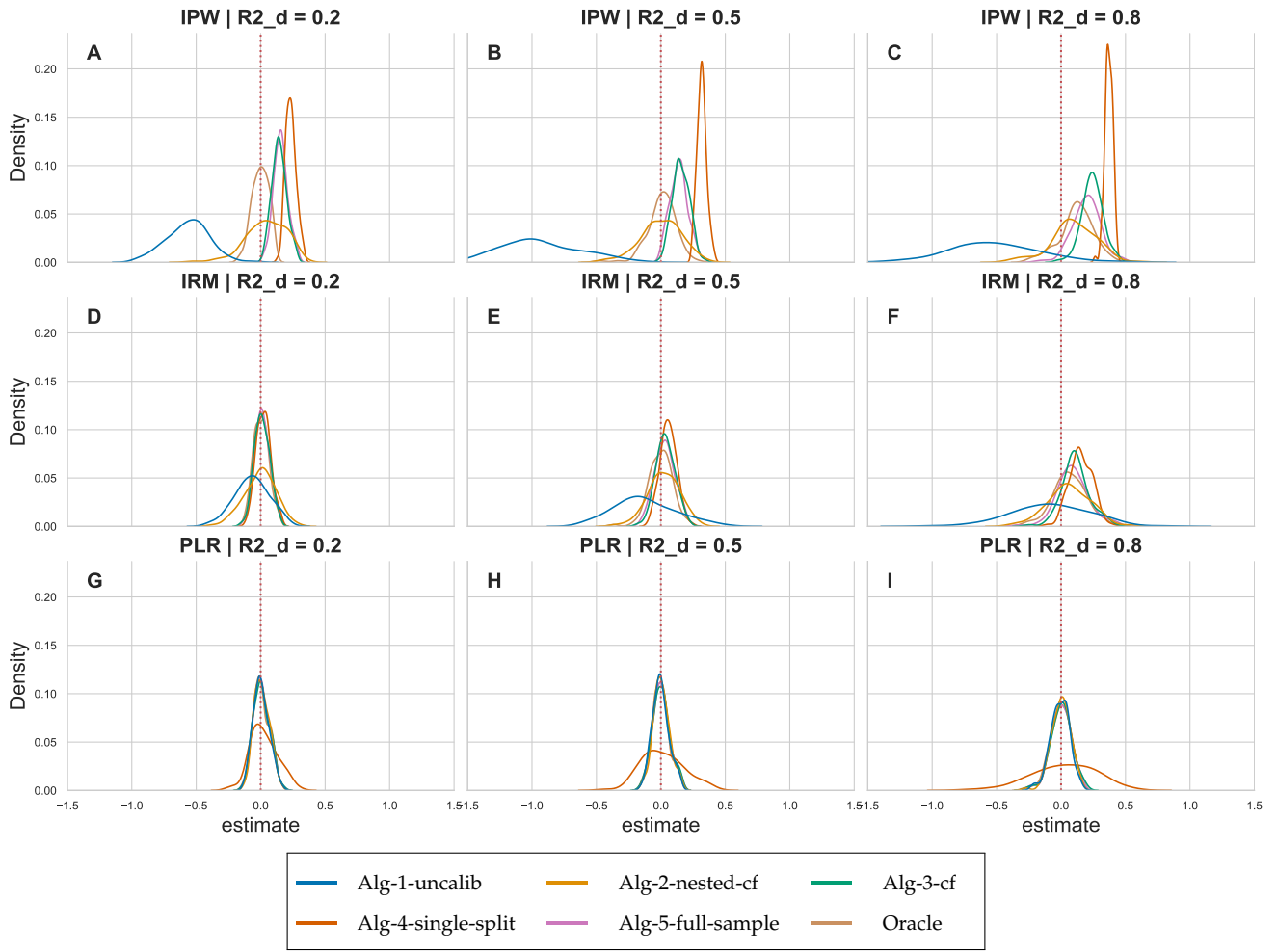

Figure S52: DGP 1 IRM, R2D on ATE, m = LGBM, g = LGBM, n = 2000, p = 20, Calibration method for Algorithms 2-5: isotonic regression, Clip = 0.01

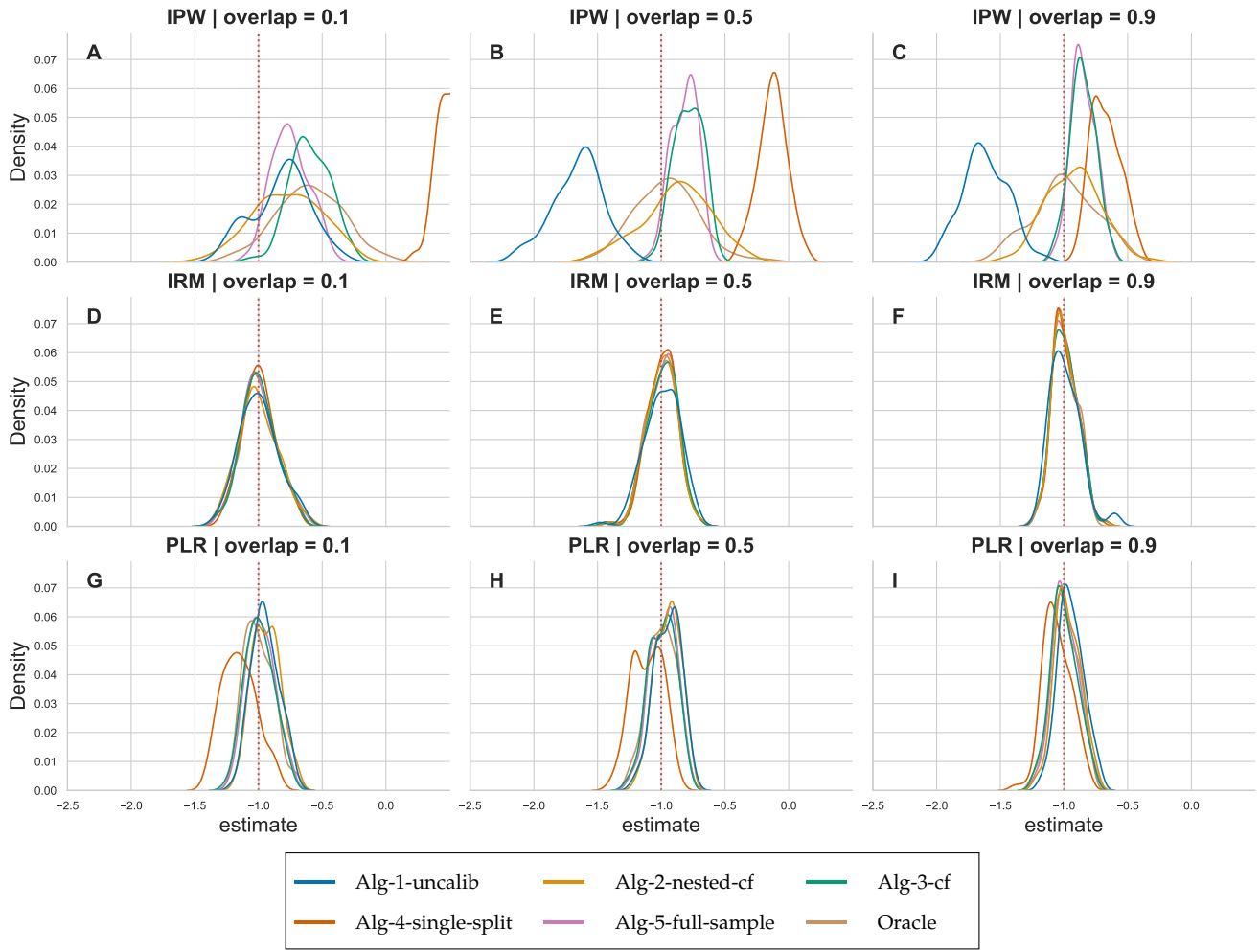

Figure S53: DGP 2 Drug, Overlap on ATE, m = LGBM, g = LGBM, n = 2000, p = 3, Calibration method for Algorithms 2-5: isotonic regression, Clip = 0.01

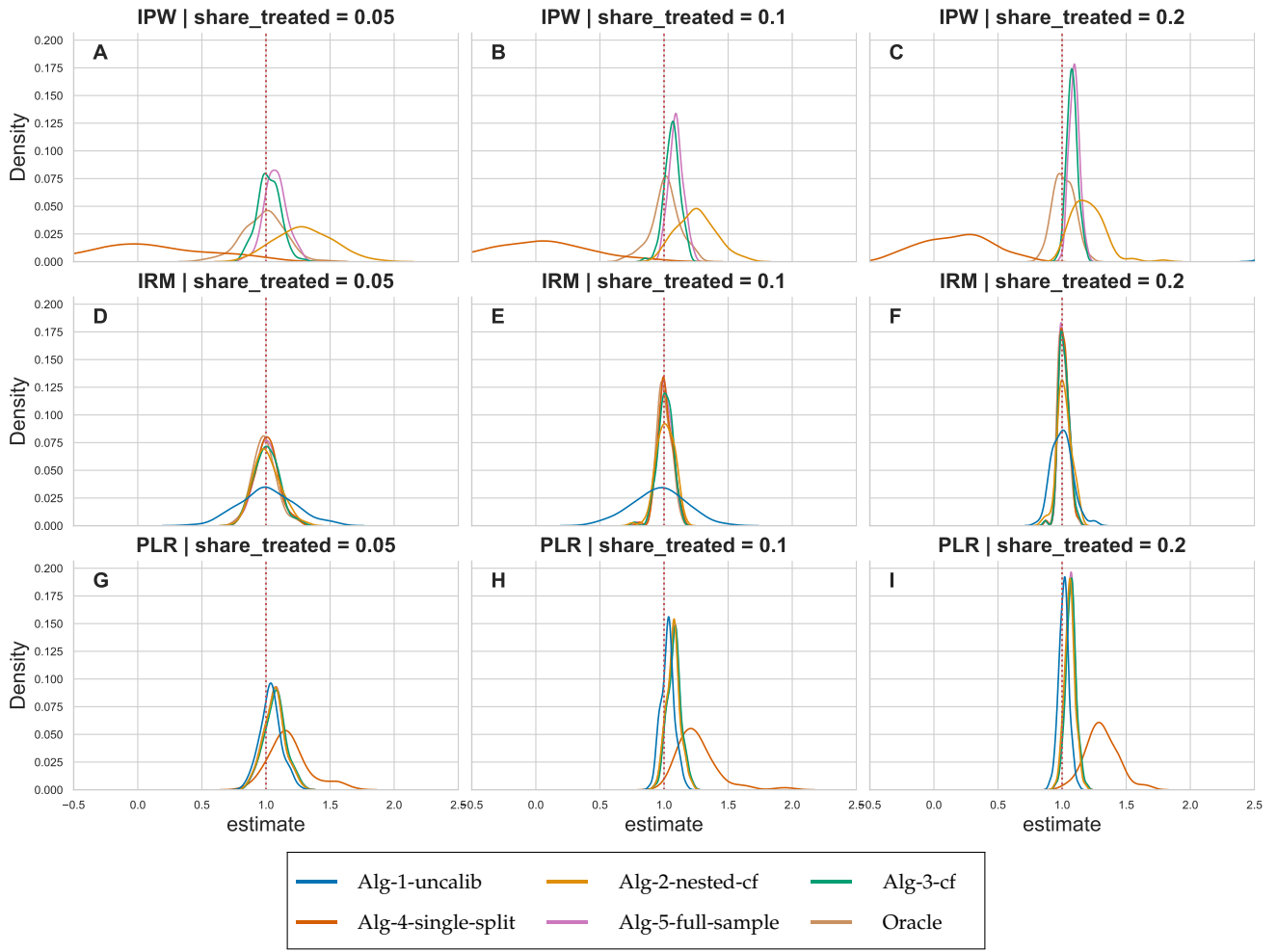

Figure S54: DGP 4 Unbalanced, Share_treated on ATE, m = LGBM, g = LGBM, n = 4000, p = 20, Calibration method for Algorithms 2-5: isotonic regression, Clip = 0.01


\fi

\end{document}